%% file: main.tex
\documentclass{article} 
\usepackage{iclr2027_conference,times}

\input{math_commands.tex}

\PassOptionsToPackage{hyphens}{url}
\usepackage{hyperref}
\usepackage{url}

\input{preamble}

\newif\ifpreprint
\preprinttrue
\hypersetup{
  pdftitle={DeclareViT: Decoupling Early Classification from Feature Refinement with Exit Query for Efficient ViTs},
  pdfauthor={Xinrui Gong, Yao Rong, Chuangtao Chen, Daniel Ochs, Ulf Schlichtmann, Bing Li, Kristian Kersting, Grace Li Zhang}
}

\title{DeclareViT: Decoupling Early Classification from Feature Refinement with Exit Query for Efficient ViTs}

\author{%
Xinrui Gong$^{1}$, Yao Rong$^{2}$, Chuangtao Chen$^{3}$, Daniel Ochs$^{1}$, Ulf Schlichtmann$^{3}$, Bing Li$^{4}$ \\
\textbf{Kristian Kersting$^{5,6,1}$, Grace Li Zhang$^{1}$} \\
$^{1}$Technische Universität Darmstadt $^{2}$Leibniz Universität Hannover $^{3}$Technische Universität München \\
$^{4}$Technische Universität Ilmenau 
$^{5}$German Research Center for AI $^{6}$The Hessian Center for AI
}

\iclrfinalcopy 
\begin{document}

\maketitle
\ifpreprint\lhead{}\fi
\input{sections/0_abstract}

\input{sections/1_introduction}

\input{sections/2_related_work}

\input{sections/3_Preliminary}

\input{sections/4_method}

\input{sections/5_experiments}

\input{sections/6_conclusion}

\input{sections/AI_use_statement}
{
    \small
    \bibliographystyle{iclr2027_conference}
    \bibliography{bibtex/short, iclr2027_conference}
}

\clearpage
\input{sections/supplement}

\end{document}

%% file: math_commands.tex
\usepackage{amsmath,amsfonts,bm}

\def\eqref#1{equation~\ref{#1}}

\def\1{\bm{1}}

\DeclareMathAlphabet{\mathsfit}{\encodingdefault}{\sfdefault}{m}{sl}
\SetMathAlphabet{\mathsfit}{bold}{\encodingdefault}{\sfdefault}{bx}{n}

\DeclareMathOperator{\sign}{sign}

%% file: preamble.tex
\usepackage{tabularx}
\usepackage{multirow}
\usepackage{booktabs}
\usepackage[per-mode=symbol, group-digits=integer, detect-all=true, input-symbols={()}]{siunitx}
\usepackage{bm}       
\usepackage{graphicx}
\usepackage{tikz}
\usetikzlibrary{fit}
\usetikzlibrary{positioning}
\usetikzlibrary{shapes.geometric, shapes, arrows, arrows.meta, bending}
\usetikzlibrary{backgrounds}
\usetikzlibrary{spy}
\usetikzlibrary{calc}

\usepackage{pgfplots}
\pgfplotsset{compat=1.17}
\usepackage[export]{adjustbox}
\usepackage[table]{xcolor}
\usepackage{tikz}
\usepackage{booktabs}
\usepackage{subcaption}

\definecolor{iccvblue}{rgb}{0.21,0.49,0.74}
\definecolor{tud0d}{RGB}{83,83,83}
\definecolor{tud0c}{RGB}{137,137,137}
\definecolor{tud0b}{RGB}{181,181,181}
\definecolor{tud0a}{RGB}{220,220,220}
\definecolor{tud1a}{RGB}{93,133,195}
\definecolor{tud2a}{RGB}{0,156,218}
\definecolor{tud3a}{RGB}{80,182,149}
\definecolor{tud4a}{RGB}{175,204,80}
\definecolor{tud5a}{RGB}{221,223,72}
\definecolor{tud6a}{RGB}{255,224,92}
\definecolor{tud7a}{RGB}{248,186,60}
\definecolor{tud8a}{RGB}{238,122,52}
\definecolor{tud9a}{RGB}{233,80,62}
\definecolor{tud10a}{RGB}{201,48,142}
\definecolor{tud11a}{RGB}{128,69,151}
\definecolor{tud1b}{RGB}{0,90,169}
\definecolor{tud2b}{RGB}{0,131,204}
\definecolor{tud3b}{RGB}{0,157,129}
\definecolor{tud4b}{RGB}{153,192,0}
\definecolor{tud5b}{RGB}{201,212,0}
\definecolor{tud6b}{RGB}{253,202,0}
\definecolor{tud7b}{RGB}{245,163,0}
\definecolor{tud8b}{RGB}{236,101,0}
\definecolor{tud9b}{RGB}{230,0,26}
\definecolor{tud10b}{RGB}{166,0,132}
\definecolor{tud11b}{RGB}{114,16,133}
\definecolor{tud1c}{RGB}{0,78,138}
\definecolor{tud2c}{RGB}{0,104,157}
\definecolor{tud3c}{RGB}{0,136,119}
\definecolor{tud4c}{RGB}{127,171,22}
\definecolor{tud5c}{RGB}{177,189,0}
\definecolor{tud6c}{RGB}{215,172,0}
\definecolor{tud7c}{RGB}{210,135,0}
\definecolor{tud8c}{RGB}{204,76,3}
\definecolor{tud9c}{RGB}{185,15,34}
\definecolor{tud10c}{RGB}{149,17,105}
\definecolor{tud11c}{RGB}{97,28,115}
\definecolor{tud1d}{RGB}{36,53,114}
\definecolor{tud2d}{RGB}{0,78,115}
\definecolor{tud3d}{RGB}{0,113,94}
\definecolor{tud4d}{RGB}{106,139,55}
\definecolor{tud5d}{RGB}{153,166,4}
\definecolor{tud6d}{RGB}{174,142,0}
\definecolor{tud7d}{RGB}{190,111,0}
\definecolor{tud8d}{RGB}{169,73,19}
\definecolor{tud9d}{RGB}{156,28,38}
\definecolor{tud10d}{RGB}{115,32,84}
\definecolor{tud11d}{RGB}{76,34,106}
\definecolor{unlabeled}{RGB}{0,0,0}
\definecolor{egovehicle}{RGB}{0,0,0}
\definecolor{rectification border}{RGB}{0,0,0}
\definecolor{outofroi}{RGB}{0,0,0}
\definecolor{static}{RGB}{0,0,0}
\definecolor{dynamic}{RGB}{111,74,0}
\definecolor{ground}{RGB}{81,0,81}
\definecolor{road}{RGB}{128,64,128}
\definecolor{sidewalk}{RGB}{244,35,232}
\definecolor{parking}{RGB}{250,170,160}
\definecolor{rail track}{RGB}{230,150,140}
\definecolor{building}{RGB}{70,70,70}
\definecolor{wall}{RGB}{102,102,156}
\definecolor{fence}{RGB}{190,153,153}
\definecolor{guard rail}{RGB}{180,165,180}
\definecolor{bridge}{RGB}{150,100,100}
\definecolor{tunnel}{RGB}{150,120,90}
\definecolor{pole}{RGB}{153,153,153}
\definecolor{polegroup}{RGB}{153,153,153}
\definecolor{trafficlight}{RGB}{250,170,30}
\definecolor{trafficsign}{RGB}{220,220,0}
\definecolor{vegetation}{RGB}{107,142,35}
\definecolor{terrain}{RGB}{152,251,152}
\definecolor{sky}{RGB}{70,130,220}
\definecolor{skylight}{RGB}{98,182,252}
\definecolor{person}{RGB}{220,20,60}
\definecolor{rider}{RGB}{255,0,0}
\definecolor{car}{RGB}{0,0,142}
\definecolor{truck}{RGB}{0,0,70}
\definecolor{bus}{RGB}{0,60,100}
\definecolor{caravan}{RGB}{0,0,90}
\definecolor{trailer}{RGB}{0,0,110}
\definecolor{train}{RGB}{0,80,100}
\definecolor{motorcycle}{RGB}{0,0,230}
\definecolor{bicycle}{RGB}{119,11,32}
\definecolor{licenseplate}{RGB}{0,0,142}

\usepackage{float}
\usepackage{xcolor}
\usepackage{tikz}
\usepackage{pgfplots}
\usepackage{subcaption}
\usetikzlibrary{plotmarks}

\definecolor{oursblue}{RGB}{31,119,180}
\definecolor{viteeorange}{RGB}{255,127,14}
\definecolor{byotgreen}{RGB}{44,160,44}
\definecolor{sdnred}{RGB}{214,39,40}
\definecolor{lgvitpurple}{RGB}{148,103,189}
\definecolor{dmpobrown}{RGB}{140,86,75}
\definecolor{coseepink}{RGB}{227,119,194}
\definecolor{envitblack}{RGB}{60,60,60}
\definecolor{backbonegray}{RGB}{125,125,125}

\pgfplotsset{compat=1.18}

\pgfplotsset{
    tradeoff/.style={
        scale only axis,
        width=0.85\linewidth,
        height=0.63\linewidth,
        xlabel={$\mathbf{S}_{\mathrm{MAC}}$},
        ylabel={Accuracy (\%)},
        xlabel style={
            font=\footnotesize,
            yshift=1pt
        },
        ylabel style={
            font=\footnotesize,
            at={(axis description cs:-0.085,0.5)},
            anchor=south
        },
        tick label style={
            font=\tiny
        },
        yticklabel style={
            xshift=1pt
        },
        xticklabel={
            \pgfmathprintnumber[
                fixed,
                fixed zerofill,
                precision=1
            ]{\tick}
        },
        axis lines=left,
        tick align=outside,
        major tick length=2pt,
        minor tick num=0,
        grid=major,
        major grid style={
            dashed,
            gray!22,
            line width=0.30pt
        },
        axis line style={
            black,
            line width=0.45pt
        },
        tick style={
            black,
            line width=0.40pt
        },
        clip=false
    },
    tradeofflegend/.style={
        legend columns=-1,
        legend style={
            draw=gray!55,
            line width=0.30pt,
            fill=white,
            fill opacity=0.96,
            text opacity=1,
            font=\fontsize{6pt}{6.5pt}\selectfont,
            cells={anchor=west},
            inner xsep=2pt,
            inner ysep=1.2pt,
            /tikz/every even column/.append style={
                column sep=3pt
            }
        },
        legend image post style={
            scale=0.70
        },
        legend cell align=left
    },
    ours/.style={
        color=oursblue,
        line width=0.75pt,
        mark=o,
        mark size=1.35pt,
        mark options={
            solid,
            fill=white,
            line width=0.55pt
        }
    },
    envit/.style={
        color=envitblack,
        line width=0.65pt,
        mark=star,
        mark size=1.45pt
    },
    vitee/.style={
        color=viteeorange,
        line width=0.60pt,
        mark=square,
        mark size=1.30pt,
        mark options={
            solid,
            fill=white,
            line width=0.50pt
        }
    },
    byot/.style={
        color=byotgreen,
        line width=0.60pt,
        mark=triangle,
        mark size=1.40pt,
        mark options={
            solid,
            fill=white,
            line width=0.50pt
        }
    },
    sdn/.style={
        color=sdnred,
        line width=0.60pt,
        mark=diamond,
        mark size=1.40pt,
        mark options={
            solid,
            fill=white,
            line width=0.50pt
        }
    },
    lgvit/.style={
        color=lgvitpurple,
        line width=0.60pt,
        mark=triangle,
        mark size=1.40pt,
        mark options={
            solid,
            fill=white,
            rotate=180,
            line width=0.50pt
        }
    },
    dmpo/.style={
        color=dmpobrown,
        line width=0.60pt,
        mark=pentagon,
        mark size=1.30pt,
        mark options={
            solid,
            fill=white,
            line width=0.50pt
        }
    },
    cosee/.style={
        color=coseepink,
        line width=0.60pt,
        mark=asterisk,
        mark size=1.40pt
    },
    backbone/.style={
        color=backbonegray,
        only marks,
        mark=triangle,
        mark size=1.85pt,
        mark options={
            solid,
            fill=white,
            line width=0.70pt
        }
    }
}
\newcommand{\revcolor}{}

\input{overview_layout.tex}

\usepackage{placeins}

%% file: sections/0_abstract.tex
\begin{abstract}
Vision Transformers achieve remarkable recognition performance but incur substantial computational cost during inference, limiting their use in latency- and compute-constrained deployment scenarios. Early-exit inference offers a promising solution by allowing easy samples to terminate at intermediate layers. 
However, existing approaches typically couple early classification with the shared backbone, requiring intermediate representations to support both early classification and subsequent feature refinement.
Under joint training, the corresponding optimization objectives jointly update shared backbone parameters, which can lead to conflicting gradient directions. 
{\revcolor Motivated by this observation}, we introduce DeclareViT, a framework that
decouples early classification from feature refinement. Each exit uses a dedicated query to read intermediate features through one-way cross-attention without modifying the backbone token stream. The query branches are optimized for early classification while the backbone and final classifier remain frozen. Across ViT, DeiT, and Swin on CIFAR-100, ImageNet-1K, and Food-101, DeclareViT consistently establishes a state-of-the-art accuracy--efficiency trade-off: 
It achieves
$1.58\times$--$2.57\times$ multiply--accumulate (MAC)-based speed-up with only a $0.39$--$1.91$ percentage-point accuracy drop. Wall-clock measurements on CPU and GPU further demonstrate practical inference acceleration. Code is available at
\url{https://github.com/TUDa-HWAI/DeclareViT}.

\end{abstract}

%% file: sections/1_introduction.tex
\section{Introduction}
\label{sec:introduction}

Vision Transformers (ViTs) have achieved impressive performance across diverse visual-recognition tasks~\citep{Dosovitskiy:2021:AII, Touvron:2021:TDI, Liu:2021:STH}.
However, standard ViTs process every input through all Transformer blocks, which can be inefficient in latency-sensitive and compute-constrained deployment scenarios.
\textit{Early-exit} techniques address this limitation by inserting classifiers at intermediate layers and terminating computation once a sufficiently confident prediction is reached~\citep{Teerapittayanon:2016:BFI,
Huang:2018:MSD}. This enables efficient inference by allocating less computation to easy samples while retaining full model capacity 
for difficult ones.

Recent work improves early-exit prediction through specialized
heads~\citep{Bakhtiarnia:2021:MVT,Xu:2023:LDE},
self-distillation and structured dropout~\citep{Shen:2023:MEV,Dong:2026:EET},
prediction reuse~\citep{Zhou:2020:BLP,Zhang:2022:PAB,
Wang:2024:ECE,Bajpai:2025:BBP}, and routing-aware
supervision~\citep{Han:2022:LTW,He:2025:CCS}. 
Other designs combine early termination with adaptive spatial
granularity or model capacity~\citep{Wang:2021:NAI, Chen:2023:AGC,
Hu:2024:LRS,Hojjat:2026:TMT}.

\begin{figure}[t]
    \centering
    \input{artwork/teaser}

    \begin{subfigure}[t]{0.49\linewidth}
        \centering
        \DrawTeaserBaseline
        \vspace{0.55em}
        \subcaption{Conventional [\textsc{cls}]-based early exits.}
        \label{fig:teaser_baseline}
    \end{subfigure}
    \hfill
    \begin{subfigure}[t]{0.49\linewidth}
        \centering
        \DrawTeaserOurs
        \vspace{0.55em}
        \subcaption{DeclareViT.}
        \label{fig:teaser_ours}
    \end{subfigure}

    \caption{
    Comparison between conventional  [\textsc{cls}]-based early exits and DeclareViT. (a) Conventional [\textsc{cls}]-based early exits using the shared backbone
[\textsc{cls}] representation for early prediction.
(b) DeclareViT with read-only queries $\mathbf q_1$ and $\mathbf q_2$
progressively updated across backbone blocks.}
    \label{fig:teaser}
\end{figure}
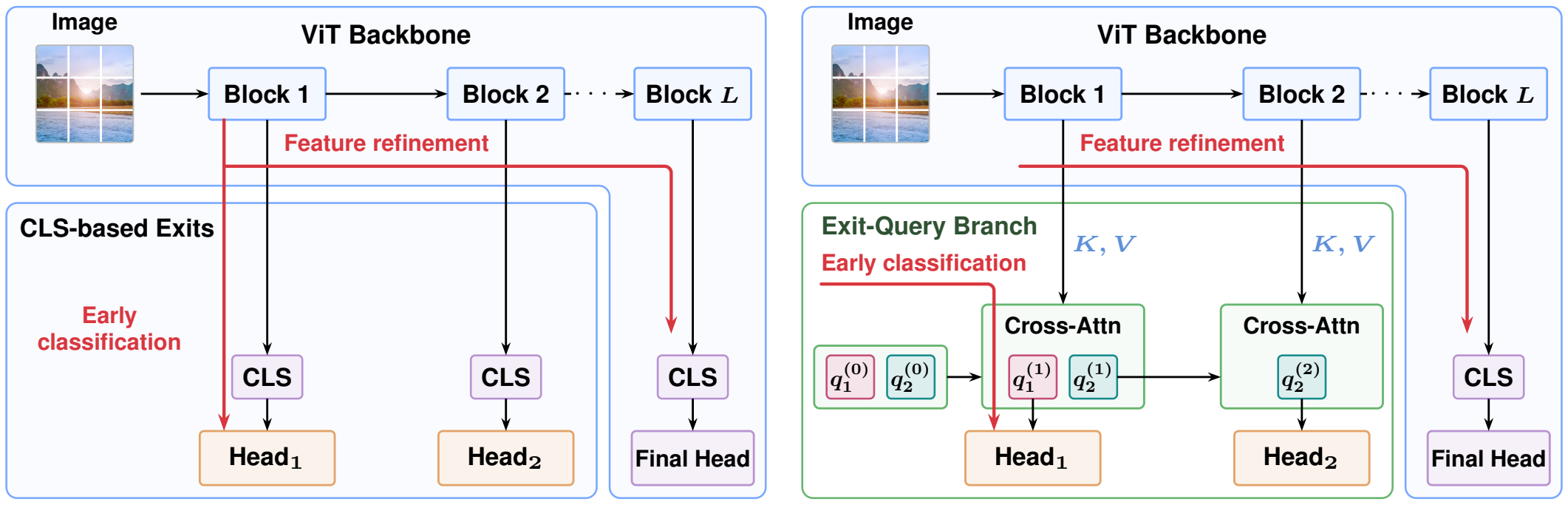

Despite these advances, the interaction between early classification and the backbone representations required for deeper feature refinement remains largely coupled. In particular, many early-exit ViTs directly classify the intermediate class token, denoted as [\textsc{cls}], and jointly optimize the intermediate exit classifiers and the final classifier as shown in Figure~\ref{fig:teaser_baseline}~\citep{Bakhtiarnia:2021:MVT, Xu:2023:LDE, Dong:2026:EET}.
This design assigns two roles to the same representation: the intermediate [\textsc{cls}] token must support both early classification and feature refinement by subsequent Transformer blocks. We refer to this coupling as \textit{[\textsc{cls}] role coupling}.
Under joint training, the corresponding optimization
objectives jointly update shared backbone parameters, which can lead to conflicting
gradient directions. 
Existing decoupling approaches alleviate this problem through separate
classification branches~\citep{Han:2023:DPF}, feature
partitioning~\citep{Gong:2024:DFS}, or adapters and bypass
predictors~\citep{Zhang:2024:DTD,Liu:2026:MTT,Luo:2025:DMO}.
However, none of these decoupling methods establishes a fully read-only and gradient-isolated interface between early classification and shared feature refinement.

To address the challenge above, we introduce \textbf{DeclareViT}, a framework for \textbf{DE}coupling early \textbf{CLA}ssification from feature
\textbf{RE}finement in early-exit Vision Transformers.
As shown in Figure~\ref{fig:teaser_ours}, in DeclareViT, each exit branch maintains an independent learnable \emph{exit query} outside the backbone token sequence.
As an input propagates through the backbone, each query reads intermediate features through one-way cross-attention and accumulates information through residual updates.
At each exit, a feed-forward network (FFN) refines the corresponding query, and a linear head produces the class logits.
After an exit produces its prediction, the corresponding query is discarded.
Training proceeds in two stages. We first optimize the backbone and final head using only the full-depth training objective. Then we freeze the backbone and final classifier and optimize only the exit-query branches.
DeclareViT decouples early classification from feature refinement in both directions: in the forward pass, the queries read the intermediate features in the backbone but never modify them; in the backward pass, gradients from the exit-query losses do not propagate into the backbone.

In summary, our contributions are threefold:
\begin{itemize}
    \item
    We characterize [\textsc{cls}] role coupling in early-exit ViTs and {\revcolor use gradient-conflict analysis as a diagnostic of optimization competition among prediction objectives.}
    \item
    We propose \textbf{DeclareViT}, a framework for \textbf{DE}coupling early \textbf{CLA}ssification from feature \textbf{RE}finement. It combines independent exit queries,
    one-way cross-attention, and two-stage training to separate
    early classification from backbone feature refinement in both
    the forward and backward passes.
    \item
    We systematically evaluate DeclareViT with ViT, DeiT, and Swin on CIFAR-100, ImageNet-1K, and Food-101.
    Across all combinations, DeclareViT outperforms state-of-the-art early-exit methods in the
    accuracy--efficiency trade-off, achieving
    $1.58\times$--$2.57\times$ multiply–accumulate (MAC)-based speed-up, with an accuracy drop of 0.39–1.91 percentage points.
    Wall-clock measurements on GPU and CPU further demonstrate its practical computational efficiency.
\end{itemize}

%% file: artwork/teaser.tex
%

\providecommand{\teaserAssetDir}{artwork/teaser-assets}

\makeatletter
\newcommand{\teaserFontPlusOne}[1]{%
  #1%
  \fontsize{\dimexpr\f@size pt+1pt\relax}%
           {\dimexpr\f@baselineskip+1pt\relax}\selectfont
}
\makeatother

\definecolor{panelborder}{HTML}{D3D6DB}
\definecolor{panelfillleft}{HTML}{FBFBFC}
\definecolor{panelfillright}{HTML}{F6FBF6}
\definecolor{accentleft}{HTML}{8F939A}
\definecolor{accentright}{HTML}{66B56F}

\definecolor{tokenpurple}{HTML}{B791D1}
\definecolor{tokenpurplelight}{HTML}{E7D6F1}
\definecolor{tokenblue}{HTML}{6FA8FF}

\definecolor{blockfill}{HTML}{F3F7FF}
\definecolor{blockborder}{HTML}{79A9FF}

\definecolor{headfill}{HTML}{FAF5FF}
\definecolor{headborder}{HTML}{BFA1DE}

\definecolor{exitgreen}{HTML}{66B56F}
\definecolor{querygreen}{HTML}{9ABE8B}
\definecolor{queryorange}{HTML}{E2A469}
\definecolor{querypurple}{HTML}{B994D0}
\definecolor{crossattncolor}{HTML}{C43C8C}

\tikzset{
  every node/.style={font=\sffamily},
  flow/.style={
    -{Stealth[length=2.3mm,width=1.5mm]},
    draw=black!75,
    line width=.42pt
  },
  transformer/.style={
    draw=blockborder,
    fill=blockfill,
    rounded corners=1.2pt,
    minimum width=1.85cm,
    minimum height=.72cm,
    inner sep=1.2pt,
    font=\sffamily\bfseries\teaserFontPlusOne{\scriptsize}
  },
  linearhead/.style={
    draw=headborder,
    fill=headfill,
    rounded corners=2pt,
    minimum width=1.48cm,
    minimum height=.90cm,
    align=center,
    inner sep=1.4pt,
    font=\sffamily\bfseries\teaserFontPlusOne{\scriptsize}
  },
  querylinearhead/.style={
    linearhead,
    draw=#1,
    fill=#1!13
  },
  crossattention/.style={
    -{Stealth[length=2.5mm,width=1.7mm]},
    draw=crossattncolor,
    line width=.75pt
  },
  crossattentionlabel/.style={
    font=\sffamily\bfseries\teaserFontPlusOne{\scriptsize},
    text=crossattncolor,
    fill=white,
    inner sep=.6pt
  },
  kvlabel/.style={
    font=\sffamily\bfseries\teaserFontPlusOne{\scriptsize},
    text=crossattncolor,
    fill=white,
    inner sep=.25pt
  },
  exitpoint/.style={
    draw=exitgreen,
    densely dashed,
    fill=white,
    rounded corners=2pt,
    minimum width=1.45cm,
    minimum height=.74cm,
    align=center,
    inner sep=1.2pt,
    font=\sffamily\bfseries\teaserFontPlusOne{\scriptsize}
  },
  query/.style={
    draw=#1,
    fill=#1!13,
    rounded corners=1.6pt,
    minimum width=1.42cm,
    minimum height=.63cm,
    align=center,
    inner sep=1pt,
    font=\sffamily\bfseries\teaserFontPlusOne{\scriptsize}
  }
}


\newcommand{\teaserLandscapeIcon}{%
  \begin{scope}
    \clip[rounded corners=1.5pt] (0,0) rectangle (2.35,2.35);
    \node[anchor=south west,inner sep=0pt] at (0,0) {%
      \includegraphics[width=2.35cm,height=2.35cm]{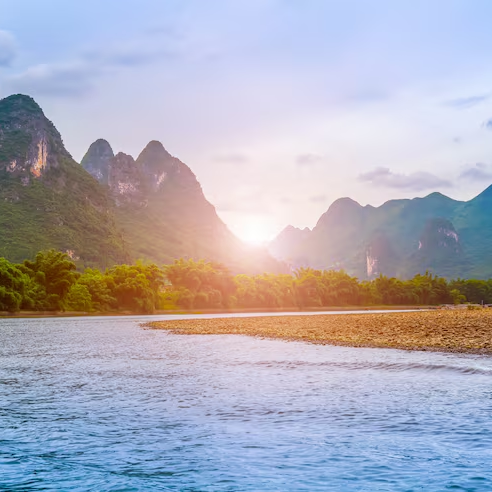}%
    };
  \end{scope}
  \draw[black!30,rounded corners=1.5pt,line width=.35pt]
    (0,0) rectangle (2.35,2.35);
}

\newcommand{\teaserPatchEmbedding}{%
  \fill[tokenpurplelight!35,rounded corners=3pt]
    (.08,.28) rectangle (2.07,2.92);
  \draw[tokenpurple!80,rounded corners=3pt,line width=.35pt]
    (.08,.28) rectangle (2.07,2.92);

  \foreach \row/\yy in {1/2.16,2/1.36,3/.56}{%
    \foreach \col/\xx in {1/.25,2/.84,3/1.43}{%
      \node[anchor=south west,inner sep=0pt] at (\xx,\yy) {%
        \includegraphics[width=.47cm,height=.47cm]{%
          \teaserAssetDir/patch-r\row-c\col.png}%
      };
      \draw[tokenpurple!70,line width=.25pt]
        (\xx,\yy) rectangle ++(.47,.47);
    }%
  }%
}

\newcommand{\teaserTokenSequence}[1]{%
  \node[font=\teaserFontPlusOne{\tiny},align=right] at (#1-0.8,5.17) {CLS\\token};
  \fill[tokenpurple] (#1-.22,4.95) rectangle (#1+.22,5.39);
  \draw[tokenpurple!90] (#1-.22,4.95) rectangle (#1+.22,5.39);

  \foreach \yy in {4.38,3.76}{%
    \draw[tokenblue,line width=.55pt] (#1-.22,\yy-.22) rectangle (#1+.22,\yy+.22);
  }%
  \node[font=\teaserFontPlusOne{\scriptsize}] at (#1,3.3) {$\vdots$};
  \draw[tokenblue,line width=.55pt] (#1-.22,2.43) rectangle (#1+.22,2.87);
  \node[font=\teaserFontPlusOne{\tiny},align=right] at (#1-0.8,2.65) {Patch\\tokens};
}

\newcommand{\teaserCrossAttentionKV}[2]{%
  \draw[crossattention]
    ([xshift=-1.3mm]#1.south)--
    node[pos=.50,xshift=-3pt,kvlabel]{K}
    ([xshift=-1.3mm]#2.north);
  \draw[crossattention]
    ([xshift=1.3mm]#1.south)--
    node[pos=.50,xshift=3pt,kvlabel]{V}
    ([xshift=1.3mm]#2.north);
  \node[crossattentionlabel,anchor=west]
    at ([xshift=3.8mm,yshift=-3.5mm]#1.south) {Cross-Attn};
}

\newcommand{\teaserPanelFrame}[2]{%
  \fill[#1,rounded corners=5pt]
    (0,2.10) rectangle (16.60,8.05);

  \fill[#2,rounded corners=5pt]
    (0,8.37) rectangle (16.60,8.05);

  \draw[panelborder,rounded corners=5pt,line width=.50pt]
    (0,2.10) rectangle (16.60,8.05);
}
\newcommand{\teaserCommonInput}{%
  \node[font=\sffamily\bfseries\teaserFontPlusOne{\scriptsize},align=center] at (1.43,6.4) {Image};
  \begin{scope}[shift={(.25,3.15)}]\teaserLandscapeIcon\end{scope}

  \node[font=\sffamily\bfseries\teaserFontPlusOne{\scriptsize},align=center] at (4.50,6.4) {Patches};
  \begin{scope}[shift={(3.42,2.72)}]\teaserPatchEmbedding\end{scope}

  \node[font=\sffamily\bfseries\teaserFontPlusOne{\scriptsize}] at (7.34,6.4) {Token Sequence};
  \teaserTokenSequence{7.34}

  \draw[flow] (2.72,4.31)--(3.27,4.31);
  \draw[flow] (5.69,4.31)--(6.33,4.31);
}

\newcommand{\teaserTransformerChain}[1]{%
  \node[transformer] (#1b1) at (9.35,5.17) {Block 1};
  \node[transformer] (#1b2) at (12.08,5.17) {Block 2};
  \node[font=\bfseries\teaserFontPlusOne{\small}] at (13.63,5.17) {$\cdots$};
  \node[transformer] (#1bL) at (15.20,5.17) {Block L};

  \draw[flow] (7.62,5.17)--(#1b1.west);
  \draw[flow] (#1b1.east)--(#1b2.west);
  \draw[flow] (#1b2.east)--(13.37,5.17);
  \draw[flow] (13.88,5.17)--(#1bL.west);
}

\newcommand{\DrawTeaserBaseline}{%
\begingroup
\definecolor{tbBack}{HTML}{79A9FF}
\definecolor{tbBackFill}{HTML}{F3F7FF}
\definecolor{tbInk}{HTML}{000000}
\definecolor{tbMuted}{HTML}{000000}
\definecolor{tbRule}{HTML}{C9C9C9}
\definecolor{tbQueryOne}{HTML}{C44E78}
\definecolor{tbQueryTwo}{HTML}{168C8C}
\definecolor{tbCLS}{HTML}{B791D1}
\definecolor{tbFinal}{HTML}{B994D0}
\definecolor{tbExitHead}{HTML}{E2A469}
\definecolor{tbTokenBlue}{HTML}{6FA8FF}
\definecolor{tbReader}{HTML}{66B56F}
\definecolor{tbQueryFill}{HTML}{EFF7F0}
\definecolor{tbTaskRed}{HTML}{D9363E}
\resizebox{\linewidth}{!}{%
\begin{tikzpicture}[x=1mm,y=-1mm,
  every node/.style={font=\sffamily\bfseries\boldmath\fontsize{6.5}{7.5}\selectfont,
    text=tbInk,inner sep=0pt,outer sep=0pt,align=center},
  tb flow/.style={-{Stealth[length=1.2mm,width=.85mm]},
    draw=black,line width=.55pt},
  tb memory/.style={-{Stealth[length=1.2mm,width=.85mm]},
    draw=black,line width=.55pt},
  tb task arrow/.style={-{Stealth[length=1.6mm,width=1.1mm]},
    draw=tbTaskRed,line width=.85pt,line cap=round,line join=round},
  tb task label/.style={font=\sffamily\bfseries\fontsize{6}{7}\selectfont,
    text=tbTaskRed,align=center,inner sep=0pt},
  tb block/.style={draw=tbBack,fill=tbBackFill,line width=.55pt,
    rounded corners=.3mm,minimum width=10.8mm,minimum height=4.8mm},
  tb reader/.style={draw=tbReader,fill=tbQueryFill!48,line width=.55pt,
    rounded corners=.5mm,minimum width=15mm,minimum height=9.6mm},
  tb chip/.style={draw=##1,fill=##1!15,line width=.5pt,
    rounded corners=.4mm,minimum width=4.4mm,minimum height=4mm,
    font=\sffamily\bfseries\boldmath\fontsize{6.5}{7.5}\selectfont,text=black},
  tb head/.style={draw=##1,fill=##1!13,line width=.55pt,
    rounded corners=.4mm,minimum width=12.5mm,minimum height=5mm,
    inner xsep=.7pt,font=\sffamily\bfseries\boldmath\fontsize{6.5}{7.5}\selectfont},
  tb title/.style={font=\sffamily\bfseries\fontsize{6.8}{7.8}\selectfont,
    anchor=west},
  tb note/.style={font=\sffamily\bfseries\fontsize{6}{7}\selectfont,
    text=tbMuted},
  tb kv/.style={font=\sffamily\bfseries\boldmath\fontsize{6}{7}\selectfont,
    text=tbTokenBlue!85!black,fill=white,inner sep=.5pt}
]
\path[use as bounding box] (0,0) rectangle (70.8,46.0);

\filldraw[fill=tbBackFill!45,draw=tbBack,line width=.5pt,
  rounded corners=1mm]
  (.25,.25)--(70.55,.25)--(70.55,45.7)--(56.1,45.7)
  --(56.1,16.8)--(.25,16.8)--cycle;
\node[tb title,anchor=center] at (35.4,2.8) {ViT Backbone};

\node[tb note] at (7.5,1.8) {Image};
\begin{scope}
  \clip[rounded corners=.3mm] (3,3.8) rectangle (12,12.8);
  \node[inner sep=0pt] (tbPatchedImage) at (7.5,8.3)
    {\includegraphics[width=9mm,height=9mm]{\teaserAssetDir/input-square.png}};
  \foreach \xx in {6,9}{%
    \draw[white,line width=.55pt] (\xx,3.8)--(\xx,12.8);
  }
  \foreach \yy in {6.8,9.8}{%
    \draw[white,line width=.55pt] (3,\yy)--(12,\yy);
  }
\end{scope}
\draw[black!28,line width=.35pt,rounded corners=.3mm]
  (3,3.8) rectangle (12,12.8);
\node[tb block] (tbBOne) at (24.4,8.3) {Block 1};
\node[tb block] (tbBTwo) at (46.5,8.3) {Block 2};
\node[tb block] (tbBL) at (63.8,8.3) {Block $L$};
\draw[tb flow] (12.7,8.3)--(tbBOne.west);
\draw[tb flow] (tbBOne.east)--(tbBTwo.west);
\draw[tb flow] (tbBTwo.east)--(tbBL.west);
\node[font=\fontsize{6}{6}\selectfont,fill=tbBackFill!45,inner xsep=.5pt]
  at (54.6,8.3) {$\cdots$};

\filldraw[fill=tbBackFill!45,draw=tbBack,line width=.5pt,
  rounded corners=1mm] (.25,18.4) rectangle (54.9,45.7);
\node[tb title,font=\sffamily\bfseries\fontsize{6.5}{7.5}\selectfont]
  at (1.5,20.7) {CLS-based Exits};

\node[tb chip=tbCLS,minimum width=6.5mm,
  font=\sffamily\bfseries\fontsize{6.5}{7.5}\selectfont] (tbCLSOne) at (24.4,34.5) {CLS};
\node[tb chip=tbCLS,minimum width=6.5mm,
  font=\sffamily\bfseries\fontsize{6.5}{7.5}\selectfont] (tbCLSTwo) at (46.5,34.5) {CLS};
\node[tb head=tbExitHead] (tbHeadOne) at (24.4,42.0) {Head$_1$};
\node[tb head=tbExitHead] (tbHeadTwo) at (46.5,42.0) {Head$_2$};
\draw[tb flow] (tbBOne.south)--(tbCLSOne.north);
\draw[tb flow] (tbBTwo.south)--(tbCLSTwo.north);
\draw[tb flow] (tbCLSOne.south)--(tbHeadOne.north);
\draw[tb flow] (tbCLSTwo.south)--(tbHeadTwo.north);

\node[tb task label] at (9.8,30.0) {Early\\classification};
\draw[tb task arrow]
  ([xshift=-4mm]tbBOne.south)--([xshift=-4mm]tbHeadOne.north);

\node[tb chip=tbCLS,minimum width=6.5mm,
  font=\sffamily\bfseries\fontsize{6.5}{7.5}\selectfont] (tbFinalCLS) at (63.8,34.5) {CLS};
\node[tb head=tbFinal,minimum width=11.3mm,
  font=\sffamily\bfseries\fontsize{6}{7}\selectfont]
  (tbFinalHead) at (63.8,42.0) {Final Head};
\draw[tb flow] (tbBL.south)--(tbFinalCLS.north);
\draw[tb flow] (tbFinalCLS.south)--(tbFinalHead.north);


\node[tb task label] at (35.45,12.8) {Feature refinement};

\draw[tb task arrow,rounded corners=.6mm]
  (20.4,15.0) -| ([xshift=-2mm,yshift=2mm]tbFinalCLS.north);
\end{tikzpicture}%
}%
\endgroup
}

\newcommand{\DrawTeaserOurs}{%
\begingroup
\def\tbQueryLabel##1##2{%
  \scalebox{.6}{\sffamily\bfseries\boldmath\fontsize{10}{12}\selectfont
    $q_{##1}^{(##2)}$}%
}
\definecolor{tbBack}{HTML}{79A9FF}
\definecolor{tbBackFill}{HTML}{F3F7FF}
\definecolor{tbInk}{HTML}{000000}
\definecolor{tbMuted}{HTML}{000000}
\definecolor{tbRule}{HTML}{C9C9C9}
\definecolor{tbQueryOne}{HTML}{C44E78}
\definecolor{tbQueryTwo}{HTML}{168C8C}
\definecolor{tbCLS}{HTML}{B791D1}
\definecolor{tbFinal}{HTML}{B994D0}
\definecolor{tbExitHead}{HTML}{E2A469}
\definecolor{tbTokenBlue}{HTML}{6FA8FF}
\definecolor{tbReader}{HTML}{66B56F}
\definecolor{tbQueryFill}{HTML}{EFF7F0}
\definecolor{tbTaskRed}{HTML}{D9363E}
\resizebox{\linewidth}{!}{%
\begin{tikzpicture}[x=1mm,y=-1mm,
  every node/.style={font=\sffamily\bfseries\boldmath\fontsize{6.5}{7.5}\selectfont,
    text=tbInk,inner sep=0pt,outer sep=0pt,align=center},
  tb flow/.style={-{Stealth[length=1.2mm,width=.85mm]},
    draw=black,line width=.55pt},
  tb memory/.style={-{Stealth[length=1.2mm,width=.85mm]},
    draw=black,line width=.55pt},
  tb task arrow/.style={-{Stealth[length=1.6mm,width=1.1mm]},
    draw=tbTaskRed,line width=.85pt,line cap=round,line join=round},
  tb task label/.style={font=\sffamily\bfseries\fontsize{6}{7}\selectfont,
    text=tbTaskRed,align=center,inner sep=0pt},
  tb block/.style={draw=tbBack,fill=tbBackFill,line width=.55pt,
    rounded corners=.3mm,minimum width=10.8mm,minimum height=4.8mm},
  tb reader/.style={draw=tbReader,fill=tbQueryFill!48,line width=.55pt,
    rounded corners=.5mm,minimum width=15mm,minimum height=9.6mm},
  tb chip/.style={draw=##1,fill=##1!15,line width=.5pt,
    rounded corners=.4mm,minimum width=4.4mm,minimum height=4mm,
    font=\sffamily\bfseries\boldmath\fontsize{6}{7}\selectfont,text=black},
  tb head/.style={draw=##1,fill=##1!13,line width=.55pt,
    rounded corners=.4mm,minimum width=12.5mm,minimum height=5mm,
    inner xsep=.7pt,font=\sffamily\bfseries\boldmath\fontsize{6.5}{7.5}\selectfont},
  tb title/.style={font=\sffamily\bfseries\fontsize{6.8}{7.8}\selectfont,
    anchor=west},
  tb note/.style={font=\sffamily\bfseries\fontsize{6}{7}\selectfont,
    text=tbMuted},
  tb kv/.style={font=\sffamily\bfseries\boldmath\fontsize{6}{7}\selectfont,
    text=tbTokenBlue!85!black,fill=white,inner sep=.5pt}
]
\path[use as bounding box] (0,0) rectangle (70.8,46.0);

\filldraw[fill=tbBackFill!45,draw=tbBack,line width=.5pt,
  rounded corners=1mm]
  (.25,.25)--(70.55,.25)--(70.55,45.7)--(56.1,45.7)
  --(56.1,16.8)--(.25,16.8)--cycle;
\node[tb title,anchor=center] at (35.4,2.8) {ViT Backbone};

\node[tb note] at (7.5,1.8) {Image};
\begin{scope}
  \clip[rounded corners=.3mm] (3,3.8) rectangle (12,12.8);
  \node[inner sep=0pt] (tbPatchedImage) at (7.5,8.3)
    {\includegraphics[width=9mm,height=9mm]{\teaserAssetDir/input-square.png}};
  \foreach \xx in {6,9}{%
    \draw[white,line width=.55pt] (\xx,3.8)--(\xx,12.8);
  }
  \foreach \yy in {6.8,9.8}{%
    \draw[white,line width=.55pt] (3,\yy)--(12,\yy);
  }
\end{scope}
\draw[black!28,line width=.35pt,rounded corners=.3mm]
  (3,3.8) rectangle (12,12.8);
\node[tb block] (tbBOne) at (24.4,8.3) {Block 1};
\node[tb block] (tbBTwo) at (46.5,8.3) {Block 2};
\node[tb block] (tbBL) at (63.8,8.3) {Block $L$};
\draw[tb flow] (12.7,8.3)--(tbBOne.west);
\draw[tb flow] (tbBOne.east)--(tbBTwo.west);
\draw[tb flow] (tbBTwo.east)--(tbBL.west);
\node[font=\fontsize{6}{6}\selectfont,fill=tbBackFill!45,inner xsep=.5pt]
  at (54.6,8.3) {$\cdots$};

\filldraw[fill=white,draw=tbReader,line width=.5pt,
  rounded corners=1mm] (.25,18.4) rectangle (54.9,45.7);
\node[tb title,font=\sffamily\bfseries\fontsize{6.5}{7.5}\selectfont,text=tbReader!45!black]
  at (2,20.7) {Exit-Query Branch};
\node[tb reader] (tbReadOne) at (24.4,32.6) {};
\node[tb reader] (tbReadTwo) at (46.5,32.6) {};
\node[font=\sffamily\bfseries\fontsize{6}{7}\selectfont]
  at (24.4,29.7) {Cross-Attn};
\node[font=\sffamily\bfseries\fontsize{6}{7}\selectfont]
  at (46.5,29.7) {Cross-Attn};

\node[draw=tbReader,fill=tbQueryFill!62,rounded corners=.5mm,line width=.55pt,
  minimum width=12.3mm,minimum height=5.8mm] (tbQueryBank) at (7.5,34.5) {};
\node[tb chip=tbQueryOne] at (4.7,34.5) {\tbQueryLabel{1}{0}};
\node[tb chip=tbQueryTwo] at (10.3,34.5) {\tbQueryLabel{2}{0}};
\node[tb chip=tbQueryOne] (tbQOneState) at (21.6,34.5) {\tbQueryLabel{1}{1}};
\node[tb chip=tbQueryTwo] (tbQTwoState) at (27.2,34.5) {\tbQueryLabel{2}{1}};
\node[tb chip=tbQueryTwo] (tbQTwoFinal) at (46.5,34.5) {\tbQueryLabel{2}{2}};
\draw[tb flow] (tbQueryBank.east)--(tbReadOne.west |- tbQueryBank.east);
\draw[tb flow]
  (tbQTwoState.east)--(tbReadTwo.west |- tbQTwoState.east);

\draw[tb memory] (tbBOne.south)--
  node[pos=.68,right=.65mm,tb kv] {$K,V$} (tbReadOne.north);
\draw[tb memory] (tbBTwo.south)--
  node[pos=.68,right=.65mm,tb kv] {$K,V$} (tbReadTwo.north);

\node[tb head=tbExitHead] (tbHeadOne) at (21.6,42.0) {Head$_1$};
\node[tb head=tbExitHead] (tbHeadTwo) at (46.5,42.0) {Head$_2$};
\draw[tb flow] (tbQOneState.south)--(tbHeadOne.north);
\draw[tb flow] (tbQTwoFinal.south)--(tbHeadTwo.north);

\node[tb task label,anchor=west] at (2.0,24.1) {Early classification};
\draw[tb task arrow,rounded corners=.6mm]
  (2.0,25.9)--(18.0,25.9)--(18.0,39.5);

\node[tb chip=tbCLS,minimum width=6.5mm,
  font=\sffamily\bfseries\fontsize{6.5}{7.5}\selectfont] (tbFinalCLS) at (63.8,34.5) {CLS};
\node[tb head=tbFinal,minimum width=11.3mm,
  font=\sffamily\bfseries\fontsize{6}{7}\selectfont]
  (tbFinalHead) at (63.8,42.0) {Final Head};
\draw[tb flow] (tbBL.south)--(tbFinalCLS.north);
\draw[tb flow] (tbFinalCLS.south)--(tbFinalHead.north);

\node[tb task label] at (35.45,12.8) {Feature refinement};
\draw[tb task arrow,rounded corners=.6mm]
  (20.4,15.0) -| ([xshift=-2mm,yshift=2mm]tbFinalCLS.north);
\end{tikzpicture}%
}%
\endgroup
}

\newcommand{\DrawTeaserPair}{%
  \begin{subfigure}[t]{.4985\linewidth}
    \centering
    \DrawTeaserBaseline
    \caption{Conventional [CLS]-based early exits.}
  \end{subfigure}\hfill%
  \begin{subfigure}[t]{.4985\linewidth}
    \centering
    \DrawTeaserOurs
    \caption{DeclareViT with read-only exit queries.}
  \end{subfigure}%
}

%% file: sections/2_related_work.tex
\section{Related Work}
\label{related_work}

\paragraph{Exit-Head Design and Information Reuse.}
\citet{Bakhtiarnia:2021:MVT} investigate several ViT early-exit architectures, including MLP heads and convolutional, Transformer, MLP-Mixer, and ResMLP branches. \citet{Bakhtiarnia:2022:SLV} attach single-layer Transformer branches to CNN backbones to improve early-exit accuracy with less computational overhead.
LGViT~\citep{Xu:2023:LDE} adopts heterogeneous heads, using local-perception heads at shallow exits and global-aggregation heads at deeper exits.
For prediction reuse,  ZTW~\citep{Wolczyk:2021:ZTW} feeds previous-layer classifier logits into subsequent classifiers and combines their predictions through geometric ensembling.
BEEM~\citep{Bajpai:2025:BBP} accumulates confidence scores from consecutive exits that predict the same label and terminates inference when the accumulated score reaches a predefined threshold. Our DeclareViT instead reads backbone features into dedicated exit queries, accumulating feature-level information across depth rather than reusing earlier predictions.
\paragraph{Training and Optimization of Multi-Exit Models.}
Knowledge distillation transfers knowledge from deeper to shallower exits.
BYOT~\citep{Zhang:2019:BYO} trains shallow classifiers
to match the deepest classifier's predictions and features.
MEViT~\citep{Shen:2023:MEV} trains the final and earliest exits together with a sampled subset of intermediate exits, which learn from the final prediction through self-distillation.
EnViT~\citep{Dong:2026:EET} uses exit-aware structured dropout to
create student subnetworks and distills knowledge from the full model into them, aiming to reduce interference between exits.
Sample weighting instead addresses the mismatch between training and dynamic routing. L2W-DEN~\citep{Han:2022:LTW} uses meta-learning to predict
sample-specific loss weights for each exit, emphasizing easy samples
at shallow exits and harder samples at deeper exits. 
COSEE~\citep{He:2025:CCS} simulates early exit
under multiple thresholds during training and assigns larger loss
weights to classifiers near each sample's simulated exit.
{\revcolor LGViT~\citep{Xu:2023:LDE} trains in two stages: the backbone is first trained by alternating between the final-classifier loss and a joint multi-exit loss~\citep{Xin:2021:BEE}, and the exit heads are then trained by self-distillation with the backbone and final classifier frozen.
DeclareViT shares this frozen second stage, but trains its backbone with the full-depth objective alone.
The backbone thus never receives an exit loss, and exits can be attached to an existing checkpoint without changing its full-depth prediction.}
\paragraph{Decoupling Early Classification from Feature Refinement.}
Dynamic Perceiver~\citep{Han:2023:DPF} maintains a feature-extraction branch and a latent classification branch.
Bidirectional cross-attention allows the classification branch to
read visual features and send information back to the feature branch. 
Dyn-Adapter~\citep{Zhang:2024:DTD} freezes pretrained backbone weights while training adaptation modules and early heads. DFS~\citep{Gong:2024:DFS} partitions features at each exit depth so that the local exit loss updates the exit-specific portion, while deeper objectives update the shared portion.
MET~\citep{Liu:2026:MTT} partially decouples different exits by inserting exit-specific adapters into the shared backbone, which transform intermediate features into separate [\textsc{cls}] representations for individual exits.
DMPO~\citep{Luo:2025:DMO} decouples the role coupling through bypass modules: the original intermediate features are preserved for propagation to deeper stages, while bypass-transformed features are separately used for early classification.
However, prior early-exit decoupling methods still let early-exit learning influence the feature-refinement path through feature feedback or trainable adaptation modules. In contrast, 
DeclareViT uses read-only exit queries and two-stage training to keep early classification separate 
from backbone feature refinement in both forward and backward passes.
{\revcolor
\paragraph{Query-Based Transfer Learning.}
VQT~\citep{Tu:2023:VQT} and SynQT~\citep{Zhang:2024:PEM} read intermediate features of a frozen ViT through learned queries without modifying it, for parameter- and memory-efficient transfer learning.
Both read through the frozen attention projections of the backbone and feed features from all blocks to a single classifier, so the full backbone always runs.
DeclareViT instead uses read-only queries as an exit interface: each exit predicts from the blocks executed so far, with its own query state and learned exit-side projections.}

%% file: sections/3_Preliminary.tex
\section{[\textsc{cls}] Role Coupling  Analysis}
\label{sec:preliminaries}

\subsection{Preliminaries}

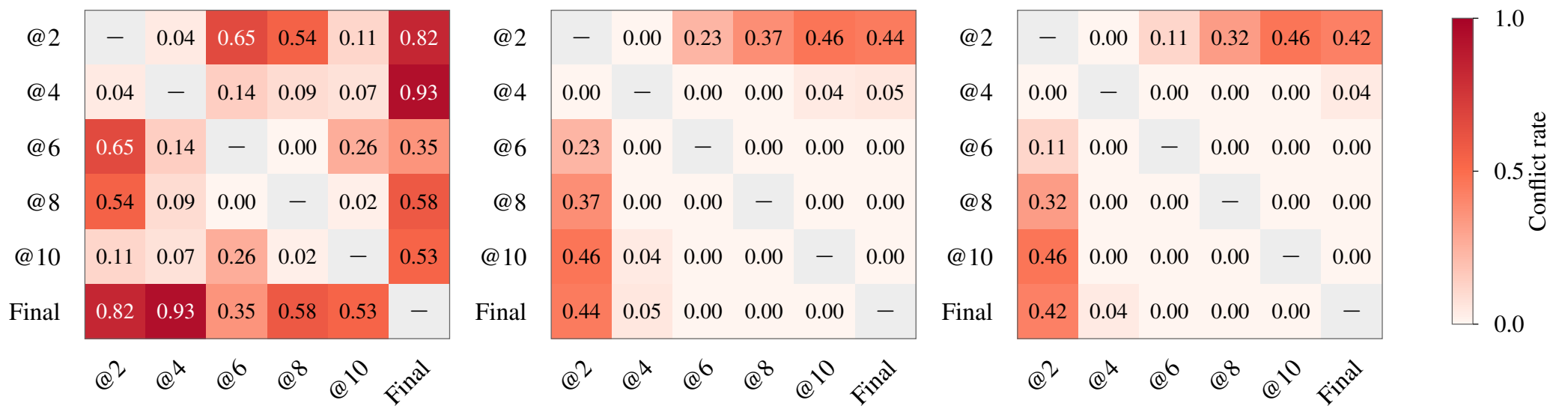
\begin{figure}[t]
\centering
\resizebox{\linewidth}{!}{%
    \input{artwork/gradient_conflict/sdn_pairwise_conflict}%
}
\caption{Pairwise gradient conflict rates on backbone block~2 of our
[\textsc{cls}]-based SDN adaptation at training epochs 1, 10,
and 100, using ViT-B/16
on ImageNet-1K. @2--@10 denote exit depths and Final denotes the full-depth objective. Early and deeper prediction objectives can favor opposing updates to the shared backbone even at a
late training checkpoint.}
\label{fig:sdn_backbone_conflict}
\end{figure}

\paragraph{Vision Transformer.}
Given an input image $\mathbf{I}\in\mathbb{R}^{H\times W\times 3}$,
a ViT~\citep{Dosovitskiy:2021:AII} projects $N$ image patches into
$D$-dimensional embeddings $\mathbf{P}^{(0)}\in\mathbb{R}^{D\times N}$
and prepends a learnable [\textsc{cls}] token
$\mathbf{x}_{\mathrm{cls}}^{(0)}\in\mathbb{R}^{D\times 1}$,
forming the input sequence
$\mathbf{X}^{(0)}=[\mathbf{x}_{\mathrm{cls}}^{(0)},\mathbf{P}^{(0)}]
\in\mathbb{R}^{D\times(N+1)}$.
An $L$-block backbone produces the final logits as
\begin{equation}
\mathbf{X}^{(\ell)}
=
\mathcal{F}_{\ell}\!\left(\mathbf{X}^{(\ell-1)}\right),
\quad \ell=1,\ldots,L,
\qquad
\mathbf{z}_{\mathrm{f}}
=
\mathbf{W}_{\mathrm{f}}
\operatorname{LN}_{\mathrm{f}}
\!\left(\mathbf{x}_{\mathrm{cls}}^{(L)}\right)
+
\mathbf{b}_{\mathrm{f}}
\in\mathbb{R}^{C},
\label{eq:vit_forward}
\end{equation}
where $\mathcal{F}_{\ell}$ denotes Transformer block $\ell$,
$\operatorname{LN}$ denotes LayerNorm, $\mathrm{f}$ identifies the final classifier, and C denotes the number of classes.
\paragraph{[\textsc{cls}]-based early exits.}
Consider $E$ early exits indexed by $e=1,\ldots,E$.
Exit $e$ is attached after backbone block $d_e$, where
$1\leq d_1<\cdots<d_E<L$.
Thus, $e$ identifies an exit, whereas $d_e$ specifies its backbone
depth; for example, five exits at blocks $2,4,6,8,10$ correspond to
$(d_1,\ldots,d_5)=(2,4,6,8,10)$.
As shown in Figure~\ref{fig:teaser_baseline}, each exit classifies
the intermediate [\textsc{cls}] token:
\begin{equation}
\mathbf h_e
=
\mathbf x_{\mathrm{cls}}^{(d_e)},
\qquad
\mathbf z_e
=
\mathcal H_e(\mathbf h_e),
\qquad e=1,\ldots,E,
\label{eq:conventional_exit}
\end{equation}
where $\mathcal{H}_e$ is a nonlinear exit head.
Joint training minimizes
\begin{equation}
\mathcal{L}_{\mathrm{joint}}
=
\lambda_{\mathrm{f}}\mathcal{L}_{\mathrm{f}}
+
\sum_{e=1}^{E}\lambda_e\mathcal{L}_e,
\label{eq:joint_early_exit_loss}
\end{equation}
where $\mathcal{L}_{\mathrm{f}}$ and $\mathcal{L}_e$ are
the full-depth and exit-specific training objectives,
with weights $\lambda_{\mathrm{f}}$ and $\lambda_e$.

\subsection{[\textsc{cls}] Role Coupling}
\label{sec:role_coupling}

\paragraph{Coupled roles and optimization.}
The same representation $\mathbf{h}_e$ is used for early classification
at exit $e$ and is subsequently propagated through the remaining
backbone blocks for further refinement.
We refer to this shared use as [\textsc{cls}] role coupling.
During joint training, prediction objectives at different depths
can impose competing demands on the shared backbone parameters.

To characterize this optimization coupling, let
$\boldsymbol{\theta}_{\ell}$ denote the parameters of backbone
block $\ell$.
For early exit $e$, we consider gradients on blocks
$1\leq\ell\leq d_e$; the final objective involves all blocks
$1\leq\ell\leq L$.
We use the individual objective gradients:
\begin{equation}
\mathbf g_{e,\ell}
=
\nabla_{\boldsymbol{\theta}_{\ell}}\mathcal L_e,
\qquad
\mathbf g_{\mathrm f,\ell}
=
\nabla_{\boldsymbol{\theta}_{\ell}}\mathcal L_{\mathrm f}.
\label{eq:exit_gradients}
\end{equation}
For distinct objectives that backpropagate through the same block,
we compute the cosine similarity:
\begin{equation}
c_{ij}^{(\ell)}
=
\frac{
\left\langle \mathbf g_{i,\ell},\mathbf g_{j,\ell}\right\rangle
}{
\left\|\mathbf g_{i,\ell}\right\|_2
\left\|\mathbf g_{j,\ell}\right\|_2+\epsilon
},
\qquad
i,j\in\{1,\ldots,E,\mathrm f\},\quad i\neq j.
\label{eq:pairwise_gradient_cosine}
\end{equation}
Here, $\epsilon=10^{-12}$.
A negative cosine indicates opposing objective-gradient
directions~\citep{Yu:2020:GSM}.
The \emph{conflict rate} is the fraction of valid sampled
training minibatches with negative cosine.

\paragraph{Gradient-conflict analysis.}
For analysis, we adapt SDN~\citep{Kaya:2019:SDN} to ViTs on ImageNet-1K by attaching
a LayerNorm followed by a linear classifier to the intermediate
[\textsc{cls}] token at blocks $\{2,4,6,8,10\}$.
The backbone, early-exit heads, and final classifier are jointly
trained using hard-label cross-entropy losses. Figure~\ref{fig:sdn_backbone_conflict} reports pairwise gradient
conflict rates on backbone block~2, through which all five
early-exit losses and the final loss backpropagate.
The @2--Final and @4--Final conflict rates are $0.82$ and $0.93$
at epoch~1.
Although many objective pairs become aligned later, @2--Final
still has conflict rates of $0.44$ and $0.42$ at epochs~10 and~100.
Thus, early and deeper prediction objectives can favor opposing
updates to the shared backbone even at a late training checkpoint.
These observations reveal persistent optimization competition in the jointly trained [\textsc{cls}]-based model. {\revcolor They are a diagnostic of the coupled design rather than a measure of its accuracy cost, and} they motivate an early-classification pathway that does not update the backbone.
Additional analyses are provided
in Appendix~\ref{app:gradient_conflict}.

%% file: artwork/gradient_conflict/sdn_pairwise_conflict.tex
\begingroup
\definecolor{GCRateLow}{RGB}{255,245,240}
\definecolor{GCRateMid}{RGB}{251,106,74}
\definecolor{GCRateHigh}{RGB}{165,0,38}
\begin{tikzpicture}[x=1cm,y=1cm]
\path[use as bounding box] (-0.08,-0.78) rectangle (14.05,3.35);
\fill[black!7] (0.66000,2.62200) rectangle ++(0.55667,0.50100);
\node[font=\fontsize{5.7}{6.7}\selectfont] at (0.93833,2.87250) {$-$};
\fill[GCRateMid!8.000!GCRateLow] (1.21667,2.62200) rectangle ++(0.55667,0.50100);
\node[font=\fontsize{5.7}{6.7}\selectfont] at (1.49500,2.87250) {0.04};
\fill[GCRateHigh!30.000!GCRateMid] (1.77333,2.62200) rectangle ++(0.55667,0.50100);
\node[font=\fontsize{5.7}{6.7}\selectfont] at (2.05167,2.87250) {\textcolor{white}{0.65}};
\fill[GCRateHigh!8.000!GCRateMid] (2.33000,2.62200) rectangle ++(0.55667,0.50100);
\node[font=\fontsize{5.7}{6.7}\selectfont] at (2.60833,2.87250) {0.54};
\fill[GCRateMid!22.000!GCRateLow] (2.88667,2.62200) rectangle ++(0.55667,0.50100);
\node[font=\fontsize{5.7}{6.7}\selectfont] at (3.16500,2.87250) {0.11};
\fill[GCRateHigh!64.000!GCRateMid] (3.44333,2.62200) rectangle ++(0.55667,0.50100);
\node[font=\fontsize{5.7}{6.7}\selectfont] at (3.72167,2.87250) {\textcolor{white}{0.82}};
\fill[GCRateMid!8.000!GCRateLow] (0.66000,2.12100) rectangle ++(0.55667,0.50100);
\node[font=\fontsize{5.7}{6.7}\selectfont] at (0.93833,2.37150) {0.04};
\fill[black!7] (1.21667,2.12100) rectangle ++(0.55667,0.50100);
\node[font=\fontsize{5.7}{6.7}\selectfont] at (1.49500,2.37150) {$-$};
\fill[GCRateMid!28.000!GCRateLow] (1.77333,2.12100) rectangle ++(0.55667,0.50100);
\node[font=\fontsize{5.7}{6.7}\selectfont] at (2.05167,2.37150) {0.14};
\fill[GCRateMid!18.000!GCRateLow] (2.33000,2.12100) rectangle ++(0.55667,0.50100);
\node[font=\fontsize{5.7}{6.7}\selectfont] at (2.60833,2.37150) {0.09};
\fill[GCRateMid!14.000!GCRateLow] (2.88667,2.12100) rectangle ++(0.55667,0.50100);
\node[font=\fontsize{5.7}{6.7}\selectfont] at (3.16500,2.37150) {0.07};
\fill[GCRateHigh!86.000!GCRateMid] (3.44333,2.12100) rectangle ++(0.55667,0.50100);
\node[font=\fontsize{5.7}{6.7}\selectfont] at (3.72167,2.37150) {\textcolor{white}{0.93}};
\fill[GCRateHigh!30.000!GCRateMid] (0.66000,1.62000) rectangle ++(0.55667,0.50100);
\node[font=\fontsize{5.7}{6.7}\selectfont] at (0.93833,1.87050) {\textcolor{white}{0.65}};
\fill[GCRateMid!28.000!GCRateLow] (1.21667,1.62000) rectangle ++(0.55667,0.50100);
\node[font=\fontsize{5.7}{6.7}\selectfont] at (1.49500,1.87050) {0.14};
\fill[black!7] (1.77333,1.62000) rectangle ++(0.55667,0.50100);
\node[font=\fontsize{5.7}{6.7}\selectfont] at (2.05167,1.87050) {$-$};
\fill[GCRateMid!0.000!GCRateLow] (2.33000,1.62000) rectangle ++(0.55667,0.50100);
\node[font=\fontsize{5.7}{6.7}\selectfont] at (2.60833,1.87050) {0.00};
\fill[GCRateMid!52.000!GCRateLow] (2.88667,1.62000) rectangle ++(0.55667,0.50100);
\node[font=\fontsize{5.7}{6.7}\selectfont] at (3.16500,1.87050) {0.26};
\fill[GCRateMid!70.000!GCRateLow] (3.44333,1.62000) rectangle ++(0.55667,0.50100);
\node[font=\fontsize{5.7}{6.7}\selectfont] at (3.72167,1.87050) {0.35};
\fill[GCRateHigh!8.000!GCRateMid] (0.66000,1.11900) rectangle ++(0.55667,0.50100);
\node[font=\fontsize{5.7}{6.7}\selectfont] at (0.93833,1.36950) {0.54};
\fill[GCRateMid!18.000!GCRateLow] (1.21667,1.11900) rectangle ++(0.55667,0.50100);
\node[font=\fontsize{5.7}{6.7}\selectfont] at (1.49500,1.36950) {0.09};
\fill[GCRateMid!0.000!GCRateLow] (1.77333,1.11900) rectangle ++(0.55667,0.50100);
\node[font=\fontsize{5.7}{6.7}\selectfont] at (2.05167,1.36950) {0.00};
\fill[black!7] (2.33000,1.11900) rectangle ++(0.55667,0.50100);
\node[font=\fontsize{5.7}{6.7}\selectfont] at (2.60833,1.36950) {$-$};
\fill[GCRateMid!4.000!GCRateLow] (2.88667,1.11900) rectangle ++(0.55667,0.50100);
\node[font=\fontsize{5.7}{6.7}\selectfont] at (3.16500,1.36950) {0.02};
\fill[GCRateHigh!16.000!GCRateMid] (3.44333,1.11900) rectangle ++(0.55667,0.50100);
\node[font=\fontsize{5.7}{6.7}\selectfont] at (3.72167,1.36950) {0.58};
\fill[GCRateMid!22.000!GCRateLow] (0.66000,0.61800) rectangle ++(0.55667,0.50100);
\node[font=\fontsize{5.7}{6.7}\selectfont] at (0.93833,0.86850) {0.11};
\fill[GCRateMid!14.000!GCRateLow] (1.21667,0.61800) rectangle ++(0.55667,0.50100);
\node[font=\fontsize{5.7}{6.7}\selectfont] at (1.49500,0.86850) {0.07};
\fill[GCRateMid!52.000!GCRateLow] (1.77333,0.61800) rectangle ++(0.55667,0.50100);
\node[font=\fontsize{5.7}{6.7}\selectfont] at (2.05167,0.86850) {0.26};
\fill[GCRateMid!4.000!GCRateLow] (2.33000,0.61800) rectangle ++(0.55667,0.50100);
\node[font=\fontsize{5.7}{6.7}\selectfont] at (2.60833,0.86850) {0.02};
\fill[black!7] (2.88667,0.61800) rectangle ++(0.55667,0.50100);
\node[font=\fontsize{5.7}{6.7}\selectfont] at (3.16500,0.86850) {$-$};
\fill[GCRateHigh!6.000!GCRateMid] (3.44333,0.61800) rectangle ++(0.55667,0.50100);
\node[font=\fontsize{5.7}{6.7}\selectfont] at (3.72167,0.86850) {0.53};
\fill[GCRateHigh!64.000!GCRateMid] (0.66000,0.11700) rectangle ++(0.55667,0.50100);
\node[font=\fontsize{5.7}{6.7}\selectfont] at (0.93833,0.36750) {\textcolor{white}{0.82}};
\fill[GCRateHigh!86.000!GCRateMid] (1.21667,0.11700) rectangle ++(0.55667,0.50100);
\node[font=\fontsize{5.7}{6.7}\selectfont] at (1.49500,0.36750) {\textcolor{white}{0.93}};
\fill[GCRateMid!70.000!GCRateLow] (1.77333,0.11700) rectangle ++(0.55667,0.50100);
\node[font=\fontsize{5.7}{6.7}\selectfont] at (2.05167,0.36750) {0.35};
\fill[GCRateHigh!16.000!GCRateMid] (2.33000,0.11700) rectangle ++(0.55667,0.50100);
\node[font=\fontsize{5.7}{6.7}\selectfont] at (2.60833,0.36750) {0.58};
\fill[GCRateHigh!6.000!GCRateMid] (2.88667,0.11700) rectangle ++(0.55667,0.50100);
\node[font=\fontsize{5.7}{6.7}\selectfont] at (3.16500,0.36750) {0.53};
\fill[black!7] (3.44333,0.11700) rectangle ++(0.55667,0.50100);
\node[font=\fontsize{5.7}{6.7}\selectfont] at (3.72167,0.36750) {$-$};
\draw[black!60,line width=.25pt] (0.66000,0.11700) rectangle ++(3.34,3.006);
\node[anchor=north east,font=\fontsize{6.5}{7.5}\selectfont,rotate=45] at (0.93833,.12) {@2};
\node[anchor=east,font=\fontsize{6.5}{7.5}\selectfont] at (0.56000,2.87250) {@2};
\node[anchor=north east,font=\fontsize{6.5}{7.5}\selectfont,rotate=45] at (1.49500,.12) {@4};
\node[anchor=east,font=\fontsize{6.5}{7.5}\selectfont] at (0.56000,2.37150) {@4};
\node[anchor=north east,font=\fontsize{6.5}{7.5}\selectfont,rotate=45] at (2.05167,.12) {@6};
\node[anchor=east,font=\fontsize{6.5}{7.5}\selectfont] at (0.56000,1.87050) {@6};
\node[anchor=north east,font=\fontsize{6.5}{7.5}\selectfont,rotate=45] at (2.60833,.12) {@8};
\node[anchor=east,font=\fontsize{6.5}{7.5}\selectfont] at (0.56000,1.36950) {@8};
\node[anchor=north east,font=\fontsize{6.5}{7.5}\selectfont,rotate=45] at (3.16500,.12) {@10};
\node[anchor=east,font=\fontsize{6.5}{7.5}\selectfont] at (0.56000,0.86850) {@10};
\node[anchor=north east,font=\fontsize{6.5}{7.5}\selectfont,rotate=45] at (3.72167,.12) {Final};
\node[anchor=east,font=\fontsize{6.5}{7.5}\selectfont] at (0.56000,0.36750) {Final};
\node[font=\small] at (2.33000,-0.7) {(a) Epoch 1};
\fill[black!7] (4.93000,2.62200) rectangle ++(0.55667,0.50100);
\node[font=\fontsize{5.7}{6.7}\selectfont] at (5.20833,2.87250) {$-$};
\fill[GCRateMid!0.000!GCRateLow] (5.48667,2.62200) rectangle ++(0.55667,0.50100);
\node[font=\fontsize{5.7}{6.7}\selectfont] at (5.76500,2.87250) {0.00};
\fill[GCRateMid!46.000!GCRateLow] (6.04333,2.62200) rectangle ++(0.55667,0.50100);
\node[font=\fontsize{5.7}{6.7}\selectfont] at (6.32167,2.87250) {0.23};
\fill[GCRateMid!74.000!GCRateLow] (6.60000,2.62200) rectangle ++(0.55667,0.50100);
\node[font=\fontsize{5.7}{6.7}\selectfont] at (6.87833,2.87250) {0.37};
\fill[GCRateMid!92.000!GCRateLow] (7.15667,2.62200) rectangle ++(0.55667,0.50100);
\node[font=\fontsize{5.7}{6.7}\selectfont] at (7.43500,2.87250) {0.46};
\fill[GCRateMid!88.000!GCRateLow] (7.71333,2.62200) rectangle ++(0.55667,0.50100);
\node[font=\fontsize{5.7}{6.7}\selectfont] at (7.99167,2.87250) {0.44};
\fill[GCRateMid!0.000!GCRateLow] (4.93000,2.12100) rectangle ++(0.55667,0.50100);
\node[font=\fontsize{5.7}{6.7}\selectfont] at (5.20833,2.37150) {0.00};
\fill[black!7] (5.48667,2.12100) rectangle ++(0.55667,0.50100);
\node[font=\fontsize{5.7}{6.7}\selectfont] at (5.76500,2.37150) {$-$};
\fill[GCRateMid!0.000!GCRateLow] (6.04333,2.12100) rectangle ++(0.55667,0.50100);
\node[font=\fontsize{5.7}{6.7}\selectfont] at (6.32167,2.37150) {0.00};
\fill[GCRateMid!0.000!GCRateLow] (6.60000,2.12100) rectangle ++(0.55667,0.50100);
\node[font=\fontsize{5.7}{6.7}\selectfont] at (6.87833,2.37150) {0.00};
\fill[GCRateMid!8.000!GCRateLow] (7.15667,2.12100) rectangle ++(0.55667,0.50100);
\node[font=\fontsize{5.7}{6.7}\selectfont] at (7.43500,2.37150) {0.04};
\fill[GCRateMid!10.000!GCRateLow] (7.71333,2.12100) rectangle ++(0.55667,0.50100);
\node[font=\fontsize{5.7}{6.7}\selectfont] at (7.99167,2.37150) {0.05};
\fill[GCRateMid!46.000!GCRateLow] (4.93000,1.62000) rectangle ++(0.55667,0.50100);
\node[font=\fontsize{5.7}{6.7}\selectfont] at (5.20833,1.87050) {0.23};
\fill[GCRateMid!0.000!GCRateLow] (5.48667,1.62000) rectangle ++(0.55667,0.50100);
\node[font=\fontsize{5.7}{6.7}\selectfont] at (5.76500,1.87050) {0.00};
\fill[black!7] (6.04333,1.62000) rectangle ++(0.55667,0.50100);
\node[font=\fontsize{5.7}{6.7}\selectfont] at (6.32167,1.87050) {$-$};
\fill[GCRateMid!0.000!GCRateLow] (6.60000,1.62000) rectangle ++(0.55667,0.50100);
\node[font=\fontsize{5.7}{6.7}\selectfont] at (6.87833,1.87050) {0.00};
\fill[GCRateMid!0.000!GCRateLow] (7.15667,1.62000) rectangle ++(0.55667,0.50100);
\node[font=\fontsize{5.7}{6.7}\selectfont] at (7.43500,1.87050) {0.00};
\fill[GCRateMid!0.000!GCRateLow] (7.71333,1.62000) rectangle ++(0.55667,0.50100);
\node[font=\fontsize{5.7}{6.7}\selectfont] at (7.99167,1.87050) {0.00};
\fill[GCRateMid!74.000!GCRateLow] (4.93000,1.11900) rectangle ++(0.55667,0.50100);
\node[font=\fontsize{5.7}{6.7}\selectfont] at (5.20833,1.36950) {0.37};
\fill[GCRateMid!0.000!GCRateLow] (5.48667,1.11900) rectangle ++(0.55667,0.50100);
\node[font=\fontsize{5.7}{6.7}\selectfont] at (5.76500,1.36950) {0.00};
\fill[GCRateMid!0.000!GCRateLow] (6.04333,1.11900) rectangle ++(0.55667,0.50100);
\node[font=\fontsize{5.7}{6.7}\selectfont] at (6.32167,1.36950) {0.00};
\fill[black!7] (6.60000,1.11900) rectangle ++(0.55667,0.50100);
\node[font=\fontsize{5.7}{6.7}\selectfont] at (6.87833,1.36950) {$-$};
\fill[GCRateMid!0.000!GCRateLow] (7.15667,1.11900) rectangle ++(0.55667,0.50100);
\node[font=\fontsize{5.7}{6.7}\selectfont] at (7.43500,1.36950) {0.00};
\fill[GCRateMid!0.000!GCRateLow] (7.71333,1.11900) rectangle ++(0.55667,0.50100);
\node[font=\fontsize{5.7}{6.7}\selectfont] at (7.99167,1.36950) {0.00};
\fill[GCRateMid!92.000!GCRateLow] (4.93000,0.61800) rectangle ++(0.55667,0.50100);
\node[font=\fontsize{5.7}{6.7}\selectfont] at (5.20833,0.86850) {0.46};
\fill[GCRateMid!8.000!GCRateLow] (5.48667,0.61800) rectangle ++(0.55667,0.50100);
\node[font=\fontsize{5.7}{6.7}\selectfont] at (5.76500,0.86850) {0.04};
\fill[GCRateMid!0.000!GCRateLow] (6.04333,0.61800) rectangle ++(0.55667,0.50100);
\node[font=\fontsize{5.7}{6.7}\selectfont] at (6.32167,0.86850) {0.00};
\fill[GCRateMid!0.000!GCRateLow] (6.60000,0.61800) rectangle ++(0.55667,0.50100);
\node[font=\fontsize{5.7}{6.7}\selectfont] at (6.87833,0.86850) {0.00};
\fill[black!7] (7.15667,0.61800) rectangle ++(0.55667,0.50100);
\node[font=\fontsize{5.7}{6.7}\selectfont] at (7.43500,0.86850) {$-$};
\fill[GCRateMid!0.000!GCRateLow] (7.71333,0.61800) rectangle ++(0.55667,0.50100);
\node[font=\fontsize{5.7}{6.7}\selectfont] at (7.99167,0.86850) {0.00};
\fill[GCRateMid!88.000!GCRateLow] (4.93000,0.11700) rectangle ++(0.55667,0.50100);
\node[font=\fontsize{5.7}{6.7}\selectfont] at (5.20833,0.36750) {0.44};
\fill[GCRateMid!10.000!GCRateLow] (5.48667,0.11700) rectangle ++(0.55667,0.50100);
\node[font=\fontsize{5.7}{6.7}\selectfont] at (5.76500,0.36750) {0.05};
\fill[GCRateMid!0.000!GCRateLow] (6.04333,0.11700) rectangle ++(0.55667,0.50100);
\node[font=\fontsize{5.7}{6.7}\selectfont] at (6.32167,0.36750) {0.00};
\fill[GCRateMid!0.000!GCRateLow] (6.60000,0.11700) rectangle ++(0.55667,0.50100);
\node[font=\fontsize{5.7}{6.7}\selectfont] at (6.87833,0.36750) {0.00};
\fill[GCRateMid!0.000!GCRateLow] (7.15667,0.11700) rectangle ++(0.55667,0.50100);
\node[font=\fontsize{5.7}{6.7}\selectfont] at (7.43500,0.36750) {0.00};
\fill[black!7] (7.71333,0.11700) rectangle ++(0.55667,0.50100);
\node[font=\fontsize{5.7}{6.7}\selectfont] at (7.99167,0.36750) {$-$};
\draw[black!60,line width=.25pt] (4.93000,0.11700) rectangle ++(3.34,3.006);
\node[anchor=north east,font=\fontsize{6.5}{7.5}\selectfont,rotate=45] at (5.20833,.12) {@2};
\node[anchor=east,font=\fontsize{6.5}{7.5}\selectfont] at (4.83000,2.87250) {@2};
\node[anchor=north east,font=\fontsize{6.5}{7.5}\selectfont,rotate=45] at (5.76500,.12) {@4};
\node[anchor=east,font=\fontsize{6.5}{7.5}\selectfont] at (4.83000,2.37150) {@4};
\node[anchor=north east,font=\fontsize{6.5}{7.5}\selectfont,rotate=45] at (6.32167,.12) {@6};
\node[anchor=east,font=\fontsize{6.5}{7.5}\selectfont] at (4.83000,1.87050) {@6};
\node[anchor=north east,font=\fontsize{6.5}{7.5}\selectfont,rotate=45] at (6.87833,.12) {@8};
\node[anchor=east,font=\fontsize{6.5}{7.5}\selectfont] at (4.83000,1.36950) {@8};
\node[anchor=north east,font=\fontsize{6.5}{7.5}\selectfont,rotate=45] at (7.43500,.12) {@10};
\node[anchor=east,font=\fontsize{6.5}{7.5}\selectfont] at (4.83000,0.86850) {@10};
\node[anchor=north east,font=\fontsize{6.5}{7.5}\selectfont,rotate=45] at (7.99167,.12) {Final};
\node[anchor=east,font=\fontsize{6.5}{7.5}\selectfont] at (4.83000,0.36750) {Final};
\node[font=\small] at (6.60000,-0.7) {(b) Epoch 10};
\fill[black!7] (9.20000,2.62200) rectangle ++(0.55667,0.50100);
\node[font=\fontsize{5.7}{6.7}\selectfont] at (9.47833,2.87250) {$-$};
\fill[GCRateMid!0.000!GCRateLow] (9.75667,2.62200) rectangle ++(0.55667,0.50100);
\node[font=\fontsize{5.7}{6.7}\selectfont] at (10.03500,2.87250) {0.00};
\fill[GCRateMid!22.000!GCRateLow] (10.31333,2.62200) rectangle ++(0.55667,0.50100);
\node[font=\fontsize{5.7}{6.7}\selectfont] at (10.59167,2.87250) {0.11};
\fill[GCRateMid!64.000!GCRateLow] (10.87000,2.62200) rectangle ++(0.55667,0.50100);
\node[font=\fontsize{5.7}{6.7}\selectfont] at (11.14833,2.87250) {0.32};
\fill[GCRateMid!92.000!GCRateLow] (11.42667,2.62200) rectangle ++(0.55667,0.50100);
\node[font=\fontsize{5.7}{6.7}\selectfont] at (11.70500,2.87250) {0.46};
\fill[GCRateMid!84.000!GCRateLow] (11.98333,2.62200) rectangle ++(0.55667,0.50100);
\node[font=\fontsize{5.7}{6.7}\selectfont] at (12.26167,2.87250) {0.42};
\fill[GCRateMid!0.000!GCRateLow] (9.20000,2.12100) rectangle ++(0.55667,0.50100);
\node[font=\fontsize{5.7}{6.7}\selectfont] at (9.47833,2.37150) {0.00};
\fill[black!7] (9.75667,2.12100) rectangle ++(0.55667,0.50100);
\node[font=\fontsize{5.7}{6.7}\selectfont] at (10.03500,2.37150) {$-$};
\fill[GCRateMid!0.000!GCRateLow] (10.31333,2.12100) rectangle ++(0.55667,0.50100);
\node[font=\fontsize{5.7}{6.7}\selectfont] at (10.59167,2.37150) {0.00};
\fill[GCRateMid!0.000!GCRateLow] (10.87000,2.12100) rectangle ++(0.55667,0.50100);
\node[font=\fontsize{5.7}{6.7}\selectfont] at (11.14833,2.37150) {0.00};
\fill[GCRateMid!0.000!GCRateLow] (11.42667,2.12100) rectangle ++(0.55667,0.50100);
\node[font=\fontsize{5.7}{6.7}\selectfont] at (11.70500,2.37150) {0.00};
\fill[GCRateMid!8.000!GCRateLow] (11.98333,2.12100) rectangle ++(0.55667,0.50100);
\node[font=\fontsize{5.7}{6.7}\selectfont] at (12.26167,2.37150) {0.04};
\fill[GCRateMid!22.000!GCRateLow] (9.20000,1.62000) rectangle ++(0.55667,0.50100);
\node[font=\fontsize{5.7}{6.7}\selectfont] at (9.47833,1.87050) {0.11};
\fill[GCRateMid!0.000!GCRateLow] (9.75667,1.62000) rectangle ++(0.55667,0.50100);
\node[font=\fontsize{5.7}{6.7}\selectfont] at (10.03500,1.87050) {0.00};
\fill[black!7] (10.31333,1.62000) rectangle ++(0.55667,0.50100);
\node[font=\fontsize{5.7}{6.7}\selectfont] at (10.59167,1.87050) {$-$};
\fill[GCRateMid!0.000!GCRateLow] (10.87000,1.62000) rectangle ++(0.55667,0.50100);
\node[font=\fontsize{5.7}{6.7}\selectfont] at (11.14833,1.87050) {0.00};
\fill[GCRateMid!0.000!GCRateLow] (11.42667,1.62000) rectangle ++(0.55667,0.50100);
\node[font=\fontsize{5.7}{6.7}\selectfont] at (11.70500,1.87050) {0.00};
\fill[GCRateMid!0.000!GCRateLow] (11.98333,1.62000) rectangle ++(0.55667,0.50100);
\node[font=\fontsize{5.7}{6.7}\selectfont] at (12.26167,1.87050) {0.00};
\fill[GCRateMid!64.000!GCRateLow] (9.20000,1.11900) rectangle ++(0.55667,0.50100);
\node[font=\fontsize{5.7}{6.7}\selectfont] at (9.47833,1.36950) {0.32};
\fill[GCRateMid!0.000!GCRateLow] (9.75667,1.11900) rectangle ++(0.55667,0.50100);
\node[font=\fontsize{5.7}{6.7}\selectfont] at (10.03500,1.36950) {0.00};
\fill[GCRateMid!0.000!GCRateLow] (10.31333,1.11900) rectangle ++(0.55667,0.50100);
\node[font=\fontsize{5.7}{6.7}\selectfont] at (10.59167,1.36950) {0.00};
\fill[black!7] (10.87000,1.11900) rectangle ++(0.55667,0.50100);
\node[font=\fontsize{5.7}{6.7}\selectfont] at (11.14833,1.36950) {$-$};
\fill[GCRateMid!0.000!GCRateLow] (11.42667,1.11900) rectangle ++(0.55667,0.50100);
\node[font=\fontsize{5.7}{6.7}\selectfont] at (11.70500,1.36950) {0.00};
\fill[GCRateMid!0.000!GCRateLow] (11.98333,1.11900) rectangle ++(0.55667,0.50100);
\node[font=\fontsize{5.7}{6.7}\selectfont] at (12.26167,1.36950) {0.00};
\fill[GCRateMid!92.000!GCRateLow] (9.20000,0.61800) rectangle ++(0.55667,0.50100);
\node[font=\fontsize{5.7}{6.7}\selectfont] at (9.47833,0.86850) {0.46};
\fill[GCRateMid!0.000!GCRateLow] (9.75667,0.61800) rectangle ++(0.55667,0.50100);
\node[font=\fontsize{5.7}{6.7}\selectfont] at (10.03500,0.86850) {0.00};
\fill[GCRateMid!0.000!GCRateLow] (10.31333,0.61800) rectangle ++(0.55667,0.50100);
\node[font=\fontsize{5.7}{6.7}\selectfont] at (10.59167,0.86850) {0.00};
\fill[GCRateMid!0.000!GCRateLow] (10.87000,0.61800) rectangle ++(0.55667,0.50100);
\node[font=\fontsize{5.7}{6.7}\selectfont] at (11.14833,0.86850) {0.00};
\fill[black!7] (11.42667,0.61800) rectangle ++(0.55667,0.50100);
\node[font=\fontsize{5.7}{6.7}\selectfont] at (11.70500,0.86850) {$-$};
\fill[GCRateMid!0.000!GCRateLow] (11.98333,0.61800) rectangle ++(0.55667,0.50100);
\node[font=\fontsize{5.7}{6.7}\selectfont] at (12.26167,0.86850) {0.00};
\fill[GCRateMid!84.000!GCRateLow] (9.20000,0.11700) rectangle ++(0.55667,0.50100);
\node[font=\fontsize{5.7}{6.7}\selectfont] at (9.47833,0.36750) {0.42};
\fill[GCRateMid!8.000!GCRateLow] (9.75667,0.11700) rectangle ++(0.55667,0.50100);
\node[font=\fontsize{5.7}{6.7}\selectfont] at (10.03500,0.36750) {0.04};
\fill[GCRateMid!0.000!GCRateLow] (10.31333,0.11700) rectangle ++(0.55667,0.50100);
\node[font=\fontsize{5.7}{6.7}\selectfont] at (10.59167,0.36750) {0.00};
\fill[GCRateMid!0.000!GCRateLow] (10.87000,0.11700) rectangle ++(0.55667,0.50100);
\node[font=\fontsize{5.7}{6.7}\selectfont] at (11.14833,0.36750) {0.00};
\fill[GCRateMid!0.000!GCRateLow] (11.42667,0.11700) rectangle ++(0.55667,0.50100);
\node[font=\fontsize{5.7}{6.7}\selectfont] at (11.70500,0.36750) {0.00};
\fill[black!7] (11.98333,0.11700) rectangle ++(0.55667,0.50100);
\node[font=\fontsize{5.7}{6.7}\selectfont] at (12.26167,0.36750) {$-$};
\draw[black!60,line width=.25pt] (9.20000,0.11700) rectangle ++(3.34,3.006);
\node[anchor=north east,font=\fontsize{6.5}{7.5}\selectfont,rotate=45] at (9.47833,.12) {@2};
\node[anchor=east,font=\fontsize{6.5}{7.5}\selectfont] at (9.10000,2.87250) {@2};
\node[anchor=north east,font=\fontsize{6.5}{7.5}\selectfont,rotate=45] at (10.03500,.12) {@4};
\node[anchor=east,font=\fontsize{6.5}{7.5}\selectfont] at (9.10000,2.37150) {@4};
\node[anchor=north east,font=\fontsize{6.5}{7.5}\selectfont,rotate=45] at (10.59167,.12) {@6};
\node[anchor=east,font=\fontsize{6.5}{7.5}\selectfont] at (9.10000,1.87050) {@6};
\node[anchor=north east,font=\fontsize{6.5}{7.5}\selectfont,rotate=45] at (11.14833,.12) {@8};
\node[anchor=east,font=\fontsize{6.5}{7.5}\selectfont] at (9.10000,1.36950) {@8};
\node[anchor=north east,font=\fontsize{6.5}{7.5}\selectfont,rotate=45] at (11.70500,.12) {@10};
\node[anchor=east,font=\fontsize{6.5}{7.5}\selectfont] at (9.10000,0.86850) {@10};
\node[anchor=north east,font=\fontsize{6.5}{7.5}\selectfont,rotate=45] at (12.26167,.12) {Final};
\node[anchor=east,font=\fontsize{6.5}{7.5}\selectfont] at (9.10000,0.36750) {Final};
\node[font=\small] at (10.87000,-0.7) {(c) Epoch 100};
\shade[bottom color=GCRateLow,top color=GCRateHigh,middle color=GCRateMid] (13.18,.25) rectangle (13.37,3.05);
\draw[black!60,line width=.25pt] (13.18,.25) rectangle (13.37,3.05);
\draw[line width=.25pt] (13.37,0.2500) -- (13.42,0.2500);
\node[anchor=west,font=\fontsize{6.3}{7.3}\selectfont] at (13.44,0.2500) {0.0};
\draw[line width=.25pt] (13.37,1.6500) -- (13.42,1.6500);
\node[anchor=west,font=\fontsize{6.3}{7.3}\selectfont] at (13.44,1.6500) {0.5};
\draw[line width=.25pt] (13.37,3.0500) -- (13.42,3.0500);
\node[anchor=west,font=\fontsize{6.3}{7.3}\selectfont] at (13.44,3.0500) {1.0};
\node[rotate=90,font=\fontsize{6.3}{7.3}\selectfont] at (13.96,1.65) {Conflict rate};
\end{tikzpicture}
\endgroup

%% file: sections/4_method.tex
\section{Decoupling early classification from feature refinement}
\label{sec:method}

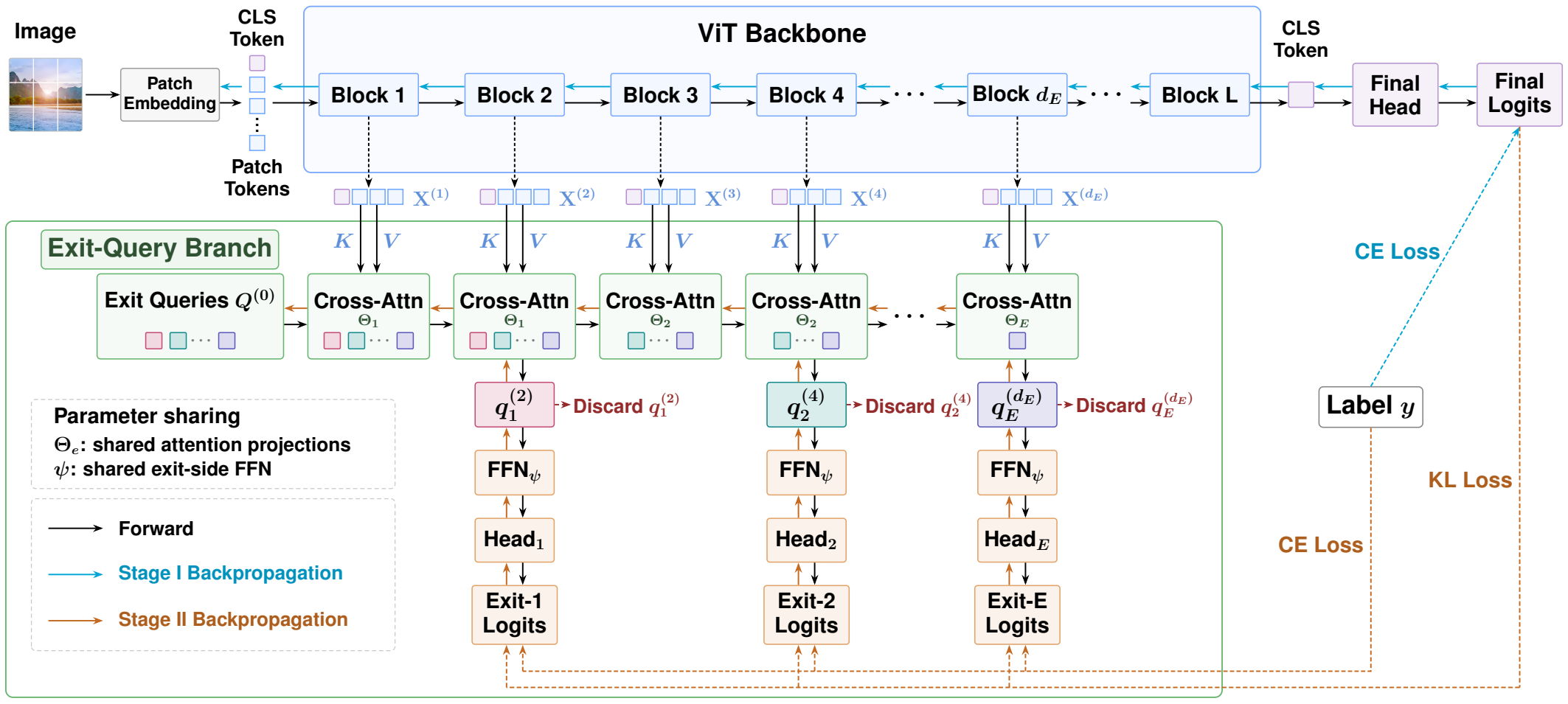
\begin{figure*}[t]
    \centering
    \input{artwork/methodology}
    \caption{DeclareViT with progressively updated read-only exit queries, discarded after prediction.
Stage I trains the backbone and final classifier with cross-entropy;
Stage II freezes them and trains only the exit branches with
cross-entropy and KL distillation from the final prediction.}
    \label{fig:method}
\end{figure*}

Motivated by the shared-backbone competition in
Section~\ref{sec:preliminaries}, DeclareViT combines read-only
exit queries with two-stage training to decouple early classification
from feature refinement
(Figure~\ref{fig:method}).
We next describe the query architecture, training procedure,
and dynamic inference.

\subsection{Read-Only Exit Queries}
\label{sec:exit_query}

\paragraph{Query initialization.}
Each exit $e$ has a learnable initial query
$\mathbf q_e^{(0)}\in\mathbb R^{D\times1}$
serving as its dedicated early-classification state outside the backbone token sequence.
Stacking these queries gives
$\mathbf Q^{(0)}
=[\mathbf q_1^{(0)},\ldots,\mathbf q_E^{(0)}]
\in\mathbb R^{D\times E}$.
$\mathbf q_e^{(\ell)}$ denotes the state of query $e$
after reading backbone block $\ell$.
Each query is updated until its exit depth $d_e$.
\paragraph{Extracting and accumulating information.}
At backbone block $\ell$, each active query $e$ with
$\ell\leq d_e$ reads the intermediate features $\mathbf X^{(\ell)}$.
Its query, key, and value projections are
\begin{equation}
\mathbf{Q}_e^{(\ell)}
=
\mathbf{W}_{Q,\ell}
\operatorname{LN}_{q,\ell}
\!\left(\mathbf{q}_e^{(\ell-1)}\right),
\;
\mathbf{K}^{(\ell)}
=
\mathbf{W}_{K,\ell}
\operatorname{LN}_{b,\ell}
\!\left(\mathbf{X}^{(\ell)}\right),
\;
\mathbf{V}^{(\ell)}
=
\mathbf{W}_{V,\ell}
\operatorname{LN}_{b,\ell}
\!\left(\mathbf{X}^{(\ell)}\right).
\label{eq:query_qkv}
\end{equation}

Here,
$\operatorname{LN}_{q,\ell}$ and $\operatorname{LN}_{b,\ell}$ are separate
LayerNorm operations for the query and backbone features, respectively.
The matrices
$\mathbf{W}_{Q,\ell}$,
$\mathbf{W}_{K,\ell}$, and
$\mathbf{W}_{V,\ell}\in\mathbb{R}^{D\times D}$
are the corresponding projection matrices.
For a standard ViT,
$\mathbf{Q}_e^{(\ell)}$ has shape $D\times 1$, $\mathbf{K}^{(\ell)}$ and $\mathbf{V}^{(\ell)}$ each have shape
$D\times(N+1)$.

Each query accumulates information through a residual
cross-attention update:
\begin{equation}
\mathbf q_e^{(\ell)}
=
\mathbf q_e^{(\ell-1)}
+
\operatorname{MHA}_{\ell}\!\left(
\mathbf Q_e^{(\ell)},\mathbf K^{(\ell)},\mathbf V^{(\ell)}
\right),
\qquad
1\leq\ell\leq d_e,
\label{eq:query_update}
\end{equation}
where $\operatorname{MHA}_{\ell}$ denotes multi-head cross-attention~\citep{Vaswani:2017:AIA} without
applying additional query, key, or value projections.
Throughout this process, the query remains read-only: it attends to
$\mathbf{X}^{(\ell)}$ without modifying the backbone features.  For hierarchical backbones such as Swin~\citep{Liu:2021:STH}, where repeatedly reading high-resolution intermediate features would incur substantial overhead, we use an exit-local non-accumulative reader; details are provided in Appendix~\ref{app:backbone_adaptation}.

\paragraph{Exit readout and discard.}
When the backbone reaches block $\ell=d_e$, query $e$ produces
its prediction through an exit-side FFN and an exit-specific
linear classifier:
\begin{equation}
\widetilde{\mathbf q}_e
=
\mathbf q_e^{(d_e)}
+
\operatorname{FFN}_{\psi}\!\left(
\operatorname{LN}_{\mathrm{ffn}}\!\left(
\mathbf q_e^{(d_e)}
\right)\right),
\qquad
\mathbf z_e
=
\mathbf W_e\widetilde{\mathbf q}_e+\mathbf b_e.
\label{eq:exit_query_logits}
\end{equation}
Here, $\psi$ denotes the FFN parameters, while
$\mathbf{W}_e$ and $\mathbf{b}_e$ form the linear classifier.
After producing $\mathbf{z}_e$, query $\mathbf{q}_e^{(d_e)}$
is discarded, i.e., excluded from the query matrices used by subsequent readers.

\subsection{Two-Stage Decoupled Training}
\label{sec:decoupled_optimization}
\paragraph{Stage I: full-depth training.}

This stage learns the backbone trajectory without early-exit supervision. For an image--label pair $(\mathbf{I},y)$, Stage~I optimizes only the full-depth classification objective:
\begin{equation}
    \left(
        \theta^\star,
        \omega^\star
    \right)
    =
    \operatorname*{arg\,min}_{\theta,\omega}
    \;
    \mathbb{E}_{(\mathbf{I},y)}
    \left[
        \mathcal{L}_{\mathrm{CE}}
        \left(
            \mathbf{z}_{\mathrm{f}}
            (\mathbf{I};\theta,\omega),
            y
        \right)
    \right],
    \label{eq:stage_one}
\end{equation}
where $\theta$ and $\omega$ denote the parameters of the backbone and final classifier, respectively.

\paragraph{Stage II: exit-branch training.}
We then freeze $\theta^\star$ and $\omega^\star$ and optimize only the
exit-query branch parameters $\varphi$, including the learnable queries,
cross-attention projections, exit-side FFN, and exit classifiers.
The loss for exit $e$ combines ground-truth supervision with self-distillation
from the frozen final classifier:
\begin{equation}
    \mathcal{L}_e
    =
    (1-\alpha)
    \mathcal{L}_{\mathrm{CE}}
    \left(
        \mathbf{z}_e,y
    \right)
    +
    \alpha T^2
    \mathcal{L}_{\mathrm{KL}}
    \left(
        \mathbf{p}_{\mathrm{f}}^{(T)}
        \,\middle\|\,
        \mathbf{p}_e^{(T)}
    \right),
    \label{eq:query_loss}
\end{equation}
where $\mathcal{L}_{\mathrm{KL}}$ denotes the Kullback--Leibler divergence,
$\mathbf{p}_e^{(T)}=\operatorname{softmax}(\mathbf{z}_e/T)$ and
$\mathbf{p}_{\mathrm{f}}^{(T)}
=\operatorname{softmax}(\mathbf{z}_{\mathrm{f}}/T)$.
Here, $\alpha\in[0,1]$ controls the distillation weight and $T>0$ is the
temperature.

The Stage~II objective is
\begin{equation}
    \varphi^\star
    =
    \operatorname*{arg\,min}_{\varphi}
    \;
    \mathbb{E}_{(\mathbf{I},y)}
    \left[
        \sum_{e=1}^{E}
        \lambda_e
        \mathcal{L}_e
    \right],
    \qquad
    \theta=\theta^\star,\quad
    \omega=\omega^\star.
    \label{eq:stage_two}
\end{equation}

\subsection{Dynamic Inference and Parameter Sharing}
\label{sec:dynamic_inference}
\paragraph{Dynamic inference.}
DeclareViT evaluates exits in depth order.
For $\mathbf{p}_e(\mathbf{I})
=\operatorname{softmax}(\mathbf{z}_e(\mathbf{I}))$,
prediction uncertainty is measured by normalized entropy~\citep{Teerapittayanon:2016:BFI}:
\begin{equation}
    H_e(\mathbf{I})
    = -\frac{1}{\log C}
    \sum_{c=1}^{C}
    p_{e,c}(\mathbf{I})\log p_{e,c}(\mathbf{I}),
    \;
    H_e(\mathbf{I})\in[0,1].
    \label{eq:normalized_entropy}
\end{equation}

At exit $e$, inference terminates when
$H_e(\mathbf I)\leq \tau_e$.
Otherwise, computation continues to the next exit.
If no early exit meets its threshold, the final classifier produces the prediction.
Here, $\tau_e$ denotes the entropy threshold for exit $e$.
\paragraph{Parameter sharing.}
To reduce parameter overhead introduced by the exit branches,
all exits share an FFN and its LayerNorm, while query states and classifiers remain exit-specific.
We further adopt inter-exit segment sharing for cross-attention.
With $d_0=0$, segment $e$ contains backbone blocks
$\ell=d_{e-1}+1,\ldots,d_e$, for $e=1,\ldots,E$.
All readers within each segment share the projection matrices
$W_Q$, $W_K$, $W_V$, and $W_O$.
The query-side and memory-side LayerNorms are layer-specific and shared by all active exit queries at the same block.

%% file: artwork/methodology.tex
\begingroup
\bfseries
\boldmath

\newcommand{\methodFontBody}{\fontsize{15}{17}\selectfont}
\newcommand{\methodFontNote}{\fontsize{13.5}{15.5}\selectfont}
\newcommand{\methodFontTiny}{\fontsize{11}{12.5}\selectfont}
\newcommand{\methodFontTitle}{\fontsize{18}{20}\selectfont}

\providecommand{\methodAssetDir}{artwork/teaser-assets}
\providecommand{\methodExitDepthOne}{2}
\providecommand{\methodExitDepthTwo}{4}
\providecommand{\methodExitDepthThree}{6}
\providecommand{\methodExitDepthLast}{d_E}
\def\methodKX{25.20}
\def\methodArrowOffset{2.0mm}

\definecolor{methodpanelborder}{HTML}{C9C9C9}
\definecolor{methodrightheader}{HTML}{EFF7F0}
\definecolor{methodtokenpurple}{HTML}{B791D1}
\definecolor{methodtokenpurplelight}{HTML}{E7D6F1}
\definecolor{methodtokenblue}{HTML}{6FA8FF}
\definecolor{methodblockfill}{HTML}{F3F7FF}
\definecolor{methodblockborder}{HTML}{79A9FF}
\definecolor{methodheadfill}{HTML}{FAF5FF}
\definecolor{methodheadborder}{HTML}{BFA1DE}
\definecolor{methodexitgreen}{HTML}{66B56F}
\definecolor{methodquerygreen}{HTML}{C44E78}
\definecolor{methodqueryorange}{HTML}{168C8C}
\definecolor{methodffnorange}{HTML}{E2A469}
\definecolor{methodquerypurple}{HTML}{B994D0}
\definecolor{methodquerycyan}{HTML}{68BFC0}
\definecolor{methodqueryblue}{HTML}{5E5ABD}
\definecolor{methodflameorange}{HTML}{D97706}
\definecolor{methodretirered}{HTML}{EF4444}
\definecolor{methodstageone}{HTML}{00A6D6}
\definecolor{methodstagetwo}{HTML}{C56A20}

\tikzset{
  every node/.append style={line width=.80pt},
  method arrow/.style={
    -{Stealth[length=2.5mm,width=1.8mm]},line width=1.0pt
  },
  method flow/.style={method arrow,draw=black},
  method query flow/.style={method flow},
  method memory/.style={
    method arrow,draw=black,dash pattern=on 2.2pt off 1.5pt
  },
  method memory link/.style={method arrow,draw=black},
  method memory label/.style={
    text=methodtokenblue!85!black,fill=white,rounded corners=1pt,
    inner sep=1.1pt,font=\sffamily\bfseries\methodFontNote
  },
  method panel title/.style={
    draw=methodexitgreen,fill=methodrightheader,rounded corners=3pt,
    minimum height=.96cm,inner xsep=1.5mm,inner ysep=1.3mm,
    font=\sffamily\bfseries\methodFontTitle,
    text=methodexitgreen!45!black
  },
  method retire/.style={
    method arrow,draw=methodretirered!62!black,densely dashed
  },
  method stage one grad/.style={method arrow,draw=methodstageone},
  method stage two grad/.style={method arrow,draw=methodstagetwo},
  method stage one loss link/.style={
    method arrow,draw=methodstageone,dash pattern=on 3.0pt off 1.6pt
  },
  method stage two loss link/.style={
    method arrow,draw=methodstagetwo,dash pattern=on 3.0pt off 1.6pt
  },
  method stage two loss stem/.style={
    draw=methodstagetwo,dash pattern=on 3.0pt off 1.6pt,line width=1.0pt
  },
  method distill feed/.style={
    method arrow,draw=methodstagetwo,dash pattern=on 3.2pt off 1.7pt
  },
  method distill stem/.style={
    draw=methodstagetwo,dash pattern=on 3.2pt off 1.7pt,line width=1.0pt
  },
  method label/.style={
    draw=black!48,fill=white,rounded corners=3pt,
    minimum width=2.40cm,minimum height=1.00cm,
    align=center,inner sep=5pt,font=\sffamily\bfseries\methodFontTitle
  },
  method loss annotation/.style={
    fill=white,rounded corners=1.2pt,inner sep=1.2pt,
    font=\sffamily\bfseries\methodFontBody
  },
  method block/.style={
    draw=methodblockborder,fill=methodblockfill,rounded corners=1.5pt,
    minimum width=2.45cm,minimum height=1.05cm,
    inner sep=1.5pt,font=\sffamily\bfseries\methodFontBody
  },
  method embedding/.style={
    draw=black!38,fill=black!3,rounded corners=2pt,
    minimum width=2.25cm,minimum height=1.30cm,
    align=center,inner sep=2pt,
    font=\sffamily\bfseries\fontsize{12}{14}\selectfont
  },
  method cross/.style={
    draw=methodexitgreen,fill=methodrightheader!48,
    rounded corners=2.5pt,minimum width=3.00cm,minimum height=2.10cm,
    inner sep=1pt
  },
  method shared projection label/.style={
    font=\sffamily\bfseries\methodFontTiny,
    text=methodexitgreen!45!black,inner sep=.5pt
  },
  method ffn/.style={
    draw=methodffnorange,fill=methodffnorange!13,rounded corners=2pt,
    minimum width=1.95cm,minimum height=1.10cm,
    align=center,inner sep=1.2pt,font=\sffamily\bfseries\methodFontBody
  },
  method exit head/.style={method ffn,minimum height=1.08cm},
  method logits/.style={
    method ffn,minimum width=2.10cm,minimum height=1.45cm
  },
  method final head/.style={
    draw=methodquerypurple,fill=methodquerypurple!13,rounded corners=2pt,
    minimum width=2.10cm,minimum height=1.55cm,
    align=center,inner sep=1.4pt,font=\sffamily\bfseries\methodFontBody
  },
  method query/.style={
    draw=#1,fill=#1!15,rounded corners=2pt,
    minimum width=.86cm,minimum height=.72cm,
    inner sep=1pt,font=\sffamily\bfseries\methodFontBody
  },
  method exit query/.style={
    draw=#1,fill=#1!16,rounded corners=2pt,
    minimum width=1.95cm,minimum height=1.10cm,
    inner xsep=2pt,inner ysep=1.5pt,
    font=\sffamily\bfseries\methodFontTitle
  },
  method query bank/.style={
    draw=methodexitgreen,fill=methodrightheader!62,rounded corners=2.5pt,
    minimum width=4.60cm,minimum height=2.10cm,inner sep=1pt
  },
  method chip/.style={
    draw=#1,fill=#1!22,rounded corners=.8pt,
    minimum width=.40cm,minimum height=.40cm,inner sep=0pt
  },
  method query ellipsis/.style={
    font=\sffamily\bfseries\methodFontTiny,text=black!52,
    minimum width=.40cm,minimum height=.40cm,
    inner sep=0pt,outer sep=0pt
  },
  method patch/.style={
    draw=methodtokenblue,fill=methodtokenblue!9,
    minimum width=.37cm,minimum height=.37cm,inner sep=0pt
  },
  method cls/.style={
    draw=methodtokenpurple,fill=methodtokenpurple!15,rounded corners=1pt,
    minimum width=.37cm,minimum height=.37cm,inner sep=0pt
  },
  method op/.style={
    draw=black!35,fill=black!3,rounded corners=1.2pt,
    minimum width=1.03cm,minimum height=.50cm,
    inner sep=1pt,font=\sffamily\bfseries\methodFontNote
  },
  method matmul/.style={
    method op,minimum width=1.10cm,minimum height=.68cm,inner sep=1.2pt
  },
  method memory source/.style={
    draw=methodtokenblue,fill=methodtokenblue!7,rounded corners=2pt,
    minimum width=7.00cm,minimum height=.90cm,align=center,
    inner sep=1.5pt,font=\sffamily\bfseries\methodFontNote,
    text=methodtokenblue!82!black
  },
  method projection/.style={
    draw=methodexitgreen,fill=methodexitgreen!15,rounded corners=1pt,
    minimum width=.54cm,minimum height=.54cm,
    inner sep=.5pt,font=\sffamily\bfseries\methodFontTiny
  }
}

\newcommand{\methodInputImage}{%
  \begin{scope}
    \clip[rounded corners=1.5pt] (0,0) rectangle (1.80,1.80);
    \IfFileExists{\methodAssetDir/input-square.png}{%
      \node[anchor=south west,inner sep=0pt] at (0,0)
        {\includegraphics[width=1.80cm,height=1.80cm]
          {\methodAssetDir/input-square.png}};
    }{%
      \fill[methodtokenblue!12] (0,0) rectangle (1.80,1.80);
      \node[font=\sffamily\bfseries\methodFontTiny] at (.90,.90) {Image};
    }%
    \foreach \p in {.60,1.20}{%
      \draw[white,line width=.65pt] (\p,0)--(\p,1.80);
      \draw[white,line width=.65pt] (0,\p)--(1.80,\p);
    }%
  \end{scope}
  \draw[black!28,rounded corners=1.5pt,line width=.35pt]
    (0,0) rectangle (1.80,1.80);
}

\newcommand{\methodTokenStack}{%
  \node[method cls] (methodInputCLS) at (0,.80) {};
  \node[method patch] at (0,.27) {};
  \node[method patch] at (0,-.26) {};
  \node[font=\sffamily\bfseries\methodFontTiny] at (0,-.62) {$\vdots$};
  \node[method patch] at (0,-1.17) {};
  \node[font=\sffamily\bfseries\methodFontNote,align=center]
    at (0,1.67) {CLS\\Token};
  \node[font=\sffamily\bfseries\methodFontNote,align=center]
    at (0,-2.02) {Patch\\Tokens};
}

\newcommand{\methodPatchMemory}[1]{%
  \node[method cls] at (-.66,0) {};
  \foreach \xx in {-.22,.22,.66}{\node[method patch] at (\xx,0) {};}
  \node[font=\sffamily\bfseries\methodFontNote,
    text=methodtokenblue!85!black,anchor=west]
    at (.96,0) {$\mathbf{X}^{(#1)}$};
}

\newcommand{\methodCrossCell}[3]{%
  \node[method cross] (#1) at (#2,20.00) {};
  \node[font=\sffamily\bfseries\methodFontBody]
    at ([yshift=4.0mm]#1.center) {Cross-Attn};
  \node[method shared projection label]
    at ([yshift=-1.1mm]#1.center) {$\Theta_{#3}$};
}

\newcommand{\methodAllQueryChips}[1]{%
  \node[method chip=methodquerygreen]
    at ([xshift=-9.0mm,yshift=-6.0mm]#1.center) {};
  \node[method chip=methodqueryorange]
    at ([xshift=-3.0mm,yshift=-6.0mm]#1.center) {};
  \node[method query ellipsis]
    at ([xshift=3.0mm,yshift=-6.0mm]#1.center) {$\cdots$};
  \node[method chip=methodqueryblue]
    at ([xshift=9.0mm,yshift=-6.0mm]#1.center) {};
}

\newcommand{\methodPostFirstQueryChips}[1]{%
  \node[method chip=methodqueryorange]
    at ([xshift=-6.0mm,yshift=-6.0mm]#1.center) {};
  \node[method query ellipsis]
    at ([yshift=-6.0mm]#1.center) {$\cdots$};
  \node[method chip=methodqueryblue]
    at ([xshift=6.0mm,yshift=-6.0mm]#1.center) {};
}

\newcommand{\methodOneQueryChip}[1]{%
  \node[method chip=methodqueryblue]
    at ([yshift=-6.0mm]#1.center) {};
}

\newcommand{\methodExitBranch}[4]{%
  \node[method exit query=#2,below=5.5mm of #1]
    (#1ExitQuery) {$q_{#3}^{(#4)}$};
  \draw[method flow]
    ([xshift=\methodArrowOffset]#1.south)--
    ([xshift=\methodArrowOffset]#1ExitQuery.north);
  \node[method ffn,below=5.5mm of #1ExitQuery]
    (#1FFN) {FFN$_{\psi}$};
  \node[method exit head,below=5.5mm of #1FFN]
    (#1Head) {Head$_{#3}$};
  \node[method logits,below=5.5mm of #1Head]
    (#1Logits) {Exit-#3\\Logits};
  \foreach \from/\to in {ExitQuery/FFN,FFN/Head,Head/Logits}{%
    \draw[method flow]
      ([xshift=\methodArrowOffset]#1\from.south)--
      ([xshift=\methodArrowOffset]#1\to.north);
  }%
  \draw[method retire] (#1ExitQuery.east)--++(.40,0)
    node[anchor=west,align=left,inner sep=1.2pt,
      text=methodretirered!62!black,
      font=\sffamily\bfseries\methodFontNote]
      {Discard $q_{#3}^{(#4)}$};
}

\newcommand{\methodBackbonePanel}{%
  \node[font=\sffamily\bfseries\methodFontBody] at (1.23,27.00) {Image};
  \begin{scope}[shift={(.33,24.57)}]\methodInputImage\end{scope}
  \node[method embedding] (methodEmbedding) at (4.30,25.47)
    {Patch\\Embedding};
  \draw[method flow] (2.22,25.47)--(methodEmbedding.west);
  \begin{scope}[shift={(6.45,25.45)}]\methodTokenStack\end{scope}
  \coordinate (methodInputTokenWest) at (6.07,25.47);
  \coordinate (methodInputTokenEast) at (6.83,25.47);
  \draw[method flow]
    ([yshift=-\methodArrowOffset]methodEmbedding.east)--
    ([yshift=-\methodArrowOffset]methodInputTokenWest);

  \fill[methodblockfill!45,rounded corners=5pt]
    (7.60,23.55) rectangle (31.20,27.65);
  \draw[methodblockborder,rounded corners=5pt,line width=.95pt]
    (7.60,23.55) rectangle (31.20,27.65);
  \node[font=\sffamily\bfseries\methodFontTitle,inner sep=0pt,outer sep=0pt]
    (methodBackboneTitleText) at (19.40,27.00) {ViT Backbone};
  \node[method block] (methodB1) at (9.20,25.47) {Block 1};
  \node[method block] (methodB2) at (12.80,25.47) {Block 2};
  \node[method block] (methodB3) at (16.40,25.47) {Block 3};
  \node[method block] (methodB4) at (20.00,25.47) {Block 4};
  \node[method block] (methodBLm) at (\methodKX,25.47) {Block $d_E$};
  \node[method block] (methodBL) at (29.70,25.47) {Block L};
  \node[font=\sffamily\bfseries\methodFontTitle,inner xsep=.6mm,inner ysep=0pt]
    (methodGapBFourK) at ($(methodB4.east)!.5!(methodBLm.west)$) {$\cdots$};
  \node[font=\sffamily\bfseries\methodFontTitle,inner xsep=.6mm,inner ysep=0pt]
    (methodGapKL) at ($(methodBLm.east)!.5!(methodBL.west)$) {$\cdots$};
  \foreach \from/\to in {
    methodInputTokenEast/methodB1.west,
    methodB1.east/methodB2.west,
    methodB2.east/methodB3.west,
    methodB3.east/methodB4.west,
    methodB4.east/methodGapBFourK.west,
    methodGapBFourK.east/methodBLm.west,
    methodBLm.east/methodGapKL.west,
    methodGapKL.east/methodBL.west}{%
    \draw[method flow]
      ([yshift=-\methodArrowOffset]\from)--
      ([yshift=-\methodArrowOffset]\to);
  }%
  \foreach \bb/\xx/\lab in {
    methodB1/9.20/1,methodB2/12.80/2,methodB3/16.40/3,
    methodB4/20.00/4,methodBLm/\methodKX/{d_E}}{%
    \draw[method memory] (\bb.south)--(\xx,23.17);
    \begin{scope}[shift={(\xx,22.95)}]\methodPatchMemory{\lab}\end{scope}
  }%
  \coordinate (methodMemoryOne) at (9.20,22.75);
  \coordinate (methodMemoryTwo) at (12.80,22.75);
  \coordinate (methodMemoryThree) at (16.40,22.75);
  \coordinate (methodMemoryFour) at (20.00,22.75);
  \coordinate (methodMemoryK) at (\methodKX,22.75);
  \node[method cls,minimum width=.62cm,minimum height=.62cm]
    (methodFinalCLS) at (32.20,25.47) {};
  \node[font=\sffamily\bfseries\methodFontNote,align=center]
    at (32.20,26.85) {CLS\\Token};
  \node[method final head] (methodFinalHead) at (34.525,25.47)
    {Final\\Head};
  \node[method final head] (methodFinalLogits) at (37.590,25.47)
    {Final\\Logits};
  \foreach \from/\to in {
    methodBL/methodFinalCLS,
    methodFinalCLS/methodFinalHead,
    methodFinalHead/methodFinalLogits}{%
    \draw[method flow]
      ([yshift=-\methodArrowOffset]\from.east)--
      ([yshift=-\methodArrowOffset]\to.west);
  }%
}

\newcommand{\methodQueryPanel}{%
  \fill[white,rounded corners=5pt] (.25,10.60) rectangle (30.25,22.35);
  \draw[methodexitgreen,rounded corners=5pt,line width=.95pt]
    (.25,10.60) rectangle (30.25,22.35);
  \node[method panel title,minimum width=0pt,outer sep=0pt]
    (methodQueryTitleText) at (4.05,21.65) {Exit-Query Branch};

  \methodCrossCell{methodCAOne}{9.20}{1}
  \methodCrossCell{methodCATwo}{12.80}{1}
  \methodCrossCell{methodCAThree}{16.40}{2}
  \methodCrossCell{methodCAFour}{20.00}{2}
  \methodCrossCell{methodCAK}{\methodKX}{E}

  \foreach \memory/\cell in {
    methodMemoryOne/methodCAOne,
    methodMemoryTwo/methodCATwo,
    methodMemoryThree/methodCAThree,
    methodMemoryFour/methodCAFour,
    methodMemoryK/methodCAK}{%
    \draw[method memory link]
      ([xshift=-\methodArrowOffset]\memory)--
      node[pos=.52,left=1mm,method memory label] {$K$}
      ([xshift=-\methodArrowOffset]\cell.north);
    \draw[method memory link]
      ([xshift=\methodArrowOffset]\memory)--
      node[pos=.52,right=1mm,method memory label] {$V$}
      ([xshift=\methodArrowOffset]\cell.north);
  }%
  \node[method query bank] (methodQueryBank) at (4.80,20.00) {};
  \node[font=\sffamily\bfseries\methodFontBody,align=center,text=black]
    at ([yshift=4.0mm]methodQueryBank.center) {Exit Queries $Q^{(0)}$};
  \methodAllQueryChips{methodQueryBank}
  \methodAllQueryChips{methodCAOne}
  \methodAllQueryChips{methodCATwo}
  \methodPostFirstQueryChips{methodCAThree}
  \methodPostFirstQueryChips{methodCAFour}
  \methodOneQueryChip{methodCAK}
  \node[font=\sffamily\bfseries\methodFontTitle,text=black,
    inner xsep=1.2mm,inner ysep=0pt]
    (methodGapQuery) at ($(methodCAFour.east)!.5!(methodCAK.west)$)
    {$\cdots$};
  \foreach \from/\to in {
    methodQueryBank/methodCAOne,
    methodCAOne/methodCATwo,
    methodCATwo/methodCAThree,
    methodCAThree/methodCAFour,
    methodCAFour/methodGapQuery,
    methodGapQuery/methodCAK}{%
    \draw[method query flow]
      ([yshift=-\methodArrowOffset]\from.east)--
      ([yshift=-\methodArrowOffset]\to.west);
  }%
  \methodExitBranch{methodCATwo}{methodquerygreen}{1}{\methodExitDepthOne}
  \methodExitBranch{methodCAFour}{methodqueryorange}{2}{\methodExitDepthTwo}
  \methodExitBranch{methodCAK}{methodqueryblue}{E}{\methodExitDepthLast}
}

\newcommand{\methodStageTwoBranchBackprop}[1]{%
  \foreach \from/\to in {Logits/Head,Head/FFN,FFN/ExitQuery}{%
    \draw[method stage two grad]
      ([xshift=-\methodArrowOffset]#1\from.north)--
      ([xshift=-\methodArrowOffset]#1\to.south);
  }%
  \draw[method stage two grad]
    ([xshift=-\methodArrowOffset]#1ExitQuery.north)--
    ([xshift=-\methodArrowOffset]#1.south);
}

\newcommand{\methodGradientFlowOverlay}{%
  \coordinate (methodKDBusY) at ([yshift=-10.5mm]methodCATwoLogits.south);
  \coordinate (methodKDTeacherTop) at (methodFinalLogits.south);
  \coordinate (methodKDTeacherBottom) at (methodKDTeacherTop |- methodKDBusY);
  \coordinate (methodKDTeacherMid) at
    ($(methodKDTeacherTop)!.5!(methodKDTeacherBottom)$);
  \coordinate (methodQueryFrameRight) at (30.25,0);
  \coordinate (methodQueryFrameRightAtKDMid) at
    (methodQueryFrameRight |- methodKDTeacherMid);
  \node[method label] (methodLabel) at
    ($(methodQueryFrameRightAtKDMid)!.5!(methodKDTeacherMid)$) {Label $y$};
  \draw[method stage one loss link]
    (methodLabel.north)--
    node[pos=.52,left=1.4mm,method loss annotation,
      text=methodstageone!88!black] {CE Loss}
    (methodFinalLogits.south);

  \foreach \from/\to in {
    methodFinalLogits.west/methodFinalHead.east,
    methodFinalHead.west/methodFinalCLS.east,
    methodFinalCLS.west/methodBL.east,
    methodBL.west/methodGapKL.east,
    methodGapKL.west/methodBLm.east,
    methodBLm.west/methodGapBFourK.east,
    methodGapBFourK.west/methodB4.east,
    methodB4.west/methodB3.east,
    methodB3.west/methodB2.east,
    methodB2.west/methodB1.east,
    methodB1.west/methodInputTokenEast,
    methodInputTokenWest/methodEmbedding.east}{%
    \draw[method stage one grad]
      ([yshift=\methodArrowOffset]\from)--
      ([yshift=\methodArrowOffset]\to);
  }%

  \coordinate (methodCEBusY) at ([yshift=-6.5mm]methodCATwoLogits.south);
  \coordinate (methodCEOne) at
    ([xshift=\methodArrowOffset]methodCATwoLogits.south);
  \coordinate (methodCETwo) at
    ([xshift=\methodArrowOffset]methodCAFourLogits.south);
  \coordinate (methodCEE) at
    ([xshift=\methodArrowOffset]methodCAKLogits.south);
  \draw[method stage two loss stem]
    (methodLabel.south)--
    node[pos=.48,left=1.2mm,method loss annotation,
      text=methodstagetwo!88!black] {CE Loss}
    (methodLabel.south |- methodCEBusY)--(methodCEOne |- methodCEBusY);
  \foreach \port in {methodCEOne,methodCETwo,methodCEE}{%
    \draw[method stage two loss link] (\port |- methodCEBusY)--(\port);
  }%

  \draw[method distill stem]
    (methodKDTeacherTop)--
    node[pos=.63,left=1.2mm,method loss annotation,
      text=methodstagetwo!88!black] {KL Loss}
    (methodKDTeacherBottom);
  \coordinate (methodKDOne) at
    ([xshift=-\methodArrowOffset]methodCATwoLogits.south);
  \coordinate (methodKDTwo) at
    ([xshift=-\methodArrowOffset]methodCAFourLogits.south);
  \coordinate (methodKDE) at
    ([xshift=-\methodArrowOffset]methodCAKLogits.south);
  \draw[method distill stem]
    (methodKDTeacherBottom)--(methodKDOne |- methodKDBusY);
  \foreach \port in {methodKDOne,methodKDTwo,methodKDE}{%
    \draw[method distill feed] (\port |- methodKDBusY)--(\port);
  }%

  \methodStageTwoBranchBackprop{methodCATwo}
  \methodStageTwoBranchBackprop{methodCAFour}
  \methodStageTwoBranchBackprop{methodCAK}
  \foreach \from/\to in {
    methodCAOne/methodQueryBank,
    methodCATwo/methodCAOne,
    methodCAThree/methodCATwo,
    methodCAFour/methodCAThree,
    methodCAK/methodGapQuery,
    methodGapQuery/methodCAFour}{%
    \draw[method stage two grad]
      ([yshift=\methodArrowOffset]\from.west)--
      ([yshift=\methodArrowOffset]\to.east);
  }%
}

\newcommand{\methodLegends}{%
  \fill[white,rounded corners=3pt] (.88,15.75) rectangle (8.35,18.45);
  \draw[methodpanelborder,densely dashed,rounded corners=3pt,line width=.65pt]
    (.88,15.75) rectangle (9.85,17.95);
  \node[anchor=west,font=\sffamily\bfseries\methodFontBody]
    at (1.30,17.50) {Parameter sharing};
  \node[anchor=west,font=\sffamily\bfseries\methodFontNote]
    at (1.30,16.80) {$\Theta_e$: shared attention projections};
  \node[anchor=west,font=\sffamily\bfseries\methodFontNote]
    at (1.30,16.20) {$\psi$: shared exit-side FFN};

  \fill[white,rounded corners=3pt] (.88,11.75) rectangle (8.35,15.50);
  \draw[methodpanelborder,densely dashed,rounded corners=3pt,line width=.65pt]
    (.88,11.75) rectangle (9.85,15.50);
  \draw[method flow] (1.30,14.77)--(2.65,14.77);
  \node[anchor=west,font=\sffamily\bfseries\methodFontNote]
    at (2.90,14.77) {Forward};
  \draw[method stage one grad] (1.30,13.63)--(2.65,13.63);
  \node[anchor=west,font=\sffamily\bfseries\methodFontNote,
    text=methodstageone!88!black] at (2.90,13.63) {Stage I Backpropagation};
  \draw[method stage two grad] (1.30,12.49)--(2.65,12.49);
  \node[anchor=west,font=\sffamily\bfseries\methodFontNote,
    text=methodstagetwo!88!black] at (2.90,12.49) {Stage II Backpropagation};
}

\newcommand{\methodModuleLegend}{%
  \begin{scope}[yshift=4.20cm]
    \fill[methodrightheader!28,rounded corners=5pt]
      (7.95,.40) rectangle (26.00,5.80);
    \fill[methodrightheader,rounded corners=5pt]
      (7.95,4.92) rectangle (26.00,5.80);
    \fill[methodrightheader] (7.95,4.92) rectangle (26.00,5.12);
    \draw[methodexitgreen,rounded corners=5pt,line width=.60pt]
      (7.95,.40) rectangle (26.00,5.80);
    \node[font=\sffamily\bfseries\methodFontTitle,text=methodexitgreen!45!black]
      at (16.98,5.38) {Read-only Cross-Attention Update};
    \node[method query=methodquerygreen,minimum width=.95cm,minimum height=.75cm]
      (methodDetailQ) at (9.45,3.35) {$Q^{(\ell-1)}$};
    \node[method projection] (methodWQ) at (10.95,3.35) {$W_Q$};
    \node[method matmul,font=\sffamily\bfseries\fontsize{12}{14}\selectfont]
      (methodDot) at (13.05,3.35) {MatMul};
    \node[method op,font=\sffamily\bfseries\fontsize{12}{14}\selectfont]
      (methodSoftmax) at (15.15,3.35) {Softmax};
    \node[method matmul,font=\sffamily\bfseries\fontsize{12}{14}\selectfont]
      (methodMix) at (17.25,3.35) {MatMul};
    \node[method projection] (methodWO) at (19.35,3.35) {$W_O$};
    \node[draw=black!35,fill=white,circle,minimum size=.58cm,
      inner sep=0pt,font=\sffamily\bfseries\methodFontBody]
      (methodResidualPlus) at (20.95,3.35) {$+$};
    \node[method query=methodquerygreen,minimum width=.95cm,minimum height=.75cm]
      (methodUpdatedQ) at (22.60,3.35) {$Q^{(\ell)}$};
    \node[anchor=west,font=\sffamily\bfseries\methodFontNote,
      align=left,text=methodexitgreen!55!black]
      at (23.22,3.35) {Updated\\Query};
    \foreach \from/\to in {
      methodDetailQ/methodWQ,methodWQ/methodDot,
      methodDot/methodSoftmax,methodSoftmax/methodMix,
      methodMix/methodWO,methodWO/methodResidualPlus,
      methodResidualPlus/methodUpdatedQ}{%
      \draw[method flow] (\from.east)--(\to.west);
    }%
    \node[font=\sffamily\bfseries\methodFontTiny,fill=white,inner sep=.6pt]
      at (14.10,4.03) {$1/\sqrt{d_k}$};
    \draw[method flow]
      (methodDetailQ.north)--(9.45,4.50)--(20.95,4.50)--(methodResidualPlus.north);
    \node[method memory source] (methodDetailMemory) at (15.15,1.23) {};
    \node[method cls,minimum width=.30cm,minimum height=.30cm]
      at (12.18,1.23) {};
    \foreach \xx in {12.57,12.96,13.35}{%
      \node[method patch,minimum width=.30cm,minimum height=.30cm]
        at (\xx,1.23) {};
    }%
    \node[font=\sffamily\bfseries\fontsize{12}{14}\selectfont,
      text=methodtokenblue!82!black,align=center]
      at (16.05,1.23) {Backbone States $X^{(\ell)}$};
    \coordinate (methodKeyPort) at (13.05,1.68);
    \coordinate (methodValuePort) at (17.25,1.68);
    \node[method projection] (methodWK)
      at ($(methodKeyPort)!.5!(methodDot.south)$) {$W_K$};
    \node[method projection] (methodWV)
      at ($(methodValuePort)!.5!(methodMix.south)$) {$W_V$};
    \draw[method memory] (methodKeyPort)--(methodWK.south);
    \draw[method memory] (methodWK.north)--
      node[midway,right=2pt,font=\sffamily\bfseries\methodFontTiny] {$K^\top$}
      (methodDot.south);
    \draw[method memory] (methodValuePort)--(methodWV.south);
    \draw[method memory] (methodWV.north)--(methodMix.south);
    \node[font=\sffamily\bfseries\methodFontTiny,align=center,
      text=methodexitgreen!55!black] at (9.68,1.30)
      {Projections\\$\Theta_e$};
  \end{scope}
}

\resizebox{\linewidth}{!}{%
  \begin{tikzpicture}[x=1cm,y=1cm]
    \methodBackbonePanel
    \methodQueryPanel
    \methodGradientFlowOverlay
    \methodLegends
  \end{tikzpicture}%
}%
\endgroup

%% file: sections/5_experiments.tex
\begin{figure}[t]
\centering
\begingroup
\captionof{table}{
Comparison of early-exit methods on CIFAR-100, ImageNet-1K,
and Food-101 using ViT, DeiT, and Swin backbones with seed 2024.
Top-1 accuracy and $\mathbf{S}_{\mathrm{MAC}}$ are reported.
\textbf{Bold} and \underline{underlining} mark the best and second-best results within each dataset--backbone setting;
\textcolor{gray}{full-depth references are excluded from this ranking.}
$^{\dagger}$ denotes externally reported EnViT results.
}
\label{tab:tripd_main}
\input{tables/4_1_main_table}
\par\endgroup
\vspace{\floatsep}
\nextfloat
\makebox[\linewidth][c]{%
    \pgfplotslegendfromname{legend-vit}%
}\par

\makebox[\linewidth][c]{%
    \hspace*{-6pt}
    \scalebox{0.95}{%
        \makebox[\linewidth][c]{%
            \input{artwork/Curve/cifar_vit}\unskip
            \hspace{0.0135\linewidth}%
            \input{artwork/Curve/imagenet_vit}\unskip
            \hspace{0.0135\linewidth}%
            \input{artwork/Curve/food_vit}\unskip
        }%
    }%
    \hspace*{6pt}%
}\par
\caption{
Accuracy--$\mathbf{S}_{\mathrm{MAC}}$ trade-offs for ViT across three datasets.
}
\label{fig:tradeoff-vit}
\end{figure}
\section{Experiments\label{sec:experiments}}
\subsection{Experimental Setup}
\label{sec:experimental_setup}

\paragraph{Datasets and backbones.}
We evaluate on CIFAR-100~\citep{Krizhevsky:2009:CIFAR},
ImageNet-1K~\citep{Deng:2009:ImageNet}, and
Food-101~\citep{Bossard:2014:Food101}, using
ViT-B/16~\citep{Dosovitskiy:2021:AII},
distilled DeiT-B/16~\citep{Touvron:2021:TDI}, and
Swin-B~\citep{Liu:2021:STH}.
We refer to these backbones as ViT, DeiT, and Swin.
ViT and DeiT have 12 blocks, and Swin has 24.
\paragraph{Metrics.}
We report Top-1 accuracy, MAC-based speed-up $\mathbf{S}_{\mathrm{MAC}}$, and
wall-clock speed-up $\mathbf{S}_{\mathrm{wall}}$.
Following prior work~\citep{Dong:2026:EET}, we define
$\mathbf{S}_{\mathrm{MAC}}=N_s C_{\mathrm{base}}/\sum_j n_j C_j$
and
$\mathbf{S}_{\mathrm{wall}}=T_{\mathrm{base}}/T_{\mathrm{dyn}}$.
Here, $j$ indexes all early-exit and final prediction paths,
$n_j$ is the number of samples terminating at path $j$,
and $N_s=\sum_j n_j$ is the total number of samples.
$C_j$ is the cumulative MAC cost of path $j$, including all backbone
and exit-branch operations executed before termination, while
$C_{\mathrm{base}}$ denotes the MAC cost of the unbranched reference.
$T_{\mathrm{base}}$ and $T_{\mathrm{dyn}}$ are measured on the same hardware under a unified benchmarking protocol.
\paragraph{Implementation details.}
All inputs are processed at $224\times224$ resolution.
DeclareViT uses five early exits at blocks $\{2,4,6,8,10\}$ for ViT and DeiT
and global blocks $\{4,8,12,16,20\}$ for Swin,
with one query per exit.
Thresholds are calibrated on a held-out training subset
and applied to the final evaluation split.
Further dataset, backbone, architecture, training, hardware, and calibration
details are provided in Appendix~\ref{app:further_implementation}.

\subsection{Main Accuracy--Efficiency Trade-off}

Table~\ref{tab:tripd_main} compares six locally evaluated baselines across nine dataset--backbone settings and includes EnViT as an external ViT reference. Following prior work~\citep{Xu:2023:LDE, Dong:2026:EET}, it provides representative operating points as a compact summary of the overall accuracy–efficiency trade-off.
At the reported operating points, DeclareViT  attains higher accuracy and
larger MAC-based speed-up than the locally evaluated early-exit baselines
across all nine settings.
Relative to the corresponding full-depth references, it provides
$1.58\times$--$2.57\times$ $\mathbf{S}_{\mathrm{MAC}}$  with accuracy reductions of
$0.39$--$1.91$ percentage points.
Figure~\ref{fig:tradeoff-vit} complements Table~\ref{tab:tripd_main} with the corresponding ViT threshold-sweep curves. Appendix~\ref{app:additional_results} provides calibration-targeted budget comparisons,
multi-seed results, additional DeiT/Swin trade-off curves, {\revcolor and results on two efficient backbones, DeiT-S and
EfficientFormerV2-L~\citep{Li:2023:RVT}}.

\subsection{Decoupling Analysis in DeclareViT}
\label{sec:gradient_conflict}

We analyze DeclareViT with ViT on ImageNet-1K. During Stage~II, the backbone and final classifier remain frozen.
Trainable parameters shared across exits are therefore confined
to exit-side modules, including the segment-shared cross-attention
projections and the shared exit-side FFN.
We examine the remaining inter-exit gradient interactions on the
cross-attention projections shared across reader blocks~1--2,
using the diagnostic defined in Section~\ref{sec:role_coupling}.

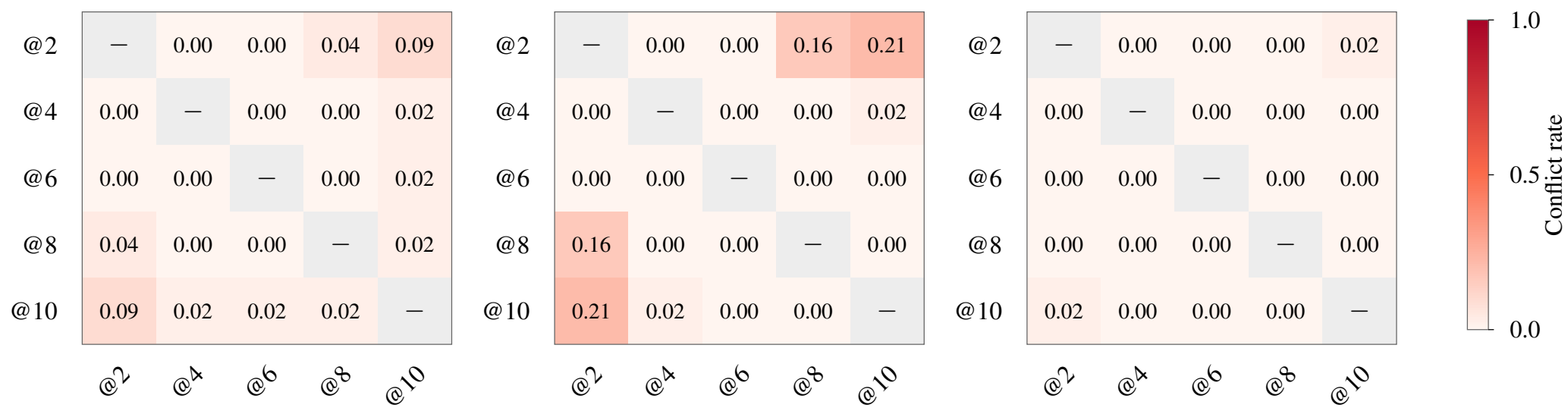
\begin{figure}[t]
\centering
\resizebox{\linewidth}{!}{%
    \input{artwork/gradient_conflict/declarevit_pairwise_conflict}%
}
\caption{Pairwise gradient conflict rates between DeclareViT exit
objectives on the cross-attention projections shared across reader
blocks~1--2 during Stage~II.
@2--@10 denote exit depths.}
\label{fig:declarevit_branch_conflict}
\end{figure}

Figure~\ref{fig:declarevit_branch_conflict} shows that inter-exit
conflict rates are already low during epoch~1, ranging from
$0.00$ to $0.09$ in the measured shared attention group.
Although a few exit pairs exhibit higher rates during epoch~10,
including $0.21$ for @2--@10, most pairs remain at or near zero.
During epoch~100, all reported rates range from $0.00$ to $0.02$.
Thus, low conflict rates are already observed early in training.
These measurements characterize residual gradient competition
within the shared exit-side parameters, while the frozen,
read-only design prevents exit learning from altering the
backbone feature-refinement path. {\revcolor They differ from Figure~\ref{fig:sdn_backbone_conflict} in parameter group and training objective, so the two analyses are complementary diagnostics rather than a controlled comparison of conflict reduction.}
Appendix~\ref{app:gradient_conflict} extends the analysis to
additional shared modules, while
Appendix~\ref{app:attention_visualization} provides qualitative
comparisons of [\textsc{cls}] and exit-query attention.
\subsection{Wall-clock speed-up}
\label{sec:wallclock}

\begin{table}[t]
\centering
\caption{
Wall-clock measurement on ImageNet-1K with ViT.
Wall-clock times are reported as the mean $\pm$ standard deviation over three independent runs, and $\mathbf{S}_{\mathrm{wall}}$ is measured relative to the
unbranched ViT on each device. \textbf{Bold} marks the best results for each metric among early-exit methods.
}
\label{tab:baseline_wallclock_imagenet_vit}
\input{tables/Further_Wall-clock_time_report}
\end{table}
Table~\ref{tab:baseline_wallclock_imagenet_vit} reports inference latency
on ImageNet-1K with ViT at a batch size of 1, measured on an NVIDIA
Quadro RTX 6000 GPU and an Intel Xeon Platinum 8480+ CPU using a single CPU thread.
DeclareViT achieves the highest accuracy and the largest wall-clock
speed-up among the evaluated early-exit methods on both GPU and CPU,
consistent with its advantage in MAC-based efficiency. {\revcolor Appendix~\ref{app:wallclock} provides further measurement details and results, including an evaluation of batched GPU inference.}
\subsection{Ablation Study}
\label{sec:ablation}
We ablate DeclareViT with ViT on ImageNet-1K.
Appendix~\ref{app:further_ablation} provides
further studies.
\paragraph{Exit-head design.}
Table~\ref{tab:ablation_exit_head} compares readouts with the same frozen
backbone, exit locations, and training objective. Exit queries improve block-2 accuracy from $29.70\%$ to
$60.10\%$ over [\textsc{cls}]+MLP and outperform Transformer-encoder
heads with 25M fewer parameters.
{\revcolor No variant sends gradients to the backbone and all see fixed features, so these differences are due to the reader architecture alone.}
\begin{table*}[t]
\centering
\caption{
Comparison of early-exit head designs.
All variants use the same frozen backbone, exit locations, and training objective. Per-exit accuracy evaluates each exit independently;
Acc. denotes accuracy under dynamic exiting. \textbf{Bold} and \underline{underlining} mark the best and second-best results.
}
\label{tab:ablation_exit_head}
\input{tables/4_4_1_Exit_Head_Type}
\end{table*}
\paragraph{Query-branch components.}
Table~\ref{tab:ablation_compact}(a) compares three ablated variants with
backbone and final-classifier parameters frozen.
Removing the exit-side FFN reduces accuracy from $80.55\%$ to
$80.41\%$ and $\mathbf{S}_{\mathrm{MAC}}$ from $1.79\times$ to $1.60\times$.
The pooling readout replaces exit queries and cross-attention
with mean pooling over normalized detached [\textsc{cls}] and patch
tokens, reaching $80.53\%$ at $1.48\times$.
The token-injection variant inserts learnable exit tokens into
the backbone sequence and lets them interact with backbone tokens
through standard self-attention; it achieves $79.82\%$ at
$1.46\times$.
These results favor the read-only query design with an exit-side FFN.
\begin{table}[t]
\centering
\caption{
Ablations on ImageNet-1K with ViT:
(a) query-branch components,
(b) query accumulation,
(c) parameter sharing, and
(d) training protocol.
\textbf{Bold} marks the best result within each panel.
}
\label{tab:ablation_compact}
\label{tab:ablation_query_components}
\label{tab:ablation_query_accumulation}
\label{tab:ablation_parameter_sharing}
\label{tab:training_protocol_ablation}
\input{tables/results_ablations_compact}
\end{table}
\paragraph{Query accumulation.}
Compared with reading features only at the designated exit,
progressive accumulation improves accuracy from $80.44\%$ to
$80.55\%$ and $\mathbf S_{\mathrm{MAC}}$ from $1.71\times$ to
$1.79\times$ (Table~\ref{tab:ablation_compact}(b)).
This supports integrating information across backbone depths.
\paragraph{Parameter sharing.}
Sharing attention projections within inter-exit segments and the FFN
across exits reduces parameters from 138M to 107M, with modest changes
in accuracy ($80.79\%$ to $80.55\%$) and $S_{\mathrm{MAC}}$
($1.83\times$ to $1.79\times$; Table~4(c)).
We adopt this compact configuration by default.
Notably, the unshared variant indicates additional performance gains when
the parameter budget is relaxed. {\revcolor For memory-constrained deployments, we also provide a lightweight
configuration in Appendix~\ref{app:lite}.}
\paragraph{Training protocol.}
We compare two joint-training variants initialized from the same backbone
and final classifier. All variants use the same exit-query architecture,
initialization, exit objective, distillation settings, and exit-training
budget, differing only in how the backbone and exit branches are optimized. In \emph{Joint-attached}, exit losses backpropagate
through the intermediate features and update the backbone together with the
main classification objective, yielding $80.12\%$ accuracy at
$1.67\times$. In \emph{Joint-detached}, stop-gradient is applied at the
backbone--exit interface, so exit losses no longer update the backbone,
while the backbone and exits are still optimized simultaneously. This
improves performance to $80.23\%$ at $1.72\times$. {\revcolor Both variants keep training the backbone, so features are
non-stationary in both and this gain comes from gradient isolation alone.} Our two-stage protocol
additionally freezes the trained backbone and final classifier before exit
optimization, exposing the exit modules to a stationary representation
space. It further improves the result to $80.55\%$ at $1.79\times$
(Table~\ref{tab:ablation_compact}(d)). {\revcolor Exit gradients are blocked in both Joint-detached and
two-stage training, so this further gain comes from freezing the backbone.}
These results support gradient isolation and are consistent with additional benefits from training the exits on fixed backbone representations.

%% file: tables/4_1_main_table.tex
\small
\setlength{\tabcolsep}{1.6pt}
\renewcommand{\arraystretch}{1.0}

\begin{tabular*}{\linewidth}{@{\extracolsep{\fill}}llcccccc@{}}
\toprule
\multirow{2}{*}{\textbf{Backbone}}
& \multirow{2}{*}{\textbf{Method}}
& \multicolumn{2}{c}{\textbf{CIFAR-100}}
& \multicolumn{2}{c}{\textbf{ImageNet-1K}}
& \multicolumn{2}{c}{\textbf{Food-101}} \\

\cmidrule(lr){3-4}
\cmidrule(lr){5-6}
\cmidrule(lr){7-8}

& & \textbf{Acc. (\%)} $\uparrow$
  & $\mathbf{S}_{\mathrm{MAC}}\uparrow$
  & \textbf{Acc. (\%)} $\uparrow$
  & $\mathbf{S}_{\mathrm{MAC}}\uparrow$
  & \textbf{Acc. (\%)} $\uparrow$
  & $\mathbf{S}_{\mathrm{MAC}}\uparrow$ \\
\midrule

\multirow{9}{*}{\textbf{ViT}}
& \textcolor{gray}{ViT~\citep{Dosovitskiy:2021:AII}}
& $91.93$ & $1.00\times$
& $81.77$ & $1.00\times$
& $91.04$ & $1.00\times$ \\

& SDN~\citep{Kaya:2019:SDN}
& $88.29$ & $1.89\times$
& $79.20$ & $1.63\times$
& $89.22$ & $2.01\times$ \\

& BYOT~\citep{Zhang:2019:BYO}
& $88.31$ & $1.70\times$
& $79.53$ & $1.64\times$
& $88.73$ & $1.93\times$ \\

& ViT-EE~\citep{Bakhtiarnia:2021:MVT}
& $88.53$ & $1.48\times$
& $78.56$ & $1.36\times$
& $88.72$ & $1.66\times$ \\

& LGViT~\citep{Xu:2023:LDE}
& $88.18$ & $1.75\times$
& $79.21$ & $1.36\times$
& $\underline{89.85}$ & $1.60\times$ \\

& COSEE~\citep{He:2025:CCS}
& $88.48$ & $1.96\times$
& $79.38$ & $1.66\times$
& $89.44$ & $2.07\times$ \\

& DMPO~\citep{Luo:2025:DMO}
& $\underline{89.98}$ & $\underline{2.07\times}$
& $\underline{80.52}$ & $1.68\times$
& $89.62$ & $\underline{2.20\times}$ \\

& EnViT$^{\dagger}$~\citep{Dong:2026:EET}
& $89.23$ & $2.05\times$
& $80.45$ & $\underline{1.72\times}$
& $89.15$ & $\underline{2.20\times}$ \\

& \textbf{Ours}
& $\mathbf{90.02}$ & $\mathbf{2.10\times}$
& $\mathbf{80.55}$ & $\mathbf{1.79\times}$
& $\mathbf{90.31}$ & $\mathbf{2.57\times}$ \\
\midrule

\multirow{8}{*}{\textbf{DeiT}}
& \textcolor{gray}{DeiT}~\citep{Touvron:2021:TDI}
& $89.90$ & $1.00\times$
& $83.38$ & $1.00\times$
& $89.96$ & $1.00\times$ \\

& SDN~\citep{Kaya:2019:SDN}
& $88.53$ & $\underline{1.61\times}$
& $78.50$ & $1.57\times$
& $\underline{89.01}$ & $1.68\times$ \\

& BYOT~\citep{Zhang:2019:BYO}
& $87.06$ & $1.51\times$
& $78.03$ & $1.54\times$
& $88.35$ & $\underline{1.74\times}$ \\

& ViT-EE~\citep{Bakhtiarnia:2021:MVT}
& $88.19$ & $1.24\times$
& $79.13$ & $1.20\times$
& $88.53$ & $1.42\times$ \\

& LGViT~\citep{Xu:2023:LDE}
& $88.46$ & $1.40\times$
& $78.85$ & $1.36\times$
& $88.77$ & $1.51\times$ \\

& COSEE~\citep{He:2025:CCS}
& $88.61$ & $1.55\times$
& $78.70$ & $\underline{1.58\times}$
& $88.80$ & $1.70\times$ \\

& DMPO~\citep{Luo:2025:DMO}
& $\underline{88.87}$ & $1.55\times$
& $\underline{80.99}$ & $1.35\times$
& $88.89$ & $1.67\times$ \\

& \textbf{Ours}
& $\mathbf{88.95}$ & $\mathbf{1.69\times}$
& $\mathbf{81.96}$ & $\mathbf{1.59\times}$
& $\mathbf{89.57}$ & $\mathbf{2.15\times}$ \\
\midrule

\multirow{8}{*}{\textbf{Swin}}
& \textcolor{gray}{Swin}~\citep{Liu:2021:STH}
& $92.47$ & $1.00\times$
& $83.59$ & $1.00\times$
& $93.68$ & $1.00\times$ \\

& SDN~\citep{Kaya:2019:SDN}
& $\underline{90.65}$ & $1.23\times$
& $79.31$ & $\underline{1.57\times}$
& $91.07$ & $1.72\times$ \\

& BYOT~\citep{Zhang:2019:BYO}
& $87.39$ & $1.20\times$
& $78.40$ & $1.56\times$
& $89.83$ & $1.50\times$ \\

& ViT-EE~\citep{Bakhtiarnia:2021:MVT}
& $90.61$ & $1.42\times$
& $80.51$ & $1.52\times$
& $90.42$ & $2.07\times$ \\

& LGViT~\citep{Xu:2023:LDE}
& $90.55$ & $1.38\times$
& $81.23$ & $1.50\times$
& $\underline{91.75}$ & $1.84\times$ \\

& COSEE~\citep{He:2025:CCS}
& $88.25$ & $1.35\times$
& $79.66$ & $\underline{1.57\times}$
& $90.04$ & $1.67\times$ \\

& DMPO~\citep{Luo:2025:DMO}
& $90.62$ & $\underline{1.62\times}$
& $\underline{81.59}$ & $1.44\times$
& $91.39$ & $\underline{2.20\times}$ \\

& \textbf{Ours}
& $\mathbf{90.81}$ & $\mathbf{1.73\times}$
& $\mathbf{82.71}$ & $\mathbf{1.58\times}$
& $\mathbf{92.37}$ & $\mathbf{2.49\times}$ \\

\bottomrule
\end{tabular*}

%% file: artwork/Curve/cifar_vit.tex
\begin{subfigure}[t]{0.332\textwidth}
\centering
\begin{tikzpicture}
\begin{axis}[
    tradeoff,
    xlabel style={font=\footnotesize,yshift=1pt,name=vitCifarXLabel},
    tradeofflegend,
    legend to name=legend-vit,
    xmin=0.9,
    xmax=2.5,
    ymin=80,
    ymax=92,
    xtick={1.0,1.5,2.0,2.5},
    ytick={80,82,84,86,88,90,92}
]

\addplot[ours] coordinates {
    (1.121688,91.98)
    (1.371600,91.88)
    (1.646650,91.57)
    (1.887822,91.13)
    (2.098700,90.02)
    (2.435329,88.03)
};
\addlegendentry{Ours}

\addplot[envit] coordinates {
    (1.76,90.01)
    (1.87,89.69)
    (1.98,89.39)
    (2.05,89.23)
    (2.11,89.04)
    (2.18,88.72)
    (2.30,88.12)
};
\addlegendentry{EnViT}

\addplot[vitee] coordinates {
    (1.000317,91.60)
    (1.196587,91.01)
    (1.476701,88.53)
    (1.597333,86.99)
    (1.768984,84.63)
    (1.992269,81.14)
};
\addlegendentry{ViT-EE}

\addplot[byot] coordinates {
    (1.137672,90.66)
    (1.360565,90.47)
    (1.695973,88.31)
    (1.833436,87.03)
    (2.084314,83.94)
    (2.233626,81.83)
};
\addlegendentry{BYOT}

\addplot[sdn] coordinates {
    (1.177279,91.91)
    (1.432530,91.58)
    (1.687774,90.18)
    (1.885111,88.29)
    (2.178282,84.09)
    (2.363147,81.45)
};
\addlegendentry{SDN}

\addplot[lgvit] coordinates {
    (1.343383,92.00)
    (1.525318,90.92)
    (1.754699,88.18)
    (1.878671,86.60)
    (2.072687,83.12)
    (2.180796,81.00)
};
\addlegendentry{LGViT}

\addplot[dmpo] coordinates {
    (1.227661,91.91)
    (1.469679,91.84)
    (1.719713,91.42)
    (2.067767,89.98)
    (2.227005,88.90)
    (2.466448,86.64)
};
\addlegendentry{DMPO}

\addplot[cosee] coordinates {
    (1.157248,91.71)
    (1.438466,91.50)
    (1.678300,90.72)
    (1.960754,88.48)
    (2.184427,85.81)
    (2.455975,82.26)
};
\addlegendentry{COSEE}

\addplot[backbone] coordinates {
    (1.00,91.93)
};
\addlegendentry{ViT}

\end{axis}
\node[anchor=north,inner sep=0pt,outer sep=0pt,font=\small]
    at ([yshift=-1mm]vitCifarXLabel.south) {(a) CIFAR-100};
\end{tikzpicture}

\label{fig:tradeoff-vit-cifar100}
\end{subfigure}

%% file: artwork/Curve/imagenet_vit.tex
\begin{subfigure}[t]{0.332\textwidth}
\centering
\begin{tikzpicture}
\begin{axis}[
    tradeoff,
    xlabel style={font=\footnotesize,yshift=1pt,name=vitImageNetXLabel},
    xmin=0.9,
    xmax=2.5,
    ymin=71,
    ymax=83,
    xtick={1.0,1.5,2.0,2.5},
    ytick={71,73,75,77,79,81,83}
]

\addplot[ours] coordinates {
    (1.250426,82.02)
    (1.493825,81.69)
    (1.791913,80.55)
    (1.940907,79.90)
    (2.242030,77.93)
    (2.449793,76.53)
};

\addplot[envit] coordinates {
    (1.40,80.92)
    (1.55,80.72)
    (1.72,80.45)
    (1.92,79.48)
    (2.02,78.19)
    (2.29,75.87)
};

\addplot[vitee] coordinates {
    (1.003503,81.83)
    (1.156298,80.94)
    (1.363610,78.56)
    (1.455376,77.27)
    (1.640774,74.33)
    (1.776143,72.41)
};

\addplot[byot] coordinates {
    (1.467637,80.03)
    (1.636685,79.53)
    (1.749322,78.85)
    (2.015358,76.63)
    (2.155453,75.20)
    (2.356160,72.89)
};

\addplot[sdn] coordinates {
    (1.234977,81.42)
    (1.412766,80.95)
    (1.583257,79.67)
    (1.682441,78.54)
    (1.916641,75.14)
    (2.102175,72.07)
};

\addplot[dmpo] coordinates {
    (1.409659,81.98)
    (1.565007,81.27)
    (1.719627,80.20)
    (1.866673,79.04)
    (2.106515,76.81)
    (2.417180,73.44)
};

\addplot[cosee] coordinates {
    (1.214832,81.44)
    (1.398501,81.17)
    (1.592326,80.01)
    (1.702564,78.95)
    (1.978991,75.45)
    (2.158519,72.79)
};

\addplot[lgvit] coordinates {
    (1.172354,81.94)
    (1.234959,81.21)
    (1.288430,80.41)
    (1.405456,78.11)
    (1.523861,75.60)
    (1.642638,72.52)
};

\addplot[backbone] coordinates {
    (1.00,81.77)
};

\end{axis}
\node[anchor=north,inner sep=0pt,outer sep=0pt,font=\small]
    at ([yshift=-1mm]vitImageNetXLabel.south) {(b) ImageNet-1K};
\end{tikzpicture}

\label{fig:tradeoff-vit-imagenet}
\end{subfigure}

%% file: artwork/Curve/food_vit.tex
\begin{subfigure}[t]{0.332\textwidth}
\centering
\begin{tikzpicture}
\begin{axis}[
    tradeoff,
    xlabel style={font=\footnotesize,yshift=1pt,name=vitFoodXLabel},
    xmin=0.9,
    xmax=3.0,
    ymin=80,
    ymax=92,
    xtick={1.0,1.5,2.0,2.5,3.0},
    ytick={80,82,84,86,88,90,92}
]

\addplot[ours] coordinates {
    (2.030426,91.47)
    (2.220022,91.17)
    (2.382322,90.91)
    (2.573527,90.31)
    (2.796736,89.50)
    (2.924627,88.95)
};

\addplot[envit] coordinates {
    (1.52,90.38)
    (1.93,89.98)
    (2.10,89.44)
    (2.20,89.15)
    (2.41,88.18)
    (2.78,87.65)
};

\addplot[vitee] coordinates {
    (1.155000,90.66)
    (1.417380,90.16)
    (1.663367,88.72)
    (1.991896,85.64)
    (2.265775,82.80)
    (2.373093,81.51)
};

\addplot[byot] coordinates {
    (1.274120,90.36)
    (1.602488,90.04)
    (1.934970,88.73)
    (2.294143,86.23)
    (2.638607,83.18)
    (2.822942,81.36)
};

\addplot[sdn] coordinates {
    (1.409525,90.93)
    (1.709477,90.67)
    (2.014595,89.22)
    (2.336999,86.63)
    (2.622725,83.75)
    (2.825348,81.37)
};

\addplot[lgvit] coordinates {
    (1.335081,90.89)
    (1.598092,89.85)
    (1.811471,88.05)
    (2.024731,86.02)
    (2.237608,83.71)
    (2.440838,81.20)
};

\addplot[dmpo] coordinates {
    (1.347777,90.77)
    (1.663991,90.60)
    (1.991771,90.18)
    (2.202362,89.62)
    (2.629357,87.48)
    (2.921466,85.47)
};

\addplot[cosee] coordinates {
    (1.682637,90.56)
    (1.937601,90.03)
    (2.070700,89.44)
    (2.407922,87.31)
    (2.735296,84.41)
    (2.951899,82.18)
};

\addplot[backbone] coordinates {
    (1.00,91.04)
};

\end{axis}
\node[anchor=north,inner sep=0pt,outer sep=0pt,font=\small]
    at ([yshift=-1mm]vitFoodXLabel.south) {(c) Food-101};
\end{tikzpicture}

\label{fig:tradeoff-vit-food101}
\end{subfigure}

%% file: artwork/gradient_conflict/declarevit_pairwise_conflict.tex
\begingroup
\definecolor{GCRateLow}{RGB}{255,245,240}
\definecolor{GCRateMid}{RGB}{251,106,74}
\definecolor{GCRateHigh}{RGB}{165,0,38}
\begin{tikzpicture}[x=1cm,y=1cm]
\path[use as bounding box] (-0.08,-0.78) rectangle (14.05,3.35);
\fill[black!7] (0.66000,2.52180) rectangle ++(0.66800,0.60120);
\node[font=\fontsize{5.7}{6.7}\selectfont] at (0.99400,2.82240) {$-$};
\fill[GCRateMid!0.000!GCRateLow] (1.32800,2.52180) rectangle ++(0.66800,0.60120);
\node[font=\fontsize{5.7}{6.7}\selectfont] at (1.66200,2.82240) {0.00};
\fill[GCRateMid!0.000!GCRateLow] (1.99600,2.52180) rectangle ++(0.66800,0.60120);
\node[font=\fontsize{5.7}{6.7}\selectfont] at (2.33000,2.82240) {0.00};
\fill[GCRateMid!8.000!GCRateLow] (2.66400,2.52180) rectangle ++(0.66800,0.60120);
\node[font=\fontsize{5.7}{6.7}\selectfont] at (2.99800,2.82240) {0.04};
\fill[GCRateMid!18.000!GCRateLow] (3.33200,2.52180) rectangle ++(0.66800,0.60120);
\node[font=\fontsize{5.7}{6.7}\selectfont] at (3.66600,2.82240) {0.09};
\fill[GCRateMid!0.000!GCRateLow] (0.66000,1.92060) rectangle ++(0.66800,0.60120);
\node[font=\fontsize{5.7}{6.7}\selectfont] at (0.99400,2.22120) {0.00};
\fill[black!7] (1.32800,1.92060) rectangle ++(0.66800,0.60120);
\node[font=\fontsize{5.7}{6.7}\selectfont] at (1.66200,2.22120) {$-$};
\fill[GCRateMid!0.000!GCRateLow] (1.99600,1.92060) rectangle ++(0.66800,0.60120);
\node[font=\fontsize{5.7}{6.7}\selectfont] at (2.33000,2.22120) {0.00};
\fill[GCRateMid!0.000!GCRateLow] (2.66400,1.92060) rectangle ++(0.66800,0.60120);
\node[font=\fontsize{5.7}{6.7}\selectfont] at (2.99800,2.22120) {0.00};
\fill[GCRateMid!4.000!GCRateLow] (3.33200,1.92060) rectangle ++(0.66800,0.60120);
\node[font=\fontsize{5.7}{6.7}\selectfont] at (3.66600,2.22120) {0.02};
\fill[GCRateMid!0.000!GCRateLow] (0.66000,1.31940) rectangle ++(0.66800,0.60120);
\node[font=\fontsize{5.7}{6.7}\selectfont] at (0.99400,1.62000) {0.00};
\fill[GCRateMid!0.000!GCRateLow] (1.32800,1.31940) rectangle ++(0.66800,0.60120);
\node[font=\fontsize{5.7}{6.7}\selectfont] at (1.66200,1.62000) {0.00};
\fill[black!7] (1.99600,1.31940) rectangle ++(0.66800,0.60120);
\node[font=\fontsize{5.7}{6.7}\selectfont] at (2.33000,1.62000) {$-$};
\fill[GCRateMid!0.000!GCRateLow] (2.66400,1.31940) rectangle ++(0.66800,0.60120);
\node[font=\fontsize{5.7}{6.7}\selectfont] at (2.99800,1.62000) {0.00};
\fill[GCRateMid!4.000!GCRateLow] (3.33200,1.31940) rectangle ++(0.66800,0.60120);
\node[font=\fontsize{5.7}{6.7}\selectfont] at (3.66600,1.62000) {0.02};
\fill[GCRateMid!8.000!GCRateLow] (0.66000,0.71820) rectangle ++(0.66800,0.60120);
\node[font=\fontsize{5.7}{6.7}\selectfont] at (0.99400,1.01880) {0.04};
\fill[GCRateMid!0.000!GCRateLow] (1.32800,0.71820) rectangle ++(0.66800,0.60120);
\node[font=\fontsize{5.7}{6.7}\selectfont] at (1.66200,1.01880) {0.00};
\fill[GCRateMid!0.000!GCRateLow] (1.99600,0.71820) rectangle ++(0.66800,0.60120);
\node[font=\fontsize{5.7}{6.7}\selectfont] at (2.33000,1.01880) {0.00};
\fill[black!7] (2.66400,0.71820) rectangle ++(0.66800,0.60120);
\node[font=\fontsize{5.7}{6.7}\selectfont] at (2.99800,1.01880) {$-$};
\fill[GCRateMid!4.000!GCRateLow] (3.33200,0.71820) rectangle ++(0.66800,0.60120);
\node[font=\fontsize{5.7}{6.7}\selectfont] at (3.66600,1.01880) {0.02};
\fill[GCRateMid!18.000!GCRateLow] (0.66000,0.11700) rectangle ++(0.66800,0.60120);
\node[font=\fontsize{5.7}{6.7}\selectfont] at (0.99400,0.41760) {0.09};
\fill[GCRateMid!4.000!GCRateLow] (1.32800,0.11700) rectangle ++(0.66800,0.60120);
\node[font=\fontsize{5.7}{6.7}\selectfont] at (1.66200,0.41760) {0.02};
\fill[GCRateMid!4.000!GCRateLow] (1.99600,0.11700) rectangle ++(0.66800,0.60120);
\node[font=\fontsize{5.7}{6.7}\selectfont] at (2.33000,0.41760) {0.02};
\fill[GCRateMid!4.000!GCRateLow] (2.66400,0.11700) rectangle ++(0.66800,0.60120);
\node[font=\fontsize{5.7}{6.7}\selectfont] at (2.99800,0.41760) {0.02};
\fill[black!7] (3.33200,0.11700) rectangle ++(0.66800,0.60120);
\node[font=\fontsize{5.7}{6.7}\selectfont] at (3.66600,0.41760) {$-$};
\draw[black!60,line width=.25pt] (0.66000,0.11700) rectangle ++(3.34,3.006);
\node[anchor=north east,font=\fontsize{6.5}{7.5}\selectfont,rotate=45] at (0.99400,.12) {@2};
\node[anchor=east,font=\fontsize{6.5}{7.5}\selectfont] at (0.56000,2.82240) {@2};
\node[anchor=north east,font=\fontsize{6.5}{7.5}\selectfont,rotate=45] at (1.66200,.12) {@4};
\node[anchor=east,font=\fontsize{6.5}{7.5}\selectfont] at (0.56000,2.22120) {@4};
\node[anchor=north east,font=\fontsize{6.5}{7.5}\selectfont,rotate=45] at (2.33000,.12) {@6};
\node[anchor=east,font=\fontsize{6.5}{7.5}\selectfont] at (0.56000,1.62000) {@6};
\node[anchor=north east,font=\fontsize{6.5}{7.5}\selectfont,rotate=45] at (2.99800,.12) {@8};
\node[anchor=east,font=\fontsize{6.5}{7.5}\selectfont] at (0.56000,1.01880) {@8};
\node[anchor=north east,font=\fontsize{6.5}{7.5}\selectfont,rotate=45] at (3.66600,.12) {@10};
\node[anchor=east,font=\fontsize{6.5}{7.5}\selectfont] at (0.56000,0.41760) {@10};
\node[font=\small] at (2.33000,-0.7) {(a) Epoch 1};
\fill[black!7] (4.93000,2.52180) rectangle ++(0.66800,0.60120);
\node[font=\fontsize{5.7}{6.7}\selectfont] at (5.26400,2.82240) {$-$};
\fill[GCRateMid!0.000!GCRateLow] (5.59800,2.52180) rectangle ++(0.66800,0.60120);
\node[font=\fontsize{5.7}{6.7}\selectfont] at (5.93200,2.82240) {0.00};
\fill[GCRateMid!0.000!GCRateLow] (6.26600,2.52180) rectangle ++(0.66800,0.60120);
\node[font=\fontsize{5.7}{6.7}\selectfont] at (6.60000,2.82240) {0.00};
\fill[GCRateMid!32.000!GCRateLow] (6.93400,2.52180) rectangle ++(0.66800,0.60120);
\node[font=\fontsize{5.7}{6.7}\selectfont] at (7.26800,2.82240) {0.16};
\fill[GCRateMid!42.000!GCRateLow] (7.60200,2.52180) rectangle ++(0.66800,0.60120);
\node[font=\fontsize{5.7}{6.7}\selectfont] at (7.93600,2.82240) {0.21};
\fill[GCRateMid!0.000!GCRateLow] (4.93000,1.92060) rectangle ++(0.66800,0.60120);
\node[font=\fontsize{5.7}{6.7}\selectfont] at (5.26400,2.22120) {0.00};
\fill[black!7] (5.59800,1.92060) rectangle ++(0.66800,0.60120);
\node[font=\fontsize{5.7}{6.7}\selectfont] at (5.93200,2.22120) {$-$};
\fill[GCRateMid!0.000!GCRateLow] (6.26600,1.92060) rectangle ++(0.66800,0.60120);
\node[font=\fontsize{5.7}{6.7}\selectfont] at (6.60000,2.22120) {0.00};
\fill[GCRateMid!0.000!GCRateLow] (6.93400,1.92060) rectangle ++(0.66800,0.60120);
\node[font=\fontsize{5.7}{6.7}\selectfont] at (7.26800,2.22120) {0.00};
\fill[GCRateMid!4.000!GCRateLow] (7.60200,1.92060) rectangle ++(0.66800,0.60120);
\node[font=\fontsize{5.7}{6.7}\selectfont] at (7.93600,2.22120) {0.02};
\fill[GCRateMid!0.000!GCRateLow] (4.93000,1.31940) rectangle ++(0.66800,0.60120);
\node[font=\fontsize{5.7}{6.7}\selectfont] at (5.26400,1.62000) {0.00};
\fill[GCRateMid!0.000!GCRateLow] (5.59800,1.31940) rectangle ++(0.66800,0.60120);
\node[font=\fontsize{5.7}{6.7}\selectfont] at (5.93200,1.62000) {0.00};
\fill[black!7] (6.26600,1.31940) rectangle ++(0.66800,0.60120);
\node[font=\fontsize{5.7}{6.7}\selectfont] at (6.60000,1.62000) {$-$};
\fill[GCRateMid!0.000!GCRateLow] (6.93400,1.31940) rectangle ++(0.66800,0.60120);
\node[font=\fontsize{5.7}{6.7}\selectfont] at (7.26800,1.62000) {0.00};
\fill[GCRateMid!0.000!GCRateLow] (7.60200,1.31940) rectangle ++(0.66800,0.60120);
\node[font=\fontsize{5.7}{6.7}\selectfont] at (7.93600,1.62000) {0.00};
\fill[GCRateMid!32.000!GCRateLow] (4.93000,0.71820) rectangle ++(0.66800,0.60120);
\node[font=\fontsize{5.7}{6.7}\selectfont] at (5.26400,1.01880) {0.16};
\fill[GCRateMid!0.000!GCRateLow] (5.59800,0.71820) rectangle ++(0.66800,0.60120);
\node[font=\fontsize{5.7}{6.7}\selectfont] at (5.93200,1.01880) {0.00};
\fill[GCRateMid!0.000!GCRateLow] (6.26600,0.71820) rectangle ++(0.66800,0.60120);
\node[font=\fontsize{5.7}{6.7}\selectfont] at (6.60000,1.01880) {0.00};
\fill[black!7] (6.93400,0.71820) rectangle ++(0.66800,0.60120);
\node[font=\fontsize{5.7}{6.7}\selectfont] at (7.26800,1.01880) {$-$};
\fill[GCRateMid!0.000!GCRateLow] (7.60200,0.71820) rectangle ++(0.66800,0.60120);
\node[font=\fontsize{5.7}{6.7}\selectfont] at (7.93600,1.01880) {0.00};
\fill[GCRateMid!42.000!GCRateLow] (4.93000,0.11700) rectangle ++(0.66800,0.60120);
\node[font=\fontsize{5.7}{6.7}\selectfont] at (5.26400,0.41760) {0.21};
\fill[GCRateMid!4.000!GCRateLow] (5.59800,0.11700) rectangle ++(0.66800,0.60120);
\node[font=\fontsize{5.7}{6.7}\selectfont] at (5.93200,0.41760) {0.02};
\fill[GCRateMid!0.000!GCRateLow] (6.26600,0.11700) rectangle ++(0.66800,0.60120);
\node[font=\fontsize{5.7}{6.7}\selectfont] at (6.60000,0.41760) {0.00};
\fill[GCRateMid!0.000!GCRateLow] (6.93400,0.11700) rectangle ++(0.66800,0.60120);
\node[font=\fontsize{5.7}{6.7}\selectfont] at (7.26800,0.41760) {0.00};
\fill[black!7] (7.60200,0.11700) rectangle ++(0.66800,0.60120);
\node[font=\fontsize{5.7}{6.7}\selectfont] at (7.93600,0.41760) {$-$};
\draw[black!60,line width=.25pt] (4.93000,0.11700) rectangle ++(3.34,3.006);
\node[anchor=north east,font=\fontsize{6.5}{7.5}\selectfont,rotate=45] at (5.26400,.12) {@2};
\node[anchor=east,font=\fontsize{6.5}{7.5}\selectfont] at (4.83000,2.82240) {@2};
\node[anchor=north east,font=\fontsize{6.5}{7.5}\selectfont,rotate=45] at (5.93200,.12) {@4};
\node[anchor=east,font=\fontsize{6.5}{7.5}\selectfont] at (4.83000,2.22120) {@4};
\node[anchor=north east,font=\fontsize{6.5}{7.5}\selectfont,rotate=45] at (6.60000,.12) {@6};
\node[anchor=east,font=\fontsize{6.5}{7.5}\selectfont] at (4.83000,1.62000) {@6};
\node[anchor=north east,font=\fontsize{6.5}{7.5}\selectfont,rotate=45] at (7.26800,.12) {@8};
\node[anchor=east,font=\fontsize{6.5}{7.5}\selectfont] at (4.83000,1.01880) {@8};
\node[anchor=north east,font=\fontsize{6.5}{7.5}\selectfont,rotate=45] at (7.93600,.12) {@10};
\node[anchor=east,font=\fontsize{6.5}{7.5}\selectfont] at (4.83000,0.41760) {@10};
\node[font=\small] at (6.60000,-0.7) {(b) Epoch 10};
\fill[black!7] (9.20000,2.52180) rectangle ++(0.66800,0.60120);
\node[font=\fontsize{5.7}{6.7}\selectfont] at (9.53400,2.82240) {$-$};
\fill[GCRateMid!0.000!GCRateLow] (9.86800,2.52180) rectangle ++(0.66800,0.60120);
\node[font=\fontsize{5.7}{6.7}\selectfont] at (10.20200,2.82240) {0.00};
\fill[GCRateMid!0.000!GCRateLow] (10.53600,2.52180) rectangle ++(0.66800,0.60120);
\node[font=\fontsize{5.7}{6.7}\selectfont] at (10.87000,2.82240) {0.00};
\fill[GCRateMid!0.000!GCRateLow] (11.20400,2.52180) rectangle ++(0.66800,0.60120);
\node[font=\fontsize{5.7}{6.7}\selectfont] at (11.53800,2.82240) {0.00};
\fill[GCRateMid!4.000!GCRateLow] (11.87200,2.52180) rectangle ++(0.66800,0.60120);
\node[font=\fontsize{5.7}{6.7}\selectfont] at (12.20600,2.82240) {0.02};
\fill[GCRateMid!0.000!GCRateLow] (9.20000,1.92060) rectangle ++(0.66800,0.60120);
\node[font=\fontsize{5.7}{6.7}\selectfont] at (9.53400,2.22120) {0.00};
\fill[black!7] (9.86800,1.92060) rectangle ++(0.66800,0.60120);
\node[font=\fontsize{5.7}{6.7}\selectfont] at (10.20200,2.22120) {$-$};
\fill[GCRateMid!0.000!GCRateLow] (10.53600,1.92060) rectangle ++(0.66800,0.60120);
\node[font=\fontsize{5.7}{6.7}\selectfont] at (10.87000,2.22120) {0.00};
\fill[GCRateMid!0.000!GCRateLow] (11.20400,1.92060) rectangle ++(0.66800,0.60120);
\node[font=\fontsize{5.7}{6.7}\selectfont] at (11.53800,2.22120) {0.00};
\fill[GCRateMid!0.000!GCRateLow] (11.87200,1.92060) rectangle ++(0.66800,0.60120);
\node[font=\fontsize{5.7}{6.7}\selectfont] at (12.20600,2.22120) {0.00};
\fill[GCRateMid!0.000!GCRateLow] (9.20000,1.31940) rectangle ++(0.66800,0.60120);
\node[font=\fontsize{5.7}{6.7}\selectfont] at (9.53400,1.62000) {0.00};
\fill[GCRateMid!0.000!GCRateLow] (9.86800,1.31940) rectangle ++(0.66800,0.60120);
\node[font=\fontsize{5.7}{6.7}\selectfont] at (10.20200,1.62000) {0.00};
\fill[black!7] (10.53600,1.31940) rectangle ++(0.66800,0.60120);
\node[font=\fontsize{5.7}{6.7}\selectfont] at (10.87000,1.62000) {$-$};
\fill[GCRateMid!0.000!GCRateLow] (11.20400,1.31940) rectangle ++(0.66800,0.60120);
\node[font=\fontsize{5.7}{6.7}\selectfont] at (11.53800,1.62000) {0.00};
\fill[GCRateMid!0.000!GCRateLow] (11.87200,1.31940) rectangle ++(0.66800,0.60120);
\node[font=\fontsize{5.7}{6.7}\selectfont] at (12.20600,1.62000) {0.00};
\fill[GCRateMid!0.000!GCRateLow] (9.20000,0.71820) rectangle ++(0.66800,0.60120);
\node[font=\fontsize{5.7}{6.7}\selectfont] at (9.53400,1.01880) {0.00};
\fill[GCRateMid!0.000!GCRateLow] (9.86800,0.71820) rectangle ++(0.66800,0.60120);
\node[font=\fontsize{5.7}{6.7}\selectfont] at (10.20200,1.01880) {0.00};
\fill[GCRateMid!0.000!GCRateLow] (10.53600,0.71820) rectangle ++(0.66800,0.60120);
\node[font=\fontsize{5.7}{6.7}\selectfont] at (10.87000,1.01880) {0.00};
\fill[black!7] (11.20400,0.71820) rectangle ++(0.66800,0.60120);
\node[font=\fontsize{5.7}{6.7}\selectfont] at (11.53800,1.01880) {$-$};
\fill[GCRateMid!0.000!GCRateLow] (11.87200,0.71820) rectangle ++(0.66800,0.60120);
\node[font=\fontsize{5.7}{6.7}\selectfont] at (12.20600,1.01880) {0.00};
\fill[GCRateMid!4.000!GCRateLow] (9.20000,0.11700) rectangle ++(0.66800,0.60120);
\node[font=\fontsize{5.7}{6.7}\selectfont] at (9.53400,0.41760) {0.02};
\fill[GCRateMid!0.000!GCRateLow] (9.86800,0.11700) rectangle ++(0.66800,0.60120);
\node[font=\fontsize{5.7}{6.7}\selectfont] at (10.20200,0.41760) {0.00};
\fill[GCRateMid!0.000!GCRateLow] (10.53600,0.11700) rectangle ++(0.66800,0.60120);
\node[font=\fontsize{5.7}{6.7}\selectfont] at (10.87000,0.41760) {0.00};
\fill[GCRateMid!0.000!GCRateLow] (11.20400,0.11700) rectangle ++(0.66800,0.60120);
\node[font=\fontsize{5.7}{6.7}\selectfont] at (11.53800,0.41760) {0.00};
\fill[black!7] (11.87200,0.11700) rectangle ++(0.66800,0.60120);
\node[font=\fontsize{5.7}{6.7}\selectfont] at (12.20600,0.41760) {$-$};
\draw[black!60,line width=.25pt] (9.20000,0.11700) rectangle ++(3.34,3.006);
\node[anchor=north east,font=\fontsize{6.5}{7.5}\selectfont,rotate=45] at (9.53400,.12) {@2};
\node[anchor=east,font=\fontsize{6.5}{7.5}\selectfont] at (9.10000,2.82240) {@2};
\node[anchor=north east,font=\fontsize{6.5}{7.5}\selectfont,rotate=45] at (10.20200,.12) {@4};
\node[anchor=east,font=\fontsize{6.5}{7.5}\selectfont] at (9.10000,2.22120) {@4};
\node[anchor=north east,font=\fontsize{6.5}{7.5}\selectfont,rotate=45] at (10.87000,.12) {@6};
\node[anchor=east,font=\fontsize{6.5}{7.5}\selectfont] at (9.10000,1.62000) {@6};
\node[anchor=north east,font=\fontsize{6.5}{7.5}\selectfont,rotate=45] at (11.53800,.12) {@8};
\node[anchor=east,font=\fontsize{6.5}{7.5}\selectfont] at (9.10000,1.01880) {@8};
\node[anchor=north east,font=\fontsize{6.5}{7.5}\selectfont,rotate=45] at (12.20600,.12) {@10};
\node[anchor=east,font=\fontsize{6.5}{7.5}\selectfont] at (9.10000,0.41760) {@10};
\node[font=\small] at (10.87000,-0.7) {(c) Epoch 100};
\shade[bottom color=GCRateLow,top color=GCRateHigh,middle color=GCRateMid] (13.18,.25) rectangle (13.37,3.05);
\draw[black!60,line width=.25pt] (13.18,.25) rectangle (13.37,3.05);
\draw[line width=.25pt] (13.37,0.2500) -- (13.42,0.2500);
\node[anchor=west,font=\fontsize{6.3}{7.3}\selectfont] at (13.44,0.2500) {0.0};
\draw[line width=.25pt] (13.37,1.6500) -- (13.42,1.6500);
\node[anchor=west,font=\fontsize{6.3}{7.3}\selectfont] at (13.44,1.6500) {0.5};
\draw[line width=.25pt] (13.37,3.0500) -- (13.42,3.0500);
\node[anchor=west,font=\fontsize{6.3}{7.3}\selectfont] at (13.44,3.0500) {1.0};
\node[rotate=90,font=\fontsize{6.3}{7.3}\selectfont] at (13.96,1.65) {Conflict rate};
\end{tikzpicture}
\endgroup

%% file: tables/Further_Wall-clock_time_report.tex
\begingroup
\small
\setlength{\tabcolsep}{2.2pt}
\renewcommand{\arraystretch}{1.0}

\begin{tabular*}{\linewidth}{@{\extracolsep{\fill}}lcccccc@{}}
\toprule
\multirow{2}{*}{\textbf{Method}}
& \multirow{2}{*}{\textbf{Acc. (\%) $\uparrow$}}
& \multirow{2}{*}{$\mathbf{S}_{\mathrm{MAC}}\uparrow$}
& \multicolumn{2}{c}{\textbf{GPU}}
& \multicolumn{2}{c}{\textbf{CPU}} \\
\cmidrule(lr){4-5}
\cmidrule(lr){6-7}
&
&
& \shortstack{\textbf{Wall-clock time $\downarrow$}}
& $\mathbf{S}_{\mathrm{wall}}\uparrow$
& \shortstack{\textbf{Wall-clock time $\downarrow$}}
& $\mathbf{S}_{\mathrm{wall}}\uparrow$ \\
\midrule

ViT
& $81.77$
& $1.00\times$
& $14.62 \pm 0.04\,\mathrm{ms}$
& $1.00\times$
& $246.42 \pm 3.85\,\mathrm{ms}$
& $1.00\times$ \\

SDN
& $79.20$
& $1.63\times$
& $9.43 \pm 0.02\,\mathrm{ms}$
& $1.55\times$
& $152.40 \pm 1.47\,\mathrm{ms}$
& $1.62\times$ \\

BYOT
& $79.53$
& $1.64\times$
& $9.41 \pm 0.02\,\mathrm{ms}$
& $1.55\times$
& $152.60 \pm 1.48\,\mathrm{ms}$
& $1.61\times$ \\

ViT-EE
& $78.56$
& $1.36\times$
& $11.19 \pm 0.01\,\mathrm{ms}$
& $1.31\times$
& $184.33 \pm 2.00\,\mathrm{ms}$
& $1.34\times$ \\

LGViT
& $79.21$
& $1.36\times$
& $12.17 \pm 0.00\,\mathrm{ms}$
& $1.20\times$
& $196.38 \pm 1.79\,\mathrm{ms}$
& $1.25\times$ \\

COSEE
& $79.38$
& $1.66\times$
& $ 9.29 \pm 0.01\,\mathrm{ms}$
& $1.57\times$
& $149.37 \pm 0.63\,\mathrm{ms}$
& $1.65\times$ \\

DMPO
& $80.52$
& $1.68\times$
& $9.85 \pm 0.03\,\mathrm{ms}$
& $1.48\times$
& $151.41 \pm 0.32\,\mathrm{ms}$
& $1.63\times$ \\

\textbf{Ours}
& $\mathbf{80.55}$
& $\mathbf{1.79\times}$
& $\mathbf{9.20 \pm 0.00}\,\mathrm{ms}$
& $\mathbf{1.59\times}$
& $\mathbf{144.63 \pm 0.70}\,\mathrm{ms}$
& $\mathbf{1.70\times}$ \\

\bottomrule
\end{tabular*}
\endgroup

%% file: tables/4_4_1_Exit_Head_Type.tex
\small
\setlength{\tabcolsep}{3.15pt}
\renewcommand{\arraystretch}{1.0}

\begin{tabular}{lccccccccc}
\toprule

\multirow{2}{*}{\textbf{Exit Head}}
& \multicolumn{6}{c}{\textbf{Per-Exit Accuracy (\%)}}
& \multirow{2}{*}{\textbf{Acc. (\%)} $\uparrow$}
& \multirow{2}{*}{$\mathbf{S}_{\mathrm{MAC}}\uparrow$}
& \multirow{2}{*}{\textbf{\#Params.} $\downarrow$} \\

\cmidrule(lr){2-7}

& \textbf{@2}
& \textbf{@4}
& \textbf{@6}
& \textbf{@8}
& \textbf{@10}
& \textbf{Final}
& & & \\

\midrule

[\textsc{cls}] + Linear
& $19.05$
& $42.56$
& $53.49$
& $62.57$
& $71.67$
& $81.77$
& $78.40$
& $1.28\times$
& $\mathbf{90}\mathrm{M}$ \\

[\textsc{cls}] + MLP
& $29.70$
& $50.33$
& $60.32$
& $67.12$
& $73.83$
& $81.77$
& $79.01$
& $1.43\times$
& $\underline{96}\mathrm{M}$ \\

Transformer Encoder
& $\underline{60.02}$
& $\underline{69.75}$
& $\underline{75.02}$
& $\underline{78.29}$
& $\underline{80.56}$
& $81.77$
& $\underline{79.66}$
& $\underline{1.56\times}$
& $132\mathrm{M}$ \\

Patch Tokens + CNN
& $51.29$
& $61.92$
& $69.26$
& $74.83$
& $79.60$
& $81.77$
& $79.63$
& $1.48\times$
& $158\mathrm{M}$ \\

\textbf{Ours}
& $\mathbf{60.10}$
& $\mathbf{70.08}$
& $\mathbf{75.71}$
& $\mathbf{79.03}$
& $\mathbf{81.20}$
& $81.77$
& $\mathbf{80.55}$
& $\mathbf{1.79\times}$
& $107\mathrm{M}$ \\

\bottomrule
\end{tabular}

%% file: tables/results_ablations_compact.tex
\begingroup
\small
\renewcommand{\arraystretch}{1.0}

\begin{minipage}[t]{0.49\linewidth}
\vspace{0pt}
\centering
\setlength{\tabcolsep}{2.8pt}

\begin{tabular*}{\linewidth}{@{\extracolsep{\fill}}lrr@{}}
\toprule
\multicolumn{3}{@{}l}{\textbf{(a) Components}} \\
\textbf{Variant}
& \textbf{Acc. (\%)}\,$\uparrow$
& $\mathbf{S}_{\mathrm{MAC}}$\,$\uparrow$ \\
\midrule

w/o FFN
& \multicolumn{1}{c}{$80.41$}
& \multicolumn{1}{c}{$1.60\times$} \\

Pooling readout
& \multicolumn{1}{c}{$80.53$}
& \multicolumn{1}{c}{$1.48\times$} \\

Token injection
& \multicolumn{1}{c}{$79.82$}
& \multicolumn{1}{c}{$1.46\times$} \\

\textbf{Ours}
& \multicolumn{1}{c}{$\mathbf{80.55}$}
& \multicolumn{1}{c}{$\mathbf{1.79\times}$} \\

\midrule
\multicolumn{3}{@{}l}{\textbf{(b) Query accumulation}} \\
\textbf{Readout}
& \textbf{Acc. (\%)}\,$\uparrow$
& $\mathbf{S}_{\mathrm{MAC}}$\,$\uparrow$ \\
\midrule

One-shot
& \multicolumn{1}{c}{$80.44$}
& \multicolumn{1}{c}{$1.71\times$} \\

\textbf{Accumulative (ours)}
& \multicolumn{1}{c}{$\mathbf{80.55}$}
& \multicolumn{1}{c}{$\mathbf{1.79\times}$} \\

\bottomrule
\end{tabular*}
\end{minipage}
\hfill
\begin{minipage}[t]{0.47\linewidth}
\vspace{0pt}
\centering
\setlength{\tabcolsep}{2pt}

\begin{tabular*}{\linewidth}{@{\extracolsep{\fill}}lrrr@{}}
\toprule
\multicolumn{4}{@{}l}{\textbf{(c) Parameter sharing}} \\
\textbf{Sharing}
& \textbf{\#Params.}\,$\downarrow$
& \textbf{Acc. (\%)}\,$\uparrow$
& $\mathbf{S}_{\mathrm{MAC}}$\,$\uparrow$ \\
\midrule

None
& \multicolumn{1}{c}{$138\mathrm{M}$}
& \multicolumn{1}{c}{$\mathbf{80.79}$}
& \multicolumn{1}{c}{$\mathbf{1.83\times}$} \\

Cross-attn.
& \multicolumn{1}{c}{$126\mathrm{M}$}
& \multicolumn{1}{c}{$80.66$}
& \multicolumn{1}{c}{$\mathbf{1.83\times}$} \\

FFN
& \multicolumn{1}{c}{$119\mathrm{M}$}
& \multicolumn{1}{c}{$80.65$}
& \multicolumn{1}{c}{$1.79\times$} \\

\textbf{Both (ours)}
& \multicolumn{1}{c}{$\mathbf{107\mathrm{M}}$}
& \multicolumn{1}{c}{$80.55$}
& \multicolumn{1}{c}{$1.79\times$} \\

\midrule
\multicolumn{4}{@{}l}{\textbf{(d) Training protocol}} \\
\multicolumn{2}{@{}l}{\textbf{Protocol}}
& \textbf{Acc. (\%)}\,$\uparrow$
& $\mathbf{S}_{\mathrm{MAC}}$\,$\uparrow$ \\
\midrule

\multicolumn{2}{@{}l}{Joint-attached}
& \multicolumn{1}{c}{$80.12$}
& \multicolumn{1}{c}{$1.67\times$} \\

\multicolumn{2}{@{}l}{Joint-detached}
& \multicolumn{1}{c}{$80.23$}
& \multicolumn{1}{c}{$1.72\times$} \\

\multicolumn{2}{@{}l}{\textbf{Two-stage (ours)}}
& \multicolumn{1}{c}{$\mathbf{80.55}$}
& \multicolumn{1}{c}{$\mathbf{1.79\times}$} \\

\bottomrule
\end{tabular*}
\end{minipage}

\endgroup

%% file: sections/6_conclusion.tex


\section{Conclusion\label{sec:conclusion}}
DeclareViT decouples early classification from feature refinement
using independent read-only exit queries and two-stage training,
preserving the backbone trajectory and full-depth prediction.
Experiments across three backbones and three datasets demonstrate
improved accuracy--efficiency trade-offs and practical GPU/CPU
acceleration.
These results show that effective early classification can be learned
without reshaping the backbone for deeper predictions.
\paragraph{Limitations.} DeclareViT introduces additional exit-query parameters beyond the backbone. Parameter sharing {\revcolor and a lightweight configuration} substantially {\revcolor reduce} this overhead, although the model remains larger than the unbranched backbone. In addition, DeclareViT currently uses manually specified exit locations. Automatically adapting the number and placement of exits to different backbones and deployment budgets remains a promising direction for future work.



%% file: sections/AI_use_statement.tex
\subsection*{AI use statement}

We used generative AI tools to assist with language editing, proofreading, and
the presentation of the manuscript, as well as with code-level assistance such
as debugging, refactoring, and improving implementation clarity. Generative AI
tools were not used to generate experimental data or results, and all scientific
claims, experimental designs, analyses, and conclusions were reviewed and
validated by the authors. All AI-assisted text was manually inspected and revised
by the authors. AI-assisted code was reviewed, executed, and tested for
correctness of the intended implementation. References, factual statements,
and comparisons to prior work were checked by the authors. We take responsibility for the final content of this work, including
text, claims, code, and other artifacts produced with the aid of generative AI.




\subsection*{Reproducibility statement}

We provide the implementation of DeclareViT and all locally evaluated baselines
in the accompanying code repository \url{https://github.com/TUDa-HWAI/DeclareViT} to support reproducibility. The main paper specifies the
model architecture, training objectives, dynamic inference rule, and evaluation
metrics. Appendix~\ref{app:further_implementation} provides detailed architecture, optimization, dataset split,
preprocessing, and threshold-calibration settings; Appendix~\ref{app:baseline} documents the
implementation provenance and training protocols of the reproduced baselines;
and Appendix~\ref{app:mac_accounting} describes the MAC-counting convention and gives an explicit
analytical example. Additional calibration-targeted budget and multi-seed results are reported
in Appendix~\ref{app:additional_results}, while Appendix~\ref{app:wallclock} specifies the wall-clock evaluation protocol.

The released repository contains the dataset and training configurations,
DeclareViT and baseline model implementations, threshold calibration, dynamic
evaluation, and MAC-accounting code used for the main locally evaluated
accuracy--efficiency experiments. For auxiliary analyses and ablation studies,
the paper and appendices provide the corresponding experimental protocols and
implementation details. 

\subsubsection*{Acknowledgments}
This work is funded by the European Union - European Research Council (ERC) Starting Grant -
Project-ID 101219243. Views and opinions expressed are however those of the author(s) only and do
not necessarily reflect those of the European Union or the European Research Council. Neither the
European Union nor the granting authority can be held responsible for them. 

%% file: sections/supplement.tex
\appendix
\let\origsection\section
\renewcommand{\section}{\FloatBarrier\origsection}

\input{sections/supplement/Further_Implementation_Details}
\section{Additional Accuracy--Efficiency Results}
\label{app:additional_results}

We complement the operating-point results in
Table~\ref{tab:tripd_main} with calibration-targeted budget comparisons,
multi-seed evaluation, additional accuracy--efficiency curves, {\revcolor and
results on two efficient backbones}.

\input{sections/supplement/Fixed_Speedup}

\input{sections/supplement/Multi-Seed_Results_and_Additional_Trade-off_Curves}

\input{sections/supplement/Wall-clock_time}

\input{sections/supplement/Knowledge_Distillation_Ablation}

\section{Baseline Implementations and Reproduction}
\label{app:baseline}

\input{sections/supplement/Baseline_Methods}

\input{sections/supplement/Baseline_Reproduction_and_Provenance}

\input{sections/supplement/MAC_Accounting}

\input{sections/supplement/Gradient_Conflict_Measurement}

\input{sections/supplement/Exit_Queries_and_CLS_Token}

%% file: sections/supplement/Further_Implementation_Details.tex
\section{Further Implementation Details}
\label{app:further_implementation}

\subsection{Datasets and Data Protocol}
\label{app:data_protocol}

\paragraph{Datasets and splits.}
We use the classification datasets introduced in Section~\ref{sec:experimental_setup}:
CIFAR-100 contains low-resolution natural images, ImageNet-1K provides
large-scale visual categories, and Food-101 contains food categories.
Table~\ref{tab:app_data_splits} summarizes their class counts and evaluation
splits. We reserve 10\% of each official training set for checkpoint selection
and threshold calibration. The remaining samples
are used for local training. 
\begin{table}[!htbp]
\centering
\caption{Dataset partitions under the released configuration. Local training
and selection/calibration subsets partition the official training set;
the counts follow the released splitting rules on complete official datasets.}
\label{tab:app_data_splits}
\small
\setlength{\tabcolsep}{5pt}
\renewcommand{\arraystretch}{1.12}
\begin{tabular*}{\linewidth}{@{\extracolsep{\fill}}lrrrl@{}}
\toprule
Dataset & Classes & Local training & Selection/calibration & Final evaluation \\
\midrule
CIFAR-100 & 100 & 45,000 & 5,000 & 10,000 (test) \\
ImageNet-1K & 1,000 & 1,153,050 & 128,117 & 50,000 (validation) \\
Food-101 & 101 & 68,175 & 7,575 & 25,250 (test) \\
\bottomrule
\end{tabular*}
\end{table}

For ImageNet-1K, the holdout is excluded from local exit-branch training;
this does not imply that it was excluded from pretraining of the public
backbone checkpoint. The official validation set is the final evaluation
split, distinct from this training-set holdout.

\paragraph{Preprocessing.}
All models use $224\times224$ inputs. Training applies random resized cropping,
random horizontal flipping, and model-specific normalization.
Evaluation and calibration use the resize, crop, interpolation, and
normalization settings resolved from the selected \texttt{timm} checkpoint.

\subsection{Backbones and Backbone-Specific Exit Branches}
\label{app:backbone_adaptation}

\paragraph{Backbones and initialization.}
We use ViT-B/16, distilled DeiT-B/16, and Swin-B, denoted as ViT, DeiT,
and Swin. ViT and DeiT maintain a constant-width token sequence, whereas
Swin uses hierarchical spatial features. Table~\ref{tab:app_backbone_exits}
collects the backbone and reader settings. The corresponding \texttt{timm}
checkpoint~\citep{rw2019timm} identifiers are:
\begin{flushleft}
\small
ViT: \path{vit_base_patch16_224.orig_in21k_ft_in1k}\\
DeiT: \path{deit_base_distilled_patch16_224.fb_in1k}\\
Swin: \path{swin_base_patch4_window7_224.ms_in1k}.
\end{flushleft}

\begin{table}[!htbp]
\centering
\caption{Backbone and exit-reader settings. Block indices are one-based;
Swin indices count Transformer blocks globally across stages, excluding
patch-merging operations. All variants use one query per exit, eight reader
attention heads, an FFN expansion ratio of four, and a shared exit-side FFN.}
\label{tab:app_backbone_exits}
\small
\setlength{\tabcolsep}{4pt}
\renewcommand{\arraystretch}{1.16}
\begin{tabularx}{\linewidth}{@{}l>{\raggedright\arraybackslash}X>{\raggedright\arraybackslash}X>{\raggedright\arraybackslash}X@{}}
\toprule
Setting & ViT & DeiT & Swin \\
\midrule
Backbone depth & 12 & 12 & $(2,2,18,2)$ by stage \\
Backbone width & 768 & 768 & $(128,256,512,1024)$ by stage \\
Exit blocks & $\{2,4,6,8,10\}$ & $\{2,4,6,8,10\}$ & $\{4,8,12,16,20\}$ \\
Query width $D_q$ & 768 & 768 & 384 \\
Reader memory & Patches + [\textsc{cls}] & Patches + [\textsc{cls}] + distillation token & Spatial tokens \\
Query updates & Every block up to its exit & Every block up to its exit & Once at its exit block \\
Attention sharing & Inter-exit segments & Inter-exit segments & None across readers \\
\bottomrule
\end{tabularx}
\end{table}

\paragraph{ViT implementation.}
ViT follows the readout architecture in the main paper. 

\paragraph{DeiT adaptation.}
The distilled DeiT backbone retains both its class and distillation tokens,
and both are included in the reader memory. The progressive query updates
are otherwise unchanged. The final prediction averages the logits from the
class and distillation heads. This averaged-logit prediction also provides the frozen
teacher for exit training; the early-exit heads are not duplicated for the
two prefix tokens.
\paragraph{Swin adaptation.}
At each configured exit, we flatten the intermediate spatial feature map
into tokens and apply an exit-specific trainable linear projection into
the common query space:
\begin{equation}
\widehat{\mathbf X}_e
=
\operatorname{Proj}_e\!\left(
\operatorname{sg}\!\left(
\operatorname{Flatten}\!\left(\mathbf X^{(d_e)}\right)
\right)
\right)
\in\mathbb R^{D_q\times N_e},
\label{eq:app_swin_memory}
\end{equation}
where $N_e$ is the number of spatial tokens at exit $e$,
$D_q$ is the query dimension, and $\operatorname{sg}$ denotes
stop-gradient.
Detachment is applied before the projection so that its parameters
remain trainable.
Each reader attends to all projected spatial tokens.

Unlike ViT, Swin maintains hierarchical spatial representations.
For Swin-B at $224\times224$ resolution, the first two blocks operate on
$56\times56=3136$ tokens with feature dimension $128$.
If progressive accumulation were applied, all five active queries would
read these high-resolution features. With query dimension $D_q=384$,
the reader cost at each of these blocks is approximately
\[
3136(128)(384)
+2(3136)(384^2)
+2(5)(384^2)
+2(5)(3136)(384)
\approx 1.09~\text{GMAC}.
\]
This already exceeds the cost of a first-stage
Swin-B block. We therefore use non-accumulative, exit-local reading for Swin:
  each exit-specific query is updated only once, after its assigned
  block $d_e\in\{4,8,12,16,20\}$.
{\revcolor\paragraph{EfficientFormerV2 adaptation.} EfficientFormerV2-L~\citep{Li:2023:RVT} is a hierarchical hybrid whose first two stages are purely convolutional. We use the exit-local reader of the Swin variant, with five exits in stages 3 and 4 at global blocks $\{13,17,22,27,31\}$ of 35.}

\paragraph{Exit-branch initialization.}
Query embeddings use a truncated-normal initializer with standard-deviation
parameter $0.02$. Exit modules use their own trainable projections and
normalizations; they do not reuse the backbone's attention parameters.

\subsection{Training and Checkpoint Selection}
\label{app:training_protocol}

\paragraph{Training schedule.}
We use the objectives defined in the main paper, with optimization settings
in Table~\ref{tab:app_training}. When Stage~I is run, the final classifier is
initialized for the target label space. The selected Stage-I backbone and classifier weights initialize Stage II, while the exit branches are initialized afresh.
During Stage~II, the backbone and final classifier are kept in evaluation
mode; the Swin feature projections remain trainable as part of the exit branch.

\begin{table}[!htbp]
\centering
\caption{DeclareViT optimization settings. Panel (a) lists shared settings;
panel (b) lists peak learning rates. CIFAR-100 and Food-101 are grouped
because their schedules are identical. A dash denotes skipping Stage~I and
retaining the public pretrained backbone and original final classifier.}
\label{tab:app_training}
\small
\setlength{\tabcolsep}{4pt}
\renewcommand{\arraystretch}{1.12}
\begin{tabularx}{\linewidth}{@{}lX@{}}
\toprule
\multicolumn{2}{@{}l}{\textbf{(a) Shared settings}}\\
\midrule
Optimizer / weight decay & AdamW / $0.05$ \\
Epochs per executed stage & 100 \\
Global batch size & 1,024 (256 per GPU) \\
Schedule & Cosine decay with 10 epochs of linear warmup \\
Minimum learning rate & $10^{-6}$ \\
Precision & BF16 mixed precision \\
Distillation & $\alpha=0.5$, $T=4$ \\
Exit loss weights & $\lambda_e=1$ \\
Training hardware & 4 NVIDIA H100 80GB GPUs \\
\bottomrule
\end{tabularx}
\par\smallskip
\begin{tabular*}{\linewidth}{@{\extracolsep{\fill}}lcccc@{}}
\toprule
\multicolumn{5}{@{}l}{\textbf{(b) Peak learning rates}}\\
\midrule
Dataset & \multicolumn{3}{c}{Stage I} & Stage II \\
\cmidrule(lr){2-4}
& ViT & DeiT & Swin & All backbones \\
\midrule
CIFAR-100 / Food-101 & $4\times10^{-5}$ & $8\times10^{-5}$ & $4\times10^{-5}$ & $4\times10^{-4}$ \\
ImageNet-1K & -- & -- & -- & $8\times10^{-4}$ \\
\bottomrule
\end{tabular*}
\end{table}

\paragraph{Optimizer and checkpoint selection.}
Stage~I selects the checkpoint with the highest
full-depth validation accuracy; Stage~II selects the checkpoint with the
lowest validation exit-training loss. Both use the calibration
subset defined in Appendix~\ref{app:data_protocol}. The same held-out subset is subsequently reused for
threshold calibration. Stage~II teacher logits
are detached.

\paragraph{Randomness and repeated runs.}
We repeat training with seeds 2024, 2025, and 2026.
For CIFAR-100 and Food-101, each run starts from the Stage-I checkpoint
trained with the corresponding seed; ImageNet-1K runs use the same public
pretrained checkpoint.
Each seed controls the initialization of newly constructed
exit-branch parameters, minibatch ordering, data partition and stochastic
data augmentation.
Thresholds are recalibrated for each trained checkpoint,
and operating points are selected using the same rule
across seeds.
{\revcolor\paragraph{Training cost.} On four H100 GPUs, Stage II takes about 8 hours on ImageNet-1K. On CIFAR-100, Stages I and II each take about 30 minutes, and on Food-101 about 50 minutes. When a trained full-depth model is available, only Stage II is run.}

\subsection{Threshold Calibration and Evaluation}
\label{app:threshold_calibration}

We follow the dynamic-threshold protocol of \citet{Kubaty:2026:RCF}.
For $E$ early exits and one final classifier, the target path proportions are
\begin{equation}
\pi_e(\gamma)
=
\frac{\exp(\gamma(e-1))}
{\sum_{j=1}^{E+1}\exp(\gamma(j-1))},
\qquad e=1,\ldots,E+1.
\label{eq:app_target_exit_shares}
\end{equation}
We use 101 logarithmically spaced magnitudes from $10^{-2}$ to $10^2$
with both signs, excluding zero, for 202 candidates before deduplicating
identical routing outcomes.

Thresholds are calibrated sequentially on the selection/calibration subset.
For $N_{\mathrm{cal}}$ samples, exit $e$ targets
$a_e=\lfloor N_{\mathrm{cal}}\pi_e(\gamma)\rfloor$ samples.
Among samples that have not exited, its threshold is the $a_e$-th smallest
entropy when this count lies strictly between zero and the remaining
sample count. A zero target disables the exit, while a target at least as
large as the remaining count accepts all remaining samples. Empty remaining
sets require no further calibration. Inclusive threshold ties can produce
more exits than targeted; unresolved samples use the final classifier.

Threshold quantiles are fitted using entropy scores without class labels.
For each candidate, the resulting thresholds are applied unchanged to the
final evaluation split. Computation is calculated from the realized routing
counts, not the target path proportions. Budget-specific aggregation is
documented with the additional accuracy--efficiency results.

\paragraph{Representative operating points.} 

For Tables~\ref{tab:tripd_main} and \ref{tab:multi_seed_results}, we construct the accuracy–\(S_{\mathrm{MAC}}\) Pareto frontier from the final-evaluation results of the threshold sweep and manually select one representative Pareto-optimal operating point for each method. Thus, threshold fitting uses only the selection/calibration subset, whereas the representative points reported in Tables 1 and 9 are selected post hoc from the final-evaluation Pareto frontier.

%% file: sections/supplement/Fixed_Speedup.tex
\subsection{Accuracy at Calibration-Targeted Speed-Ups}
\label{app:fix_speedup}

To compare methods under controlled computation targets,
Table~\ref{tab:fixed_speedup_accuracy_cal} reports final-evaluation
performance at target $\mathbf{S}_{\mathrm{MAC}}$ values of
$1.50\times$, $2.00\times$, and $2.50\times$ across all nine
dataset--backbone settings{\revcolor, as the mean and sample standard deviation
over seeds 2024, 2025, and 2026}.
For each method and target $\mathbf{S}_{\mathrm{target}}$, we select on
the calibration subset the candidate that maximizes calibration accuracy
among those satisfying
$\mathbf{S}_{\mathrm{MAC,cal}}\ge
\mathbf{S}_{\mathrm{target}}$.
The selected thresholds are then fixed and applied unchanged to the final
evaluation split.
Each cell reports the resulting final-evaluation accuracy together with the
realized final-evaluation $\mathbf{S}_{\mathrm{MAC}}$ in parentheses.
The realized speed-up may differ from the nominal target because the
calibration constraint need not be tight, the candidate set is discrete,
and routing statistics can differ across splits.
No final-evaluation statistic is used for operating-point selection, and no
interpolation is performed.
EnViT is excluded because only externally reported operating points are
available.

Across the resulting 27 dataset--backbone--budget settings, DeclareViT
achieves the highest {\revcolor mean} final-evaluation accuracy in 25 settings and ranks second
{\revcolor to DMPO} in the remaining two{\revcolor, both on ImageNet-1K (ViT at $1.50\times$ and
Swin at $2.50\times$)}. It ranks first in all nine DeiT settings {\revcolor and in eight of
nine ViT and Swin settings each}.
In a few settings, dynamic early-exit accuracy slightly exceeds the
corresponding full-depth accuracy, consistent with the overthinking
phenomenon~\citep{Kaya:2019:SDN}, where additional computation can change an already correct
intermediate prediction into an incorrect final prediction.
Overall, these results show that DeclareViT's accuracy--efficiency advantage
persists across backbones, datasets, and calibration-targeted compute budgets.


\begin{table}[!htbp]
\centering
\revcolor
\caption{\revcolor Top-1 accuracy (\%) at calibration-targeted $\mathbf{S}_{\mathrm{MAC}}$ on CIFAR-100, ImageNet-1K, and Food-101 using ViT, DeiT, and Swin. Operating points maximize calibration accuracy
subject to
$\mathbf{S}_{\mathrm{MAC,cal}}\ge\mathbf{S}_{\mathrm{target}}$
and are then evaluated unchanged on the final evaluation split. Each cell reports final-evaluation accuracy with the realized $\mathbf{S}_{\mathrm{MAC}}$ in parentheses, both as the mean $\pm$ sample standard deviation over seeds 2024, 2025, and 2026. \textbf{Bold} and \underline{underlining} indicate the best and second-best mean accuracies, respectively. Actual speed-ups may differ from the targets.}
\label{tab:fixed_speedup_accuracy_cal}
\input{tables/results_fixed_speedup_cal}
\end{table}


%% file: tables/results_fixed_speedup_cal.tex
\small
\setlength{\tabcolsep}{2.5pt}
\renewcommand{\arraystretch}{1.12}
\resizebox{\linewidth}{!}{%
\begin{tabular}{@{}l*{9}{c}@{}}
\toprule
\multirow{2}{*}{\textbf{Method}}
& \multicolumn{3}{c}{\textbf{CIFAR-100}} & \multicolumn{3}{c}{\textbf{ImageNet-1K}} & \multicolumn{3}{c}{\textbf{Food-101}} \\
\cmidrule(lr){2-4}\cmidrule(lr){5-7}\cmidrule(lr){8-10}
& $\mathbf{1.50\times}$ & $\mathbf{2.00\times}$ & $\mathbf{2.50\times}$ & $\mathbf{1.50\times}$ & $\mathbf{2.00\times}$ & $\mathbf{2.50\times}$ & $\mathbf{1.50\times}$ & $\mathbf{2.00\times}$ & $\mathbf{2.50\times}$ \\
\midrule
\multicolumn{10}{c}{\textbf{ViT}} \\
\midrule
SDN & \shortstack{\underline{91.43{\scriptsize$\pm$0.23}}\\{\tiny(\ensuremath{1.50{\pm}0.02\times})}} & \shortstack{87.09{\scriptsize$\pm$0.57}\\{\tiny(\ensuremath{1.99{\pm}0.03\times})}} & \shortstack{79.53{\scriptsize$\pm$0.84}\\{\tiny(\ensuremath{2.49{\pm}0.04\times})}} & \shortstack{80.70{\scriptsize$\pm$0.14}\\{\tiny(\ensuremath{1.46{\pm}0.00\times})}} & \shortstack{74.31{\scriptsize$\pm$0.31}\\{\tiny(\ensuremath{1.95{\pm}0.00\times})}} & \shortstack{66.09{\scriptsize$\pm$0.39}\\{\tiny(\ensuremath{2.45{\pm}0.01\times})}} & \shortstack{\underline{90.92{\scriptsize$\pm$0.15}}\\{\tiny(\ensuremath{1.58{\pm}0.03\times})}} & \shortstack{88.42{\scriptsize$\pm$0.23}\\{\tiny(\ensuremath{2.14{\pm}0.02\times})}} & \shortstack{82.75{\scriptsize$\pm$0.21}\\{\tiny(\ensuremath{2.71{\pm}0.02\times})}} \\
BYOT & \shortstack{89.61{\scriptsize$\pm$0.30}\\{\tiny(\ensuremath{1.50{\pm}0.01\times})}} & \shortstack{84.21{\scriptsize$\pm$0.81}\\{\tiny(\ensuremath{1.98{\pm}0.03\times})}} & \shortstack{76.50{\scriptsize$\pm$1.11}\\{\tiny(\ensuremath{2.47{\pm}0.04\times})}} & \shortstack{80.09{\scriptsize$\pm$0.10}\\{\tiny(\ensuremath{1.47{\pm}0.00\times})}} & \shortstack{77.11{\scriptsize$\pm$0.27}\\{\tiny(\ensuremath{1.95{\pm}0.01\times})}} & \shortstack{71.36{\scriptsize$\pm$0.68}\\{\tiny(\ensuremath{2.45{\pm}0.01\times})}} & \shortstack{90.20{\scriptsize$\pm$0.18}\\{\tiny(\ensuremath{1.57{\pm}0.01\times})}} & \shortstack{86.39{\scriptsize$\pm$0.89}\\{\tiny(\ensuremath{2.15{\pm}0.01\times})}} & \shortstack{80.52{\scriptsize$\pm$1.62}\\{\tiny(\ensuremath{2.71{\pm}0.01\times})}} \\
ViT-EE & \shortstack{88.79{\scriptsize$\pm$0.68}\\{\tiny(\ensuremath{1.48{\pm}0.02\times})}} & \shortstack{81.08{\scriptsize$\pm$1.46}\\{\tiny(\ensuremath{2.05{\pm}0.03\times})}} & \shortstack{75.33{\scriptsize$\pm$1.42}\\{\tiny(\ensuremath{2.49{\pm}0.02\times})}} & \shortstack{77.44{\scriptsize$\pm$0.25}\\{\tiny(\ensuremath{1.46{\pm}0.00\times})}} & \shortstack{69.08{\scriptsize$\pm$0.51}\\{\tiny(\ensuremath{2.03{\pm}0.00\times})}} & \shortstack{63.62{\scriptsize$\pm$0.67}\\{\tiny(\ensuremath{2.48{\pm}0.01\times})}} & \shortstack{89.43{\scriptsize$\pm$0.57}\\{\tiny(\ensuremath{1.60{\pm}0.02\times})}} & \shortstack{82.93{\scriptsize$\pm$1.18}\\{\tiny(\ensuremath{2.27{\pm}0.01\times})}} & \shortstack{77.97{\scriptsize$\pm$1.52}\\{\tiny(\ensuremath{2.74{\pm}0.01\times})}} \\
LGViT & \shortstack{91.02{\scriptsize$\pm$0.03}\\{\tiny(\ensuremath{1.51{\pm}0.00\times})}} & \shortstack{83.85{\scriptsize$\pm$0.35}\\{\tiny(\ensuremath{2.02{\pm}0.01\times})}} & \shortstack{74.83{\scriptsize$\pm$0.47}\\{\tiny(\ensuremath{2.49{\pm}0.01\times})}} & \shortstack{75.61{\scriptsize$\pm$1.36}\\{\tiny(\ensuremath{1.47{\pm}0.00\times})}} & \shortstack{61.89{\scriptsize$\pm$2.33}\\{\tiny(\ensuremath{1.99{\pm}0.00\times})}} & \shortstack{51.93{\scriptsize$\pm$2.71}\\{\tiny(\ensuremath{2.49{\pm}0.00\times})}} & \shortstack{90.05{\scriptsize$\pm$0.25}\\{\tiny(\ensuremath{1.59{\pm}0.01\times})}} & \shortstack{85.59{\scriptsize$\pm$0.58}\\{\tiny(\ensuremath{2.14{\pm}0.01\times})}} & \shortstack{79.67{\scriptsize$\pm$0.99}\\{\tiny(\ensuremath{2.58{\pm}0.01\times})}} \\
COSEE & \shortstack{91.28{\scriptsize$\pm$0.06}\\{\tiny(\ensuremath{1.51{\pm}0.02\times})}} & \shortstack{\underline{88.16{\scriptsize$\pm$0.24}}\\{\tiny(\ensuremath{1.99{\pm}0.02\times})}} & \shortstack{81.99{\scriptsize$\pm$0.91}\\{\tiny(\ensuremath{2.48{\pm}0.05\times})}} & \shortstack{80.98{\scriptsize$\pm$0.20}\\{\tiny(\ensuremath{1.47{\pm}0.00\times})}} & \shortstack{75.67{\scriptsize$\pm$0.18}\\{\tiny(\ensuremath{1.95{\pm}0.00\times})}} & \shortstack{68.25{\scriptsize$\pm$0.35}\\{\tiny(\ensuremath{2.45{\pm}0.01\times})}} & \shortstack{90.54{\scriptsize$\pm$0.19}\\{\tiny(\ensuremath{1.57{\pm}0.02\times})}} & \shortstack{88.99{\scriptsize$\pm$0.30}\\{\tiny(\ensuremath{2.15{\pm}0.01\times})}} & \shortstack{84.07{\scriptsize$\pm$0.47}\\{\tiny(\ensuremath{2.72{\pm}0.02\times})}} \\
DMPO & \shortstack{91.30{\scriptsize$\pm$0.95}\\{\tiny(\ensuremath{1.50{\pm}0.02\times})}} & \shortstack{88.06{\scriptsize$\pm$3.91}\\{\tiny(\ensuremath{1.99{\pm}0.03\times})}} & \shortstack{\underline{82.67{\scriptsize$\pm$6.16}}\\{\tiny(\ensuremath{2.54{\pm}0.02\times})}} & \shortstack{\textbf{81.85}{\scriptsize$\pm$\textbf{0.08}}\\{\tiny(\ensuremath{1.46{\pm}0.00\times})}} & \shortstack{\underline{78.39{\scriptsize$\pm$0.12}}\\{\tiny(\ensuremath{1.95{\pm}0.00\times})}} & \shortstack{\underline{72.39{\scriptsize$\pm$0.18}}\\{\tiny(\ensuremath{2.51{\pm}0.00\times})}} & \shortstack{90.87{\scriptsize$\pm$0.19}\\{\tiny(\ensuremath{1.58{\pm}0.05\times})}} & \shortstack{\underline{90.16{\scriptsize$\pm$0.38}}\\{\tiny(\ensuremath{2.12{\pm}0.03\times})}} & \shortstack{\underline{87.32{\scriptsize$\pm$0.40}}\\{\tiny(\ensuremath{2.71{\pm}0.01\times})}} \\
EnViT$^{\dagger}$ & -- & \shortstack{89.23\\{\tiny(\ensuremath{2.05\times})}} & -- & -- & \shortstack{80.45\\{\tiny(\ensuremath{1.72\times})}} & -- & -- & \shortstack{89.15\\{\tiny(\ensuremath{2.20\times})}} & -- \\
\textbf{Ours} & \shortstack{\textbf{91.75}{\scriptsize$\pm$\textbf{0.13}}\\{\tiny(\ensuremath{1.52{\pm}0.10\times})}} & \shortstack{\textbf{90.76}{\scriptsize$\pm$\textbf{0.28}}\\{\tiny(\ensuremath{1.98{\pm}0.05\times})}} & \shortstack{\textbf{87.64}{\scriptsize$\pm$\textbf{0.31}}\\{\tiny(\ensuremath{2.49{\pm}0.05\times})}} & \shortstack{\underline{81.81{\scriptsize$\pm$0.05}}\\{\tiny(\ensuremath{1.44{\pm}0.00\times})}} & \shortstack{\textbf{79.91}{\scriptsize$\pm$\textbf{0.12}}\\{\tiny(\ensuremath{1.94{\pm}0.00\times})}} & \shortstack{\textbf{76.35}{\scriptsize$\pm$\textbf{0.19}}\\{\tiny(\ensuremath{2.45{\pm}0.00\times})}} & \shortstack{\textbf{91.18}{\scriptsize$\pm$\textbf{0.15}}\\{\tiny(\ensuremath{1.73{\pm}0.30\times})}} & \shortstack{\textbf{91.20}{\scriptsize$\pm$\textbf{0.04}}\\{\tiny(\ensuremath{2.17{\pm}0.05\times})}} & \shortstack{\textbf{89.61}{\scriptsize$\pm$\textbf{0.09}}\\{\tiny(\ensuremath{2.76{\pm}0.04\times})}} \\
\midrule
\multicolumn{10}{c}{\textbf{DeiT}} \\
\midrule
SDN & \shortstack{89.03{\scriptsize$\pm$0.23}\\{\tiny(\ensuremath{1.50{\pm}0.01\times})}} & \shortstack{84.08{\scriptsize$\pm$0.13}\\{\tiny(\ensuremath{1.99{\pm}0.02\times})}} & \shortstack{76.99{\scriptsize$\pm$0.25}\\{\tiny(\ensuremath{2.48{\pm}0.03\times})}} & \shortstack{79.19{\scriptsize$\pm$0.12}\\{\tiny(\ensuremath{1.47{\pm}0.00\times})}} & \shortstack{72.83{\scriptsize$\pm$0.36}\\{\tiny(\ensuremath{1.97{\pm}0.00\times})}} & \shortstack{64.94{\scriptsize$\pm$0.42}\\{\tiny(\ensuremath{2.48{\pm}0.01\times})}} & \shortstack{89.37{\scriptsize$\pm$0.18}\\{\tiny(\ensuremath{1.58{\pm}0.03\times})}} & \shortstack{86.27{\scriptsize$\pm$0.12}\\{\tiny(\ensuremath{2.15{\pm}0.01\times})}} & \shortstack{79.88{\scriptsize$\pm$0.23}\\{\tiny(\ensuremath{2.72{\pm}0.00\times})}} \\
BYOT & \shortstack{87.29{\scriptsize$\pm$0.30}\\{\tiny(\ensuremath{1.50{\pm}0.02\times})}} & \shortstack{80.85{\scriptsize$\pm$0.72}\\{\tiny(\ensuremath{1.98{\pm}0.03\times})}} & \shortstack{72.72{\scriptsize$\pm$1.48}\\{\tiny(\ensuremath{2.48{\pm}0.05\times})}} & \shortstack{78.29{\scriptsize$\pm$0.10}\\{\tiny(\ensuremath{1.47{\pm}0.00\times})}} & \shortstack{73.99{\scriptsize$\pm$0.06}\\{\tiny(\ensuremath{1.96{\pm}0.00\times})}} & \shortstack{\underline{67.01{\scriptsize$\pm$0.34}}\\{\tiny(\ensuremath{2.46{\pm}0.00\times})}} & \shortstack{88.44{\scriptsize$\pm$0.44}\\{\tiny(\ensuremath{1.58{\pm}0.01\times})}} & \shortstack{83.79{\scriptsize$\pm$1.79}\\{\tiny(\ensuremath{2.15{\pm}0.02\times})}} & \shortstack{76.72{\scriptsize$\pm$2.86}\\{\tiny(\ensuremath{2.71{\pm}0.02\times})}} \\
ViT-EE & \shortstack{84.85{\scriptsize$\pm$0.33}\\{\tiny(\ensuremath{1.48{\pm}0.02\times})}} & \shortstack{75.07{\scriptsize$\pm$0.42}\\{\tiny(\ensuremath{2.05{\pm}0.02\times})}} & \shortstack{67.96{\scriptsize$\pm$0.63}\\{\tiny(\ensuremath{2.50{\pm}0.00\times})}} & \shortstack{75.13{\scriptsize$\pm$0.17}\\{\tiny(\ensuremath{1.46{\pm}0.00\times})}} & \shortstack{65.92{\scriptsize$\pm$0.36}\\{\tiny(\ensuremath{2.05{\pm}0.01\times})}} & \shortstack{60.24{\scriptsize$\pm$0.44}\\{\tiny(\ensuremath{2.50{\pm}0.01\times})}} & \shortstack{87.12{\scriptsize$\pm$0.20}\\{\tiny(\ensuremath{1.61{\pm}0.01\times})}} & \shortstack{78.78{\scriptsize$\pm$0.42}\\{\tiny(\ensuremath{2.26{\pm}0.01\times})}} & \shortstack{73.05{\scriptsize$\pm$0.48}\\{\tiny(\ensuremath{2.72{\pm}0.02\times})}} \\
LGViT & \shortstack{87.62{\scriptsize$\pm$0.18}\\{\tiny(\ensuremath{1.51{\pm}0.00\times})}} & \shortstack{78.51{\scriptsize$\pm$0.37}\\{\tiny(\ensuremath{2.02{\pm}0.01\times})}} & \shortstack{68.20{\scriptsize$\pm$1.03}\\{\tiny(\ensuremath{2.50{\pm}0.01\times})}} & \shortstack{76.49{\scriptsize$\pm$0.55}\\{\tiny(\ensuremath{1.46{\pm}0.00\times})}} & \shortstack{64.22{\scriptsize$\pm$1.05}\\{\tiny(\ensuremath{2.00{\pm}0.00\times})}} & \shortstack{54.73{\scriptsize$\pm$1.11}\\{\tiny(\ensuremath{2.50{\pm}0.00\times})}} & \shortstack{87.99{\scriptsize$\pm$0.38}\\{\tiny(\ensuremath{1.59{\pm}0.01\times})}} & \shortstack{83.74{\scriptsize$\pm$0.78}\\{\tiny(\ensuremath{2.14{\pm}0.01\times})}} & \shortstack{77.94{\scriptsize$\pm$0.95}\\{\tiny(\ensuremath{2.58{\pm}0.01\times})}} \\
COSEE & \shortstack{88.60{\scriptsize$\pm$0.18}\\{\tiny(\ensuremath{1.51{\pm}0.01\times})}} & \shortstack{\underline{84.68{\scriptsize$\pm$0.21}}\\{\tiny(\ensuremath{1.99{\pm}0.02\times})}} & \shortstack{\underline{78.39{\scriptsize$\pm$0.26}}\\{\tiny(\ensuremath{2.49{\pm}0.02\times})}} & \shortstack{79.39{\scriptsize$\pm$0.04}\\{\tiny(\ensuremath{1.47{\pm}0.00\times})}} & \shortstack{73.40{\scriptsize$\pm$0.19}\\{\tiny(\ensuremath{1.97{\pm}0.00\times})}} & \shortstack{65.38{\scriptsize$\pm$0.17}\\{\tiny(\ensuremath{2.48{\pm}0.00\times})}} & \shortstack{89.07{\scriptsize$\pm$0.17}\\{\tiny(\ensuremath{1.59{\pm}0.05\times})}} & \shortstack{\underline{86.98{\scriptsize$\pm$0.10}}\\{\tiny(\ensuremath{2.15{\pm}0.01\times})}} & \shortstack{\underline{81.48{\scriptsize$\pm$0.10}}\\{\tiny(\ensuremath{2.73{\pm}0.01\times})}} \\
DMPO & \shortstack{\underline{89.12{\scriptsize$\pm$0.44}}\\{\tiny(\ensuremath{1.49{\pm}0.02\times})}} & \shortstack{84.49{\scriptsize$\pm$1.73}\\{\tiny(\ensuremath{1.99{\pm}0.04\times})}} & \shortstack{77.15{\scriptsize$\pm$2.83}\\{\tiny(\ensuremath{2.55{\pm}0.04\times})}} & \shortstack{\underline{80.13{\scriptsize$\pm$0.05}}\\{\tiny(\ensuremath{1.46{\pm}0.00\times})}} & \shortstack{\underline{74.42{\scriptsize$\pm$0.43}}\\{\tiny(\ensuremath{1.96{\pm}0.00\times})}} & \shortstack{66.57{\scriptsize$\pm$0.65}\\{\tiny(\ensuremath{2.53{\pm}0.00\times})}} & \shortstack{\underline{89.63{\scriptsize$\pm$0.26}}\\{\tiny(\ensuremath{1.56{\pm}0.02\times})}} & \shortstack{86.65{\scriptsize$\pm$1.16}\\{\tiny(\ensuremath{2.12{\pm}0.02\times})}} & \shortstack{80.70{\scriptsize$\pm$1.55}\\{\tiny(\ensuremath{2.72{\pm}0.01\times})}} \\
\textbf{Ours} & \shortstack{\textbf{89.49}{\scriptsize$\pm$\textbf{0.07}}\\{\tiny(\ensuremath{1.47{\pm}0.05\times})}} & \shortstack{\textbf{86.59}{\scriptsize$\pm$\textbf{0.10}}\\{\tiny(\ensuremath{1.98{\pm}0.06\times})}} & \shortstack{\textbf{81.36}{\scriptsize$\pm$\textbf{0.12}}\\{\tiny(\ensuremath{2.49{\pm}0.04\times})}} & \shortstack{\textbf{82.74}{\scriptsize$\pm$\textbf{0.08}}\\{\tiny(\ensuremath{1.43{\pm}0.00\times})}} & \shortstack{\textbf{78.81}{\scriptsize$\pm$\textbf{0.06}}\\{\tiny(\ensuremath{1.94{\pm}0.00\times})}} & \shortstack{\textbf{73.40}{\scriptsize$\pm$\textbf{0.05}}\\{\tiny(\ensuremath{2.46{\pm}0.01\times})}} & \shortstack{\textbf{90.19}{\scriptsize$\pm$\textbf{0.13}}\\{\tiny(\ensuremath{1.57{\pm}0.05\times})}} & \shortstack{\textbf{89.45}{\scriptsize$\pm$\textbf{0.15}}\\{\tiny(\ensuremath{2.15{\pm}0.05\times})}} & \shortstack{\textbf{86.16}{\scriptsize$\pm$\textbf{0.05}}\\{\tiny(\ensuremath{2.75{\pm}0.02\times})}} \\
\midrule
\multicolumn{10}{c}{\textbf{Swin}} \\
\midrule
SDN & \shortstack{87.42{\scriptsize$\pm$0.27}\\{\tiny(\ensuremath{1.49{\pm}0.01\times})}} & \shortstack{77.31{\scriptsize$\pm$0.37}\\{\tiny(\ensuremath{1.95{\pm}0.01\times})}} & \shortstack{64.74{\scriptsize$\pm$0.69}\\{\tiny(\ensuremath{2.52{\pm}0.02\times})}} & \shortstack{79.78{\scriptsize$\pm$0.02}\\{\tiny(\ensuremath{1.44{\pm}0.00\times})}} & \shortstack{76.47{\scriptsize$\pm$0.08}\\{\tiny(\ensuremath{1.91{\pm}0.00\times})}} & \shortstack{70.40{\scriptsize$\pm$0.32}\\{\tiny(\ensuremath{2.41{\pm}0.00\times})}} & \shortstack{91.77{\scriptsize$\pm$0.11}\\{\tiny(\ensuremath{1.56{\pm}0.01\times})}} & \shortstack{87.89{\scriptsize$\pm$0.21}\\{\tiny(\ensuremath{2.12{\pm}0.01\times})}} & \shortstack{79.91{\scriptsize$\pm$0.20}\\{\tiny(\ensuremath{2.77{\pm}0.02\times})}} \\
BYOT & \shortstack{83.71{\scriptsize$\pm$0.66}\\{\tiny(\ensuremath{1.50{\pm}0.01\times})}} & \shortstack{73.14{\scriptsize$\pm$1.26}\\{\tiny(\ensuremath{1.97{\pm}0.02\times})}} & \shortstack{60.37{\scriptsize$\pm$1.06}\\{\tiny(\ensuremath{2.55{\pm}0.03\times})}} & \shortstack{78.45{\scriptsize$\pm$0.08}\\{\tiny(\ensuremath{1.47{\pm}0.00\times})}} & \shortstack{76.47{\scriptsize$\pm$0.16}\\{\tiny(\ensuremath{1.94{\pm}0.00\times})}} & \shortstack{72.41{\scriptsize$\pm$0.17}\\{\tiny(\ensuremath{2.43{\pm}0.00\times})}} & \shortstack{89.74{\scriptsize$\pm$0.19}\\{\tiny(\ensuremath{1.58{\pm}0.01\times})}} & \shortstack{85.52{\scriptsize$\pm$0.63}\\{\tiny(\ensuremath{2.13{\pm}0.01\times})}} & \shortstack{78.63{\scriptsize$\pm$0.58}\\{\tiny(\ensuremath{2.77{\pm}0.02\times})}} \\
ViT-EE & \shortstack{90.22{\scriptsize$\pm$0.19}\\{\tiny(\ensuremath{1.47{\pm}0.01\times})}} & \shortstack{83.28{\scriptsize$\pm$0.48}\\{\tiny(\ensuremath{1.97{\pm}0.00\times})}} & \shortstack{73.85{\scriptsize$\pm$0.87}\\{\tiny(\ensuremath{2.56{\pm}0.01\times})}} & \shortstack{80.81{\scriptsize$\pm$0.02}\\{\tiny(\ensuremath{1.43{\pm}0.00\times})}} & \shortstack{77.22{\scriptsize$\pm$0.17}\\{\tiny(\ensuremath{1.93{\pm}0.01\times})}} & \shortstack{71.41{\scriptsize$\pm$0.41}\\{\tiny(\ensuremath{2.53{\pm}0.01\times})}} & \shortstack{92.90{\scriptsize$\pm$0.15}\\{\tiny(\ensuremath{1.57{\pm}0.00\times})}} & \shortstack{90.09{\scriptsize$\pm$0.59}\\{\tiny(\ensuremath{2.15{\pm}0.01\times})}} & \shortstack{84.26{\scriptsize$\pm$1.01}\\{\tiny(\ensuremath{2.79{\pm}0.02\times})}} \\
LGViT & \shortstack{89.98{\scriptsize$\pm$0.08}\\{\tiny(\ensuremath{1.43{\pm}0.00\times})}} & \shortstack{82.39{\scriptsize$\pm$0.35}\\{\tiny(\ensuremath{1.88{\pm}0.01\times})}} & \shortstack{73.57{\scriptsize$\pm$0.43}\\{\tiny(\ensuremath{2.36{\pm}0.03\times})}} & \shortstack{\underline{81.96{\scriptsize$\pm$0.05}}\\{\tiny(\ensuremath{1.42{\pm}0.00\times})}} & \shortstack{76.29{\scriptsize$\pm$0.16}\\{\tiny(\ensuremath{1.89{\pm}0.00\times})}} & \shortstack{68.68{\scriptsize$\pm$0.05}\\{\tiny(\ensuremath{2.41{\pm}0.00\times})}} & \shortstack{\underline{93.37{\scriptsize$\pm$0.06}}\\{\tiny(\ensuremath{1.50{\pm}0.01\times})}} & \shortstack{90.51{\scriptsize$\pm$0.12}\\{\tiny(\ensuremath{1.99{\pm}0.00\times})}} & \shortstack{85.24{\scriptsize$\pm$0.09}\\{\tiny(\ensuremath{2.53{\pm}0.01\times})}} \\
COSEE & \shortstack{86.48{\scriptsize$\pm$0.38}\\{\tiny(\ensuremath{1.50{\pm}0.01\times})}} & \shortstack{78.10{\scriptsize$\pm$0.66}\\{\tiny(\ensuremath{1.97{\pm}0.01\times})}} & \shortstack{66.45{\scriptsize$\pm$0.46}\\{\tiny(\ensuremath{2.53{\pm}0.01\times})}} & \shortstack{79.89{\scriptsize$\pm$0.15}\\{\tiny(\ensuremath{1.44{\pm}0.00\times})}} & \shortstack{77.46{\scriptsize$\pm$0.22}\\{\tiny(\ensuremath{1.91{\pm}0.00\times})}} & \shortstack{72.30{\scriptsize$\pm$0.36}\\{\tiny(\ensuremath{2.41{\pm}0.00\times})}} & \shortstack{90.60{\scriptsize$\pm$0.09}\\{\tiny(\ensuremath{1.57{\pm}0.01\times})}} & \shortstack{86.38{\scriptsize$\pm$0.51}\\{\tiny(\ensuremath{2.13{\pm}0.01\times})}} & \shortstack{78.14{\scriptsize$\pm$0.91}\\{\tiny(\ensuremath{2.78{\pm}0.02\times})}} \\
DMPO & \shortstack{\underline{91.49{\scriptsize$\pm$0.15}}\\{\tiny(\ensuremath{1.47{\pm}0.01\times})}} & \shortstack{\underline{86.99{\scriptsize$\pm$0.45}}\\{\tiny(\ensuremath{1.99{\pm}0.02\times})}} & \shortstack{\underline{80.39{\scriptsize$\pm$0.18}}\\{\tiny(\ensuremath{2.48{\pm}0.02\times})}} & \shortstack{81.62{\scriptsize$\pm$0.04}\\{\tiny(\ensuremath{1.43{\pm}0.00\times})}} & \shortstack{\underline{80.24{\scriptsize$\pm$0.13}}\\{\tiny(\ensuremath{1.94{\pm}0.00\times})}} & \shortstack{\textbf{76.54}{\scriptsize$\pm$\textbf{0.28}}\\{\tiny(\ensuremath{2.44{\pm}0.01\times})}} & \shortstack{92.84{\scriptsize$\pm$0.11}\\{\tiny(\ensuremath{1.54{\pm}0.01\times})}} & \shortstack{\underline{91.60{\scriptsize$\pm$0.06}}\\{\tiny(\ensuremath{2.13{\pm}0.01\times})}} & \shortstack{\underline{88.01{\scriptsize$\pm$0.16}}\\{\tiny(\ensuremath{2.65{\pm}0.01\times})}} \\
\textbf{Ours} & \shortstack{\textbf{92.03}{\scriptsize$\pm$\textbf{0.04}}\\{\tiny(\ensuremath{1.45{\pm}0.00\times})}} & \shortstack{\textbf{89.11}{\scriptsize$\pm$\textbf{0.30}}\\{\tiny(\ensuremath{1.95{\pm}0.02\times})}} & \shortstack{\textbf{84.20}{\scriptsize$\pm$\textbf{0.57}}\\{\tiny(\ensuremath{2.40{\pm}0.04\times})}} & \shortstack{\textbf{83.37}{\scriptsize$\pm$\textbf{0.02}}\\{\tiny(\ensuremath{1.42{\pm}0.00\times})}} & \shortstack{\textbf{80.52}{\scriptsize$\pm$\textbf{0.04}}\\{\tiny(\ensuremath{1.90{\pm}0.00\times})}} & \shortstack{\underline{76.23{\scriptsize$\pm$0.19}}\\{\tiny(\ensuremath{2.37{\pm}0.00\times})}} & \shortstack{\textbf{93.66}{\scriptsize$\pm$\textbf{0.02}}\\{\tiny(\ensuremath{1.53{\pm}0.00\times})}} & \shortstack{\textbf{93.36}{\scriptsize$\pm$\textbf{0.02}}\\{\tiny(\ensuremath{2.07{\pm}0.01\times})}} & \shortstack{\textbf{91.91}{\scriptsize$\pm$\textbf{0.08}}\\{\tiny(\ensuremath{2.60{\pm}0.01\times})}} \\
\bottomrule
\end{tabular}%
}

%% file: sections/supplement/Multi-Seed_Results_and_Additional_Trade-off_Curves.tex
\subsection{Multi-Seed Results {\revcolor at Representative Operating Points}}
\label{app:multi_seed}

Table~\ref{tab:multi_seed_results} reports three-seed results for
DeclareViT and the evaluated baselines across three datasets and three
backbones{\revcolor, at the representative operating points of
Table~\ref{tab:tripd_main}}. DeclareViT maintains strong accuracy--efficiency trade-offs across
different training seeds, with mean $\mathbf{S}_{\mathrm{MAC}}$ ranging
from $1.58\times$ to $2.57\times$. With ViT, it achieves
$90.14 \pm 0.11\%$ at $2.09 \pm 0.01\times$ on CIFAR-100,
$80.56 \pm 0.04\%$ at $1.81 \pm 0.02\times$ on ImageNet-1K,
and $90.28 \pm 0.11\%$ at $2.57 \pm 0.05\times$ on Food-101.
The same stability is observed with DeiT and Swin, indicating that
the accuracy--efficiency gains are robust to training randomness.

\begin{table*}[!htbp]
\centering
\caption{
Multi-seed comparison on CIFAR-100, ImageNet-1K, and Food-101 using
ViT, DeiT, and Swin.
Top-1 accuracy (\%) and $\mathbf{S}_{\mathrm{MAC}}$ are reported as the mean $\pm$ sample
standard deviation over seeds 2024, 2025, and 2026. \textbf{Bold} and \underline{underlining} mark the best and
second-best results, respectively. Full-depth references are excluded from ranking. Full-depth reference rows reproduce the seed-2024 results from Table 1.
}
\label{tab:multi_seed_results}
\input{tables/Multi-seed_results}
\end{table*}

\subsection{Additional Trade-off Curves}
\label{app:additional_curves} 

\begin{figure}[!htbp]
\centering

\makebox[\linewidth][c]{%
    \pgfplotslegendfromname{legend-deit-swin}%
}\par

\makebox[\linewidth][c]{%
    \scalebox{0.93}{%
        \makebox[\linewidth][c]{%
            \input{artwork/Curve/cifar_deit}\unskip
            \hspace{0.0135\linewidth}%
            \input{artwork/Curve/imagenet_deit}\unskip
            \hspace{0.0135\linewidth}%
            \input{artwork/Curve/food_deit}\unskip
        }%
    }%
}\par

\makebox[\linewidth][c]{%
    \scalebox{0.93}{%
        \makebox[\linewidth][c]{%
            \input{artwork/Curve/cifar_swin}\unskip
            \hspace{0.0135\linewidth}%
            \input{artwork/Curve/imagenet_swin}\unskip
            \hspace{0.0135\linewidth}%
            \input{artwork/Curve/food_swin}\unskip
        }%
    }%
}\par

\caption{
Accuracy--$\mathbf{S}_{\mathrm{MAC}}$ trade-offs for DeiT (top)
and Swin (bottom) across three datasets.
}
\label{fig:tradeoff-deit-swin}
\end{figure}
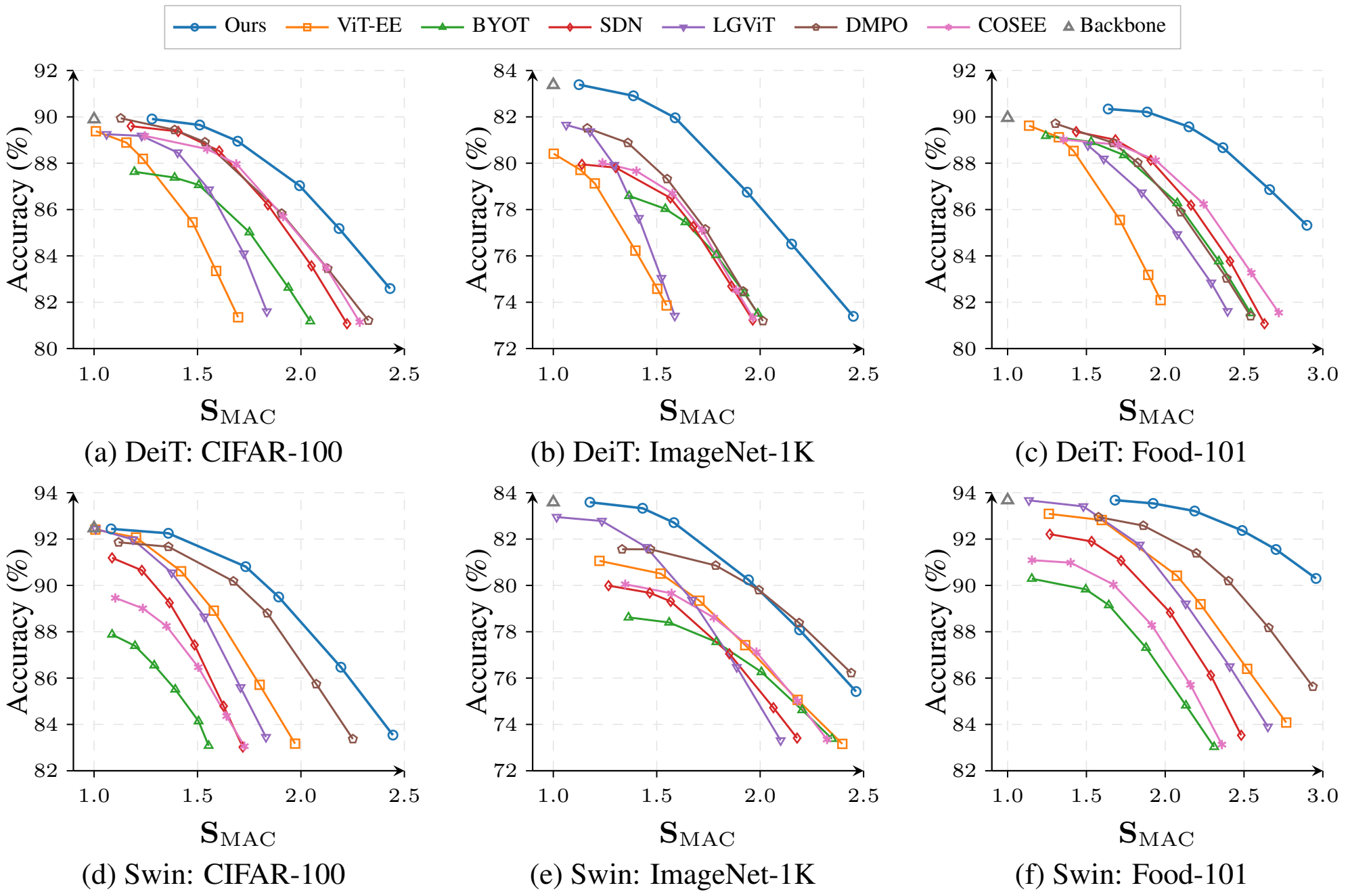

Figure~\ref{fig:tradeoff-deit-swin} presents the complete
accuracy--efficiency trade-offs for DeiT and Swin,
complementing the ViT curves in Figure~\ref{fig:tradeoff-vit}.
Across the six additional dataset--backbone settings, DeclareViT
generally forms the upper accuracy--efficiency frontier over the
evaluated speed-up range.

Together, the calibration-targeted budget comparison, multi-seed evaluation, and complete trade-off curves show that DeclareViT’s gains persist across computation budgets, training seeds, and inference operating points.

\begingroup\revcolor
\subsection{Portability to Efficient Backbones}
\label{app:compact_backbones}
 
\begin{table}[!htbp]
\centering
\revcolor
\caption{\revcolor Backbones on their own and with DeclareViT attached on
ImageNet-1K using seed 2024.
Operating points are selected on the calibration split at the
$1.50\times$ and $2.00\times$ targets.
GMAC is the average cost per image.}
\label{tab:compact_backbones}
\resizebox{\linewidth}{!}{%
\begin{tabular}{lccccccccc}
\toprule
 & & \multicolumn{2}{c}{Backbone only} & \multicolumn{3}{c}{+ DeclareViT, $1.50\times$ target} & \multicolumn{3}{c}{+ DeclareViT, $2.00\times$ target} \\
\cmidrule(lr){3-4}\cmidrule(lr){5-7}\cmidrule(lr){8-10}
Backbone & \#Params & Acc.\ (\%) & GMAC & Acc.\ (\%) & GMAC & $\mathbf{S}_{\mathrm{MAC}}$ & Acc.\ (\%) & GMAC & $\mathbf{S}_{\mathrm{MAC}}$ \\
\midrule
ViT-B & 86.6M & 81.77 & 17.56 & 81.77 & 12.21 & $1.44\times$ & 79.90 & 9.05 & $1.94\times$ \\
DeiT-B & 87.3M & 83.38 & 17.66 & 82.74 & 12.32 & $1.43\times$ & 78.75 & 9.11 & $1.94\times$ \\
Swin-B & 87.8M & 83.59 & 15.43 & 83.35 & 10.88 & $1.42\times$ & 80.57 & 8.13 & $1.90\times$ \\
\midrule
DeiT-S & 22.4M & 81.21 & 4.62 & 80.43 & 3.21 & $1.44\times$ & 76.89 & 2.38 & $1.94\times$ \\
EfficientFormerV2-L & 26.3M & 83.36 & 2.55 & 80.97 & 1.76 & $1.45\times$ & 73.66 & 1.29 & $1.98\times$ \\
\bottomrule
\end{tabular}%
}
\end{table}
 
To examine portability, we attach DeclareViT to two public efficient
backbones, DeiT-S and EfficientFormerV2-L~\citep{Li:2023:RVT}, without
retraining them; only the exit branches are trained.
DeiT-S uses the DeiT reader with exits at blocks $\{2,4,6,8,10\}$, and
EfficientFormerV2-L uses the exit-local reader.
As shown in Table~\ref{tab:compact_backbones}, DeiT-S loses 0.78 and 4.32
points at $1.44\times$ and $1.94\times$, similar to DeiT-B (0.64 and 4.63).
On EfficientFormerV2-L, the loss is larger (2.39 and 9.70 points), since
its two convolutional stages offer no exit.
In all cases the backbone and its full-depth prediction are unchanged.
Efficient static architectures and early exiting are therefore
complementary: DeclareViT adds a run-time adjustable budget to an
existing model.
\endgroup

%% file: tables/Multi-seed_results.tex
\small
\setlength{\tabcolsep}{1.2pt}
\renewcommand{\arraystretch}{1.10}
\begin{tabular*}{\linewidth}{@{\extracolsep{\fill}}lcccccc@{}}
\toprule
\multirow{2}{*}{\textbf{Method}} & \multicolumn{2}{c}{\textbf{CIFAR-100}} & \multicolumn{2}{c}{\textbf{ImageNet-1K}} & \multicolumn{2}{c}{\textbf{Food-101}} \\
\cmidrule(lr){2-3}\cmidrule(lr){4-5}\cmidrule(lr){6-7}
& \textbf{Acc. (\%)} $\uparrow$ & $\mathbf{S}_{\mathrm{MAC}}\uparrow$ & \textbf{Acc. (\%)} $\uparrow$ & $\mathbf{S}_{\mathrm{MAC}}\uparrow$ & \textbf{Acc. (\%)} $\uparrow$ & $\mathbf{S}_{\mathrm{MAC}}\uparrow$ \\
\midrule
\multicolumn{7}{c}{\textbf{ViT}} \\
\midrule
ViT & $91.93$ & $1.00\times$ & $81.77$ & $1.00\times$ & $91.04$ & $1.00\times$ \\
SDN & $88.26 \pm 0.03$ & $1.89 \pm 0.01\times$ & $79.19 \pm 0.01$ & $1.63 \pm 0.01\times$ & $89.19 \pm 0.03$ & $2.03 \pm 0.01\times$ \\
BYOT & $88.31 \pm 0.01$ & $1.64 \pm 0.06\times$ & $79.52 \pm 0.00$ & $1.64 \pm 0.00\times$ & $88.74 \pm 0.01$ & $1.86 \pm 0.07\times$ \\
ViT-EE & $88.65 \pm 0.11$ & $1.50 \pm 0.05\times$ & $78.63 \pm 0.06$ & $1.37 \pm 0.02\times$ & $88.71 \pm 0.08$ & $1.70 \pm 0.10\times$ \\
LGViT & $88.20 \pm 0.18$ & $1.74 \pm 0.02\times$ & $79.20 \pm 0.02$ & $1.32 \pm 0.03\times$ & $\underline{89.83 \pm 0.04}$ & $1.62 \pm 0.04\times$ \\
COSEE & $88.48 \pm 0.10$ & $\underline{1.96 \pm 0.02\times}$ & $79.42 \pm 0.06$ & $1.66 \pm 0.02\times$ & $89.43 \pm 0.04$ & $2.05 \pm 0.08\times$ \\
DMPO & $\underline{89.99 \pm 0.01}$ & $1.88 \pm 0.32\times$ & $\underline{80.54 \pm 0.04}$ & $\underline{1.68 \pm 0.00\times}$ & $89.60 \pm 0.07$ & $\underline{2.30 \pm 0.08\times}$ \\
\textbf{Ours} & $\mathbf{90.14 \pm 0.11}$ & $\mathbf{2.09 \pm 0.01\times}$ & $\mathbf{80.56 \pm 0.04}$ & $\mathbf{1.81 \pm 0.02\times}$ & $\mathbf{90.28 \pm 0.11}$ & $\mathbf{2.57 \pm 0.05\times}$ \\
\midrule
\multicolumn{7}{c}{\textbf{DeiT}} \\
\midrule
DeiT & $89.90$ & $1.00\times$ & $83.38$ & $1.00\times$ & $89.96$ & $1.00\times$ \\
SDN & $88.52 \pm 0.01$ & $\underline{1.59 \pm 0.03\times}$ & $78.51 \pm 0.03$ & $\underline{1.55 \pm 0.01\times}$ & $\underline{89.02 \pm 0.01}$ & $1.71 \pm 0.03\times$ \\
BYOT & $87.07 \pm 0.03$ & $1.52 \pm 0.01\times$ & $78.03 \pm 0.00$ & $1.54 \pm 0.00\times$ & $88.36 \pm 0.01$ & $1.62 \pm 0.12\times$ \\
ViT-EE & $88.19 \pm 0.02$ & $1.23 \pm 0.01\times$ & $79.13 \pm 0.03$ & $1.19 \pm 0.02\times$ & $88.57 \pm 0.04$ & $1.43 \pm 0.01\times$ \\
LGViT & $88.46 \pm 0.04$ & $1.43 \pm 0.02\times$ & $78.79 \pm 0.11$ & $1.36 \pm 0.01\times$ & $88.71 \pm 0.10$ & $1.40 \pm 0.09\times$ \\
COSEE & $88.62 \pm 0.01$ & $1.49 \pm 0.07\times$ & $78.69 \pm 0.02$ & $\mathbf{1.58 \pm 0.01\times}$ & $88.79 \pm 0.02$ & $1.76 \pm 0.05\times$ \\
DMPO & $\underline{88.87 \pm 0.01}$ & $1.54 \pm 0.06\times$ & $\underline{81.00 \pm 0.02}$ & $1.33 \pm 0.02\times$ & $88.86 \pm 0.05$ & $\underline{1.79 \pm 0.13\times}$ \\
\textbf{Ours} & $\mathbf{88.95 \pm 0.02}$ & $\mathbf{1.64 \pm 0.05\times}$ & $\mathbf{81.96 \pm 0.02}$ & $\mathbf{1.58 \pm 0.01\times}$ & $\mathbf{89.54 \pm 0.02}$ & $\mathbf{2.13 \pm 0.02\times}$ \\
\midrule
\multicolumn{7}{c}{\textbf{Swin}} \\
\midrule
Swin & $92.47$ & $1.00\times$ & $83.59$ & $1.00\times$ & $93.68$ & $1.00\times$ \\
SDN & $90.61 \pm 0.04$ & $1.24 \pm 0.01\times$ & $79.31 \pm 0.01$ & $\underline{1.55 \pm 0.01\times}$ & $91.08 \pm 0.02$ & $1.72 \pm 0.02\times$ \\
BYOT & $87.37 \pm 0.09$ & $1.22 \pm 0.04\times$ & $78.40 \pm 0.00$ & $1.50 \pm 0.05\times$ & $89.84 \pm 0.01$ & $1.54 \pm 0.05\times$ \\
ViT-EE & $90.61 \pm 0.01$ & $1.43 \pm 0.02\times$ & $80.49 \pm 0.02$ & $1.52 \pm 0.01\times$ & $90.29 \pm 0.12$ & $2.13 \pm 0.08\times$ \\
LGViT & $90.54 \pm 0.01$ & $1.38 \pm 0.01\times$ & $81.22 \pm 0.02$ & $1.51 \pm 0.00\times$ & $\underline{91.74 \pm 0.07}$ & $1.84 \pm 0.00\times$ \\
COSEE & $88.23 \pm 0.06$ & $1.34 \pm 0.03\times$ & $79.65 \pm 0.00$ & $1.54 \pm 0.06\times$ & $90.04 \pm 0.01$ & $1.69 \pm 0.04\times$ \\
DMPO & $\underline{90.62 \pm 0.06}$ & $\underline{1.63 \pm 0.01\times}$ & $\underline{81.59 \pm 0.00}$ & $1.52 \pm 0.07\times$ & $91.42 \pm 0.10$ & $\underline{2.16 \pm 0.04\times}$ \\
\textbf{Ours} & $\mathbf{90.82 \pm 0.03}$ & $\mathbf{1.72 \pm 0.02\times}$ & $\mathbf{82.71 \pm 0.01}$ & $\mathbf{1.59 \pm 0.01\times}$ & $\mathbf{92.38 \pm 0.11}$ & $\mathbf{2.47 \pm 0.04\times}$ \\
\bottomrule
\end{tabular*}

%% file: artwork/Curve/cifar_deit.tex
\begin{subfigure}[t]{0.332\textwidth}
\centering
\begin{tikzpicture}
\begin{axis}[
    tradeoff,
    width=0.75\linewidth,
    tradeofflegend,
    legend to name=legend-deit-swin,
    xmin=0.9, xmax=2.5,
    ymin=80, ymax=92,
    xtick={1.0,1.5,2.0,2.5},
    ytick={80,82,84,86,88,90,92}
]

\addplot[ours] coordinates {
    (1.279945,89.91)
    (1.510686,89.65)
    (1.694512,88.95)
    (1.993540,87.03)
    (2.184397,85.18)
    (2.430928,82.59)
};
\addlegendentry{Ours}

\addplot[vitee] coordinates {
    (1.008715,89.38)
    (1.156430,88.89)
    (1.235786,88.19)
    (1.475889,85.45)
    (1.590177,83.35)
    (1.696402,81.35)
};
\addlegendentry{ViT-EE}

\addplot[byot] coordinates {
    (1.194940,87.63)
    (1.389380,87.38)
    (1.506414,87.06)
    (1.750951,85.02)
    (1.939307,82.63)
    (2.045104,81.18)
};
\addlegendentry{BYOT}

\addplot[sdn] coordinates {
    (1.178202,89.60)
    (1.406694,89.37)
    (1.605020,88.53)
    (1.840602,86.20)
    (2.051077,83.57)
    (2.222985,81.07)
};
\addlegendentry{SDN}

\addplot[lgvit] coordinates {
    (1.060335,89.25)
    (1.230205,89.18)
    (1.403893,88.46)
    (1.555593,86.86)
    (1.723884,84.10)
    (1.835276,81.60)
};
\addlegendentry{LGViT}

\addplot[dmpo] coordinates {
    (1.129117,89.94)
    (1.389225,89.44)
    (1.536169,88.91)
    (1.906444,85.83)
    (2.129009,83.45)
    (2.326315,81.21)
};
\addlegendentry{DMPO}

\addplot[cosee] coordinates {
    (1.245176,89.18)
    (1.547009,88.61)
    (1.687542,87.97)
    (1.913864,85.70)
    (2.122588,83.49)
    (2.284194,81.14)
};
\addlegendentry{COSEE}

\addplot[backbone] coordinates {
    (1.00,89.90)
};
\addlegendentry{Backbone}

\end{axis}
\end{tikzpicture}

\caption{DeiT: CIFAR-100}
\label{fig:tradeoff-deit-cifar100}
\end{subfigure}

%% file: artwork/Curve/imagenet_deit.tex
\begin{subfigure}[t]{0.332\textwidth}
\centering
\begin{tikzpicture}
\begin{axis}[
    tradeoff,
    width=0.75\linewidth,
    xmin=0.9, xmax=2.5,
    ymin=72, ymax=84,
    xtick={1.0,1.5,2.0,2.5},
    ytick={72,74,76,78,80,82,84}
]

\addplot[ours] coordinates {
    (1.124155,83.39)
    (1.387101,82.91)
    (1.589056,81.96)
    (1.937457,78.75)
    (2.152103,76.51)
    (2.450371,73.39)
};

\addplot[vitee] coordinates {
    (1.001959,80.41)
    (1.131421,79.71)
    (1.200475,79.13)
    (1.396908,76.23)
    (1.502739,74.58)
    (1.547603,73.86)
};

\addplot[byot] coordinates {
    (1.366579,78.59)
    (1.541478,78.03)
    (1.639159,77.46)
    (1.787416,76.04)
    (1.924096,74.39)
    (1.988828,73.51)
};

\addplot[sdn] coordinates {
    (1.138388,79.95)
    (1.301795,79.81)
    (1.566299,78.50)
    (1.676877,77.27)
    (1.861422,74.69)
    (1.964136,73.23)
};

\addplot[dmpo] coordinates {
    (1.164369,81.51)
    (1.360479,80.88)
    (1.550324,79.32)
    (1.734723,77.15)
    (1.918106,74.47)
    (2.013381,73.19)
};

\addplot[cosee] coordinates {
    (1.237580,80.02)
    (1.401966,79.66)
    (1.576212,78.70)
    (1.718854,77.11)
    (1.884490,74.53)
    (1.965312,73.33)
};

\addplot[lgvit] coordinates {
    (1.062377,81.65)
    (1.178115,81.37)
    (1.294617,79.92)
    (1.413380,77.63)
    (1.522480,75.04)
    (1.587288,73.41)
};

\addplot[backbone] coordinates {
    (1.00,83.38)
};

\end{axis}
\end{tikzpicture}

\caption{DeiT: ImageNet-1K}
\label{fig:tradeoff-deit-imagenet}
\end{subfigure}

%% file: artwork/Curve/food_deit.tex
\begin{subfigure}[t]{0.332\textwidth}
\centering
\begin{tikzpicture}
\begin{axis}[
    tradeoff,
    width=0.75\linewidth,
    xmin=0.9, xmax=3.0,
    ymin=80, ymax=92,
    xtick={1.0,1.5,2.0,2.5,3.0},
    ytick={80,82,84,86,88,90,92}
]

\addplot[ours] coordinates {
    (1.638015,90.34)
    (1.885598,90.21)
    (2.150566,89.57)
    (2.366027,88.67)
    (2.662618,86.86)
    (2.899185,85.32)
};

\addplot[vitee] coordinates {
    (1.136323,89.62)
    (1.324777,89.12)
    (1.418352,88.53)
    (1.712058,85.55)
    (1.892472,83.18)
    (1.970955,82.09)
};

\addplot[byot] coordinates {
    (1.242442,89.18)
    (1.530005,88.92)
    (1.738026,88.35)
    (2.076569,86.28)
    (2.340429,83.77)
    (2.542356,81.53)
};

\addplot[sdn] coordinates {
    (1.434886,89.36)
    (1.684071,89.01)
    (1.908068,88.13)
    (2.163664,86.19)
    (2.410766,83.77)
    (2.629917,81.07)
};

\addplot[lgvit] coordinates {
    (1.509547,88.77)
    (1.609963,88.19)
    (1.851402,86.73)
    (2.076401,84.93)
    (2.291763,82.84)
    (2.395989,81.61)
};

\addplot[dmpo] coordinates {
    (1.303496,89.71)
    (1.669064,88.88)
    (1.825155,88.02)
    (2.098856,85.89)
    (2.390352,83.03)
    (2.541774,81.40)
};

\addplot[cosee] coordinates {
    (1.354489,89.02)
    (1.703010,88.80)
    (1.939199,88.13)
    (2.243830,86.24)
    (2.546432,83.27)
    (2.720420,81.55)
};

\addplot[backbone] coordinates {
    (1.00,89.96)
};

\end{axis}
\end{tikzpicture}

\caption{DeiT: Food-101}
\label{fig:tradeoff-deit-food101}
\end{subfigure}

%% file: artwork/Curve/cifar_swin.tex
\begin{subfigure}[t]{0.332\textwidth}
\centering
\begin{tikzpicture}
\begin{axis}[
    tradeoff,
    width=0.75\linewidth,
    xmin=0.9, xmax=2.5,
    ymin=82, ymax=94,
    xtick={1.0,1.5,2.0,2.5},
    ytick={82,84,86,88,90,92,94}
]

\addplot[ours] coordinates {
    (1.082134,92.44)
    (1.359072,92.25)
    (1.732902,90.81)
    (1.892542,89.50)
    (2.193088,86.47)
    (2.444802,83.54)
};

\addplot[vitee] coordinates {
    (1.006147,92.41)
    (1.202757,92.07)
    (1.420641,90.61)
    (1.578770,88.91)
    (1.800367,85.71)
    (1.971611,83.17)
};

\addplot[byot] coordinates {
    (1.086958,87.88)
    (1.197778,87.39)
    (1.290985,86.55)
    (1.392461,85.51)
    (1.505118,84.14)
    (1.553834,83.08)
};

\addplot[sdn] coordinates {
    (1.086922,91.19)
    (1.231015,90.65)
    (1.364626,89.25)
    (1.485496,87.43)
    (1.626279,84.79)
    (1.719168,83.02)
};

\addplot[lgvit] coordinates {
    (1.007876,92.44)
    (1.188191,91.98)
    (1.375472,90.55)
    (1.533375,88.65)
    (1.708034,85.60)
    (1.831024,83.45)
};

\addplot[dmpo] coordinates {
    (1.118488,91.86)
    (1.359575,91.67)
    (1.673491,90.18)
    (1.837716,88.80)
    (2.074223,85.74)
    (2.251170,83.37)
};

\addplot[cosee] coordinates {
    (1.102484,89.46)
    (1.235658,89.01)
    (1.349663,88.25)
    (1.503782,86.47)
    (1.640542,84.36)
    (1.726759,83.06)
};

\addplot[backbone] coordinates {
    (1.00,92.47)
};

\end{axis}
\end{tikzpicture}

\caption{Swin: CIFAR-100}
\label{fig:tradeoff-swin-cifar100}
\end{subfigure}

%% file: artwork/Curve/imagenet_swin.tex
\begin{subfigure}[t]{0.332\textwidth}
\centering
\begin{tikzpicture}
\begin{axis}[
    tradeoff,
    width=0.75\linewidth,
    xmin=0.9, xmax=2.5,
    ymin=72, ymax=84,
    xtick={1.0,1.5,2.0,2.5},
    ytick={72,74,76,78,80,82,84}
]

\addplot[ours] coordinates {
    (1.177017,83.59)
    (1.431659,83.33)
    (1.583017,82.71)
    (1.943758,80.24)
    (2.190518,78.08)
    (2.463533,75.42)
};

\addplot[vitee] coordinates {
    (1.223079,81.06)
    (1.518914,80.51)
    (1.706553,79.34)
    (1.927859,77.42)
    (2.181175,75.06)
    (2.397106,73.16)
};

\addplot[byot] coordinates {
    (1.363589,78.62)
    (1.560063,78.40)
    (1.787926,77.56)
    (2.006189,76.26)
    (2.203278,74.62)
    (2.347187,73.38)
};

\addplot[sdn] coordinates {
    (1.266988,79.99)
    (1.466424,79.68)
    (1.567655,79.31)
    (1.851157,77.05)
    (2.064145,74.72)
    (2.178604,73.41)
};

\addplot[dmpo] coordinates {
    (1.332532,81.56)
    (1.468961,81.56)
    (1.785019,80.86)
    (1.994312,79.80)
    (2.188432,78.38)
    (2.440264,76.22)
};

\addplot[cosee] coordinates {
    (1.347471,80.05)
    (1.571595,79.66)
    (1.775944,78.61)
    (1.982177,77.12)
    (2.179960,75.01)
    (2.323337,73.35)
};

\addplot[lgvit] coordinates {
    (1.016243,82.95)
    (1.234880,82.78)
    (1.449851,81.65)
    (1.670971,79.37)
    (1.886995,76.47)
    (2.099049,73.31)
};

\addplot[backbone] coordinates {
    (1.00,83.59)
};

\end{axis}
\end{tikzpicture}

\caption{Swin: ImageNet-1K}
\label{fig:tradeoff-swin-imagenet}
\end{subfigure}

%% file: artwork/Curve/food_swin.tex
\begin{subfigure}[t]{0.332\textwidth}
\centering
\begin{tikzpicture}
\begin{axis}[
    tradeoff,
    width=0.75\linewidth,
    xmin=0.9, xmax=3.0,
    ymin=82, ymax=94,
    xtick={1.0,1.5,2.0,2.5,3.0},
    ytick={82,84,86,88,90,92,94}
]

\addplot[ours] coordinates {
    (1.681667,93.68)
    (1.923288,93.54)
    (2.186667,93.21)
    (2.488419,92.37)
    (2.703038,91.55)
    (2.956982,90.30)
};

\addplot[vitee] coordinates {
    (1.261718,93.09)
    (1.597210,92.82)
    (2.074149,90.42)
    (2.224034,89.19)
    (2.520019,86.40)
    (2.767955,84.08)
};

\addplot[byot] coordinates {
    (1.153492,90.29)
    (1.495853,89.83)
    (1.642190,89.15)
    (1.878442,87.31)
    (2.131638,84.82)
    (2.310181,83.03)
};

\addplot[sdn] coordinates {
    (1.267534,92.21)
    (1.533012,91.90)
    (1.719918,91.07)
    (2.031470,88.83)
    (2.289184,86.12)
    (2.482493,83.53)
};

\addplot[lgvit] coordinates {
    (1.135383,93.67)
    (1.480621,93.41)
    (1.837767,91.75)
    (2.129699,89.21)
    (2.409136,86.50)
    (2.651434,83.90)
};

\addplot[dmpo] coordinates {
    (1.576344,92.95)
    (1.861227,92.58)
    (2.197302,91.39)
    (2.401542,90.19)
    (2.656626,88.17)
    (2.936947,85.63)
};

\addplot[cosee] coordinates {
    (1.154265,91.09)
    (1.400218,90.98)
    (1.670921,90.04)
    (1.914513,88.29)
    (2.159859,85.71)
    (2.359748,83.13)
};

\addplot[backbone] coordinates {
    (1.00,93.68)
};

\end{axis}
\end{tikzpicture}

\caption{Swin: Food-101}
\label{fig:tradeoff-swin-food101}
\end{subfigure}

%% file: sections/supplement/Wall-clock_time.tex
\section{Wall-Clock Evaluation Protocol and Additional Results}
\label{app:wallclock}

\paragraph{Evaluation protocol.}
For the wall-clock evaluation in Section~\ref{sec:wallclock},
we measure batch-one inference latency on an NVIDIA Quadro RTX 6000 GPU
and an Intel Xeon Platinum 8480+ CPU using a single CPU thread.
GPU inference uses FP16 autocast mixed precision, whereas CPU inference uses FP32. All models are evaluated in \texttt{eval} mode under \texttt{inference\_mode} with PyTorch eager execution. Inputs are preprocessed to $224\times224$ and resident on the target device before timing, excluding data loading, preprocessing, and host-to-device transfer.

For GPU measurements, we synchronize the device immediately before and after each forward pass and measure synchronized wall-clock latency with a host timer. Dynamic inference terminates immediately once an exit satisfies its entropy threshold; otherwise, computation continues to the final classifier. Each process is warmed up on 500 samples and then
evaluated over the complete evaluation set. We repeat the measurement in three fresh processes and report the mean per-image latency and sample standard deviation.

All locally evaluated methods follow the same batch size, precision, warm-up, synchronization, and timing protocol while retaining their method-specific implementations. For DMPO, LoRA weights are merged into the corresponding linear layers before wall-clock evaluation.

\paragraph{Additional Results.}
As shown in Table~\ref{tab:wall_clock_speedup}, DeclareViT achieves
GPU wall-clock speed-ups of $1.86\times$, $1.59\times$, and $2.26\times$, and CPU wall-clock
speed-ups of $1.99\times$, $1.70\times$, and $2.44\times$ on CIFAR-100,
ImageNet-1K, and Food-101, respectively.
Wall-clock speed-ups are generally smaller than the corresponding
MAC-based speed-ups because MAC accounting does not capture runtime
overheads such as routing, synchronization, and kernel launches.

\begin{table*}[!htbp]
\centering
\caption{
Wall-clock evaluation of DeclareViT and the unbranched ViT-B/16 reference.
Wall-clock time is reported as the mean $\pm$ standard deviation over three independent runs at
batch size one, and $\mathbf{S}_{\mathrm{wall}}$ is measured relative
to the corresponding unbranched ViT.
}
\label{tab:wall_clock_speedup}
\input{tables/4_4_Wall-block_time}
\end{table*}

\begingroup\revcolor
\paragraph{Batched inference.}
We further measure GPU inference at batch sizes 4, 8, 16, 32, and 64 for all methods, using the protocol above and the operating points of Table~\ref{tab:baseline_wallclock_imagenet_vit}. For batched inference, every method uses in-batch compaction, where samples that exit are removed from the batch, and cross-batch regrouping, where the remaining samples are merged with those from other batches. We report the wall-clock time per image and $\mathbf{S}_{\mathrm{wall}}$ relative to ViT at the same batch size.
 
As shown in Table~\ref{tab:wallclock_batched}, DeclareViT is $1.48\times$ to $1.71\times$ faster than ViT at the same batch size. It is the fastest early-exit method at batch sizes 32 and 64, and on par with DMPO at batch size 16. At smaller batch sizes it is slower: SDN, BYOT, COSEE, and DMPO are faster at batch size 4, and COSEE and DMPO at batch size 8.
This gap comes from the query branch. The baselines add computation only at their exits, whereas DeclareViT also updates its exit queries after each of the first ten blocks. Each update has a fixed cost per call that does not depend on the batch size. In a small batch, this cost is shared by only a few images; in a larger batch, it is shared by more images. As a result, the speed-up of DeclareViT increases from batch size 4 to 64.
 
\begin{table}[!htbp]
\centering
\revcolor
\caption{\revcolor Batched GPU inference on ImageNet-1K with ViT.
Each batch-size cell reports $\mathbf{S}_{\mathrm{wall}}$ relative to ViT at the
same batch size and, in parentheses, the wall-clock time per image in ms.
\textbf{Bold} marks the best result among early-exit methods.}
\label{tab:wallclock_batched}
\resizebox{\linewidth}{!}{%
\begin{tabular}{lccccccc}
\toprule
Method & Acc.\ (\%) & $\mathbf{S}_{\mathrm{MAC}}$ & $B{=}4$ & $B{=}8$ & $B{=}16$ & $B{=}32$ & $B{=}64$ \\
\midrule
ViT & 81.77 & 1.00$\times$ & 1.00$\times$ (7.66) & 1.00$\times$ (6.91) & 1.00$\times$ (6.26) & 1.00$\times$ (5.93) & 1.00$\times$ (5.78) \\
\midrule
SDN & 79.20 & 1.63$\times$ & 1.55$\times$ (4.95) & 1.57$\times$ (4.40) & 1.58$\times$ (3.95) & 1.59$\times$ (3.73) & 1.60$\times$ (3.62) \\
BYOT & 79.53 & 1.64$\times$ & 1.55$\times$ (4.94) & 1.57$\times$ (4.39) & 1.59$\times$ (3.94) & 1.59$\times$ (3.72) & 1.60$\times$ (3.61) \\
ViT-EE & 78.56 & 1.36$\times$ & 1.31$\times$ (5.85) & 1.32$\times$ (5.21) & 1.33$\times$ (4.69) & 1.34$\times$ (4.44) & 1.34$\times$ (4.30) \\
LGViT & 79.21 & 1.36$\times$ & 1.18$\times$ (6.50) & 1.17$\times$ (5.90) & 1.16$\times$ (5.37) & 1.16$\times$ (5.11) & 1.16$\times$ (4.97) \\
COSEE & 79.38 & 1.66$\times$ & \textbf{1.58$\times$ (4.87)} & 1.60$\times$ (4.32) & 1.61$\times$ (3.88) & 1.62$\times$ (3.67) & 1.62$\times$ (3.56) \\
DMPO & 80.52 & 1.68$\times$ & \textbf{1.58$\times$ (4.87)} & \textbf{1.62$\times$ (4.26)} & \textbf{1.64$\times$ (3.81)} & 1.65$\times$ (3.60) & 1.66$\times$ (3.48) \\
Ours & \textbf{80.55} & \textbf{1.79$\times$} & 1.48$\times$ (5.17) & 1.57$\times$ (4.40) & 1.64$\times$ (3.82) & \textbf{1.68$\times$ (3.53)} & \textbf{1.71$\times$ (3.38)} \\
\bottomrule
\end{tabular}%
}
\end{table}
\endgroup

%% file: tables/4_4_Wall-block_time.tex
\small
\setlength{\tabcolsep}{3pt}
\renewcommand{\arraystretch}{1.12}

\begin{tabular}{llcccccc}
\toprule
\multirow{2}{*}{\textbf{Dataset}}
& \multirow{2}{*}{\textbf{Method}}
& \multirow{2}{*}{\textbf{Acc. (\%) $\uparrow$}}
& \multirow{2}{*}{$\mathbf{S}_{\mathrm{MAC}}\uparrow$}
& \multicolumn{2}{c}{\textbf{GPU}}
& \multicolumn{2}{c}{\textbf{CPU}} \\
\cmidrule(lr){5-6}
\cmidrule(lr){7-8}
&
&
&
& \textbf{Wall-clock time $\downarrow$}
& $\mathbf{S}_{\mathrm{wall}}\uparrow$
& \textbf{Wall-clock time $\downarrow$}
& $\mathbf{S}_{\mathrm{wall}}\uparrow$ \\
\midrule

\multirow{2}{*}{CIFAR-100}
& ViT
& $91.93$
& $1.00\times$
& $14.56 \pm 0.01\,\mathrm{ms}$
& $1.00\times$
& $245.28 \pm 1.27\,\mathrm{ms}$
& $1.00\times$ \\

& \textbf{Ours}
& $90.02$
& $2.10\times$
& $7.84 \pm 0.01\,\mathrm{ms}$
& $1.86\times$
& $123.39 \pm 1.11\,\mathrm{ms}$
& $1.99\times$ \\

\midrule

\multirow{2}{*}{ImageNet-1K}
& ViT
& $81.77$
& $1.00\times$
& $14.62 \pm 0.04\,\mathrm{ms}$
& $1.00\times$
& $246.42 \pm 3.85\,\mathrm{ms}$
& $1.00\times$ \\

& \textbf{Ours}
& $80.55$
& $1.79\times$
& $9.20 \pm 0.00\,\mathrm{ms}$
& $1.59\times$
& $144.63 \pm 0.70\,\mathrm{ms}$
& $1.70\times$ \\

\midrule

\multirow{2}{*}{Food-101}
& ViT
& $91.04$
& $1.00\times$
& $14.56 \pm 0.00\,\mathrm{ms}$
& $1.00\times$
& $243.75 \pm 1.92\,\mathrm{ms}$
& $1.00\times$ \\

& \textbf{Ours}
& $90.31$
& $2.57\times$
& $6.45 \pm 0.01\,\mathrm{ms}$
& $2.26\times$
& $99.95 \pm 0.65\,\mathrm{ms}$
& $2.44\times$ \\

\bottomrule
\end{tabular}

%% file: sections/supplement/Knowledge_Distillation_Ablation.tex
\section{Further Ablation Studies}
\label{app:further_ablation}

\subsection{Knowledge Distillation}

\begin{table}[!htbp]
\centering
\caption{Further ablations on ImageNet-1K:
(a) knowledge distillation,
(b) independent vs.\ shared query states,
(c) backbone K/V reuse, and
(d) training protocol with ViT;
(e) query accumulation with Swin.
Per-exit accuracies are evaluated independently without dynamic
routing, whereas Acc.\ denotes dynamic inference accuracy.
\textbf{Bold} marks the best value for each metric within each panel.}
\label{tab:further_ablation}
\input{tables/KD_ablation}
\end{table}

Table~\ref{tab:further_ablation}(a) evaluates knowledge distillation
from the frozen final classifier.
Adding distillation improves dynamic accuracy from $80.05\%$ to
$80.55\%$ and $\mathbf{S}_{\mathrm{MAC}}$ from $1.50\times$ to
$1.79\times$, supporting the use of full-depth predictions to
guide early-exit learning.

\subsection{Independent vs. Shared Query States}

The shared variant maintains a single recurrent query state used
by all exits, whereas DeclareViT maintains a separate query state for each exit.
As shown in Table~\ref{tab:further_ablation}(b), independent states
improve dynamic accuracy from $80.16\%$ to $80.55\%$ and
$\mathbf{S}_{\mathrm{MAC}}$ from $1.69\times$ to $1.79\times$.
These results support maintaining exit-specific query states rather than sharing a single state across prediction depths.

\subsection{Backbone Key/Value Reuse}

Table~\ref{tab:further_ablation}(c) evaluates reusing backbone keys
and values instead of computing them through separate, trainable
exit-side projections. Backbone K/V reuse yields $80.05\%$ dynamic
accuracy at $1.71\times$ $\mathbf{S}_{\mathrm{MAC}}$, compared with $80.55\%$ at
$1.79\times$ for our default design. These results support learning
a separate exit-side reading interface rather than directly reusing
backbone K/V representations for early classification.

\subsection{Further Analysis of Training Protocol}

In Section~\ref{sec:ablation}, we compare Joint-attached, Joint-detached,
and our two-stage protocol under matched architecture, initialization,
exit objective, distillation settings, and exit-training budget. For both
joint-training variants, we use separate parameter groups with a learning
rate of \(4.0\times10^{-5}\) for the backbone and final classifier and
\(8.0\times10^{-4}\) for the query branch. Here, we
further report per-exit and full-depth accuracies. Joint-attached allows
exit losses to update the backbone, Joint-detached blocks these gradients
while still training the backbone jointly, and Two-stage additionally
freezes the backbone and final classifier during exit optimization.

As shown in Table~\ref{tab:further_ablation}(d), Joint-attached reduces the
full-depth accuracy from $81.77\%$ to $81.39\%$. Applying stop-gradient
improves the shallow exits, e.g., from $55.85\%$ to $60.25\%$ at block 2
and from $67.53\%$ to $69.94\%$ at block 4, while recovering the final
accuracy to $81.50\%$. Two-stage further improves the deeper exits and
fully preserves the original $81.77\%$ full-depth accuracy, yielding
$80.55\%$ dynamic accuracy at $1.79\times$ $\mathbf{S}_{\mathrm{MAC}}$.
These results support gradient isolation and are consistent with the benefits of training exits on fixed backbone representations.

\subsection{Query Accumulation in Swin}

Table~\ref {tab:further_ablation}(e) evaluates the Swin-specific choice
of exit-local non-accumulative reading against progressive query
accumulation. Although accumulation improves several standalone exit
accuracies, particularly at shallow depths, it requires every active
query to repeatedly attend to high-resolution intermediate feature maps.
For example, the first two Swin-B blocks contain $56\times56=3136$
spatial tokens, and updating all five active queries with $D_q=384$
incurs approximately $1.09$ GMAC of reader computation per block,
already exceeding the cost of a first-stage Swin-B block.
Consequently, the accumulative variant achieves only $81.74\%$ dynamic
accuracy at $1.38\times$ $\mathbf{S}_{\mathrm{MAC}}$, whereas our
exit-local design reaches $82.71\%$ at $1.58\times$.
These results complement the computational analysis in
Appendix~\ref{app:backbone_adaptation} and show that, for hierarchical
Swin representations, the improvement in shallow-exit accuracy does
not compensate for the repeated high-resolution reader computation.

\begingroup\revcolor
\subsection{Lightweight Configuration}
\label{app:lite}
 
\begin{table}[!htbp]
\centering
\revcolor
\caption{\revcolor Default and lightweight DeclareViT with ViT-B/16 on
ImageNet-1K using seed 2024.
Overheads are relative to the original ViT-B/16.
Final-path GMAC is the cost of a sample that reaches the final classifier.
Operating points are selected on the calibration split.}
\label{tab:lite}
\resizebox{\linewidth}{!}{%
\begin{tabular}{lcccccc}
\toprule
 & & & \multicolumn{2}{c}{$1.50\times$ target} & \multicolumn{2}{c}{$2.00\times$ target} \\
\cmidrule(lr){4-5}\cmidrule(lr){6-7}
Configuration & \#Params (extra) & Final-path GMAC (overhead)
& Acc.\ (\%) & $\mathbf{S}_{\mathrm{MAC}}$ & Acc.\ (\%) & $\mathbf{S}_{\mathrm{MAC}}$ \\
\midrule
ViT-B/16 & 86.6M (--) & 17.56 (--)
& \multicolumn{4}{c}{81.77 (full depth, $1.00\times$)} \\
DeclareViT ($D_q{=}768$, FFN ratio 4) & 107.0M (+20.4M) & 19.96 (+13.6\%)
& 81.77 & $1.44\times$ & 79.90 & $1.94\times$ \\
DeclareViT-Lite ($D_q{=}384$, FFN ratio 2) & 93.5M (+7.0M) & 18.74 (+6.7\%)
& 81.59 & $1.44\times$ & 78.63 & $1.93\times$ \\
\bottomrule
\end{tabular}%
}
\end{table}
 
For memory-constrained deployments, we provide a lightweight
configuration, DeclareViT-Lite.
It reduces the query width from $D_q=768$ to $D_q=384$, with the key and value projections mapping the
768-dimensional backbone features to this width, and uses an exit-side FFN
expansion ratio of 2.
All other settings follow the default ViT configuration.
As shown in Table~\ref{tab:lite}, DeclareViT-Lite reduces the added
parameters from 20.4M to 7.0M and the final-path MAC overhead from
$13.6\%$ to $6.7\%$.
Accuracy is 0.18 and 1.27 points lower at the $1.50\times$ and
$2.00\times$ calibration targets, at nearly the same realized
$\mathbf{S}_{\mathrm{MAC}}$, while the full-depth accuracy is unchanged.
\endgroup

%% file: tables/KD_ablation.tex

\begingroup
\small
\renewcommand{\arraystretch}{1.08}


\begin{minipage}[t]{0.305\linewidth}
\vspace{0pt}
\centering
\setlength{\tabcolsep}{1.0pt}

\begin{tabular*}{\linewidth}{@{\extracolsep{\fill}}lrr@{}}
\toprule
\multicolumn{3}{@{}l}{
    \textbf{(a) Knowledge distillation}
} \\
\textbf{Variant}
& \textbf{Acc.}\,$\uparrow$
& $\mathbf{S}_{\mathrm{MAC}}$\,$\uparrow$ \\
\midrule
w/o KD
& $80.05$
& $1.50\times$ \\
\textbf{w/ KD (ours)}
& $\mathbf{80.55}$
& $\mathbf{1.79\times}$ \\
\bottomrule
\end{tabular*}
\end{minipage}%
\hfill
\begin{minipage}[t]{0.350\linewidth}
\vspace{0pt}
\centering
\setlength{\tabcolsep}{1.5pt}

\begin{tabular*}{\linewidth}{@{\extracolsep{\fill}}lrr@{}}
\toprule
\multicolumn{3}{@{}l}{
    \textbf{(b) Query states}
} \\
\textbf{Variant}
& \textbf{Acc.}\,$\uparrow$
& $\mathbf{S}_{\mathrm{MAC}}$\,$\uparrow$ \\
\midrule
Shared
& $80.16$
& $1.69\times$ \\
\textbf{Independent (ours)}
& $\mathbf{80.55}$
& $\mathbf{1.79\times}$ \\
\bottomrule
\end{tabular*}
\end{minipage}%
\hfill
\begin{minipage}[t]{0.290\linewidth}
\vspace{0pt}
\centering
\setlength{\tabcolsep}{1.0pt}

\begin{tabular*}{\linewidth}{@{\extracolsep{\fill}}lrr@{}}
\toprule
\multicolumn{3}{@{}l}{
    \textbf{(c) K/V reuse}
} \\
\textbf{Variant}
& \textbf{Acc.}\,$\uparrow$
& $\mathbf{S}_{\mathrm{MAC}}$\,$\uparrow$ \\
\midrule
Reused K/V
& $80.05$
& $1.71\times$ \\
\textbf{Ours}
& $\mathbf{80.55}$
& $\mathbf{1.79\times}$ \\
\bottomrule
\end{tabular*}
\end{minipage}

\par\vspace{0.7em}


\begin{minipage}[t]{\linewidth}
\vspace{0pt}
\centering
\setlength{\tabcolsep}{3.0pt}

\begin{tabular*}{\linewidth}{
    @{\extracolsep{\fill}}lrrrrrrrr@{}
}
\toprule
\multicolumn{9}{@{}l}{
    \textbf{(d) Training protocol}
} \\
\multirow{2}{*}{\textbf{Protocol}}
& \multicolumn{6}{c}{\textbf{Per-Exit Accuracy (\%)}}
& \multirow{2}{*}{\textbf{Acc. (\%)}\,$\uparrow$}
& \multirow{2}{*}{$\mathbf{S}_{\mathrm{MAC}}$\,$\uparrow$} \\
\cmidrule(lr){2-7}
& \textbf{@2}
& \textbf{@4}
& \textbf{@6}
& \textbf{@8}
& \textbf{@10}
& \textbf{Final}
&
& \\
\midrule
Joint-attached
& 55.85
& 67.53
& 74.00
& 77.69
& 80.53
& 81.39
& 80.12
& $1.67\times$ \\

Joint-detached
& \textbf{60.25}
& 69.94
& 75.59
& 78.67
& 80.74
& 81.50
& 80.23
& $1.72\times$ \\

\textbf{Two-stage (ours)}
& 60.10
& \textbf{70.08}
& \textbf{75.71}
& \textbf{79.03}
& \textbf{81.20}
& \textbf{81.77}
& \textbf{80.55}
& $\mathbf{1.79\times}$ \\
\bottomrule
\end{tabular*}
\end{minipage}

\par\vspace{0.7em}


\begin{minipage}[t]{\linewidth}
\vspace{0pt}
\centering
\setlength{\tabcolsep}{3.0pt}

\begin{tabular*}{\linewidth}{
    @{\extracolsep{\fill}}lrrrrrrrr@{}
}
\toprule
\multicolumn{9}{@{}l}{
    \textbf{(e) Query accumulation (Swin)}
} \\
\multirow{2}{*}{\textbf{Readout}}
& \multicolumn{6}{c}{\textbf{Per-Exit Accuracy (\%)}}
& \multirow{2}{*}{\textbf{Acc. (\%)}\,$\uparrow$}
& \multirow{2}{*}{$\mathbf{S}_{\mathrm{MAC}}$\,$\uparrow$} \\
\cmidrule(lr){2-7}
& \textbf{@4}
& \textbf{@8}
& \textbf{@12}
& \textbf{@16}
& \textbf{@20}
& \textbf{Final}
&
& \\
\midrule
Accumulative
& \textbf{59.80}
& \textbf{71.39}
& \textbf{77.02}
& \textbf{80.13}
& 82.44
& 83.59
& 81.74
& $1.38\times$ \\

\textbf{Non-accumulative (ours)}
& 55.22
& 68.05
& 75.47
& 79.99
& \textbf{82.79}
& 83.59
& \textbf{82.71}
& $\mathbf{1.58\times}$ \\
\bottomrule
\end{tabular*}
\end{minipage}

\endgroup

%% file: sections/supplement/Baseline_Methods.tex
\subsection{Baseline Methods}
\label{app:baseline_methods}

\paragraph{SDN.}
Shallow-Deep Networks~\citep{Kaya:2019:SDN} attach internal classifiers
to intermediate representations of a deep network.
These classifiers allow confident inputs to terminate early and
expose overthinking, where additional computation can turn a correct
intermediate prediction into an incorrect final prediction.

\paragraph{BYOT.}
BYOT~\citep{Zhang:2019:BYO} trains shallow classifiers through online
self-distillation within a single network.
The deepest classifier supplies soft prediction targets and feature
supervision to earlier branches, complementing their ground-truth
classification losses without requiring a separately trained teacher.

\paragraph{ViT-EE.}
ViT-EE~\citep{Bakhtiarnia:2021:MVT} studies multi-exit vision transformers
with different intermediate classifier designs, including MLP,
convolutional, and Transformer-based branches.
These branches provide alternative ways to extract class information
from intermediate token representations for dynamic inference.

\paragraph{LGViT.}
LGViT~\citep{Xu:2023:LDE} introduces heterogeneous exit heads tailored
to network depth: local-perception heads use convolutional operators,
while global-aggregation heads use spatially reduced self-attention.
Its two-stage training first optimizes the multi-exit network and then
freezes the backbone and final classifier while refining the exits
through self-distillation.

\paragraph{COSEE.}
COSEE~\citep{He:2025:CCS} addresses the mismatch between training and
dynamic early-exit inference in pretrained language models.
Its sample weighting mechanism simulates exiting under multiple
thresholds and emphasizes classifiers near each sample's estimated exit.
Online signal calibration further separates the exiting signals of
correctly and incorrectly classified samples.

\paragraph{DMPO.}
DMPO~\citep{Luo:2025:DMO} combines parameter-efficient backbone tuning
with bypass-assisted, stage-wise predictors, using high-order predictors
to strengthen early classification.
Its representation-to-discrimination schedule progressively shifts
supervision from later to earlier stages, while an additional
loss factor regulates inter-stage optimization.

\paragraph{EnViT.}
EnViT~\citep{Dong:2026:EET} combines exit-aware structured dropout
with self-distillation to improve multi-exit vision transformers.
It drops entire Transformer layers on a per-sample basis, forming
virtual student networks supervised by the full model.
The exit-dependent dropout schedule is designed to reduce gradient
conflicts between exit objectives.

%% file: sections/supplement/Baseline_Reproduction_and_Provenance.tex
\subsection{Reproduction and Provenance}
\label{app:baseline_provenance}

Our framework contains local implementations of SDN, BYOT, ViT-EE,
LGViT, COSEE, and DMPO.
We adapt their architectures and training procedures as described below;
EnViT is included using externally reported results because its official
implementation was not publicly available at the time of our experiments.
Table~\ref{tab:baseline_provenance} summarizes implementation origins
and prediction locations.

\begin{table}[!htbp]
\centering
\small
\setlength{\tabcolsep}{3pt}
\renewcommand{\arraystretch}{1.12}
\caption{Baseline implementation provenance and prediction locations.
Indices are one-based; Swin indices count blocks across stages.
For SDN, BYOT, ViT-EE, COSEE, and LGViT, the listed locations are
early exits preceding the final classifier.
DMPO lists all four predictors, including its full-depth predictor.}
\label{tab:baseline_provenance}
\begin{tabularx}{\linewidth}{@{}l
>{\raggedright\arraybackslash}X
>{\raggedright\arraybackslash}X
>{\raggedright\arraybackslash}X@{}}
\toprule
\textbf{Method} & \textbf{Implementation}
& \textbf{ViT / DeiT blocks} & \textbf{Swin blocks} \\
\midrule
SDN & Adapted linear-readout baseline
& $2,4,6,8,10$ & $4,8,12,16,20$ \\
BYOT & Self-distillation adaptation
& $2,4,6,8,10$ & $4,8,12,16,20$ \\
ViT-EE & Transformer-exit adaptation
& $2,4,6,8,10$ & $4,8,12,16,20$ \\
COSEE & Visual adaptation of published objectives
& $2,4,6,8,10$ & $4,8,12,16,20$ \\
LGViT & Native reimplementation
& $4,\ldots,11$ & $4,7,10,13,$ $16,19,22,23$ \\
DMPO & Adapted released implementation;
DeiT/Swin extensions
& $3,6,9,12$ & $6,12,18,24$ \\
EnViT & Externally reported
& $2,4,6,8,10$ & -- \\
\bottomrule
\end{tabularx}
\end{table}

\paragraph{Data and evaluation settings.}
Unless otherwise specified, all locally evaluated baselines use
the same backbones, input resolution, data splits, preprocessing,
and evaluation protocol as DeclareViT.
Method-specific exit configurations and training objectives
are detailed below.

\paragraph{Hyperparameter selection and adaptation policy.}
Our baseline adaptations are intentionally limited to the changes necessary for controlled comparison under a common evaluation protocol. All locally evaluated methods use the same backbone sources, dataset partitions, training/held-out split construction, preprocessing, and threshold-calibration and dynamic-evaluation procedure. Beyond these protocol-level alignments, method-specific architectures, objectives, and hyperparameters follow the original papers or released implementations whenever directly applicable and are not tuned on the final evaluation split. Modifications required to extend a method to additional backbones or to make it compatible with the unified evaluation setting are kept minimal and stated explicitly below.

\paragraph{Initialization.}
For SDN, BYOT, ViT-EE, and COSEE on CIFAR-100 and Food-101,
ViT/DeiT/Swin are initialized from the dataset-specific Stage-I backbone
and final-classifier checkpoints used by DeclareViT produced with the same seed.
 
Method-specific exits are constructed afresh; copied reader blocks
are initialized from the loaded backbone.
On ImageNet-1K, these methods load the corresponding pretrained
backbone and original classifier.
DMPO uses the same seed-matched backbone sources but discards the source classifiers
and initializes its own predictors.
LGViT has a separate initialization procedure described below.

\paragraph{SDN and ViT-EE.}
Our SDN adaptation uses LayerNorm followed by a linear classifier:
ViT reads the class token, DeiT averages predictions from its class
and distillation tokens, and Swin uses global average pooling.
It does not reproduce the original SDN exit placement or progressive
loss-weight schedule.

For ViT-EE, we use an independent backbone-compatible Transformer
block followed by normalization and three-layer MLP readouts.
ViT classifies the resulting class token, while Swin applies global
average pooling before classification.
Distilled DeiT uses separate MLPs for the class and distillation
tokens and averages their logits.
Each MLP has two hidden layers of width 768 with GELU activations.
These exit readers do not write their outputs back into the backbone.
Both baselines are jointly trained with hard-label cross-entropy:
\begin{equation}
\mathcal L_{\mathrm{SDN/ViT\text{-}EE}}
=
0.5\,\mathcal L_{\mathrm{CE}}(\mathbf z_{\mathrm f},y)
+
0.1\sum_{e=1}^{5}
\mathcal L_{\mathrm{CE}}(\mathbf z_e,y).
\label{eq:baseline_ce}
\end{equation}
No knowledge distillation is used for these two configurations.

\paragraph{BYOT.}
We adapt BYOT to the same lightweight readouts as SDN, adding
training-only linear adapters for feature alignment.
The final branch provides detached prediction and feature targets,
while the backbone and classifiers are jointly optimized.
We use a prediction-distillation weight of 0.3, temperature 3,
and feature-loss coefficient 0.03.
Our implementation uses soft-target cross-entropy without an additional
$T^2$ factor, and feature alignment uses an unsquared $L_2$ norm.
The implementation computes this norm over the global training batch.
The final classification loss has weight 0.5; each early branch has
weight 0.1, including its feature-alignment contribution.
Feature adapters are excluded from inference.

\paragraph{COSEE.}
We implement COSEE's sample weighting and online signal calibration
objectives on the SDN readout architecture and jointly train the
backbone and classifiers.
Normalized entropy is used as the exiting signal.
We sample five thresholds per training step and linearly increase
the sample-weighting sharpness from 0 to 1.
The signal-calibration loss has coefficient 0.1 and margin 0.3.
Sample-weighted classification includes all early exits and the final
classifier; signal calibration applies only to early exits and
averages over exits containing both correct and incorrect predictions.
The sparse exit layout and vision backbones are adaptations of the
original language-model setting.

\paragraph{LGViT.}
We reimplement LGViT using native \texttt{timm} backbones and its
local-perception and global-aggregation head designs.
We retain eight early exits, with four local-perception heads followed
by four global-aggregation heads.
Stage~I alternates between final-classifier cross-entropy and
depth-weighted joint cross-entropy.
Stage~II freezes the backbone and final classifier and trains the exits
with prediction-distillation weight 0.3 and temperature 2.
For ViT/DeiT, we additionally retain the heterogeneous and homogeneous
feature-distillation losses, each weighted by 0.01.
Following the released Swin Stage-II implementation, our Swin adaptation
uses prediction distillation without these feature losses.
The prediction-distillation KL term is averaged over both samples and
classes before multiplication by $T^2$.
For CIFAR-100 and Food-101, all three backbones execute both stages from
their corresponding pretrained backbone initialization; ImageNet-1K is
likewise configured to run both stages.
We retain the released learning rate of $5\times10^{-5}$ and the
method-specific loss coefficients, while using 100 epochs per locally
executed stage under our common training protocol.
\paragraph{DMPO.}
We adapt the released DMPO implementation and extend its token and
spatial-feature interfaces to distilled DeiT and Swin.
We retain rank-8 LoRA~\citep{Hu:2022:LRA}, rank-8 predictor-specific bypasses,
high-order predictors at the first two stages, and trainable prediction
scales initialized to one.
The four predictors include three early exits and a full-depth predictor;
the original classifier is not used as an additional fallback.
Pretrained core weights remain frozen, while LoRA, bypasses, predictors,
and prediction scales are optimized.
Following the default setting reported in the DMPO paper, the
representation-to-discrimination loss weights interpolate linearly from
$(0.01,0.01,1,2)$ to $(1.5,1,0.1,0.1)$.
We use the paper-reported default consistently
across all experiments.
We also retain the sigmoid loss factor based on the preceding predictor's
per-sample cross-entropy.
We use the released learning rates of $5\times10^{-4}$ for
CIFAR-100/Food-101 and $10^{-4}$ for ImageNet-1K, while using a common
100-epoch training budget under our local evaluation protocol.
DeiT and Swin are backbone extensions of the released ViT implementation.

\paragraph{Optimization.}
Local training uses AdamW with weight decay 0.05, a global batch
size of 1024, and ten epochs of linear warmup followed by cosine decay. For SDN, BYOT, ViT-EE, and COSEE, we use the same grouped learning rate
scheme as DeclareViT for the backbone and exit modules.
Unless otherwise stated above, locally trained methods use a
100-epoch budget.
All locally trained methods use BF16 mixed precision and are trained on four NVIDIA H100 80GB GPUs.

\paragraph{Dynamic evaluation and computation accounting.}
Local models use the sequential entropy-threshold calibration
procedure in Appendix~\ref{app:further_implementation}, with the
number of prediction paths set separately for each method.
Thresholds are fitted on the held-out training subset and applied
unchanged to the final evaluation split.
The sweep uses 202 signed logarithmically spaced $\gamma$ values.
Expected MACs include the backbone computation and all branch
operations executed before termination.
MAC speed-up is measured relative to the corresponding unbranched
full-depth backbone, including its classifier.

\paragraph{Externally reported EnViT results.}
We use the ViT-B/16 results from Table~2 of
\citet{Dong:2026:EET} as an external reference.
EnViT is not retrained or recalibrated in our framework, and its
reported operating points are not aligned to the local $\gamma$ grid.
Its results retain the original authors' experimental settings and
are therefore separate from comparisons under our local protocol.

%% file: sections/supplement/MAC_Accounting.tex
\section{MAC Accounting}
\label{app:mac_accounting}

\begin{table}[!htbp]
\centering
\caption{
Analytical cumulative MACs per image for DeclareViT with ViT-B/16
on ImageNet-1K at $224\times224$ resolution.
All numerical cost columns are in GMACs.
Backbone includes patch embedding and the reached Transformer blocks.
Readers include all active-query updates up to termination.
Readouts include all evaluated exit-side FFNs and classifiers,
and the final classifier when used.
Rounded components may not sum exactly to the rounded total.
}
\label{tab:exit_path_macs}
\begingroup
\small
\setlength{\tabcolsep}{4pt}
\renewcommand{\arraystretch}{1.08}
\begin{tabular*}{\linewidth}{@{\extracolsep{\fill}}lrrrrr@{}}
\toprule
\textbf{Path} & \textbf{Block} & \textbf{Backbone}
& \textbf{Readers} & \textbf{Readouts} & \textbf{Total} \\
\midrule
Exit 1 & 2  & 3.0235  & 0.4796 & 0.0055 & 3.5086  \\
Exit 2 & 4  & 5.9314  & 0.9562 & 0.0110 & 6.8986  \\
Exit 3 & 6  & 8.8393  & 1.4299 & 0.0165 & 10.2857 \\
Exit 4 & 8  & 11.7472 & 1.9006 & 0.0219 & 13.6698 \\
Exit 5 & 10 & 14.6552 & 2.3684 & 0.0274 & 17.0510 \\
Dynamic final & 12 & 17.5631 & 2.3684 & 0.0282 & 19.9596 \\
\midrule
Unbranched ViT & 12 & 17.5631 & 0.0000 & 0.0008 & 17.5638 \\
\bottomrule
\end{tabular*}
\endgroup
\end{table}

\paragraph{Counting convention.}
We count one multiply--accumulate pair as one MAC.
Our ViT estimator includes the patch-embedding convolution, linear
projections, and attention matrix products.
Bias additions, LayerNorm, softmax, nonlinear activations, residual
additions, pooling reductions, and routing comparisons are excluded.
All costs below are per image, with $1\,\mathrm{GMAC}=10^9$ MACs.

\paragraph{Cumulative inference paths.}
The cost $C_j$ is the cumulative computation required to reach
prediction path $j$.
It includes all backbone blocks, reader operations, and earlier exit
readouts executed before termination.
For DeclareViT, active queries are updated until their designated exits
and incur no further cost after being discarded.
$C_{\mathrm{base}}$ denotes the unbranched full-depth model cost.

\subsection{ViT-B/16 example}
\label{app:mac_vit_example}

We consider DeclareViT with ViT-B/16 on ImageNet-1K at
$224\times224$ resolution.
The patch size is $p=16$, giving $N=(224/16)^2=196$ patch tokens
and $M=N+1=197$ backbone and memory tokens, including
[\textsc{cls}].
The backbone has $L=12$ blocks and embedding dimension $D=768$;
both the backbone and exit-side FFNs have hidden dimension
$D_{\mathrm{ffn}}=3072$.
There are $C=1000$ classes and five exits at
$\{2,4,6,8,10\}$.
This example uses progressive accumulation, segment-shared readers with eight attention heads, and an exit-side FFN applied only at
prediction depths.

\paragraph{Patch embedding and backbone blocks.}
Patch embedding is performed once, with cost
\begin{equation}
c_{\mathrm{patch}}
= N(3p^2)D
= 115{,}605{,}504.
\label{eq:mac_patch_embedding}
\end{equation}
For each backbone block, the Q/K/V projections cost $3MD^2$,
the attention output projection costs $MD^2$,
the two attention matrix products cost $2M^2D$,
and the two FFN linear layers cost $2MDD_{\mathrm{ffn}}$.
Thus,
\begin{equation}
\begin{aligned}
c_{\mathrm{block}}
&=4MD^2+2M^2D+2MDD_{\mathrm{ffn}} \\
&=464{,}781{,}312+59{,}610{,}624+929{,}562{,}624 \\
&=1{,}453{,}954{,}560.
\end{aligned}
\label{eq:mac_vit_block}
\end{equation}
The final classifier costs $c_{\mathrm{final}}=DC=768{,}000$ MACs.
The unbranched reference therefore has cost
\begin{equation}
\begin{aligned}
C_{\mathrm{base}}
&=c_{\mathrm{patch}}+L c_{\mathrm{block}}+c_{\mathrm{final}} \\
&=17{,}563{,}828{,}224
\approx17.5638\,\mathrm{GMAC}.
\end{aligned}
\label{eq:mac_unbranched_vit}
\end{equation}

\paragraph{Cross-attention readers.}
Let $a_\ell=|\{e:\ell\leq d_e\}|$ be the number of active queries
at block $\ell$.
For the configured exits,
\begin{equation}
(a_1,\ldots,a_{12})=(5,5,4,4,3,3,2,2,1,1,0,0).
\label{eq:mac_active_queries}
\end{equation}
At each block, the reader processes all active exit queries together.
Because the memory-side LayerNorm and the K/V projection parameters
are common to these queries, the normalized memory and its K/V
projections are computed once per block, rather than once per exit.
The K/V projections therefore cost $2MD^2$.
The query and output projections cost $2a_\ell D^2$, while the two
attention products cost $2a_\ell MD$.
These terms already sum over all heads; the head count introduces
no additional multiplicative factor at fixed total dimension $D$.
The reader cost is therefore
\begin{equation}
r_\ell=
\begin{cases}
2MD^2+2a_\ell D^2+2a_\ell MD, & a_\ell>0, \\
0, & a_\ell=0.
\end{cases}
\label{eq:mac_reader_cost}
\end{equation}
Numerically, $r_\ell=232{,}390{,}656+1{,}482{,}240a_\ell$
when $a_\ell>0$.
For example, each of the first two blocks incurs
$239{,}801{,}856$ reader MACs for five queries, not five times
the cost of a separate one-query reader.
Across all ten reader-equipped blocks,
\begin{equation}
\sum_{\ell=1}^{10}r_\ell
=10(232{,}390{,}656)+30(1{,}482{,}240)
=2{,}368{,}373{,}760.
\label{eq:mac_total_readers}
\end{equation}
Sharing projection parameters across blocks does not reuse their
projected features: each block has different memory and query states,
so its matrix products are counted separately.

\paragraph{Exit readouts.}
Each exit applies its FFN to one query, followed by a linear classifier:
\begin{equation}
c_{\mathrm{exit}}
=2DD_{\mathrm{ffn}}+DC
=4{,}718{,}592+768{,}000
=5{,}486{,}592.
\label{eq:mac_exit_readout}
\end{equation}
The shared FFN is counted once at each evaluated exit, not once per
image regardless of how many exits are evaluated.

\paragraph{Static path costs.}
For early exit $e\in\{1,\ldots,E\}$,
\begin{equation}
C_e
=c_{\mathrm{patch}}+d_e c_{\mathrm{block}}
+\sum_{\ell=1}^{d_e}r_\ell+e c_{\mathrm{exit}}.
\label{eq:mac_early_path}
\end{equation}
For example, a sample terminating at block 2 has already updated
all five queries at blocks 1 and 2 and evaluated the first readout:
\begin{equation}
\begin{aligned}
C_1
&=115{,}605{,}504+2(1{,}453{,}954{,}560) \\
&\quad+2(239{,}801{,}856)+5{,}486{,}592 \\
&=3{,}508{,}604{,}928.
\end{aligned}
\label{eq:mac_first_exit_example}
\end{equation}
For a sample reaching the final classifier, path $E+1$ has cost
\begin{equation}
\begin{aligned}
C_{E+1}
&=c_{\mathrm{patch}}+L c_{\mathrm{block}}
+\sum_{\ell=1}^{d_E}r_\ell
+E c_{\mathrm{exit}}+c_{\mathrm{final}} \\
&=19{,}959{,}634{,}944
\approx19.9596\,\mathrm{GMAC}.
\end{aligned}
\label{eq:mac_dynamic_final}
\end{equation}
Table~\ref{tab:exit_path_macs} lists the cumulative costs.
They describe termination paths of the complete five-exit model,
not separately reconfigured single-exit networks.
The dynamic final path costs approximately $13.64\%$ more than the
unbranched reference because it evaluates all earlier branches.

%% file: sections/supplement/Gradient_Conflict_Measurement.tex
\section{Further Gradient Conflict Analysis}
\label{app:gradient_conflict}

We extend Sections~\ref{sec:role_coupling} and~\ref{sec:gradient_conflict} across backbone depths and shared
exit-side modules at epochs 1, 10, and 100, complementing conflict
rates with loss-weighted gradient norms.

\paragraph{Diagnostic protocol.}
At each checkpoint, we sample 57 minibatches with a global batch
size of 1024. Objective-specific gradients are computed from the
same augmented samples and forward pass, without intervening
optimizer updates, and averaged across workers before analysis.
Conflict rates measure the fraction of valid comparisons with
negative cosine, as defined in Eq.~\ref{eq:pairwise_gradient_cosine}. 
Valid comparisons require
both gradients to satisfy
$\|g\|_2^2 > \epsilon(\delta^{-1}-1)$,
with $\epsilon=10^{-12}$ and $\delta=10^{-5}$.
We additionally report the minibatch-mean loss-weighted gradient
norm $\|w_i\nabla_{\theta_g}\mathcal{L}_i\|_2$, where $w_i$ is the
training-loss weight and $\theta_g$ denotes the measured parameter
group.

\paragraph{SDN.}
Figures~\ref{fig:app_sdn_conflict_ep1}--\ref{fig:app_sdn_conflict_ep100} show that early--final conflicts occur across multiple
backbone depths initially and persist in shallow blocks at later
checkpoints. Importantly, in blocks 1--2, the earliest-exit
objective has a mean loss-weighted gradient norm comparable to
or larger than that of the final objective at later checkpoints.
Thus, persistent directional conflicts coexist with non-negligible
average exit gradients, supporting our motivation for isolating
backbone feature refinement from early-exit supervision.

\paragraph{DeclareViT.}
Figures~\ref{fig:app_declarevit_conflict_ep1}--\ref{fig:app_declarevit_conflict_ep100} show generally low inter-exit conflict rates in the
shared cross-attention projections and exit FFN, alongside
non-negligible average gradient norms.
These measurements characterize residual interactions within
the exit branches, while the frozen, read-only design prevents
exit learning from altering the backbone.
Since SDN and DeclareViT are measured on different parameter
groups and training objectives, these results are complementary
diagnostics rather than a controlled comparison of conflict
reduction. 
\input{artwork/gradient_conflict/further/appendix_figures}

%% file: artwork/gradient_conflict/further/appendix_figures.tex
\begin{figure}[p]
\centering
\begingroup
\setlength{\parskip}{0pt}
\textbf{SDN: joint training, epoch~1}\par
\vspace{3pt}
\resizebox{\linewidth}{!}{%
    \input{artwork/gradient_conflict/further/sdn_conflict_epoch001}%
}\par
\vspace{6pt}
\resizebox{\linewidth}{!}{%
    \input{artwork/gradient_conflict/further/sdn_gradient_norm_epoch001}%
}\par
\endgroup
\caption{Block-wise gradient diagnostics for the [\textsc{cls}]-based SDN adaptation at joint-training epoch~1. Top: pairwise gradient conflict rates. Bottom: mean loss-weighted gradient norms, $\|w_i\nabla_{\theta_g}\mathcal{L}_i\|_2$. Final denotes the full-depth objective.}
\label{fig:app_sdn_conflict_ep1}
\label{fig:sdn_gradnorm_epoch01}
\end{figure}

\begin{figure}[p]
\centering
\begingroup
\setlength{\parskip}{0pt}
\textbf{SDN: joint training, epoch~10}\par
\vspace{3pt}
\resizebox{\linewidth}{!}{%
    \input{artwork/gradient_conflict/further/sdn_conflict_epoch010}%
}\par
\vspace{6pt}
\resizebox{\linewidth}{!}{%
    \input{artwork/gradient_conflict/further/sdn_gradient_norm_epoch010}%
}\par
\endgroup
\caption{Block-wise gradient diagnostics for the [\textsc{cls}]-based SDN adaptation at joint-training epoch~10. Top: pairwise gradient conflict rates. Bottom: mean loss-weighted gradient norms, $\|w_i\nabla_{\theta_g}\mathcal{L}_i\|_2$. Final denotes the full-depth objective.}
\label{fig:app_sdn_conflict_ep10}
\label{fig:sdn_gradnorm_epoch10}
\end{figure}

\begin{figure}[p]
\centering
\begingroup
\setlength{\parskip}{0pt}
\textbf{SDN: joint training, epoch~100}\par
\vspace{3pt}
\resizebox{\linewidth}{!}{%
    \input{artwork/gradient_conflict/further/sdn_conflict_epoch100}%
}\par
\vspace{6pt}
\resizebox{\linewidth}{!}{%
    \input{artwork/gradient_conflict/further/sdn_gradient_norm_epoch100}%
}\par
\endgroup
\caption{Block-wise gradient diagnostics for the [\textsc{cls}]-based SDN adaptation at joint-training epoch~100. Top: pairwise gradient conflict rates. Bottom: mean loss-weighted gradient norms, $\|w_i\nabla_{\theta_g}\mathcal{L}_i\|_2$. Final denotes the full-depth objective.}
\label{fig:app_sdn_conflict_ep100}
\label{fig:sdn_gradnorm_epoch100}
\end{figure}

\begin{figure}[p]
\centering
\begingroup
\setlength{\parskip}{0pt}
\textbf{DeclareViT: Stage II, epoch~1}\par
\vspace{3pt}
\resizebox{\linewidth}{!}{%
    \input{artwork/gradient_conflict/further/declarevit_conflict_epoch001}%
}\par
\vspace{6pt}
\resizebox{\linewidth}{!}{%
    \input{artwork/gradient_conflict/further/declarevit_gradient_norm_epoch001}%
}\par
\endgroup
\caption{Gradient diagnostics for DeclareViT's segment-shared cross-attention projections and shared exit FFN at Stage~II epoch~1. Top: pairwise gradient conflict rates. Bottom: mean loss-weighted gradient norms, $\|w_i\nabla_{\theta_g}\mathcal{L}_i\|_2$.}
\label{fig:app_declarevit_conflict_ep1}
\label{fig:declarevit_gradnorm_epoch1}
\label{fig:app_declarevit_conflict_all}
\end{figure}

\begin{figure}[p]
\centering
\begingroup
\setlength{\parskip}{0pt}
\textbf{DeclareViT: Stage II, epoch~10}\par
\vspace{3pt}
\resizebox{\linewidth}{!}{%
    \input{artwork/gradient_conflict/further/declarevit_conflict_epoch010}%
}\par
\vspace{6pt}
\resizebox{\linewidth}{!}{%
    \input{artwork/gradient_conflict/further/declarevit_gradient_norm_epoch010}%
}\par
\endgroup
\caption{Gradient diagnostics for DeclareViT's segment-shared cross-attention projections and shared exit FFN at Stage~II epoch~10. Top: pairwise gradient conflict rates. Bottom: mean loss-weighted gradient norms, $\|w_i\nabla_{\theta_g}\mathcal{L}_i\|_2$.}
\label{fig:app_declarevit_conflict_ep10}
\label{fig:declarevit_gradnorm_epoch10}

\end{figure}

\begin{figure}[p]
\centering
\begingroup
\setlength{\parskip}{0pt}
\textbf{DeclareViT: Stage II, epoch~100}\par
\vspace{3pt}
\resizebox{\linewidth}{!}{%
    \input{artwork/gradient_conflict/further/declarevit_conflict_epoch100}%
}\par
\vspace{6pt}
\resizebox{\linewidth}{!}{%
    \input{artwork/gradient_conflict/further/declarevit_gradient_norm_epoch100}%
}\par
\endgroup
\caption{Gradient diagnostics for DeclareViT's segment-shared cross-attention projections and shared exit FFN at Stage~II epoch~100. Top: pairwise gradient conflict rates. Bottom: mean loss-weighted gradient norms, $\|w_i\nabla_{\theta_g}\mathcal{L}_i\|_2$.}
\label{fig:app_declarevit_conflict_ep100}
\label{fig:declarevit_gradnorm_epoch100}

\end{figure}

%% file: artwork/gradient_conflict/further/sdn_gradient_norm_epoch001.tex
\begingroup
\definecolor{GNBlue}{RGB}{0,114,178}
\definecolor{GNGreen}{RGB}{0,158,115}
\definecolor{GNPink}{RGB}{204,121,167}
\definecolor{GNOchre}{RGB}{181,137,0}
\definecolor{GNGray}{RGB}{102,102,102}
\definecolor{GNOrange}{RGB}{213,94,0}
\begin{tikzpicture}[x=1cm,y=1.05cm]
\path[use as bounding box] (-0.10,0.00) rectangle (16.70,10.90);
\node[font=\fontsize{9.4}{10.4}\selectfont\bfseries,inner sep=0pt] at (8.30,10.62) {(B) Loss-weighted gradient norms};
\node[font=\fontsize{7.5}{8.5}\selectfont,inner sep=0pt] at (8.30,0.12) {Exit objective};
\node[anchor=east,font=\fontsize{7.2}{8.2}\selectfont,inner sep=.4pt] at (0.592,7.963) {0.05};
\node[anchor=east,font=\fontsize{7.2}{8.2}\selectfont,inner sep=.4pt] at (0.592,8.327) {0.10};
\node[anchor=east,font=\fontsize{7.2}{8.2}\selectfont,inner sep=.4pt] at (0.592,8.690) {0.15};
\node[anchor=east,font=\fontsize{7.2}{8.2}\selectfont,inner sep=.4pt] at (0.592,9.053) {0.20};
\node[anchor=east,font=\fontsize{7.2}{8.2}\selectfont,inner sep=.4pt] at (0.592,9.417) {0.25};
\node[anchor=east,font=\fontsize{7.2}{8.2}\selectfont,inner sep=.4pt] at (0.592,9.780) {0.30};
\node[anchor=east,font=\fontsize{7.2}{8.2}\selectfont,inner sep=.4pt] at (0.592,7.600) {0.00};
\node[font=\fontsize{8}{9}\selectfont\bfseries,inner sep=0pt] at (2.290,10.060) {Block 1};
\draw[black!75,line width=.30pt] (0.650,7.600) -- (3.930,7.600);
\draw[black!75,line width=.30pt] (0.650,7.600) -- (0.650,9.780);
\draw[black!12,line width=.20pt] (0.650,7.963) -- (3.930,7.963);
\draw[black!12,line width=.20pt] (0.650,8.327) -- (3.930,8.327);
\draw[black!12,line width=.20pt] (0.650,8.690) -- (3.930,8.690);
\draw[black!12,line width=.20pt] (0.650,9.053) -- (3.930,9.053);
\draw[black!12,line width=.20pt] (0.650,9.417) -- (3.930,9.417);
\draw[black!12,line width=.20pt] (0.650,9.780) -- (3.930,9.780);
\fill[GNBlue] (0.988,7.600) rectangle ++(0.262,0.483);
\node[font=\fontsize{7.2}{8.2}\selectfont,inner sep=.25pt] at (1.119,8.213) {0.066};
\node[anchor=north east,rotate=45,font=\fontsize{7.2}{8.2}\selectfont,inner sep=.25pt] at (1.119,7.500) {@2};
\fill[GNGreen] (1.456,7.600) rectangle ++(0.262,0.323);
\node[font=\fontsize{7.2}{8.2}\selectfont,inner sep=.25pt] at (1.587,8.573) {0.044};
\node[anchor=north east,rotate=45,font=\fontsize{7.2}{8.2}\selectfont,inner sep=.25pt] at (1.587,7.500) {@4};
\fill[GNPink] (1.925,7.600) rectangle ++(0.262,0.204);
\node[font=\fontsize{7.2}{8.2}\selectfont,inner sep=.25pt] at (2.056,7.934) {0.028};
\node[anchor=north east,rotate=45,font=\fontsize{7.2}{8.2}\selectfont,inner sep=.25pt] at (2.056,7.500) {@6};
\fill[GNOchre] (2.393,7.600) rectangle ++(0.262,0.225);
\node[font=\fontsize{7.2}{8.2}\selectfont,inner sep=.25pt] at (2.524,8.215) {0.031};
\node[anchor=north east,rotate=45,font=\fontsize{7.2}{8.2}\selectfont,inner sep=.25pt] at (2.524,7.500) {@8};
\fill[GNGray] (2.861,7.600) rectangle ++(0.262,0.363);
\node[font=\fontsize{7.2}{8.2}\selectfont,inner sep=.25pt] at (2.993,8.613) {0.050};
\node[anchor=north east,rotate=45,font=\fontsize{7.2}{8.2}\selectfont,inner sep=.25pt] at (2.993,7.500) {@10};
\fill[GNOrange] (3.330,7.600) rectangle ++(0.262,1.926);
\node[font=\fontsize{7.2}{8.2}\selectfont,inner sep=.25pt] at (3.461,9.656) {0.265};
\node[anchor=north east,rotate=45,font=\fontsize{7.2}{8.2}\selectfont,inner sep=.25pt] at (3.461,7.500) {Final};
\node[font=\fontsize{8}{9}\selectfont\bfseries,inner sep=0pt] at (6.370,10.060) {Block 2};
\draw[black!75,line width=.30pt] (4.730,7.600) -- (8.010,7.600);
\draw[black!75,line width=.30pt] (4.730,7.600) -- (4.730,9.780);
\draw[black!12,line width=.20pt] (4.730,7.963) -- (8.010,7.963);
\draw[black!12,line width=.20pt] (4.730,8.327) -- (8.010,8.327);
\draw[black!12,line width=.20pt] (4.730,8.690) -- (8.010,8.690);
\draw[black!12,line width=.20pt] (4.730,9.053) -- (8.010,9.053);
\draw[black!12,line width=.20pt] (4.730,9.417) -- (8.010,9.417);
\draw[black!12,line width=.20pt] (4.730,9.780) -- (8.010,9.780);
\fill[GNBlue] (5.068,7.600) rectangle ++(0.262,0.459);
\node[font=\fontsize{7.2}{8.2}\selectfont,inner sep=.25pt] at (5.199,8.189) {0.063};
\node[anchor=north east,rotate=45,font=\fontsize{7.2}{8.2}\selectfont,inner sep=.25pt] at (5.199,7.500) {@2};
\fill[GNGreen] (5.536,7.600) rectangle ++(0.262,0.305);
\node[font=\fontsize{7.2}{8.2}\selectfont,inner sep=.25pt] at (5.667,8.555) {0.042};
\node[anchor=north east,rotate=45,font=\fontsize{7.2}{8.2}\selectfont,inner sep=.25pt] at (5.667,7.500) {@4};
\fill[GNPink] (6.005,7.600) rectangle ++(0.262,0.220);
\node[font=\fontsize{7.2}{8.2}\selectfont,inner sep=.25pt] at (6.136,7.950) {0.030};
\node[anchor=north east,rotate=45,font=\fontsize{7.2}{8.2}\selectfont,inner sep=.25pt] at (6.136,7.500) {@6};
\fill[GNOchre] (6.473,7.600) rectangle ++(0.262,0.219);
\node[font=\fontsize{7.2}{8.2}\selectfont,inner sep=.25pt] at (6.604,8.209) {0.030};
\node[anchor=north east,rotate=45,font=\fontsize{7.2}{8.2}\selectfont,inner sep=.25pt] at (6.604,7.500) {@8};
\fill[GNGray] (6.941,7.600) rectangle ++(0.262,0.326);
\node[font=\fontsize{7.2}{8.2}\selectfont,inner sep=.25pt] at (7.073,8.576) {0.045};
\node[anchor=north east,rotate=45,font=\fontsize{7.2}{8.2}\selectfont,inner sep=.25pt] at (7.073,7.500) {@10};
\fill[GNOrange] (7.410,7.600) rectangle ++(0.262,1.613);
\node[font=\fontsize{7.2}{8.2}\selectfont,inner sep=.25pt] at (7.541,9.343) {0.222};
\node[anchor=north east,rotate=45,font=\fontsize{7.2}{8.2}\selectfont,inner sep=.25pt] at (7.541,7.500) {Final};
\node[font=\fontsize{8}{9}\selectfont\bfseries,inner sep=0pt] at (10.450,10.060) {Block 3};
\draw[black!75,line width=.30pt] (8.810,7.600) -- (12.090,7.600);
\draw[black!75,line width=.30pt] (8.810,7.600) -- (8.810,9.780);
\draw[black!12,line width=.20pt] (8.810,7.963) -- (12.090,7.963);
\draw[black!12,line width=.20pt] (8.810,8.327) -- (12.090,8.327);
\draw[black!12,line width=.20pt] (8.810,8.690) -- (12.090,8.690);
\draw[black!12,line width=.20pt] (8.810,9.053) -- (12.090,9.053);
\draw[black!12,line width=.20pt] (8.810,9.417) -- (12.090,9.417);
\draw[black!12,line width=.20pt] (8.810,9.780) -- (12.090,9.780);
\fill[GNGreen] (9.204,7.600) rectangle ++(0.306,0.270);
\node[font=\fontsize{7.2}{8.2}\selectfont,inner sep=.25pt] at (9.357,8.000) {0.037};
\node[anchor=north east,rotate=45,font=\fontsize{7.2}{8.2}\selectfont,inner sep=.25pt] at (9.357,7.500) {@4};
\fill[GNPink] (9.751,7.600) rectangle ++(0.306,0.200);
\node[font=\fontsize{7.2}{8.2}\selectfont,inner sep=.25pt] at (9.903,8.450) {0.028};
\node[anchor=north east,rotate=45,font=\fontsize{7.2}{8.2}\selectfont,inner sep=.25pt] at (9.903,7.500) {@6};
\fill[GNOchre] (10.297,7.600) rectangle ++(0.306,0.201);
\node[font=\fontsize{7.2}{8.2}\selectfont,inner sep=.25pt] at (10.450,7.931) {0.028};
\node[anchor=north east,rotate=45,font=\fontsize{7.2}{8.2}\selectfont,inner sep=.25pt] at (10.450,7.500) {@8};
\fill[GNGray] (10.844,7.600) rectangle ++(0.306,0.258);
\node[font=\fontsize{7.2}{8.2}\selectfont,inner sep=.25pt] at (10.997,8.248) {0.036};
\node[anchor=north east,rotate=45,font=\fontsize{7.2}{8.2}\selectfont,inner sep=.25pt] at (10.997,7.500) {@10};
\fill[GNOrange] (11.391,7.600) rectangle ++(0.306,1.424);
\node[font=\fontsize{7.2}{8.2}\selectfont,inner sep=.25pt] at (11.543,9.154) {0.196};
\node[anchor=north east,rotate=45,font=\fontsize{7.2}{8.2}\selectfont,inner sep=.25pt] at (11.543,7.500) {Final};
\node[font=\fontsize{8}{9}\selectfont\bfseries,inner sep=0pt] at (14.530,10.060) {Block 4};
\draw[black!75,line width=.30pt] (12.890,7.600) -- (16.170,7.600);
\draw[black!75,line width=.30pt] (12.890,7.600) -- (12.890,9.780);
\draw[black!12,line width=.20pt] (12.890,7.963) -- (16.170,7.963);
\draw[black!12,line width=.20pt] (12.890,8.327) -- (16.170,8.327);
\draw[black!12,line width=.20pt] (12.890,8.690) -- (16.170,8.690);
\draw[black!12,line width=.20pt] (12.890,9.053) -- (16.170,9.053);
\draw[black!12,line width=.20pt] (12.890,9.417) -- (16.170,9.417);
\draw[black!12,line width=.20pt] (12.890,9.780) -- (16.170,9.780);
\fill[GNGreen] (13.284,7.600) rectangle ++(0.306,0.244);
\node[font=\fontsize{7.2}{8.2}\selectfont,inner sep=.25pt] at (13.437,7.974) {0.034};
\node[anchor=north east,rotate=45,font=\fontsize{7.2}{8.2}\selectfont,inner sep=.25pt] at (13.437,7.500) {@4};
\fill[GNPink] (13.831,7.600) rectangle ++(0.306,0.209);
\node[font=\fontsize{7.2}{8.2}\selectfont,inner sep=.25pt] at (13.983,8.459) {0.029};
\node[anchor=north east,rotate=45,font=\fontsize{7.2}{8.2}\selectfont,inner sep=.25pt] at (13.983,7.500) {@6};
\fill[GNOchre] (14.377,7.600) rectangle ++(0.306,0.195);
\node[font=\fontsize{7.2}{8.2}\selectfont,inner sep=.25pt] at (14.530,7.925) {0.027};
\node[anchor=north east,rotate=45,font=\fontsize{7.2}{8.2}\selectfont,inner sep=.25pt] at (14.530,7.500) {@8};
\fill[GNGray] (14.924,7.600) rectangle ++(0.306,0.244);
\node[font=\fontsize{7.2}{8.2}\selectfont,inner sep=.25pt] at (15.077,8.234) {0.034};
\node[anchor=north east,rotate=45,font=\fontsize{7.2}{8.2}\selectfont,inner sep=.25pt] at (15.077,7.500) {@10};
\fill[GNOrange] (15.471,7.600) rectangle ++(0.306,1.381);
\node[font=\fontsize{7.2}{8.2}\selectfont,inner sep=.25pt] at (15.623,9.111) {0.190};
\node[anchor=north east,rotate=45,font=\fontsize{7.2}{8.2}\selectfont,inner sep=.25pt] at (15.623,7.500) {Final};
\node[anchor=east,font=\fontsize{7.2}{8.2}\selectfont,inner sep=.4pt] at (0.592,4.563) {0.05};
\node[anchor=east,font=\fontsize{7.2}{8.2}\selectfont,inner sep=.4pt] at (0.592,4.927) {0.10};
\node[anchor=east,font=\fontsize{7.2}{8.2}\selectfont,inner sep=.4pt] at (0.592,5.290) {0.15};
\node[anchor=east,font=\fontsize{7.2}{8.2}\selectfont,inner sep=.4pt] at (0.592,5.653) {0.20};
\node[anchor=east,font=\fontsize{7.2}{8.2}\selectfont,inner sep=.4pt] at (0.592,6.017) {0.25};
\node[anchor=east,font=\fontsize{7.2}{8.2}\selectfont,inner sep=.4pt] at (0.592,6.380) {0.30};
\node[anchor=east,font=\fontsize{7.2}{8.2}\selectfont,inner sep=.4pt] at (0.592,4.200) {0.00};
\node[font=\fontsize{8}{9}\selectfont\bfseries,inner sep=0pt] at (2.290,6.660) {Block 5};
\draw[black!75,line width=.30pt] (0.650,4.200) -- (3.930,4.200);
\draw[black!75,line width=.30pt] (0.650,4.200) -- (0.650,6.380);
\draw[black!12,line width=.20pt] (0.650,4.563) -- (3.930,4.563);
\draw[black!12,line width=.20pt] (0.650,4.927) -- (3.930,4.927);
\draw[black!12,line width=.20pt] (0.650,5.290) -- (3.930,5.290);
\draw[black!12,line width=.20pt] (0.650,5.653) -- (3.930,5.653);
\draw[black!12,line width=.20pt] (0.650,6.017) -- (3.930,6.017);
\draw[black!12,line width=.20pt] (0.650,6.380) -- (3.930,6.380);
\fill[GNPink] (1.147,4.200) rectangle ++(0.317,0.209);
\node[font=\fontsize{7.2}{8.2}\selectfont,inner sep=.25pt] at (1.306,4.539) {0.029};
\node[anchor=north east,rotate=45,font=\fontsize{7.2}{8.2}\selectfont,inner sep=.25pt] at (1.306,4.100) {@6};
\fill[GNOchre] (1.803,4.200) rectangle ++(0.317,0.195);
\node[font=\fontsize{7.2}{8.2}\selectfont,inner sep=.25pt] at (1.962,5.045) {0.027};
\node[anchor=north east,rotate=45,font=\fontsize{7.2}{8.2}\selectfont,inner sep=.25pt] at (1.962,4.100) {@8};
\fill[GNGray] (2.459,4.200) rectangle ++(0.317,0.232);
\node[font=\fontsize{7.2}{8.2}\selectfont,inner sep=.25pt] at (2.618,4.562) {0.032};
\node[anchor=north east,rotate=45,font=\fontsize{7.2}{8.2}\selectfont,inner sep=.25pt] at (2.618,4.100) {@10};
\fill[GNOrange] (3.115,4.200) rectangle ++(0.317,1.337);
\node[font=\fontsize{7.2}{8.2}\selectfont,inner sep=.25pt] at (3.274,5.667) {0.184};
\node[anchor=north east,rotate=45,font=\fontsize{7.2}{8.2}\selectfont,inner sep=.25pt] at (3.274,4.100) {Final};
\node[font=\fontsize{8}{9}\selectfont\bfseries,inner sep=0pt] at (6.370,6.660) {Block 6};
\draw[black!75,line width=.30pt] (4.730,4.200) -- (8.010,4.200);
\draw[black!75,line width=.30pt] (4.730,4.200) -- (4.730,6.380);
\draw[black!12,line width=.20pt] (4.730,4.563) -- (8.010,4.563);
\draw[black!12,line width=.20pt] (4.730,4.927) -- (8.010,4.927);
\draw[black!12,line width=.20pt] (4.730,5.290) -- (8.010,5.290);
\draw[black!12,line width=.20pt] (4.730,5.653) -- (8.010,5.653);
\draw[black!12,line width=.20pt] (4.730,6.017) -- (8.010,6.017);
\draw[black!12,line width=.20pt] (4.730,6.380) -- (8.010,6.380);
\fill[GNPink] (5.227,4.200) rectangle ++(0.317,0.193);
\node[font=\fontsize{7.2}{8.2}\selectfont,inner sep=.25pt] at (5.386,4.523) {0.027};
\node[anchor=north east,rotate=45,font=\fontsize{7.2}{8.2}\selectfont,inner sep=.25pt] at (5.386,4.100) {@6};
\fill[GNOchre] (5.883,4.200) rectangle ++(0.317,0.220);
\node[font=\fontsize{7.2}{8.2}\selectfont,inner sep=.25pt] at (6.042,4.810) {0.030};
\node[anchor=north east,rotate=45,font=\fontsize{7.2}{8.2}\selectfont,inner sep=.25pt] at (6.042,4.100) {@8};
\fill[GNGray] (6.539,4.200) rectangle ++(0.317,0.235);
\node[font=\fontsize{7.2}{8.2}\selectfont,inner sep=.25pt] at (6.698,5.085) {0.032};
\node[anchor=north east,rotate=45,font=\fontsize{7.2}{8.2}\selectfont,inner sep=.25pt] at (6.698,4.100) {@10};
\fill[GNOrange] (7.195,4.200) rectangle ++(0.317,1.323);
\node[font=\fontsize{7.2}{8.2}\selectfont,inner sep=.25pt] at (7.354,5.653) {0.182};
\node[anchor=north east,rotate=45,font=\fontsize{7.2}{8.2}\selectfont,inner sep=.25pt] at (7.354,4.100) {Final};
\node[font=\fontsize{8}{9}\selectfont\bfseries,inner sep=0pt] at (10.450,6.660) {Block 7};
\draw[black!75,line width=.30pt] (8.810,4.200) -- (12.090,4.200);
\draw[black!75,line width=.30pt] (8.810,4.200) -- (8.810,6.380);
\draw[black!12,line width=.20pt] (8.810,4.563) -- (12.090,4.563);
\draw[black!12,line width=.20pt] (8.810,4.927) -- (12.090,4.927);
\draw[black!12,line width=.20pt] (8.810,5.290) -- (12.090,5.290);
\draw[black!12,line width=.20pt] (8.810,5.653) -- (12.090,5.653);
\draw[black!12,line width=.20pt] (8.810,6.017) -- (12.090,6.017);
\draw[black!12,line width=.20pt] (8.810,6.380) -- (12.090,6.380);
\fill[GNOchre] (9.471,4.200) rectangle ++(0.317,0.225);
\node[font=\fontsize{7.2}{8.2}\selectfont,inner sep=.25pt] at (9.630,4.555) {0.031};
\node[anchor=north east,rotate=45,font=\fontsize{7.2}{8.2}\selectfont,inner sep=.25pt] at (9.630,4.100) {@8};
\fill[GNGray] (10.291,4.200) rectangle ++(0.317,0.278);
\node[font=\fontsize{7.2}{8.2}\selectfont,inner sep=.25pt] at (10.450,4.608) {0.038};
\node[anchor=north east,rotate=45,font=\fontsize{7.2}{8.2}\selectfont,inner sep=.25pt] at (10.450,4.100) {@10};
\fill[GNOrange] (11.111,4.200) rectangle ++(0.317,1.308);
\node[font=\fontsize{7.2}{8.2}\selectfont,inner sep=.25pt] at (11.270,5.638) {0.180};
\node[anchor=north east,rotate=45,font=\fontsize{7.2}{8.2}\selectfont,inner sep=.25pt] at (11.270,4.100) {Final};
\node[font=\fontsize{8}{9}\selectfont\bfseries,inner sep=0pt] at (14.530,6.660) {Block 8};
\draw[black!75,line width=.30pt] (12.890,4.200) -- (16.170,4.200);
\draw[black!75,line width=.30pt] (12.890,4.200) -- (12.890,6.380);
\draw[black!12,line width=.20pt] (12.890,4.563) -- (16.170,4.563);
\draw[black!12,line width=.20pt] (12.890,4.927) -- (16.170,4.927);
\draw[black!12,line width=.20pt] (12.890,5.290) -- (16.170,5.290);
\draw[black!12,line width=.20pt] (12.890,5.653) -- (16.170,5.653);
\draw[black!12,line width=.20pt] (12.890,6.017) -- (16.170,6.017);
\draw[black!12,line width=.20pt] (12.890,6.380) -- (16.170,6.380);
\fill[GNOchre] (13.551,4.200) rectangle ++(0.317,0.418);
\node[font=\fontsize{7.2}{8.2}\selectfont,inner sep=.25pt] at (13.710,4.748) {0.058};
\node[anchor=north east,rotate=45,font=\fontsize{7.2}{8.2}\selectfont,inner sep=.25pt] at (13.710,4.100) {@8};
\fill[GNGray] (14.371,4.200) rectangle ++(0.317,0.237);
\node[font=\fontsize{7.2}{8.2}\selectfont,inner sep=.25pt] at (14.530,4.567) {0.033};
\node[anchor=north east,rotate=45,font=\fontsize{7.2}{8.2}\selectfont,inner sep=.25pt] at (14.530,4.100) {@10};
\fill[GNOrange] (15.191,4.200) rectangle ++(0.317,1.279);
\node[font=\fontsize{7.2}{8.2}\selectfont,inner sep=.25pt] at (15.350,5.609) {0.176};
\node[anchor=north east,rotate=45,font=\fontsize{7.2}{8.2}\selectfont,inner sep=.25pt] at (15.350,4.100) {Final};
\node[anchor=east,font=\fontsize{7.2}{8.2}\selectfont,inner sep=.4pt] at (4.672,1.163) {0.05};
\node[anchor=east,font=\fontsize{7.2}{8.2}\selectfont,inner sep=.4pt] at (4.672,1.527) {0.10};
\node[anchor=east,font=\fontsize{7.2}{8.2}\selectfont,inner sep=.4pt] at (4.672,1.890) {0.15};
\node[anchor=east,font=\fontsize{7.2}{8.2}\selectfont,inner sep=.4pt] at (4.672,2.253) {0.20};
\node[anchor=east,font=\fontsize{7.2}{8.2}\selectfont,inner sep=.4pt] at (4.672,2.617) {0.25};
\node[anchor=east,font=\fontsize{7.2}{8.2}\selectfont,inner sep=.4pt] at (4.672,2.980) {0.30};
\node[anchor=east,font=\fontsize{7.2}{8.2}\selectfont,inner sep=.4pt] at (4.672,0.800) {0.00};
\node[font=\fontsize{8}{9}\selectfont\bfseries,inner sep=0pt] at (6.370,3.260) {Block 9};
\draw[black!75,line width=.30pt] (4.730,0.800) -- (8.010,0.800);
\draw[black!75,line width=.30pt] (4.730,0.800) -- (4.730,2.980);
\draw[black!12,line width=.20pt] (4.730,1.163) -- (8.010,1.163);
\draw[black!12,line width=.20pt] (4.730,1.527) -- (8.010,1.527);
\draw[black!12,line width=.20pt] (4.730,1.890) -- (8.010,1.890);
\draw[black!12,line width=.20pt] (4.730,2.253) -- (8.010,2.253);
\draw[black!12,line width=.20pt] (4.730,2.617) -- (8.010,2.617);
\draw[black!12,line width=.20pt] (4.730,2.980) -- (8.010,2.980);
\fill[GNGray] (5.664,0.800) rectangle ++(0.317,0.292);
\node[font=\fontsize{7.2}{8.2}\selectfont,inner sep=.25pt] at (5.823,1.222) {0.040};
\node[anchor=north east,rotate=45,font=\fontsize{7.2}{8.2}\selectfont,inner sep=.25pt] at (5.823,0.700) {@10};
\fill[GNOrange] (6.758,0.800) rectangle ++(0.317,1.250);
\node[font=\fontsize{7.2}{8.2}\selectfont,inner sep=.25pt] at (6.917,2.180) {0.172};
\node[anchor=north east,rotate=45,font=\fontsize{7.2}{8.2}\selectfont,inner sep=.25pt] at (6.917,0.700) {Final};
\node[font=\fontsize{8}{9}\selectfont\bfseries,inner sep=0pt] at (10.450,3.260) {Block 10};
\draw[black!75,line width=.30pt] (8.810,0.800) -- (12.090,0.800);
\draw[black!75,line width=.30pt] (8.810,0.800) -- (8.810,2.980);
\draw[black!12,line width=.20pt] (8.810,1.163) -- (12.090,1.163);
\draw[black!12,line width=.20pt] (8.810,1.527) -- (12.090,1.527);
\draw[black!12,line width=.20pt] (8.810,1.890) -- (12.090,1.890);
\draw[black!12,line width=.20pt] (8.810,2.253) -- (12.090,2.253);
\draw[black!12,line width=.20pt] (8.810,2.617) -- (12.090,2.617);
\draw[black!12,line width=.20pt] (8.810,2.980) -- (12.090,2.980);
\fill[GNGray] (9.744,0.800) rectangle ++(0.317,0.268);
\node[font=\fontsize{7.2}{8.2}\selectfont,inner sep=.25pt] at (9.903,1.198) {0.037};
\node[anchor=north east,rotate=45,font=\fontsize{7.2}{8.2}\selectfont,inner sep=.25pt] at (9.903,0.700) {@10};
\fill[GNOrange] (10.838,0.800) rectangle ++(0.317,1.250);
\node[font=\fontsize{7.2}{8.2}\selectfont,inner sep=.25pt] at (10.997,2.180) {0.172};
\node[anchor=north east,rotate=45,font=\fontsize{7.2}{8.2}\selectfont,inner sep=.25pt] at (10.997,0.700) {Final};
\end{tikzpicture}
\endgroup

%% file: artwork/gradient_conflict/further/sdn_gradient_norm_epoch010.tex
\begingroup
\definecolor{GNBlue}{RGB}{0,114,178}
\definecolor{GNGreen}{RGB}{0,158,115}
\definecolor{GNPink}{RGB}{204,121,167}
\definecolor{GNOchre}{RGB}{181,137,0}
\definecolor{GNGray}{RGB}{102,102,102}
\definecolor{GNOrange}{RGB}{213,94,0}
\begin{tikzpicture}[x=1cm,y=1.05cm]
\path[use as bounding box] (-0.10,0.00) rectangle (16.70,10.90);
\node[font=\fontsize{9.4}{10.4}\selectfont\bfseries,inner sep=0pt] at (8.30,10.62) {(B) Loss-weighted gradient norms};
\node[font=\fontsize{7.5}{8.5}\selectfont,inner sep=0pt] at (8.30,0.12) {Exit objective};
\node[anchor=east,font=\fontsize{7.2}{8.2}\selectfont,inner sep=.4pt] at (0.592,8.036) {0.05};
\node[anchor=east,font=\fontsize{7.2}{8.2}\selectfont,inner sep=.4pt] at (0.592,8.472) {0.10};
\node[anchor=east,font=\fontsize{7.2}{8.2}\selectfont,inner sep=.4pt] at (0.592,8.908) {0.15};
\node[anchor=east,font=\fontsize{7.2}{8.2}\selectfont,inner sep=.4pt] at (0.592,9.344) {0.20};
\node[anchor=east,font=\fontsize{7.2}{8.2}\selectfont,inner sep=.4pt] at (0.592,9.780) {0.25};
\node[anchor=east,font=\fontsize{7.2}{8.2}\selectfont,inner sep=.4pt] at (0.592,7.600) {0.00};
\node[font=\fontsize{8}{9}\selectfont\bfseries,inner sep=0pt] at (2.290,10.060) {Block 1};
\draw[black!75,line width=.30pt] (0.650,7.600) -- (3.930,7.600);
\draw[black!75,line width=.30pt] (0.650,7.600) -- (0.650,9.780);
\draw[black!12,line width=.20pt] (0.650,8.036) -- (3.930,8.036);
\draw[black!12,line width=.20pt] (0.650,8.472) -- (3.930,8.472);
\draw[black!12,line width=.20pt] (0.650,8.908) -- (3.930,8.908);
\draw[black!12,line width=.20pt] (0.650,9.344) -- (3.930,9.344);
\draw[black!12,line width=.20pt] (0.650,9.780) -- (3.930,9.780);
\fill[GNBlue] (0.988,7.600) rectangle ++(0.262,1.953);
\node[font=\fontsize{7.2}{8.2}\selectfont,inner sep=.25pt] at (1.119,9.683) {0.224};
\node[anchor=north east,rotate=45,font=\fontsize{7.2}{8.2}\selectfont,inner sep=.25pt] at (1.119,7.500) {@2};
\fill[GNGreen] (1.456,7.600) rectangle ++(0.262,0.850);
\node[font=\fontsize{7.2}{8.2}\selectfont,inner sep=.25pt] at (1.587,8.580) {0.098};
\node[anchor=north east,rotate=45,font=\fontsize{7.2}{8.2}\selectfont,inner sep=.25pt] at (1.587,7.500) {@4};
\fill[GNPink] (1.925,7.600) rectangle ++(0.262,0.542);
\node[font=\fontsize{7.2}{8.2}\selectfont,inner sep=.25pt] at (2.056,8.272) {0.062};
\node[anchor=north east,rotate=45,font=\fontsize{7.2}{8.2}\selectfont,inner sep=.25pt] at (2.056,7.500) {@6};
\fill[GNOchre] (2.393,7.600) rectangle ++(0.262,0.448);
\node[font=\fontsize{7.2}{8.2}\selectfont,inner sep=.25pt] at (2.524,8.698) {0.051};
\node[anchor=north east,rotate=45,font=\fontsize{7.2}{8.2}\selectfont,inner sep=.25pt] at (2.524,7.500) {@8};
\fill[GNGray] (2.861,7.600) rectangle ++(0.262,0.426);
\node[font=\fontsize{7.2}{8.2}\selectfont,inner sep=.25pt] at (2.993,8.156) {0.049};
\node[anchor=north east,rotate=45,font=\fontsize{7.2}{8.2}\selectfont,inner sep=.25pt] at (2.993,7.500) {@10};
\fill[GNOrange] (3.330,7.600) rectangle ++(0.262,1.945);
\node[font=\fontsize{7.2}{8.2}\selectfont,inner sep=.25pt] at (3.461,9.675) {0.223};
\node[anchor=north east,rotate=45,font=\fontsize{7.2}{8.2}\selectfont,inner sep=.25pt] at (3.461,7.500) {Final};
\node[font=\fontsize{8}{9}\selectfont\bfseries,inner sep=0pt] at (6.370,10.060) {Block 2};
\draw[black!75,line width=.30pt] (4.730,7.600) -- (8.010,7.600);
\draw[black!75,line width=.30pt] (4.730,7.600) -- (4.730,9.780);
\draw[black!12,line width=.20pt] (4.730,8.036) -- (8.010,8.036);
\draw[black!12,line width=.20pt] (4.730,8.472) -- (8.010,8.472);
\draw[black!12,line width=.20pt] (4.730,8.908) -- (8.010,8.908);
\draw[black!12,line width=.20pt] (4.730,9.344) -- (8.010,9.344);
\draw[black!12,line width=.20pt] (4.730,9.780) -- (8.010,9.780);
\fill[GNBlue] (5.068,7.600) rectangle ++(0.262,1.901);
\node[font=\fontsize{7.2}{8.2}\selectfont,inner sep=.25pt] at (5.199,9.631) {0.218};
\node[anchor=north east,rotate=45,font=\fontsize{7.2}{8.2}\selectfont,inner sep=.25pt] at (5.199,7.500) {@2};
\fill[GNGreen] (5.536,7.600) rectangle ++(0.262,0.916);
\node[font=\fontsize{7.2}{8.2}\selectfont,inner sep=.25pt] at (5.667,8.646) {0.105};
\node[anchor=north east,rotate=45,font=\fontsize{7.2}{8.2}\selectfont,inner sep=.25pt] at (5.667,7.500) {@4};
\fill[GNPink] (6.005,7.600) rectangle ++(0.262,0.527);
\node[font=\fontsize{7.2}{8.2}\selectfont,inner sep=.25pt] at (6.136,8.257) {0.060};
\node[anchor=north east,rotate=45,font=\fontsize{7.2}{8.2}\selectfont,inner sep=.25pt] at (6.136,7.500) {@6};
\fill[GNOchre] (6.473,7.600) rectangle ++(0.262,0.394);
\node[font=\fontsize{7.2}{8.2}\selectfont,inner sep=.25pt] at (6.604,8.644) {0.045};
\node[anchor=north east,rotate=45,font=\fontsize{7.2}{8.2}\selectfont,inner sep=.25pt] at (6.604,7.500) {@8};
\fill[GNGray] (6.941,7.600) rectangle ++(0.262,0.358);
\node[font=\fontsize{7.2}{8.2}\selectfont,inner sep=.25pt] at (7.073,8.088) {0.041};
\node[anchor=north east,rotate=45,font=\fontsize{7.2}{8.2}\selectfont,inner sep=.25pt] at (7.073,7.500) {@10};
\fill[GNOrange] (7.410,7.600) rectangle ++(0.262,1.570);
\node[font=\fontsize{7.2}{8.2}\selectfont,inner sep=.25pt] at (7.541,9.300) {0.180};
\node[anchor=north east,rotate=45,font=\fontsize{7.2}{8.2}\selectfont,inner sep=.25pt] at (7.541,7.500) {Final};
\node[font=\fontsize{8}{9}\selectfont\bfseries,inner sep=0pt] at (10.450,10.060) {Block 3};
\draw[black!75,line width=.30pt] (8.810,7.600) -- (12.090,7.600);
\draw[black!75,line width=.30pt] (8.810,7.600) -- (8.810,9.780);
\draw[black!12,line width=.20pt] (8.810,8.036) -- (12.090,8.036);
\draw[black!12,line width=.20pt] (8.810,8.472) -- (12.090,8.472);
\draw[black!12,line width=.20pt] (8.810,8.908) -- (12.090,8.908);
\draw[black!12,line width=.20pt] (8.810,9.344) -- (12.090,9.344);
\draw[black!12,line width=.20pt] (8.810,9.780) -- (12.090,9.780);
\fill[GNGreen] (9.204,7.600) rectangle ++(0.306,0.942);
\node[font=\fontsize{7.2}{8.2}\selectfont,inner sep=.25pt] at (9.357,8.672) {0.108};
\node[anchor=north east,rotate=45,font=\fontsize{7.2}{8.2}\selectfont,inner sep=.25pt] at (9.357,7.500) {@4};
\fill[GNPink] (9.751,7.600) rectangle ++(0.306,0.521);
\node[font=\fontsize{7.2}{8.2}\selectfont,inner sep=.25pt] at (9.903,8.251) {0.060};
\node[anchor=north east,rotate=45,font=\fontsize{7.2}{8.2}\selectfont,inner sep=.25pt] at (9.903,7.500) {@6};
\fill[GNOchre] (10.297,7.600) rectangle ++(0.306,0.383);
\node[font=\fontsize{7.2}{8.2}\selectfont,inner sep=.25pt] at (10.450,8.633) {0.044};
\node[anchor=north east,rotate=45,font=\fontsize{7.2}{8.2}\selectfont,inner sep=.25pt] at (10.450,7.500) {@8};
\fill[GNGray] (10.844,7.600) rectangle ++(0.306,0.344);
\node[font=\fontsize{7.2}{8.2}\selectfont,inner sep=.25pt] at (10.997,8.074) {0.040};
\node[anchor=north east,rotate=45,font=\fontsize{7.2}{8.2}\selectfont,inner sep=.25pt] at (10.997,7.500) {@10};
\fill[GNOrange] (11.391,7.600) rectangle ++(0.306,1.491);
\node[font=\fontsize{7.2}{8.2}\selectfont,inner sep=.25pt] at (11.543,9.221) {0.171};
\node[anchor=north east,rotate=45,font=\fontsize{7.2}{8.2}\selectfont,inner sep=.25pt] at (11.543,7.500) {Final};
\node[font=\fontsize{8}{9}\selectfont\bfseries,inner sep=0pt] at (14.530,10.060) {Block 4};
\draw[black!75,line width=.30pt] (12.890,7.600) -- (16.170,7.600);
\draw[black!75,line width=.30pt] (12.890,7.600) -- (12.890,9.780);
\draw[black!12,line width=.20pt] (12.890,8.036) -- (16.170,8.036);
\draw[black!12,line width=.20pt] (12.890,8.472) -- (16.170,8.472);
\draw[black!12,line width=.20pt] (12.890,8.908) -- (16.170,8.908);
\draw[black!12,line width=.20pt] (12.890,9.344) -- (16.170,9.344);
\draw[black!12,line width=.20pt] (12.890,9.780) -- (16.170,9.780);
\fill[GNGreen] (13.284,7.600) rectangle ++(0.306,0.933);
\node[font=\fontsize{7.2}{8.2}\selectfont,inner sep=.25pt] at (13.437,8.663) {0.107};
\node[anchor=north east,rotate=45,font=\fontsize{7.2}{8.2}\selectfont,inner sep=.25pt] at (13.437,7.500) {@4};
\fill[GNPink] (13.831,7.600) rectangle ++(0.306,0.561);
\node[font=\fontsize{7.2}{8.2}\selectfont,inner sep=.25pt] at (13.983,8.291) {0.064};
\node[anchor=north east,rotate=45,font=\fontsize{7.2}{8.2}\selectfont,inner sep=.25pt] at (13.983,7.500) {@6};
\fill[GNOchre] (14.377,7.600) rectangle ++(0.306,0.396);
\node[font=\fontsize{7.2}{8.2}\selectfont,inner sep=.25pt] at (14.530,8.646) {0.045};
\node[anchor=north east,rotate=45,font=\fontsize{7.2}{8.2}\selectfont,inner sep=.25pt] at (14.530,7.500) {@8};
\fill[GNGray] (14.924,7.600) rectangle ++(0.306,0.349);
\node[font=\fontsize{7.2}{8.2}\selectfont,inner sep=.25pt] at (15.077,8.079) {0.040};
\node[anchor=north east,rotate=45,font=\fontsize{7.2}{8.2}\selectfont,inner sep=.25pt] at (15.077,7.500) {@10};
\fill[GNOrange] (15.471,7.600) rectangle ++(0.306,1.474);
\node[font=\fontsize{7.2}{8.2}\selectfont,inner sep=.25pt] at (15.623,9.204) {0.169};
\node[anchor=north east,rotate=45,font=\fontsize{7.2}{8.2}\selectfont,inner sep=.25pt] at (15.623,7.500) {Final};
\node[anchor=east,font=\fontsize{7.2}{8.2}\selectfont,inner sep=.4pt] at (0.592,4.636) {0.05};
\node[anchor=east,font=\fontsize{7.2}{8.2}\selectfont,inner sep=.4pt] at (0.592,5.072) {0.10};
\node[anchor=east,font=\fontsize{7.2}{8.2}\selectfont,inner sep=.4pt] at (0.592,5.508) {0.15};
\node[anchor=east,font=\fontsize{7.2}{8.2}\selectfont,inner sep=.4pt] at (0.592,5.944) {0.20};
\node[anchor=east,font=\fontsize{7.2}{8.2}\selectfont,inner sep=.4pt] at (0.592,6.380) {0.25};
\node[anchor=east,font=\fontsize{7.2}{8.2}\selectfont,inner sep=.4pt] at (0.592,4.200) {0.00};
\node[font=\fontsize{8}{9}\selectfont\bfseries,inner sep=0pt] at (2.290,6.660) {Block 5};
\draw[black!75,line width=.30pt] (0.650,4.200) -- (3.930,4.200);
\draw[black!75,line width=.30pt] (0.650,4.200) -- (0.650,6.380);
\draw[black!12,line width=.20pt] (0.650,4.636) -- (3.930,4.636);
\draw[black!12,line width=.20pt] (0.650,5.072) -- (3.930,5.072);
\draw[black!12,line width=.20pt] (0.650,5.508) -- (3.930,5.508);
\draw[black!12,line width=.20pt] (0.650,5.944) -- (3.930,5.944);
\draw[black!12,line width=.20pt] (0.650,6.380) -- (3.930,6.380);
\fill[GNPink] (1.147,4.200) rectangle ++(0.317,0.548);
\node[font=\fontsize{7.2}{8.2}\selectfont,inner sep=.25pt] at (1.306,4.878) {0.063};
\node[anchor=north east,rotate=45,font=\fontsize{7.2}{8.2}\selectfont,inner sep=.25pt] at (1.306,4.100) {@6};
\fill[GNOchre] (1.803,4.200) rectangle ++(0.317,0.389);
\node[font=\fontsize{7.2}{8.2}\selectfont,inner sep=.25pt] at (1.962,5.239) {0.045};
\node[anchor=north east,rotate=45,font=\fontsize{7.2}{8.2}\selectfont,inner sep=.25pt] at (1.962,4.100) {@8};
\fill[GNGray] (2.459,4.200) rectangle ++(0.317,0.345);
\node[font=\fontsize{7.2}{8.2}\selectfont,inner sep=.25pt] at (2.618,4.675) {0.040};
\node[anchor=north east,rotate=45,font=\fontsize{7.2}{8.2}\selectfont,inner sep=.25pt] at (2.618,4.100) {@10};
\fill[GNOrange] (3.115,4.200) rectangle ++(0.317,1.491);
\node[font=\fontsize{7.2}{8.2}\selectfont,inner sep=.25pt] at (3.274,5.821) {0.171};
\node[anchor=north east,rotate=45,font=\fontsize{7.2}{8.2}\selectfont,inner sep=.25pt] at (3.274,4.100) {Final};
\node[font=\fontsize{8}{9}\selectfont\bfseries,inner sep=0pt] at (6.370,6.660) {Block 6};
\draw[black!75,line width=.30pt] (4.730,4.200) -- (8.010,4.200);
\draw[black!75,line width=.30pt] (4.730,4.200) -- (4.730,6.380);
\draw[black!12,line width=.20pt] (4.730,4.636) -- (8.010,4.636);
\draw[black!12,line width=.20pt] (4.730,5.072) -- (8.010,5.072);
\draw[black!12,line width=.20pt] (4.730,5.508) -- (8.010,5.508);
\draw[black!12,line width=.20pt] (4.730,5.944) -- (8.010,5.944);
\draw[black!12,line width=.20pt] (4.730,6.380) -- (8.010,6.380);
\fill[GNPink] (5.227,4.200) rectangle ++(0.317,0.548);
\node[font=\fontsize{7.2}{8.2}\selectfont,inner sep=.25pt] at (5.386,4.878) {0.063};
\node[anchor=north east,rotate=45,font=\fontsize{7.2}{8.2}\selectfont,inner sep=.25pt] at (5.386,4.100) {@6};
\fill[GNOchre] (5.883,4.200) rectangle ++(0.317,0.385);
\node[font=\fontsize{7.2}{8.2}\selectfont,inner sep=.25pt] at (6.042,5.235) {0.044};
\node[anchor=north east,rotate=45,font=\fontsize{7.2}{8.2}\selectfont,inner sep=.25pt] at (6.042,4.100) {@8};
\fill[GNGray] (6.539,4.200) rectangle ++(0.317,0.334);
\node[font=\fontsize{7.2}{8.2}\selectfont,inner sep=.25pt] at (6.698,4.664) {0.038};
\node[anchor=north east,rotate=45,font=\fontsize{7.2}{8.2}\selectfont,inner sep=.25pt] at (6.698,4.100) {@10};
\fill[GNOrange] (7.195,4.200) rectangle ++(0.317,1.439);
\node[font=\fontsize{7.2}{8.2}\selectfont,inner sep=.25pt] at (7.354,5.769) {0.165};
\node[anchor=north east,rotate=45,font=\fontsize{7.2}{8.2}\selectfont,inner sep=.25pt] at (7.354,4.100) {Final};
\node[font=\fontsize{8}{9}\selectfont\bfseries,inner sep=0pt] at (10.450,6.660) {Block 7};
\draw[black!75,line width=.30pt] (8.810,4.200) -- (12.090,4.200);
\draw[black!75,line width=.30pt] (8.810,4.200) -- (8.810,6.380);
\draw[black!12,line width=.20pt] (8.810,4.636) -- (12.090,4.636);
\draw[black!12,line width=.20pt] (8.810,5.072) -- (12.090,5.072);
\draw[black!12,line width=.20pt] (8.810,5.508) -- (12.090,5.508);
\draw[black!12,line width=.20pt] (8.810,5.944) -- (12.090,5.944);
\draw[black!12,line width=.20pt] (8.810,6.380) -- (12.090,6.380);
\fill[GNOchre] (9.471,4.200) rectangle ++(0.317,0.368);
\node[font=\fontsize{7.2}{8.2}\selectfont,inner sep=.25pt] at (9.630,4.698) {0.042};
\node[anchor=north east,rotate=45,font=\fontsize{7.2}{8.2}\selectfont,inner sep=.25pt] at (9.630,4.100) {@8};
\fill[GNGray] (10.291,4.200) rectangle ++(0.317,0.320);
\node[font=\fontsize{7.2}{8.2}\selectfont,inner sep=.25pt] at (10.450,4.650) {0.037};
\node[anchor=north east,rotate=45,font=\fontsize{7.2}{8.2}\selectfont,inner sep=.25pt] at (10.450,4.100) {@10};
\fill[GNOrange] (11.111,4.200) rectangle ++(0.317,1.413);
\node[font=\fontsize{7.2}{8.2}\selectfont,inner sep=.25pt] at (11.270,5.743) {0.162};
\node[anchor=north east,rotate=45,font=\fontsize{7.2}{8.2}\selectfont,inner sep=.25pt] at (11.270,4.100) {Final};
\node[font=\fontsize{8}{9}\selectfont\bfseries,inner sep=0pt] at (14.530,6.660) {Block 8};
\draw[black!75,line width=.30pt] (12.890,4.200) -- (16.170,4.200);
\draw[black!75,line width=.30pt] (12.890,4.200) -- (12.890,6.380);
\draw[black!12,line width=.20pt] (12.890,4.636) -- (16.170,4.636);
\draw[black!12,line width=.20pt] (12.890,5.072) -- (16.170,5.072);
\draw[black!12,line width=.20pt] (12.890,5.508) -- (16.170,5.508);
\draw[black!12,line width=.20pt] (12.890,5.944) -- (16.170,5.944);
\draw[black!12,line width=.20pt] (12.890,6.380) -- (16.170,6.380);
\fill[GNOchre] (13.551,4.200) rectangle ++(0.317,0.377);
\node[font=\fontsize{7.2}{8.2}\selectfont,inner sep=.25pt] at (13.710,4.707) {0.043};
\node[anchor=north east,rotate=45,font=\fontsize{7.2}{8.2}\selectfont,inner sep=.25pt] at (13.710,4.100) {@8};
\fill[GNGray] (14.371,4.200) rectangle ++(0.317,0.322);
\node[font=\fontsize{7.2}{8.2}\selectfont,inner sep=.25pt] at (14.530,4.652) {0.037};
\node[anchor=north east,rotate=45,font=\fontsize{7.2}{8.2}\selectfont,inner sep=.25pt] at (14.530,4.100) {@10};
\fill[GNOrange] (15.191,4.200) rectangle ++(0.317,1.369);
\node[font=\fontsize{7.2}{8.2}\selectfont,inner sep=.25pt] at (15.350,5.699) {0.157};
\node[anchor=north east,rotate=45,font=\fontsize{7.2}{8.2}\selectfont,inner sep=.25pt] at (15.350,4.100) {Final};
\node[anchor=east,font=\fontsize{7.2}{8.2}\selectfont,inner sep=.4pt] at (4.672,1.236) {0.05};
\node[anchor=east,font=\fontsize{7.2}{8.2}\selectfont,inner sep=.4pt] at (4.672,1.672) {0.10};
\node[anchor=east,font=\fontsize{7.2}{8.2}\selectfont,inner sep=.4pt] at (4.672,2.108) {0.15};
\node[anchor=east,font=\fontsize{7.2}{8.2}\selectfont,inner sep=.4pt] at (4.672,2.544) {0.20};
\node[anchor=east,font=\fontsize{7.2}{8.2}\selectfont,inner sep=.4pt] at (4.672,2.980) {0.25};
\node[anchor=east,font=\fontsize{7.2}{8.2}\selectfont,inner sep=.4pt] at (4.672,0.800) {0.00};
\node[font=\fontsize{8}{9}\selectfont\bfseries,inner sep=0pt] at (6.370,3.260) {Block 9};
\draw[black!75,line width=.30pt] (4.730,0.800) -- (8.010,0.800);
\draw[black!75,line width=.30pt] (4.730,0.800) -- (4.730,2.980);
\draw[black!12,line width=.20pt] (4.730,1.236) -- (8.010,1.236);
\draw[black!12,line width=.20pt] (4.730,1.672) -- (8.010,1.672);
\draw[black!12,line width=.20pt] (4.730,2.108) -- (8.010,2.108);
\draw[black!12,line width=.20pt] (4.730,2.544) -- (8.010,2.544);
\draw[black!12,line width=.20pt] (4.730,2.980) -- (8.010,2.980);
\fill[GNGray] (5.664,0.800) rectangle ++(0.317,0.300);
\node[font=\fontsize{7.2}{8.2}\selectfont,inner sep=.25pt] at (5.823,1.230) {0.034};
\node[anchor=north east,rotate=45,font=\fontsize{7.2}{8.2}\selectfont,inner sep=.25pt] at (5.823,0.700) {@10};
\fill[GNOrange] (6.758,0.800) rectangle ++(0.317,1.325);
\node[font=\fontsize{7.2}{8.2}\selectfont,inner sep=.25pt] at (6.917,2.255) {0.152};
\node[anchor=north east,rotate=45,font=\fontsize{7.2}{8.2}\selectfont,inner sep=.25pt] at (6.917,0.700) {Final};
\node[font=\fontsize{8}{9}\selectfont\bfseries,inner sep=0pt] at (10.450,3.260) {Block 10};
\draw[black!75,line width=.30pt] (8.810,0.800) -- (12.090,0.800);
\draw[black!75,line width=.30pt] (8.810,0.800) -- (8.810,2.980);
\draw[black!12,line width=.20pt] (8.810,1.236) -- (12.090,1.236);
\draw[black!12,line width=.20pt] (8.810,1.672) -- (12.090,1.672);
\draw[black!12,line width=.20pt] (8.810,2.108) -- (12.090,2.108);
\draw[black!12,line width=.20pt] (8.810,2.544) -- (12.090,2.544);
\draw[black!12,line width=.20pt] (8.810,2.980) -- (12.090,2.980);
\fill[GNGray] (9.744,0.800) rectangle ++(0.317,0.296);
\node[font=\fontsize{7.2}{8.2}\selectfont,inner sep=.25pt] at (9.903,1.226) {0.034};
\node[anchor=north east,rotate=45,font=\fontsize{7.2}{8.2}\selectfont,inner sep=.25pt] at (9.903,0.700) {@10};
\fill[GNOrange] (10.838,0.800) rectangle ++(0.317,1.308);
\node[font=\fontsize{7.2}{8.2}\selectfont,inner sep=.25pt] at (10.997,2.238) {0.150};
\node[anchor=north east,rotate=45,font=\fontsize{7.2}{8.2}\selectfont,inner sep=.25pt] at (10.997,0.700) {Final};
\end{tikzpicture}
\endgroup

%% file: artwork/gradient_conflict/further/sdn_gradient_norm_epoch100.tex
\begingroup
\definecolor{GNBlue}{RGB}{0,114,178}
\definecolor{GNGreen}{RGB}{0,158,115}
\definecolor{GNPink}{RGB}{204,121,167}
\definecolor{GNOchre}{RGB}{181,137,0}
\definecolor{GNGray}{RGB}{102,102,102}
\definecolor{GNOrange}{RGB}{213,94,0}
\begin{tikzpicture}[x=1cm,y=1.05cm]
\path[use as bounding box] (-0.10,0.00) rectangle (16.70,10.90);
\node[font=\fontsize{9.4}{10.4}\selectfont\bfseries,inner sep=0pt] at (8.30,10.62) {(B) Loss-weighted gradient norms};
\node[font=\fontsize{7.5}{8.5}\selectfont,inner sep=0pt] at (8.30,0.12) {Exit objective};
\node[anchor=east,font=\fontsize{7.2}{8.2}\selectfont,inner sep=.4pt] at (0.592,7.872) {0.05};
\node[anchor=east,font=\fontsize{7.2}{8.2}\selectfont,inner sep=.4pt] at (0.592,8.145) {0.10};
\node[anchor=east,font=\fontsize{7.2}{8.2}\selectfont,inner sep=.4pt] at (0.592,8.417) {0.15};
\node[anchor=east,font=\fontsize{7.2}{8.2}\selectfont,inner sep=.4pt] at (0.592,8.690) {0.20};
\node[anchor=east,font=\fontsize{7.2}{8.2}\selectfont,inner sep=.4pt] at (0.592,8.962) {0.25};
\node[anchor=east,font=\fontsize{7.2}{8.2}\selectfont,inner sep=.4pt] at (0.592,9.235) {0.30};
\node[anchor=east,font=\fontsize{7.2}{8.2}\selectfont,inner sep=.4pt] at (0.592,9.508) {0.35};
\node[anchor=east,font=\fontsize{7.2}{8.2}\selectfont,inner sep=.4pt] at (0.592,9.780) {0.40};
\node[anchor=east,font=\fontsize{7.2}{8.2}\selectfont,inner sep=.4pt] at (0.592,7.600) {0.00};
\node[font=\fontsize{8}{9}\selectfont\bfseries,inner sep=0pt] at (2.290,10.060) {Block 1};
\draw[black!75,line width=.30pt] (0.650,7.600) -- (3.930,7.600);
\draw[black!75,line width=.30pt] (0.650,7.600) -- (0.650,9.780);
\draw[black!12,line width=.20pt] (0.650,7.872) -- (3.930,7.872);
\draw[black!12,line width=.20pt] (0.650,8.145) -- (3.930,8.145);
\draw[black!12,line width=.20pt] (0.650,8.417) -- (3.930,8.417);
\draw[black!12,line width=.20pt] (0.650,8.690) -- (3.930,8.690);
\draw[black!12,line width=.20pt] (0.650,8.962) -- (3.930,8.962);
\draw[black!12,line width=.20pt] (0.650,9.235) -- (3.930,9.235);
\draw[black!12,line width=.20pt] (0.650,9.508) -- (3.930,9.508);
\draw[black!12,line width=.20pt] (0.650,9.780) -- (3.930,9.780);
\fill[GNBlue] (0.988,7.600) rectangle ++(0.262,2.000);
\node[font=\fontsize{7.2}{8.2}\selectfont,inner sep=.25pt] at (1.119,9.730) {0.367};
\node[anchor=north east,rotate=45,font=\fontsize{7.2}{8.2}\selectfont,inner sep=.25pt] at (1.119,7.500) {@2};
\fill[GNGreen] (1.456,7.600) rectangle ++(0.262,0.768);
\node[font=\fontsize{7.2}{8.2}\selectfont,inner sep=.25pt] at (1.587,8.498) {0.141};
\node[anchor=north east,rotate=45,font=\fontsize{7.2}{8.2}\selectfont,inner sep=.25pt] at (1.587,7.500) {@4};
\fill[GNPink] (1.925,7.600) rectangle ++(0.262,0.458);
\node[font=\fontsize{7.2}{8.2}\selectfont,inner sep=.25pt] at (2.056,8.188) {0.084};
\node[anchor=north east,rotate=45,font=\fontsize{7.2}{8.2}\selectfont,inner sep=.25pt] at (2.056,7.500) {@6};
\fill[GNOchre] (2.393,7.600) rectangle ++(0.262,0.348);
\node[font=\fontsize{7.2}{8.2}\selectfont,inner sep=.25pt] at (2.524,8.598) {0.064};
\node[anchor=north east,rotate=45,font=\fontsize{7.2}{8.2}\selectfont,inner sep=.25pt] at (2.524,7.500) {@8};
\fill[GNGray] (2.861,7.600) rectangle ++(0.262,0.292);
\node[font=\fontsize{7.2}{8.2}\selectfont,inner sep=.25pt] at (2.993,8.022) {0.054};
\node[anchor=north east,rotate=45,font=\fontsize{7.2}{8.2}\selectfont,inner sep=.25pt] at (2.993,7.500) {@10};
\fill[GNOrange] (3.330,7.600) rectangle ++(0.262,1.444);
\node[font=\fontsize{7.2}{8.2}\selectfont,inner sep=.25pt] at (3.461,9.174) {0.265};
\node[anchor=north east,rotate=45,font=\fontsize{7.2}{8.2}\selectfont,inner sep=.25pt] at (3.461,7.500) {Final};
\node[font=\fontsize{8}{9}\selectfont\bfseries,inner sep=0pt] at (6.370,10.060) {Block 2};
\draw[black!75,line width=.30pt] (4.730,7.600) -- (8.010,7.600);
\draw[black!75,line width=.30pt] (4.730,7.600) -- (4.730,9.780);
\draw[black!12,line width=.20pt] (4.730,7.872) -- (8.010,7.872);
\draw[black!12,line width=.20pt] (4.730,8.145) -- (8.010,8.145);
\draw[black!12,line width=.20pt] (4.730,8.417) -- (8.010,8.417);
\draw[black!12,line width=.20pt] (4.730,8.690) -- (8.010,8.690);
\draw[black!12,line width=.20pt] (4.730,8.962) -- (8.010,8.962);
\draw[black!12,line width=.20pt] (4.730,9.235) -- (8.010,9.235);
\draw[black!12,line width=.20pt] (4.730,9.508) -- (8.010,9.508);
\draw[black!12,line width=.20pt] (4.730,9.780) -- (8.010,9.780);
\fill[GNBlue] (5.068,7.600) rectangle ++(0.262,2.076);
\node[font=\fontsize{7.2}{8.2}\selectfont,inner sep=.25pt] at (5.199,9.806) {0.381};
\node[anchor=north east,rotate=45,font=\fontsize{7.2}{8.2}\selectfont,inner sep=.25pt] at (5.199,7.500) {@2};
\fill[GNGreen] (5.536,7.600) rectangle ++(0.262,0.845);
\node[font=\fontsize{7.2}{8.2}\selectfont,inner sep=.25pt] at (5.667,8.575) {0.155};
\node[anchor=north east,rotate=45,font=\fontsize{7.2}{8.2}\selectfont,inner sep=.25pt] at (5.667,7.500) {@4};
\fill[GNPink] (6.005,7.600) rectangle ++(0.262,0.468);
\node[font=\fontsize{7.2}{8.2}\selectfont,inner sep=.25pt] at (6.136,8.198) {0.086};
\node[anchor=north east,rotate=45,font=\fontsize{7.2}{8.2}\selectfont,inner sep=.25pt] at (6.136,7.500) {@6};
\fill[GNOchre] (6.473,7.600) rectangle ++(0.262,0.323);
\node[font=\fontsize{7.2}{8.2}\selectfont,inner sep=.25pt] at (6.604,8.573) {0.059};
\node[anchor=north east,rotate=45,font=\fontsize{7.2}{8.2}\selectfont,inner sep=.25pt] at (6.604,7.500) {@8};
\fill[GNGray] (6.941,7.600) rectangle ++(0.262,0.259);
\node[font=\fontsize{7.2}{8.2}\selectfont,inner sep=.25pt] at (7.073,7.989) {0.048};
\node[anchor=north east,rotate=45,font=\fontsize{7.2}{8.2}\selectfont,inner sep=.25pt] at (7.073,7.500) {@10};
\fill[GNOrange] (7.410,7.600) rectangle ++(0.262,1.161);
\node[font=\fontsize{7.2}{8.2}\selectfont,inner sep=.25pt] at (7.541,8.891) {0.213};
\node[anchor=north east,rotate=45,font=\fontsize{7.2}{8.2}\selectfont,inner sep=.25pt] at (7.541,7.500) {Final};
\node[font=\fontsize{8}{9}\selectfont\bfseries,inner sep=0pt] at (10.450,10.060) {Block 3};
\draw[black!75,line width=.30pt] (8.810,7.600) -- (12.090,7.600);
\draw[black!75,line width=.30pt] (8.810,7.600) -- (8.810,9.780);
\draw[black!12,line width=.20pt] (8.810,7.872) -- (12.090,7.872);
\draw[black!12,line width=.20pt] (8.810,8.145) -- (12.090,8.145);
\draw[black!12,line width=.20pt] (8.810,8.417) -- (12.090,8.417);
\draw[black!12,line width=.20pt] (8.810,8.690) -- (12.090,8.690);
\draw[black!12,line width=.20pt] (8.810,8.962) -- (12.090,8.962);
\draw[black!12,line width=.20pt] (8.810,9.235) -- (12.090,9.235);
\draw[black!12,line width=.20pt] (8.810,9.508) -- (12.090,9.508);
\draw[black!12,line width=.20pt] (8.810,9.780) -- (12.090,9.780);
\fill[GNGreen] (9.204,7.600) rectangle ++(0.306,0.883);
\node[font=\fontsize{7.2}{8.2}\selectfont,inner sep=.25pt] at (9.357,8.613) {0.162};
\node[anchor=north east,rotate=45,font=\fontsize{7.2}{8.2}\selectfont,inner sep=.25pt] at (9.357,7.500) {@4};
\fill[GNPink] (9.751,7.600) rectangle ++(0.306,0.478);
\node[font=\fontsize{7.2}{8.2}\selectfont,inner sep=.25pt] at (9.903,8.208) {0.088};
\node[anchor=north east,rotate=45,font=\fontsize{7.2}{8.2}\selectfont,inner sep=.25pt] at (9.903,7.500) {@6};
\fill[GNOchre] (10.297,7.600) rectangle ++(0.306,0.321);
\node[font=\fontsize{7.2}{8.2}\selectfont,inner sep=.25pt] at (10.450,8.571) {0.059};
\node[anchor=north east,rotate=45,font=\fontsize{7.2}{8.2}\selectfont,inner sep=.25pt] at (10.450,7.500) {@8};
\fill[GNGray] (10.844,7.600) rectangle ++(0.306,0.252);
\node[font=\fontsize{7.2}{8.2}\selectfont,inner sep=.25pt] at (10.997,7.982) {0.046};
\node[anchor=north east,rotate=45,font=\fontsize{7.2}{8.2}\selectfont,inner sep=.25pt] at (10.997,7.500) {@10};
\fill[GNOrange] (11.391,7.600) rectangle ++(0.306,1.068);
\node[font=\fontsize{7.2}{8.2}\selectfont,inner sep=.25pt] at (11.543,8.798) {0.196};
\node[anchor=north east,rotate=45,font=\fontsize{7.2}{8.2}\selectfont,inner sep=.25pt] at (11.543,7.500) {Final};
\node[font=\fontsize{8}{9}\selectfont\bfseries,inner sep=0pt] at (14.530,10.060) {Block 4};
\draw[black!75,line width=.30pt] (12.890,7.600) -- (16.170,7.600);
\draw[black!75,line width=.30pt] (12.890,7.600) -- (12.890,9.780);
\draw[black!12,line width=.20pt] (12.890,7.872) -- (16.170,7.872);
\draw[black!12,line width=.20pt] (12.890,8.145) -- (16.170,8.145);
\draw[black!12,line width=.20pt] (12.890,8.417) -- (16.170,8.417);
\draw[black!12,line width=.20pt] (12.890,8.690) -- (16.170,8.690);
\draw[black!12,line width=.20pt] (12.890,8.962) -- (16.170,8.962);
\draw[black!12,line width=.20pt] (12.890,9.235) -- (16.170,9.235);
\draw[black!12,line width=.20pt] (12.890,9.508) -- (16.170,9.508);
\draw[black!12,line width=.20pt] (12.890,9.780) -- (16.170,9.780);
\fill[GNGreen] (13.284,7.600) rectangle ++(0.306,0.910);
\node[font=\fontsize{7.2}{8.2}\selectfont,inner sep=.25pt] at (13.437,8.640) {0.167};
\node[anchor=north east,rotate=45,font=\fontsize{7.2}{8.2}\selectfont,inner sep=.25pt] at (13.437,7.500) {@4};
\fill[GNPink] (13.831,7.600) rectangle ++(0.306,0.517);
\node[font=\fontsize{7.2}{8.2}\selectfont,inner sep=.25pt] at (13.983,8.247) {0.095};
\node[anchor=north east,rotate=45,font=\fontsize{7.2}{8.2}\selectfont,inner sep=.25pt] at (13.983,7.500) {@6};
\fill[GNOchre] (14.377,7.600) rectangle ++(0.306,0.337);
\node[font=\fontsize{7.2}{8.2}\selectfont,inner sep=.25pt] at (14.530,8.587) {0.062};
\node[anchor=north east,rotate=45,font=\fontsize{7.2}{8.2}\selectfont,inner sep=.25pt] at (14.530,7.500) {@8};
\fill[GNGray] (14.924,7.600) rectangle ++(0.306,0.260);
\node[font=\fontsize{7.2}{8.2}\selectfont,inner sep=.25pt] at (15.077,7.990) {0.048};
\node[anchor=north east,rotate=45,font=\fontsize{7.2}{8.2}\selectfont,inner sep=.25pt] at (15.077,7.500) {@10};
\fill[GNOrange] (15.471,7.600) rectangle ++(0.306,1.052);
\node[font=\fontsize{7.2}{8.2}\selectfont,inner sep=.25pt] at (15.623,8.782) {0.193};
\node[anchor=north east,rotate=45,font=\fontsize{7.2}{8.2}\selectfont,inner sep=.25pt] at (15.623,7.500) {Final};
\node[anchor=east,font=\fontsize{7.2}{8.2}\selectfont,inner sep=.4pt] at (0.592,4.472) {0.05};
\node[anchor=east,font=\fontsize{7.2}{8.2}\selectfont,inner sep=.4pt] at (0.592,4.745) {0.10};
\node[anchor=east,font=\fontsize{7.2}{8.2}\selectfont,inner sep=.4pt] at (0.592,5.017) {0.15};
\node[anchor=east,font=\fontsize{7.2}{8.2}\selectfont,inner sep=.4pt] at (0.592,5.290) {0.20};
\node[anchor=east,font=\fontsize{7.2}{8.2}\selectfont,inner sep=.4pt] at (0.592,5.562) {0.25};
\node[anchor=east,font=\fontsize{7.2}{8.2}\selectfont,inner sep=.4pt] at (0.592,5.835) {0.30};
\node[anchor=east,font=\fontsize{7.2}{8.2}\selectfont,inner sep=.4pt] at (0.592,6.108) {0.35};
\node[anchor=east,font=\fontsize{7.2}{8.2}\selectfont,inner sep=.4pt] at (0.592,6.380) {0.40};
\node[anchor=east,font=\fontsize{7.2}{8.2}\selectfont,inner sep=.4pt] at (0.592,4.200) {0.00};
\node[font=\fontsize{8}{9}\selectfont\bfseries,inner sep=0pt] at (2.290,6.660) {Block 5};
\draw[black!75,line width=.30pt] (0.650,4.200) -- (3.930,4.200);
\draw[black!75,line width=.30pt] (0.650,4.200) -- (0.650,6.380);
\draw[black!12,line width=.20pt] (0.650,4.472) -- (3.930,4.472);
\draw[black!12,line width=.20pt] (0.650,4.745) -- (3.930,4.745);
\draw[black!12,line width=.20pt] (0.650,5.017) -- (3.930,5.017);
\draw[black!12,line width=.20pt] (0.650,5.290) -- (3.930,5.290);
\draw[black!12,line width=.20pt] (0.650,5.562) -- (3.930,5.562);
\draw[black!12,line width=.20pt] (0.650,5.835) -- (3.930,5.835);
\draw[black!12,line width=.20pt] (0.650,6.108) -- (3.930,6.108);
\draw[black!12,line width=.20pt] (0.650,6.380) -- (3.930,6.380);
\fill[GNPink] (1.147,4.200) rectangle ++(0.317,0.501);
\node[font=\fontsize{7.2}{8.2}\selectfont,inner sep=.25pt] at (1.306,4.831) {0.092};
\node[anchor=north east,rotate=45,font=\fontsize{7.2}{8.2}\selectfont,inner sep=.25pt] at (1.306,4.100) {@6};
\fill[GNOchre] (1.803,4.200) rectangle ++(0.317,0.328);
\node[font=\fontsize{7.2}{8.2}\selectfont,inner sep=.25pt] at (1.962,5.178) {0.060};
\node[anchor=north east,rotate=45,font=\fontsize{7.2}{8.2}\selectfont,inner sep=.25pt] at (1.962,4.100) {@8};
\fill[GNGray] (2.459,4.200) rectangle ++(0.317,0.255);
\node[font=\fontsize{7.2}{8.2}\selectfont,inner sep=.25pt] at (2.618,4.585) {0.047};
\node[anchor=north east,rotate=45,font=\fontsize{7.2}{8.2}\selectfont,inner sep=.25pt] at (2.618,4.100) {@10};
\fill[GNOrange] (3.115,4.200) rectangle ++(0.317,1.052);
\node[font=\fontsize{7.2}{8.2}\selectfont,inner sep=.25pt] at (3.274,5.382) {0.193};
\node[anchor=north east,rotate=45,font=\fontsize{7.2}{8.2}\selectfont,inner sep=.25pt] at (3.274,4.100) {Final};
\node[font=\fontsize{8}{9}\selectfont\bfseries,inner sep=0pt] at (6.370,6.660) {Block 6};
\draw[black!75,line width=.30pt] (4.730,4.200) -- (8.010,4.200);
\draw[black!75,line width=.30pt] (4.730,4.200) -- (4.730,6.380);
\draw[black!12,line width=.20pt] (4.730,4.473) -- (8.010,4.473);
\draw[black!12,line width=.20pt] (4.730,4.745) -- (8.010,4.745);
\draw[black!12,line width=.20pt] (4.730,5.018) -- (8.010,5.018);
\draw[black!12,line width=.20pt] (4.730,5.290) -- (8.010,5.290);
\draw[black!12,line width=.20pt] (4.730,5.563) -- (8.010,5.563);
\draw[black!12,line width=.20pt] (4.730,5.835) -- (8.010,5.835);
\draw[black!12,line width=.20pt] (4.730,6.108) -- (8.010,6.108);
\draw[black!12,line width=.20pt] (4.730,6.380) -- (8.010,6.380);
\fill[GNPink] (5.227,4.200) rectangle ++(0.317,0.502);
\node[font=\fontsize{7.2}{8.2}\selectfont,inner sep=.25pt] at (5.386,4.832) {0.092};
\node[anchor=north east,rotate=45,font=\fontsize{7.2}{8.2}\selectfont,inner sep=.25pt] at (5.386,4.100) {@6};
\fill[GNOchre] (5.883,4.200) rectangle ++(0.317,0.325);
\node[font=\fontsize{7.2}{8.2}\selectfont,inner sep=.25pt] at (6.042,5.175) {0.060};
\node[anchor=north east,rotate=45,font=\fontsize{7.2}{8.2}\selectfont,inner sep=.25pt] at (6.042,4.100) {@8};
\fill[GNGray] (6.539,4.200) rectangle ++(0.317,0.248);
\node[font=\fontsize{7.2}{8.2}\selectfont,inner sep=.25pt] at (6.698,4.578) {0.046};
\node[anchor=north east,rotate=45,font=\fontsize{7.2}{8.2}\selectfont,inner sep=.25pt] at (6.698,4.100) {@10};
\fill[GNOrange] (7.195,4.200) rectangle ++(0.317,1.008);
\node[font=\fontsize{7.2}{8.2}\selectfont,inner sep=.25pt] at (7.354,5.338) {0.185};
\node[anchor=north east,rotate=45,font=\fontsize{7.2}{8.2}\selectfont,inner sep=.25pt] at (7.354,4.100) {Final};
\node[font=\fontsize{8}{9}\selectfont\bfseries,inner sep=0pt] at (10.450,6.660) {Block 7};
\draw[black!75,line width=.30pt] (8.810,4.200) -- (12.090,4.200);
\draw[black!75,line width=.30pt] (8.810,4.200) -- (8.810,6.380);
\draw[black!12,line width=.20pt] (8.810,4.473) -- (12.090,4.473);
\draw[black!12,line width=.20pt] (8.810,4.745) -- (12.090,4.745);
\draw[black!12,line width=.20pt] (8.810,5.018) -- (12.090,5.018);
\draw[black!12,line width=.20pt] (8.810,5.290) -- (12.090,5.290);
\draw[black!12,line width=.20pt] (8.810,5.563) -- (12.090,5.563);
\draw[black!12,line width=.20pt] (8.810,5.835) -- (12.090,5.835);
\draw[black!12,line width=.20pt] (8.810,6.108) -- (12.090,6.108);
\draw[black!12,line width=.20pt] (8.810,6.380) -- (12.090,6.380);
\fill[GNOchre] (9.471,4.200) rectangle ++(0.317,0.295);
\node[font=\fontsize{7.2}{8.2}\selectfont,inner sep=.25pt] at (9.630,4.625) {0.054};
\node[anchor=north east,rotate=45,font=\fontsize{7.2}{8.2}\selectfont,inner sep=.25pt] at (9.630,4.100) {@8};
\fill[GNGray] (10.291,4.200) rectangle ++(0.317,0.231);
\node[font=\fontsize{7.2}{8.2}\selectfont,inner sep=.25pt] at (10.450,4.561) {0.042};
\node[anchor=north east,rotate=45,font=\fontsize{7.2}{8.2}\selectfont,inner sep=.25pt] at (10.450,4.100) {@10};
\fill[GNOrange] (11.111,4.200) rectangle ++(0.317,0.981);
\node[font=\fontsize{7.2}{8.2}\selectfont,inner sep=.25pt] at (11.270,5.311) {0.180};
\node[anchor=north east,rotate=45,font=\fontsize{7.2}{8.2}\selectfont,inner sep=.25pt] at (11.270,4.100) {Final};
\node[font=\fontsize{8}{9}\selectfont\bfseries,inner sep=0pt] at (14.530,6.660) {Block 8};
\draw[black!75,line width=.30pt] (12.890,4.200) -- (16.170,4.200);
\draw[black!75,line width=.30pt] (12.890,4.200) -- (12.890,6.380);
\draw[black!12,line width=.20pt] (12.890,4.473) -- (16.170,4.473);
\draw[black!12,line width=.20pt] (12.890,4.745) -- (16.170,4.745);
\draw[black!12,line width=.20pt] (12.890,5.018) -- (16.170,5.018);
\draw[black!12,line width=.20pt] (12.890,5.290) -- (16.170,5.290);
\draw[black!12,line width=.20pt] (12.890,5.563) -- (16.170,5.563);
\draw[black!12,line width=.20pt] (12.890,5.835) -- (16.170,5.835);
\draw[black!12,line width=.20pt] (12.890,6.108) -- (16.170,6.108);
\draw[black!12,line width=.20pt] (12.890,6.380) -- (16.170,6.380);
\fill[GNOchre] (13.551,4.200) rectangle ++(0.317,0.296);
\node[font=\fontsize{7.2}{8.2}\selectfont,inner sep=.25pt] at (13.710,4.626) {0.054};
\node[anchor=north east,rotate=45,font=\fontsize{7.2}{8.2}\selectfont,inner sep=.25pt] at (13.710,4.100) {@8};
\fill[GNGray] (14.371,4.200) rectangle ++(0.317,0.225);
\node[font=\fontsize{7.2}{8.2}\selectfont,inner sep=.25pt] at (14.530,4.555) {0.041};
\node[anchor=north east,rotate=45,font=\fontsize{7.2}{8.2}\selectfont,inner sep=.25pt] at (14.530,4.100) {@10};
\fill[GNOrange] (15.191,4.200) rectangle ++(0.317,0.926);
\node[font=\fontsize{7.2}{8.2}\selectfont,inner sep=.25pt] at (15.350,5.256) {0.170};
\node[anchor=north east,rotate=45,font=\fontsize{7.2}{8.2}\selectfont,inner sep=.25pt] at (15.350,4.100) {Final};
\node[anchor=east,font=\fontsize{7.2}{8.2}\selectfont,inner sep=.4pt] at (4.672,1.072) {0.05};
\node[anchor=east,font=\fontsize{7.2}{8.2}\selectfont,inner sep=.4pt] at (4.672,1.345) {0.10};
\node[anchor=east,font=\fontsize{7.2}{8.2}\selectfont,inner sep=.4pt] at (4.672,1.617) {0.15};
\node[anchor=east,font=\fontsize{7.2}{8.2}\selectfont,inner sep=.4pt] at (4.672,1.890) {0.20};
\node[anchor=east,font=\fontsize{7.2}{8.2}\selectfont,inner sep=.4pt] at (4.672,2.162) {0.25};
\node[anchor=east,font=\fontsize{7.2}{8.2}\selectfont,inner sep=.4pt] at (4.672,2.435) {0.30};
\node[anchor=east,font=\fontsize{7.2}{8.2}\selectfont,inner sep=.4pt] at (4.672,2.708) {0.35};
\node[anchor=east,font=\fontsize{7.2}{8.2}\selectfont,inner sep=.4pt] at (4.672,2.980) {0.40};
\node[anchor=east,font=\fontsize{7.2}{8.2}\selectfont,inner sep=.4pt] at (4.672,0.800) {0.000};
\node[font=\fontsize{8}{9}\selectfont\bfseries,inner sep=0pt] at (6.370,3.260) {Block 9};
\draw[black!75,line width=.30pt] (4.730,0.800) -- (8.010,0.800);
\draw[black!75,line width=.30pt] (4.730,0.800) -- (4.730,2.980);
\draw[black!12,line width=.20pt] (4.730,1.073) -- (8.010,1.073);
\draw[black!12,line width=.20pt] (4.730,1.345) -- (8.010,1.345);
\draw[black!12,line width=.20pt] (4.730,1.618) -- (8.010,1.618);
\draw[black!12,line width=.20pt] (4.730,1.890) -- (8.010,1.890);
\draw[black!12,line width=.20pt] (4.730,2.163) -- (8.010,2.163);
\draw[black!12,line width=.20pt] (4.730,2.435) -- (8.010,2.435);
\draw[black!12,line width=.20pt] (4.730,2.708) -- (8.010,2.708);
\draw[black!12,line width=.20pt] (4.730,2.980) -- (8.010,2.980);
\fill[GNGray] (5.664,0.800) rectangle ++(0.317,0.197);
\node[font=\fontsize{7.2}{8.2}\selectfont,inner sep=.25pt] at (5.823,1.127) {0.036};
\node[anchor=north east,rotate=45,font=\fontsize{7.2}{8.2}\selectfont,inner sep=.25pt] at (5.823,0.700) {@10};
\fill[GNOrange] (6.758,0.800) rectangle ++(0.317,0.883);
\node[font=\fontsize{7.2}{8.2}\selectfont,inner sep=.25pt] at (6.917,1.813) {0.162};
\node[anchor=north east,rotate=45,font=\fontsize{7.2}{8.2}\selectfont,inner sep=.25pt] at (6.917,0.700) {Final};
\node[font=\fontsize{8}{9}\selectfont\bfseries,inner sep=0pt] at (10.450,3.260) {Block 10};
\draw[black!75,line width=.30pt] (8.810,0.800) -- (12.090,0.800);
\draw[black!75,line width=.30pt] (8.810,0.800) -- (8.810,2.980);
\draw[black!12,line width=.20pt] (8.810,1.073) -- (12.090,1.073);
\draw[black!12,line width=.20pt] (8.810,1.345) -- (12.090,1.345);
\draw[black!12,line width=.20pt] (8.810,1.618) -- (12.090,1.618);
\draw[black!12,line width=.20pt] (8.810,1.890) -- (12.090,1.890);
\draw[black!12,line width=.20pt] (8.810,2.163) -- (12.090,2.163);
\draw[black!12,line width=.20pt] (8.810,2.435) -- (12.090,2.435);
\draw[black!12,line width=.20pt] (8.810,2.708) -- (12.090,2.708);
\draw[black!12,line width=.20pt] (8.810,2.980) -- (12.090,2.980);
\fill[GNGray] (9.744,0.800) rectangle ++(0.317,0.196);
\node[font=\fontsize{7.2}{8.2}\selectfont,inner sep=.25pt] at (9.903,1.126) {0.036};
\node[anchor=north east,rotate=45,font=\fontsize{7.2}{8.2}\selectfont,inner sep=.25pt] at (9.903,0.700) {@10};
\fill[GNOrange] (10.838,0.800) rectangle ++(0.317,0.861);
\node[font=\fontsize{7.2}{8.2}\selectfont,inner sep=.25pt] at (10.997,1.791) {0.158};
\node[anchor=north east,rotate=45,font=\fontsize{7.2}{8.2}\selectfont,inner sep=.25pt] at (10.997,0.700) {Final};
\end{tikzpicture}
\endgroup

%% file: artwork/gradient_conflict/further/declarevit_conflict_epoch001.tex
\begingroup%
\definecolor{GCRateLow}{RGB}{255,245,240}%
\definecolor{GCRateMid}{RGB}{251,106,74}%
\definecolor{GCRateHigh}{RGB}{165,0,38}%
\begin{tikzpicture}[x=1cm,y=1cm]
\path[use as bounding box] (-0.10,-0.05) rectangle (16.35,9.65);
\node[font=\fontsize{9.4}{10.4}\selectfont\bfseries,inner sep=0pt] at (8.125,9.43) {(A) Pairwise gradient conflict rates};
\fill[black!7] (0.65000,7.82400) rectangle ++(0.65600,0.65600);
\node[font=\fontsize{8.0}{9.0}\selectfont,inner sep=0pt] at (0.97800,8.15200) {\textendash};
\fill[GCRateMid!0.000!GCRateLow] (1.30600,7.82400) rectangle ++(0.65600,0.65600);
\node[font=\fontsize{8.0}{9.0}\selectfont,inner sep=0pt] at (1.63400,8.15200) {0.00};
\fill[GCRateMid!0.000!GCRateLow] (1.96200,7.82400) rectangle ++(0.65600,0.65600);
\node[font=\fontsize{8.0}{9.0}\selectfont,inner sep=0pt] at (2.29000,8.15200) {0.00};
\fill[GCRateMid!8.000!GCRateLow] (2.61800,7.82400) rectangle ++(0.65600,0.65600);
\node[font=\fontsize{8.0}{9.0}\selectfont,inner sep=0pt] at (2.94600,8.15200) {0.04};
\fill[GCRateMid!18.000!GCRateLow] (3.27400,7.82400) rectangle ++(0.65600,0.65600);
\node[font=\fontsize{8.0}{9.0}\selectfont,inner sep=0pt] at (3.60200,8.15200) {0.09};
\fill[GCRateMid!0.000!GCRateLow] (0.65000,7.16800) rectangle ++(0.65600,0.65600);
\node[font=\fontsize{8.0}{9.0}\selectfont,inner sep=0pt] at (0.97800,7.49600) {0.00};
\fill[black!7] (1.30600,7.16800) rectangle ++(0.65600,0.65600);
\node[font=\fontsize{8.0}{9.0}\selectfont,inner sep=0pt] at (1.63400,7.49600) {\textendash};
\fill[GCRateMid!0.000!GCRateLow] (1.96200,7.16800) rectangle ++(0.65600,0.65600);
\node[font=\fontsize{8.0}{9.0}\selectfont,inner sep=0pt] at (2.29000,7.49600) {0.00};
\fill[GCRateMid!0.000!GCRateLow] (2.61800,7.16800) rectangle ++(0.65600,0.65600);
\node[font=\fontsize{8.0}{9.0}\selectfont,inner sep=0pt] at (2.94600,7.49600) {0.00};
\fill[GCRateMid!4.000!GCRateLow] (3.27400,7.16800) rectangle ++(0.65600,0.65600);
\node[font=\fontsize{8.0}{9.0}\selectfont,inner sep=0pt] at (3.60200,7.49600) {0.02};
\fill[GCRateMid!0.000!GCRateLow] (0.65000,6.51200) rectangle ++(0.65600,0.65600);
\node[font=\fontsize{8.0}{9.0}\selectfont,inner sep=0pt] at (0.97800,6.84000) {0.00};
\fill[GCRateMid!0.000!GCRateLow] (1.30600,6.51200) rectangle ++(0.65600,0.65600);
\node[font=\fontsize{8.0}{9.0}\selectfont,inner sep=0pt] at (1.63400,6.84000) {0.00};
\fill[black!7] (1.96200,6.51200) rectangle ++(0.65600,0.65600);
\node[font=\fontsize{8.0}{9.0}\selectfont,inner sep=0pt] at (2.29000,6.84000) {\textendash};
\fill[GCRateMid!0.000!GCRateLow] (2.61800,6.51200) rectangle ++(0.65600,0.65600);
\node[font=\fontsize{8.0}{9.0}\selectfont,inner sep=0pt] at (2.94600,6.84000) {0.00};
\fill[GCRateMid!4.000!GCRateLow] (3.27400,6.51200) rectangle ++(0.65600,0.65600);
\node[font=\fontsize{8.0}{9.0}\selectfont,inner sep=0pt] at (3.60200,6.84000) {0.02};
\fill[GCRateMid!8.000!GCRateLow] (0.65000,5.85600) rectangle ++(0.65600,0.65600);
\node[font=\fontsize{8.0}{9.0}\selectfont,inner sep=0pt] at (0.97800,6.18400) {0.04};
\fill[GCRateMid!0.000!GCRateLow] (1.30600,5.85600) rectangle ++(0.65600,0.65600);
\node[font=\fontsize{8.0}{9.0}\selectfont,inner sep=0pt] at (1.63400,6.18400) {0.00};
\fill[GCRateMid!0.000!GCRateLow] (1.96200,5.85600) rectangle ++(0.65600,0.65600);
\node[font=\fontsize{8.0}{9.0}\selectfont,inner sep=0pt] at (2.29000,6.18400) {0.00};
\fill[black!7] (2.61800,5.85600) rectangle ++(0.65600,0.65600);
\node[font=\fontsize{8.0}{9.0}\selectfont,inner sep=0pt] at (2.94600,6.18400) {\textendash};
\fill[GCRateMid!4.000!GCRateLow] (3.27400,5.85600) rectangle ++(0.65600,0.65600);
\node[font=\fontsize{8.0}{9.0}\selectfont,inner sep=0pt] at (3.60200,6.18400) {0.02};
\fill[GCRateMid!18.000!GCRateLow] (0.65000,5.20000) rectangle ++(0.65600,0.65600);
\node[font=\fontsize{8.0}{9.0}\selectfont,inner sep=0pt] at (0.97800,5.52800) {0.09};
\fill[GCRateMid!4.000!GCRateLow] (1.30600,5.20000) rectangle ++(0.65600,0.65600);
\node[font=\fontsize{8.0}{9.0}\selectfont,inner sep=0pt] at (1.63400,5.52800) {0.02};
\fill[GCRateMid!4.000!GCRateLow] (1.96200,5.20000) rectangle ++(0.65600,0.65600);
\node[font=\fontsize{8.0}{9.0}\selectfont,inner sep=0pt] at (2.29000,5.52800) {0.02};
\fill[GCRateMid!4.000!GCRateLow] (2.61800,5.20000) rectangle ++(0.65600,0.65600);
\node[font=\fontsize{8.0}{9.0}\selectfont,inner sep=0pt] at (2.94600,5.52800) {0.02};
\fill[black!7] (3.27400,5.20000) rectangle ++(0.65600,0.65600);
\node[font=\fontsize{8.0}{9.0}\selectfont,inner sep=0pt] at (3.60200,5.52800) {\textendash};
\draw[black!60,line width=.25pt] (0.65000,5.20000) rectangle ++(3.28000,3.28000);
\node[anchor=north east,font=\fontsize{7.6}{8.6}\selectfont,rotate=45,inner sep=1pt] at (0.97800,5.11000) {@2};
\node[anchor=east,font=\fontsize{7.6}{8.6}\selectfont,inner sep=1pt] at (0.57000,8.15200) {@2};
\node[anchor=north east,font=\fontsize{7.6}{8.6}\selectfont,rotate=45,inner sep=1pt] at (1.63400,5.11000) {@4};
\node[anchor=east,font=\fontsize{7.6}{8.6}\selectfont,inner sep=1pt] at (0.57000,7.49600) {@4};
\node[anchor=north east,font=\fontsize{7.6}{8.6}\selectfont,rotate=45,inner sep=1pt] at (2.29000,5.11000) {@6};
\node[anchor=east,font=\fontsize{7.6}{8.6}\selectfont,inner sep=1pt] at (0.57000,6.84000) {@6};
\node[anchor=north east,font=\fontsize{7.6}{8.6}\selectfont,rotate=45,inner sep=1pt] at (2.94600,5.11000) {@8};
\node[anchor=east,font=\fontsize{7.6}{8.6}\selectfont,inner sep=1pt] at (0.57000,6.18400) {@8};
\node[anchor=north east,font=\fontsize{7.6}{8.6}\selectfont,rotate=45,inner sep=1pt] at (3.60200,5.11000) {@10};
\node[anchor=east,font=\fontsize{7.6}{8.6}\selectfont,inner sep=1pt] at (0.57000,5.52800) {@10};
\node[align=center,font=\fontsize{8.8}{9.8}\selectfont\bfseries,inner sep=0pt] at (2.29000,8.85000) {Cross-attention\\Blocks 1--2};
\fill[black!7] (4.73000,7.66000) rectangle ++(0.82000,0.82000);
\node[font=\fontsize{8.0}{9.0}\selectfont,inner sep=0pt] at (5.14000,8.07000) {\textendash};
\fill[GCRateMid!0.000!GCRateLow] (5.55000,7.66000) rectangle ++(0.82000,0.82000);
\node[font=\fontsize{8.0}{9.0}\selectfont,inner sep=0pt] at (5.96000,8.07000) {0.00};
\fill[GCRateMid!0.000!GCRateLow] (6.37000,7.66000) rectangle ++(0.82000,0.82000);
\node[font=\fontsize{8.0}{9.0}\selectfont,inner sep=0pt] at (6.78000,8.07000) {0.00};
\fill[GCRateMid!0.000!GCRateLow] (7.19000,7.66000) rectangle ++(0.82000,0.82000);
\node[font=\fontsize{8.0}{9.0}\selectfont,inner sep=0pt] at (7.60000,8.07000) {0.00};
\fill[GCRateMid!0.000!GCRateLow] (4.73000,6.84000) rectangle ++(0.82000,0.82000);
\node[font=\fontsize{8.0}{9.0}\selectfont,inner sep=0pt] at (5.14000,7.25000) {0.00};
\fill[black!7] (5.55000,6.84000) rectangle ++(0.82000,0.82000);
\node[font=\fontsize{8.0}{9.0}\selectfont,inner sep=0pt] at (5.96000,7.25000) {\textendash};
\fill[GCRateMid!0.000!GCRateLow] (6.37000,6.84000) rectangle ++(0.82000,0.82000);
\node[font=\fontsize{8.0}{9.0}\selectfont,inner sep=0pt] at (6.78000,7.25000) {0.00};
\fill[GCRateMid!4.000!GCRateLow] (7.19000,6.84000) rectangle ++(0.82000,0.82000);
\node[font=\fontsize{8.0}{9.0}\selectfont,inner sep=0pt] at (7.60000,7.25000) {0.02};
\fill[GCRateMid!0.000!GCRateLow] (4.73000,6.02000) rectangle ++(0.82000,0.82000);
\node[font=\fontsize{8.0}{9.0}\selectfont,inner sep=0pt] at (5.14000,6.43000) {0.00};
\fill[GCRateMid!0.000!GCRateLow] (5.55000,6.02000) rectangle ++(0.82000,0.82000);
\node[font=\fontsize{8.0}{9.0}\selectfont,inner sep=0pt] at (5.96000,6.43000) {0.00};
\fill[black!7] (6.37000,6.02000) rectangle ++(0.82000,0.82000);
\node[font=\fontsize{8.0}{9.0}\selectfont,inner sep=0pt] at (6.78000,6.43000) {\textendash};
\fill[GCRateMid!4.000!GCRateLow] (7.19000,6.02000) rectangle ++(0.82000,0.82000);
\node[font=\fontsize{8.0}{9.0}\selectfont,inner sep=0pt] at (7.60000,6.43000) {0.02};
\fill[GCRateMid!0.000!GCRateLow] (4.73000,5.20000) rectangle ++(0.82000,0.82000);
\node[font=\fontsize{8.0}{9.0}\selectfont,inner sep=0pt] at (5.14000,5.61000) {0.00};
\fill[GCRateMid!4.000!GCRateLow] (5.55000,5.20000) rectangle ++(0.82000,0.82000);
\node[font=\fontsize{8.0}{9.0}\selectfont,inner sep=0pt] at (5.96000,5.61000) {0.02};
\fill[GCRateMid!4.000!GCRateLow] (6.37000,5.20000) rectangle ++(0.82000,0.82000);
\node[font=\fontsize{8.0}{9.0}\selectfont,inner sep=0pt] at (6.78000,5.61000) {0.02};
\fill[black!7] (7.19000,5.20000) rectangle ++(0.82000,0.82000);
\node[font=\fontsize{8.0}{9.0}\selectfont,inner sep=0pt] at (7.60000,5.61000) {\textendash};
\draw[black!60,line width=.25pt] (4.73000,5.20000) rectangle ++(3.28000,3.28000);
\node[anchor=north east,font=\fontsize{7.6}{8.6}\selectfont,rotate=45,inner sep=1pt] at (5.14000,5.11000) {@4};
\node[anchor=east,font=\fontsize{7.6}{8.6}\selectfont,inner sep=1pt] at (4.65000,8.07000) {@4};
\node[anchor=north east,font=\fontsize{7.6}{8.6}\selectfont,rotate=45,inner sep=1pt] at (5.96000,5.11000) {@6};
\node[anchor=east,font=\fontsize{7.6}{8.6}\selectfont,inner sep=1pt] at (4.65000,7.25000) {@6};
\node[anchor=north east,font=\fontsize{7.6}{8.6}\selectfont,rotate=45,inner sep=1pt] at (6.78000,5.11000) {@8};
\node[anchor=east,font=\fontsize{7.6}{8.6}\selectfont,inner sep=1pt] at (4.65000,6.43000) {@8};
\node[anchor=north east,font=\fontsize{7.6}{8.6}\selectfont,rotate=45,inner sep=1pt] at (7.60000,5.11000) {@10};
\node[anchor=east,font=\fontsize{7.6}{8.6}\selectfont,inner sep=1pt] at (4.65000,5.61000) {@10};
\node[align=center,font=\fontsize{8.8}{9.8}\selectfont\bfseries,inner sep=0pt] at (6.37000,8.85000) {Cross-attention\\Blocks 3--4};
\fill[black!7] (8.81000,7.38667) rectangle ++(1.09333,1.09333);
\node[font=\fontsize{8.0}{9.0}\selectfont,inner sep=0pt] at (9.35667,7.93333) {\textendash};
\fill[GCRateMid!0.000!GCRateLow] (9.90333,7.38667) rectangle ++(1.09333,1.09333);
\node[font=\fontsize{8.0}{9.0}\selectfont,inner sep=0pt] at (10.45000,7.93333) {0.00};
\fill[GCRateMid!0.000!GCRateLow] (10.99667,7.38667) rectangle ++(1.09333,1.09333);
\node[font=\fontsize{8.0}{9.0}\selectfont,inner sep=0pt] at (11.54333,7.93333) {0.00};
\fill[GCRateMid!0.000!GCRateLow] (8.81000,6.29333) rectangle ++(1.09333,1.09333);
\node[font=\fontsize{8.0}{9.0}\selectfont,inner sep=0pt] at (9.35667,6.84000) {0.00};
\fill[black!7] (9.90333,6.29333) rectangle ++(1.09333,1.09333);
\node[font=\fontsize{8.0}{9.0}\selectfont,inner sep=0pt] at (10.45000,6.84000) {\textendash};
\fill[GCRateMid!0.000!GCRateLow] (10.99667,6.29333) rectangle ++(1.09333,1.09333);
\node[font=\fontsize{8.0}{9.0}\selectfont,inner sep=0pt] at (11.54333,6.84000) {0.00};
\fill[GCRateMid!0.000!GCRateLow] (8.81000,5.20000) rectangle ++(1.09333,1.09333);
\node[font=\fontsize{8.0}{9.0}\selectfont,inner sep=0pt] at (9.35667,5.74667) {0.00};
\fill[GCRateMid!0.000!GCRateLow] (9.90333,5.20000) rectangle ++(1.09333,1.09333);
\node[font=\fontsize{8.0}{9.0}\selectfont,inner sep=0pt] at (10.45000,5.74667) {0.00};
\fill[black!7] (10.99667,5.20000) rectangle ++(1.09333,1.09333);
\node[font=\fontsize{8.0}{9.0}\selectfont,inner sep=0pt] at (11.54333,5.74667) {\textendash};
\draw[black!60,line width=.25pt] (8.81000,5.20000) rectangle ++(3.28000,3.28000);
\node[anchor=north east,font=\fontsize{7.6}{8.6}\selectfont,rotate=45,inner sep=1pt] at (9.35667,5.11000) {@6};
\node[anchor=east,font=\fontsize{7.6}{8.6}\selectfont,inner sep=1pt] at (8.73000,7.93333) {@6};
\node[anchor=north east,font=\fontsize{7.6}{8.6}\selectfont,rotate=45,inner sep=1pt] at (10.45000,5.11000) {@8};
\node[anchor=east,font=\fontsize{7.6}{8.6}\selectfont,inner sep=1pt] at (8.73000,6.84000) {@8};
\node[anchor=north east,font=\fontsize{7.6}{8.6}\selectfont,rotate=45,inner sep=1pt] at (11.54333,5.11000) {@10};
\node[anchor=east,font=\fontsize{7.6}{8.6}\selectfont,inner sep=1pt] at (8.73000,5.74667) {@10};
\node[align=center,font=\fontsize{8.8}{9.8}\selectfont\bfseries,inner sep=0pt] at (10.45000,8.85000) {Cross-attention\\Blocks 5--6};
\fill[black!7] (12.89000,6.84000) rectangle ++(1.64000,1.64000);
\node[font=\fontsize{8.0}{9.0}\selectfont,inner sep=0pt] at (13.71000,7.66000) {\textendash};
\fill[GCRateMid!0.000!GCRateLow] (14.53000,6.84000) rectangle ++(1.64000,1.64000);
\node[font=\fontsize{8.0}{9.0}\selectfont,inner sep=0pt] at (15.35000,7.66000) {0.00};
\fill[GCRateMid!0.000!GCRateLow] (12.89000,5.20000) rectangle ++(1.64000,1.64000);
\node[font=\fontsize{8.0}{9.0}\selectfont,inner sep=0pt] at (13.71000,6.02000) {0.00};
\fill[black!7] (14.53000,5.20000) rectangle ++(1.64000,1.64000);
\node[font=\fontsize{8.0}{9.0}\selectfont,inner sep=0pt] at (15.35000,6.02000) {\textendash};
\draw[black!60,line width=.25pt] (12.89000,5.20000) rectangle ++(3.28000,3.28000);
\node[anchor=north east,font=\fontsize{7.6}{8.6}\selectfont,rotate=45,inner sep=1pt] at (13.71000,5.11000) {@8};
\node[anchor=east,font=\fontsize{7.6}{8.6}\selectfont,inner sep=1pt] at (12.81000,7.66000) {@8};
\node[anchor=north east,font=\fontsize{7.6}{8.6}\selectfont,rotate=45,inner sep=1pt] at (15.35000,5.11000) {@10};
\node[anchor=east,font=\fontsize{7.6}{8.6}\selectfont,inner sep=1pt] at (12.81000,6.02000) {@10};
\node[align=center,font=\fontsize{8.8}{9.8}\selectfont\bfseries,inner sep=0pt] at (14.53000,8.85000) {Cross-attention\\Blocks 7--8};
\fill[black!7] (6.77000,3.37400) rectangle ++(0.65600,0.65600);
\node[font=\fontsize{8.0}{9.0}\selectfont,inner sep=0pt] at (7.09800,3.70200) {\textendash};
\fill[GCRateMid!0.000!GCRateLow] (7.42600,3.37400) rectangle ++(0.65600,0.65600);
\node[font=\fontsize{8.0}{9.0}\selectfont,inner sep=0pt] at (7.75400,3.70200) {0.00};
\fill[GCRateMid!8.000!GCRateLow] (8.08200,3.37400) rectangle ++(0.65600,0.65600);
\node[font=\fontsize{8.0}{9.0}\selectfont,inner sep=0pt] at (8.41000,3.70200) {0.04};
\fill[GCRateMid!10.000!GCRateLow] (8.73800,3.37400) rectangle ++(0.65600,0.65600);
\node[font=\fontsize{8.0}{9.0}\selectfont,inner sep=0pt] at (9.06600,3.70200) {0.05};
\fill[GCRateMid!38.000!GCRateLow] (9.39400,3.37400) rectangle ++(0.65600,0.65600);
\node[font=\fontsize{8.0}{9.0}\selectfont,inner sep=0pt] at (9.72200,3.70200) {0.19};
\fill[GCRateMid!0.000!GCRateLow] (6.77000,2.71800) rectangle ++(0.65600,0.65600);
\node[font=\fontsize{8.0}{9.0}\selectfont,inner sep=0pt] at (7.09800,3.04600) {0.00};
\fill[black!7] (7.42600,2.71800) rectangle ++(0.65600,0.65600);
\node[font=\fontsize{8.0}{9.0}\selectfont,inner sep=0pt] at (7.75400,3.04600) {\textendash};
\fill[GCRateMid!0.000!GCRateLow] (8.08200,2.71800) rectangle ++(0.65600,0.65600);
\node[font=\fontsize{8.0}{9.0}\selectfont,inner sep=0pt] at (8.41000,3.04600) {0.00};
\fill[GCRateMid!8.000!GCRateLow] (8.73800,2.71800) rectangle ++(0.65600,0.65600);
\node[font=\fontsize{8.0}{9.0}\selectfont,inner sep=0pt] at (9.06600,3.04600) {0.04};
\fill[GCRateMid!14.000!GCRateLow] (9.39400,2.71800) rectangle ++(0.65600,0.65600);
\node[font=\fontsize{8.0}{9.0}\selectfont,inner sep=0pt] at (9.72200,3.04600) {0.07};
\fill[GCRateMid!8.000!GCRateLow] (6.77000,2.06200) rectangle ++(0.65600,0.65600);
\node[font=\fontsize{8.0}{9.0}\selectfont,inner sep=0pt] at (7.09800,2.39000) {0.04};
\fill[GCRateMid!0.000!GCRateLow] (7.42600,2.06200) rectangle ++(0.65600,0.65600);
\node[font=\fontsize{8.0}{9.0}\selectfont,inner sep=0pt] at (7.75400,2.39000) {0.00};
\fill[black!7] (8.08200,2.06200) rectangle ++(0.65600,0.65600);
\node[font=\fontsize{8.0}{9.0}\selectfont,inner sep=0pt] at (8.41000,2.39000) {\textendash};
\fill[GCRateMid!0.000!GCRateLow] (8.73800,2.06200) rectangle ++(0.65600,0.65600);
\node[font=\fontsize{8.0}{9.0}\selectfont,inner sep=0pt] at (9.06600,2.39000) {0.00};
\fill[GCRateMid!0.000!GCRateLow] (9.39400,2.06200) rectangle ++(0.65600,0.65600);
\node[font=\fontsize{8.0}{9.0}\selectfont,inner sep=0pt] at (9.72200,2.39000) {0.00};
\fill[GCRateMid!10.000!GCRateLow] (6.77000,1.40600) rectangle ++(0.65600,0.65600);
\node[font=\fontsize{8.0}{9.0}\selectfont,inner sep=0pt] at (7.09800,1.73400) {0.05};
\fill[GCRateMid!8.000!GCRateLow] (7.42600,1.40600) rectangle ++(0.65600,0.65600);
\node[font=\fontsize{8.0}{9.0}\selectfont,inner sep=0pt] at (7.75400,1.73400) {0.04};
\fill[GCRateMid!0.000!GCRateLow] (8.08200,1.40600) rectangle ++(0.65600,0.65600);
\node[font=\fontsize{8.0}{9.0}\selectfont,inner sep=0pt] at (8.41000,1.73400) {0.00};
\fill[black!7] (8.73800,1.40600) rectangle ++(0.65600,0.65600);
\node[font=\fontsize{8.0}{9.0}\selectfont,inner sep=0pt] at (9.06600,1.73400) {\textendash};
\fill[GCRateMid!4.000!GCRateLow] (9.39400,1.40600) rectangle ++(0.65600,0.65600);
\node[font=\fontsize{8.0}{9.0}\selectfont,inner sep=0pt] at (9.72200,1.73400) {0.02};
\fill[GCRateMid!38.000!GCRateLow] (6.77000,0.75000) rectangle ++(0.65600,0.65600);
\node[font=\fontsize{8.0}{9.0}\selectfont,inner sep=0pt] at (7.09800,1.07800) {0.19};
\fill[GCRateMid!14.000!GCRateLow] (7.42600,0.75000) rectangle ++(0.65600,0.65600);
\node[font=\fontsize{8.0}{9.0}\selectfont,inner sep=0pt] at (7.75400,1.07800) {0.07};
\fill[GCRateMid!0.000!GCRateLow] (8.08200,0.75000) rectangle ++(0.65600,0.65600);
\node[font=\fontsize{8.0}{9.0}\selectfont,inner sep=0pt] at (8.41000,1.07800) {0.00};
\fill[GCRateMid!4.000!GCRateLow] (8.73800,0.75000) rectangle ++(0.65600,0.65600);
\node[font=\fontsize{8.0}{9.0}\selectfont,inner sep=0pt] at (9.06600,1.07800) {0.02};
\fill[black!7] (9.39400,0.75000) rectangle ++(0.65600,0.65600);
\node[font=\fontsize{8.0}{9.0}\selectfont,inner sep=0pt] at (9.72200,1.07800) {\textendash};
\draw[black!60,line width=.25pt] (6.77000,0.75000) rectangle ++(3.28000,3.28000);
\node[anchor=north east,font=\fontsize{7.6}{8.6}\selectfont,rotate=45,inner sep=1pt] at (7.09800,0.66000) {@2};
\node[anchor=east,font=\fontsize{7.6}{8.6}\selectfont,inner sep=1pt] at (6.69000,3.70200) {@2};
\node[anchor=north east,font=\fontsize{7.6}{8.6}\selectfont,rotate=45,inner sep=1pt] at (7.75400,0.66000) {@4};
\node[anchor=east,font=\fontsize{7.6}{8.6}\selectfont,inner sep=1pt] at (6.69000,3.04600) {@4};
\node[anchor=north east,font=\fontsize{7.6}{8.6}\selectfont,rotate=45,inner sep=1pt] at (8.41000,0.66000) {@6};
\node[anchor=east,font=\fontsize{7.6}{8.6}\selectfont,inner sep=1pt] at (6.69000,2.39000) {@6};
\node[anchor=north east,font=\fontsize{7.6}{8.6}\selectfont,rotate=45,inner sep=1pt] at (9.06600,0.66000) {@8};
\node[anchor=east,font=\fontsize{7.6}{8.6}\selectfont,inner sep=1pt] at (6.69000,1.73400) {@8};
\node[anchor=north east,font=\fontsize{7.6}{8.6}\selectfont,rotate=45,inner sep=1pt] at (9.72200,0.66000) {@10};
\node[anchor=east,font=\fontsize{7.6}{8.6}\selectfont,inner sep=1pt] at (6.69000,1.07800) {@10};
\node[align=center,font=\fontsize{8.8}{9.8}\selectfont\bfseries,inner sep=0pt] at (8.41000,4.40000) {Shared exit FFN};
\begin{scope}[shift={(-0.60,-1.752)}]
\shade[bottom color=GCRateLow,top color=GCRateHigh,middle color=GCRateMid] (13.18000,2.50200) rectangle ++(0.19000,3.00600);
\draw[black!60,line width=.25pt] (13.18000,2.50200) rectangle ++(0.19000,3.00600);
\draw[black!65,line width=.25pt] (13.37000,2.50200) -- ++(.045,0);
\node[anchor=west,font=\fontsize{6.5}{7.5}\selectfont,inner sep=1pt] at (13.47000,2.50200) {0.0};
\draw[black!65,line width=.25pt] (13.37000,4.00500) -- ++(.045,0);
\node[anchor=west,font=\fontsize{6.5}{7.5}\selectfont,inner sep=1pt] at (13.47000,4.00500) {0.5};
\draw[black!65,line width=.25pt] (13.37000,5.50800) -- ++(.045,0);
\node[anchor=west,font=\fontsize{6.5}{7.5}\selectfont,inner sep=1pt] at (13.47000,5.50800) {1.0};
\node[rotate=90,font=\fontsize{6.8}{7.8}\selectfont,inner sep=0pt] at (14.00,4.00500) {Conflict rate};

\end{scope}
\end{tikzpicture}%
\endgroup%

%% file: artwork/gradient_conflict/further/declarevit_gradient_norm_epoch001.tex
\begingroup%
\definecolor{GNBlue}{RGB}{0,114,178}%
\definecolor{GNGreen}{RGB}{0,158,115}%
\definecolor{GNPink}{RGB}{204,121,167}%
\definecolor{GNOchre}{RGB}{181,137,0}%
\definecolor{GNGray}{RGB}{102,102,102}%
\begin{tikzpicture}[x=1cm,y=1cm]
\path[use as bounding box] (-0.10,0.00) rectangle (16.35,9.25);
\node[font=\fontsize{9.4}{10.4}\selectfont\bfseries,inner sep=0pt] at (8.125,9.04) {(B) Loss-weighted gradient norms};
\node[align=center,font=\fontsize{8.4}{9.4}\selectfont\bfseries,inner sep=0pt] at (2.290,8.400) {Cross-attention\\Blocks 1--2};
\node[anchor=east,font=\fontsize{7.6}{8.6}\selectfont,inner sep=.4pt] at (0.592,4.900) {0.00};
\draw[black!12,line width=.20pt] (0.650,5.288) -- (3.930,5.288);
\node[anchor=east,font=\fontsize{7.6}{8.6}\selectfont,inner sep=.4pt] at (0.592,5.288) {0.20};
\draw[black!12,line width=.20pt] (0.650,5.675) -- (3.930,5.675);
\node[anchor=east,font=\fontsize{7.6}{8.6}\selectfont,inner sep=.4pt] at (0.592,5.675) {0.40};
\draw[black!12,line width=.20pt] (0.650,6.062) -- (3.930,6.062);
\node[anchor=east,font=\fontsize{7.6}{8.6}\selectfont,inner sep=.4pt] at (0.592,6.062) {0.60};
\draw[black!12,line width=.20pt] (0.650,6.450) -- (3.930,6.450);
\node[anchor=east,font=\fontsize{7.6}{8.6}\selectfont,inner sep=.4pt] at (0.592,6.450) {0.80};
\draw[black!12,line width=.20pt] (0.650,6.838) -- (3.930,6.838);
\node[anchor=east,font=\fontsize{7.6}{8.6}\selectfont,inner sep=.4pt] at (0.592,6.838) {1.00};
\draw[black!12,line width=.20pt] (0.650,7.225) -- (3.930,7.225);
\node[anchor=east,font=\fontsize{7.6}{8.6}\selectfont,inner sep=.4pt] at (0.592,7.225) {1.20};
\draw[black!12,line width=.20pt] (0.650,7.613) -- (3.930,7.613);
\node[anchor=east,font=\fontsize{7.6}{8.6}\selectfont,inner sep=.4pt] at (0.592,7.613) {1.40};
\draw[black!12,line width=.20pt] (0.650,8.000) -- (3.930,8.000);
\node[anchor=east,font=\fontsize{7.6}{8.6}\selectfont,inner sep=.4pt] at (0.592,8.000) {1.60};
\draw[black!75,line width=.30pt] (0.650,4.900) -- (3.930,4.900);
\draw[black!75,line width=.30pt] (0.650,4.900) -- (0.650,8.000);
\fill[GNBlue] (0.7681,4.9000) rectangle ++(0.4198,2.6737);
\node[anchor=south,font=\fontsize{7.4}{8.4}\selectfont,inner sep=.25pt] at (0.9780,7.6338) {1.380};
\node[anchor=north east,rotate=45,font=\fontsize{7.6}{8.6}\selectfont,inner sep=.25pt] at (0.9780,4.8000) {@2};
\fill[GNGreen] (1.4241,4.9000) rectangle ++(0.4198,1.3504);
\node[anchor=south,font=\fontsize{7.4}{8.4}\selectfont,inner sep=.25pt] at (1.6340,6.3104) {0.697};
\node[anchor=north east,rotate=45,font=\fontsize{7.6}{8.6}\selectfont,inner sep=.25pt] at (1.6340,4.8000) {@4};
\fill[GNPink] (2.0801,4.9000) rectangle ++(0.4198,0.7382);
\node[anchor=south,font=\fontsize{7.4}{8.4}\selectfont,inner sep=.25pt] at (2.2900,5.6982) {0.381};
\node[anchor=north east,rotate=45,font=\fontsize{7.6}{8.6}\selectfont,inner sep=.25pt] at (2.2900,4.8000) {@6};
\fill[GNOchre] (2.7361,4.9000) rectangle ++(0.4198,0.5619);
\node[anchor=south,font=\fontsize{7.4}{8.4}\selectfont,inner sep=.25pt] at (2.9460,5.5219) {0.290};
\node[anchor=north east,rotate=45,font=\fontsize{7.6}{8.6}\selectfont,inner sep=.25pt] at (2.9460,4.8000) {@8};
\fill[GNGray] (3.3921,4.9000) rectangle ++(0.4198,0.4243);
\node[anchor=south,font=\fontsize{7.4}{8.4}\selectfont,inner sep=.25pt] at (3.6020,5.3843) {0.219};
\node[anchor=north east,rotate=45,font=\fontsize{7.6}{8.6}\selectfont,inner sep=.25pt] at (3.6020,4.8000) {@10};
\node[align=center,font=\fontsize{8.4}{9.4}\selectfont\bfseries,inner sep=0pt] at (6.370,8.400) {Cross-attention\\Blocks 3--4};
\draw[black!12,line width=.20pt] (4.730,5.288) -- (8.010,5.288);
\draw[black!12,line width=.20pt] (4.730,5.675) -- (8.010,5.675);
\draw[black!12,line width=.20pt] (4.730,6.062) -- (8.010,6.062);
\draw[black!12,line width=.20pt] (4.730,6.450) -- (8.010,6.450);
\draw[black!12,line width=.20pt] (4.730,6.838) -- (8.010,6.838);
\draw[black!12,line width=.20pt] (4.730,7.225) -- (8.010,7.225);
\draw[black!12,line width=.20pt] (4.730,7.613) -- (8.010,7.613);
\draw[black!12,line width=.20pt] (4.730,8.000) -- (8.010,8.000);
\draw[black!75,line width=.30pt] (4.730,4.900) -- (8.010,4.900);
\draw[black!75,line width=.30pt] (4.730,4.900) -- (4.730,8.000);
\fill[GNGreen] (4.8776,4.9000) rectangle ++(0.5248,1.3989);
\node[anchor=south,font=\fontsize{8.0}{9.0}\selectfont,inner sep=.25pt] at (5.1400,6.3589) {0.722};
\node[anchor=north east,rotate=45,font=\fontsize{7.6}{8.6}\selectfont,inner sep=.25pt] at (5.1400,4.8000) {@4};
\fill[GNPink] (5.6976,4.9000) rectangle ++(0.5248,0.7246);
\node[anchor=south,font=\fontsize{8.0}{9.0}\selectfont,inner sep=.25pt] at (5.9600,5.6846) {0.374};
\node[anchor=north east,rotate=45,font=\fontsize{7.6}{8.6}\selectfont,inner sep=.25pt] at (5.9600,4.8000) {@6};
\fill[GNOchre] (6.5176,4.9000) rectangle ++(0.5248,0.5038);
\node[anchor=south,font=\fontsize{8.0}{9.0}\selectfont,inner sep=.25pt] at (6.7800,5.4638) {0.260};
\node[anchor=north east,rotate=45,font=\fontsize{7.6}{8.6}\selectfont,inner sep=.25pt] at (6.7800,4.8000) {@8};
\fill[GNGray] (7.3376,4.9000) rectangle ++(0.5248,0.4030);
\node[anchor=south,font=\fontsize{8.0}{9.0}\selectfont,inner sep=.25pt] at (7.6000,5.3630) {0.208};
\node[anchor=north east,rotate=45,font=\fontsize{7.6}{8.6}\selectfont,inner sep=.25pt] at (7.6000,4.8000) {@10};
\node[align=center,font=\fontsize{8.4}{9.4}\selectfont\bfseries,inner sep=0pt] at (10.450,8.400) {Cross-attention\\Blocks 5--6};
\draw[black!12,line width=.20pt] (8.810,5.288) -- (12.090,5.288);
\draw[black!12,line width=.20pt] (8.810,5.675) -- (12.090,5.675);
\draw[black!12,line width=.20pt] (8.810,6.062) -- (12.090,6.062);
\draw[black!12,line width=.20pt] (8.810,6.450) -- (12.090,6.450);
\draw[black!12,line width=.20pt] (8.810,6.838) -- (12.090,6.838);
\draw[black!12,line width=.20pt] (8.810,7.225) -- (12.090,7.225);
\draw[black!12,line width=.20pt] (8.810,7.613) -- (12.090,7.613);
\draw[black!12,line width=.20pt] (8.810,8.000) -- (12.090,8.000);
\draw[black!75,line width=.30pt] (8.810,4.900) -- (12.090,4.900);
\draw[black!75,line width=.30pt] (8.810,4.900) -- (8.810,8.000);
\fill[GNPink] (9.0068,4.9000) rectangle ++(0.6997,0.8874);
\node[anchor=south,font=\fontsize{8.0}{9.0}\selectfont,inner sep=.25pt] at (9.3567,5.8474) {0.458};
\node[anchor=north east,rotate=45,font=\fontsize{7.6}{8.6}\selectfont,inner sep=.25pt] at (9.3567,4.8000) {@6};
\fill[GNOchre] (10.1001,4.9000) rectangle ++(0.6997,0.5968);
\node[anchor=south,font=\fontsize{8.0}{9.0}\selectfont,inner sep=.25pt] at (10.4500,5.5568) {0.308};
\node[anchor=north east,rotate=45,font=\fontsize{7.6}{8.6}\selectfont,inner sep=.25pt] at (10.4500,4.8000) {@8};
\fill[GNGray] (11.1935,4.9000) rectangle ++(0.6997,0.5657);
\node[anchor=south,font=\fontsize{8.0}{9.0}\selectfont,inner sep=.25pt] at (11.5433,5.5257) {0.292};
\node[anchor=north east,rotate=45,font=\fontsize{7.6}{8.6}\selectfont,inner sep=.25pt] at (11.5433,4.8000) {@10};
\node[align=center,font=\fontsize{8.4}{9.4}\selectfont\bfseries,inner sep=0pt] at (14.530,8.400) {Cross-attention\\Blocks 7--8};
\draw[black!12,line width=.20pt] (12.890,5.288) -- (16.170,5.288);
\draw[black!12,line width=.20pt] (12.890,5.675) -- (16.170,5.675);
\draw[black!12,line width=.20pt] (12.890,6.062) -- (16.170,6.062);
\draw[black!12,line width=.20pt] (12.890,6.450) -- (16.170,6.450);
\draw[black!12,line width=.20pt] (12.890,6.838) -- (16.170,6.838);
\draw[black!12,line width=.20pt] (12.890,7.225) -- (16.170,7.225);
\draw[black!12,line width=.20pt] (12.890,7.613) -- (16.170,7.613);
\draw[black!12,line width=.20pt] (12.890,8.000) -- (16.170,8.000);
\draw[black!75,line width=.30pt] (12.890,4.900) -- (16.170,4.900);
\draw[black!75,line width=.30pt] (12.890,4.900) -- (12.890,8.000);
\fill[GNOchre] (13.1852,4.9000) rectangle ++(1.0496,0.7459);
\node[anchor=south,font=\fontsize{8.0}{9.0}\selectfont,inner sep=.25pt] at (13.7100,5.7059) {0.385};
\node[anchor=north east,rotate=45,font=\fontsize{7.6}{8.6}\selectfont,inner sep=.25pt] at (13.7100,4.8000) {@8};
\fill[GNGray] (14.8252,4.9000) rectangle ++(1.0496,0.5192);
\node[anchor=south,font=\fontsize{8.0}{9.0}\selectfont,inner sep=.25pt] at (15.3500,5.4792) {0.268};
\node[anchor=north east,rotate=45,font=\fontsize{7.6}{8.6}\selectfont,inner sep=.25pt] at (15.3500,4.8000) {@10};
\node[align=center,font=\fontsize{8.4}{9.4}\selectfont\bfseries,inner sep=0pt] at (8.410,4.140) {Shared exit FFN};
\node[anchor=east,font=\fontsize{7.6}{8.6}\selectfont,inner sep=.4pt] at (6.712,0.850) {0.00};
\draw[black!12,line width=.20pt] (6.770,1.238) -- (10.050,1.238);
\node[anchor=east,font=\fontsize{7.6}{8.6}\selectfont,inner sep=.4pt] at (6.712,1.238) {0.20};
\draw[black!12,line width=.20pt] (6.770,1.625) -- (10.050,1.625);
\node[anchor=east,font=\fontsize{7.6}{8.6}\selectfont,inner sep=.4pt] at (6.712,1.625) {0.40};
\draw[black!12,line width=.20pt] (6.770,2.013) -- (10.050,2.013);
\node[anchor=east,font=\fontsize{7.6}{8.6}\selectfont,inner sep=.4pt] at (6.712,2.013) {0.60};
\draw[black!12,line width=.20pt] (6.770,2.400) -- (10.050,2.400);
\node[anchor=east,font=\fontsize{7.6}{8.6}\selectfont,inner sep=.4pt] at (6.712,2.400) {0.80};
\draw[black!12,line width=.20pt] (6.770,2.788) -- (10.050,2.788);
\node[anchor=east,font=\fontsize{7.6}{8.6}\selectfont,inner sep=.4pt] at (6.712,2.788) {1.00};
\draw[black!12,line width=.20pt] (6.770,3.175) -- (10.050,3.175);
\node[anchor=east,font=\fontsize{7.6}{8.6}\selectfont,inner sep=.4pt] at (6.712,3.175) {1.20};
\draw[black!12,line width=.20pt] (6.770,3.563) -- (10.050,3.563);
\node[anchor=east,font=\fontsize{7.6}{8.6}\selectfont,inner sep=.4pt] at (6.712,3.563) {1.40};
\draw[black!12,line width=.20pt] (6.770,3.950) -- (10.050,3.950);
\node[anchor=east,font=\fontsize{7.6}{8.6}\selectfont,inner sep=.4pt] at (6.712,3.950) {1.60};
\draw[black!75,line width=.30pt] (6.770,0.850) -- (10.050,0.850);
\draw[black!75,line width=.30pt] (6.770,0.850) -- (6.770,3.950);
\fill[GNBlue] (6.8881,0.8500) rectangle ++(0.4198,1.0191);
\node[anchor=south,font=\fontsize{7.4}{8.4}\selectfont,inner sep=.25pt] at (7.0980,1.9291) {0.526};
\node[anchor=north east,rotate=45,font=\fontsize{7.6}{8.6}\selectfont,inner sep=.25pt] at (7.0980,0.7500) {@2};
\fill[GNGreen] (7.5441,0.8500) rectangle ++(0.4198,1.0308);
\node[anchor=south,font=\fontsize{7.4}{8.4}\selectfont,inner sep=.25pt] at (7.7540,1.9407) {0.532};
\node[anchor=north east,rotate=45,font=\fontsize{7.6}{8.6}\selectfont,inner sep=.25pt] at (7.7540,0.7500) {@4};
\fill[GNPink] (8.2001,0.8500) rectangle ++(0.4198,0.9494);
\node[anchor=south,font=\fontsize{7.4}{8.4}\selectfont,inner sep=.25pt] at (8.4100,1.8594) {0.490};
\node[anchor=north east,rotate=45,font=\fontsize{7.6}{8.6}\selectfont,inner sep=.25pt] at (8.4100,0.7500) {@6};
\fill[GNOchre] (8.8561,0.8500) rectangle ++(0.4198,0.8118);
\node[anchor=south,font=\fontsize{7.4}{8.4}\selectfont,inner sep=.25pt] at (9.0660,1.7218) {0.419};
\node[anchor=north east,rotate=45,font=\fontsize{7.6}{8.6}\selectfont,inner sep=.25pt] at (9.0660,0.7500) {@8};
\fill[GNGray] (9.5121,0.8500) rectangle ++(0.4198,0.6181);
\node[anchor=south,font=\fontsize{7.4}{8.4}\selectfont,inner sep=.25pt] at (9.7220,1.5281) {0.319};
\node[anchor=north east,rotate=45,font=\fontsize{7.6}{8.6}\selectfont,inner sep=.25pt] at (9.7220,0.7500) {@10};
\end{tikzpicture}%
\endgroup%

%% file: artwork/gradient_conflict/further/declarevit_conflict_epoch010.tex
\begingroup%
\definecolor{GCRateLow}{RGB}{255,245,240}%
\definecolor{GCRateMid}{RGB}{251,106,74}%
\definecolor{GCRateHigh}{RGB}{165,0,38}%
\begin{tikzpicture}[x=1cm,y=1cm]
\path[use as bounding box] (-0.10,-0.05) rectangle (16.35,9.65);
\node[font=\fontsize{9.4}{10.4}\selectfont\bfseries,inner sep=0pt] at (8.125,9.43) {(A) Pairwise gradient conflict rates};
\fill[black!7] (0.65000,7.82400) rectangle ++(0.65600,0.65600);
\node[font=\fontsize{8.0}{9.0}\selectfont,inner sep=0pt] at (0.97800,8.15200) {\textendash};
\fill[GCRateMid!0.000!GCRateLow] (1.30600,7.82400) rectangle ++(0.65600,0.65600);
\node[font=\fontsize{8.0}{9.0}\selectfont,inner sep=0pt] at (1.63400,8.15200) {0.00};
\fill[GCRateMid!0.000!GCRateLow] (1.96200,7.82400) rectangle ++(0.65600,0.65600);
\node[font=\fontsize{8.0}{9.0}\selectfont,inner sep=0pt] at (2.29000,8.15200) {0.00};
\fill[GCRateMid!32.000!GCRateLow] (2.61800,7.82400) rectangle ++(0.65600,0.65600);
\node[font=\fontsize{8.0}{9.0}\selectfont,inner sep=0pt] at (2.94600,8.15200) {0.16};
\fill[GCRateMid!42.000!GCRateLow] (3.27400,7.82400) rectangle ++(0.65600,0.65600);
\node[font=\fontsize{8.0}{9.0}\selectfont,inner sep=0pt] at (3.60200,8.15200) {0.21};
\fill[GCRateMid!0.000!GCRateLow] (0.65000,7.16800) rectangle ++(0.65600,0.65600);
\node[font=\fontsize{8.0}{9.0}\selectfont,inner sep=0pt] at (0.97800,7.49600) {0.00};
\fill[black!7] (1.30600,7.16800) rectangle ++(0.65600,0.65600);
\node[font=\fontsize{8.0}{9.0}\selectfont,inner sep=0pt] at (1.63400,7.49600) {\textendash};
\fill[GCRateMid!0.000!GCRateLow] (1.96200,7.16800) rectangle ++(0.65600,0.65600);
\node[font=\fontsize{8.0}{9.0}\selectfont,inner sep=0pt] at (2.29000,7.49600) {0.00};
\fill[GCRateMid!0.000!GCRateLow] (2.61800,7.16800) rectangle ++(0.65600,0.65600);
\node[font=\fontsize{8.0}{9.0}\selectfont,inner sep=0pt] at (2.94600,7.49600) {0.00};
\fill[GCRateMid!4.000!GCRateLow] (3.27400,7.16800) rectangle ++(0.65600,0.65600);
\node[font=\fontsize{8.0}{9.0}\selectfont,inner sep=0pt] at (3.60200,7.49600) {0.02};
\fill[GCRateMid!0.000!GCRateLow] (0.65000,6.51200) rectangle ++(0.65600,0.65600);
\node[font=\fontsize{8.0}{9.0}\selectfont,inner sep=0pt] at (0.97800,6.84000) {0.00};
\fill[GCRateMid!0.000!GCRateLow] (1.30600,6.51200) rectangle ++(0.65600,0.65600);
\node[font=\fontsize{8.0}{9.0}\selectfont,inner sep=0pt] at (1.63400,6.84000) {0.00};
\fill[black!7] (1.96200,6.51200) rectangle ++(0.65600,0.65600);
\node[font=\fontsize{8.0}{9.0}\selectfont,inner sep=0pt] at (2.29000,6.84000) {\textendash};
\fill[GCRateMid!0.000!GCRateLow] (2.61800,6.51200) rectangle ++(0.65600,0.65600);
\node[font=\fontsize{8.0}{9.0}\selectfont,inner sep=0pt] at (2.94600,6.84000) {0.00};
\fill[GCRateMid!0.000!GCRateLow] (3.27400,6.51200) rectangle ++(0.65600,0.65600);
\node[font=\fontsize{8.0}{9.0}\selectfont,inner sep=0pt] at (3.60200,6.84000) {0.00};
\fill[GCRateMid!32.000!GCRateLow] (0.65000,5.85600) rectangle ++(0.65600,0.65600);
\node[font=\fontsize{8.0}{9.0}\selectfont,inner sep=0pt] at (0.97800,6.18400) {0.16};
\fill[GCRateMid!0.000!GCRateLow] (1.30600,5.85600) rectangle ++(0.65600,0.65600);
\node[font=\fontsize{8.0}{9.0}\selectfont,inner sep=0pt] at (1.63400,6.18400) {0.00};
\fill[GCRateMid!0.000!GCRateLow] (1.96200,5.85600) rectangle ++(0.65600,0.65600);
\node[font=\fontsize{8.0}{9.0}\selectfont,inner sep=0pt] at (2.29000,6.18400) {0.00};
\fill[black!7] (2.61800,5.85600) rectangle ++(0.65600,0.65600);
\node[font=\fontsize{8.0}{9.0}\selectfont,inner sep=0pt] at (2.94600,6.18400) {\textendash};
\fill[GCRateMid!0.000!GCRateLow] (3.27400,5.85600) rectangle ++(0.65600,0.65600);
\node[font=\fontsize{8.0}{9.0}\selectfont,inner sep=0pt] at (3.60200,6.18400) {0.00};
\fill[GCRateMid!42.000!GCRateLow] (0.65000,5.20000) rectangle ++(0.65600,0.65600);
\node[font=\fontsize{8.0}{9.0}\selectfont,inner sep=0pt] at (0.97800,5.52800) {0.21};
\fill[GCRateMid!4.000!GCRateLow] (1.30600,5.20000) rectangle ++(0.65600,0.65600);
\node[font=\fontsize{8.0}{9.0}\selectfont,inner sep=0pt] at (1.63400,5.52800) {0.02};
\fill[GCRateMid!0.000!GCRateLow] (1.96200,5.20000) rectangle ++(0.65600,0.65600);
\node[font=\fontsize{8.0}{9.0}\selectfont,inner sep=0pt] at (2.29000,5.52800) {0.00};
\fill[GCRateMid!0.000!GCRateLow] (2.61800,5.20000) rectangle ++(0.65600,0.65600);
\node[font=\fontsize{8.0}{9.0}\selectfont,inner sep=0pt] at (2.94600,5.52800) {0.00};
\fill[black!7] (3.27400,5.20000) rectangle ++(0.65600,0.65600);
\node[font=\fontsize{8.0}{9.0}\selectfont,inner sep=0pt] at (3.60200,5.52800) {\textendash};
\draw[black!60,line width=.25pt] (0.65000,5.20000) rectangle ++(3.28000,3.28000);
\node[anchor=north east,font=\fontsize{7.6}{8.6}\selectfont,rotate=45,inner sep=1pt] at (0.97800,5.11000) {@2};
\node[anchor=east,font=\fontsize{7.6}{8.6}\selectfont,inner sep=1pt] at (0.57000,8.15200) {@2};
\node[anchor=north east,font=\fontsize{7.6}{8.6}\selectfont,rotate=45,inner sep=1pt] at (1.63400,5.11000) {@4};
\node[anchor=east,font=\fontsize{7.6}{8.6}\selectfont,inner sep=1pt] at (0.57000,7.49600) {@4};
\node[anchor=north east,font=\fontsize{7.6}{8.6}\selectfont,rotate=45,inner sep=1pt] at (2.29000,5.11000) {@6};
\node[anchor=east,font=\fontsize{7.6}{8.6}\selectfont,inner sep=1pt] at (0.57000,6.84000) {@6};
\node[anchor=north east,font=\fontsize{7.6}{8.6}\selectfont,rotate=45,inner sep=1pt] at (2.94600,5.11000) {@8};
\node[anchor=east,font=\fontsize{7.6}{8.6}\selectfont,inner sep=1pt] at (0.57000,6.18400) {@8};
\node[anchor=north east,font=\fontsize{7.6}{8.6}\selectfont,rotate=45,inner sep=1pt] at (3.60200,5.11000) {@10};
\node[anchor=east,font=\fontsize{7.6}{8.6}\selectfont,inner sep=1pt] at (0.57000,5.52800) {@10};
\node[align=center,font=\fontsize{8.8}{9.8}\selectfont\bfseries,inner sep=0pt] at (2.29000,8.85000) {Cross-attention\\Blocks 1--2};
\fill[black!7] (4.73000,7.66000) rectangle ++(0.82000,0.82000);
\node[font=\fontsize{8.0}{9.0}\selectfont,inner sep=0pt] at (5.14000,8.07000) {\textendash};
\fill[GCRateMid!0.000!GCRateLow] (5.55000,7.66000) rectangle ++(0.82000,0.82000);
\node[font=\fontsize{8.0}{9.0}\selectfont,inner sep=0pt] at (5.96000,8.07000) {0.00};
\fill[GCRateMid!0.000!GCRateLow] (6.37000,7.66000) rectangle ++(0.82000,0.82000);
\node[font=\fontsize{8.0}{9.0}\selectfont,inner sep=0pt] at (6.78000,8.07000) {0.00};
\fill[GCRateMid!0.000!GCRateLow] (7.19000,7.66000) rectangle ++(0.82000,0.82000);
\node[font=\fontsize{8.0}{9.0}\selectfont,inner sep=0pt] at (7.60000,8.07000) {0.00};
\fill[GCRateMid!0.000!GCRateLow] (4.73000,6.84000) rectangle ++(0.82000,0.82000);
\node[font=\fontsize{8.0}{9.0}\selectfont,inner sep=0pt] at (5.14000,7.25000) {0.00};
\fill[black!7] (5.55000,6.84000) rectangle ++(0.82000,0.82000);
\node[font=\fontsize{8.0}{9.0}\selectfont,inner sep=0pt] at (5.96000,7.25000) {\textendash};
\fill[GCRateMid!0.000!GCRateLow] (6.37000,6.84000) rectangle ++(0.82000,0.82000);
\node[font=\fontsize{8.0}{9.0}\selectfont,inner sep=0pt] at (6.78000,7.25000) {0.00};
\fill[GCRateMid!0.000!GCRateLow] (7.19000,6.84000) rectangle ++(0.82000,0.82000);
\node[font=\fontsize{8.0}{9.0}\selectfont,inner sep=0pt] at (7.60000,7.25000) {0.00};
\fill[GCRateMid!0.000!GCRateLow] (4.73000,6.02000) rectangle ++(0.82000,0.82000);
\node[font=\fontsize{8.0}{9.0}\selectfont,inner sep=0pt] at (5.14000,6.43000) {0.00};
\fill[GCRateMid!0.000!GCRateLow] (5.55000,6.02000) rectangle ++(0.82000,0.82000);
\node[font=\fontsize{8.0}{9.0}\selectfont,inner sep=0pt] at (5.96000,6.43000) {0.00};
\fill[black!7] (6.37000,6.02000) rectangle ++(0.82000,0.82000);
\node[font=\fontsize{8.0}{9.0}\selectfont,inner sep=0pt] at (6.78000,6.43000) {\textendash};
\fill[GCRateMid!0.000!GCRateLow] (7.19000,6.02000) rectangle ++(0.82000,0.82000);
\node[font=\fontsize{8.0}{9.0}\selectfont,inner sep=0pt] at (7.60000,6.43000) {0.00};
\fill[GCRateMid!0.000!GCRateLow] (4.73000,5.20000) rectangle ++(0.82000,0.82000);
\node[font=\fontsize{8.0}{9.0}\selectfont,inner sep=0pt] at (5.14000,5.61000) {0.00};
\fill[GCRateMid!0.000!GCRateLow] (5.55000,5.20000) rectangle ++(0.82000,0.82000);
\node[font=\fontsize{8.0}{9.0}\selectfont,inner sep=0pt] at (5.96000,5.61000) {0.00};
\fill[GCRateMid!0.000!GCRateLow] (6.37000,5.20000) rectangle ++(0.82000,0.82000);
\node[font=\fontsize{8.0}{9.0}\selectfont,inner sep=0pt] at (6.78000,5.61000) {0.00};
\fill[black!7] (7.19000,5.20000) rectangle ++(0.82000,0.82000);
\node[font=\fontsize{8.0}{9.0}\selectfont,inner sep=0pt] at (7.60000,5.61000) {\textendash};
\draw[black!60,line width=.25pt] (4.73000,5.20000) rectangle ++(3.28000,3.28000);
\node[anchor=north east,font=\fontsize{7.6}{8.6}\selectfont,rotate=45,inner sep=1pt] at (5.14000,5.11000) {@4};
\node[anchor=east,font=\fontsize{7.6}{8.6}\selectfont,inner sep=1pt] at (4.65000,8.07000) {@4};
\node[anchor=north east,font=\fontsize{7.6}{8.6}\selectfont,rotate=45,inner sep=1pt] at (5.96000,5.11000) {@6};
\node[anchor=east,font=\fontsize{7.6}{8.6}\selectfont,inner sep=1pt] at (4.65000,7.25000) {@6};
\node[anchor=north east,font=\fontsize{7.6}{8.6}\selectfont,rotate=45,inner sep=1pt] at (6.78000,5.11000) {@8};
\node[anchor=east,font=\fontsize{7.6}{8.6}\selectfont,inner sep=1pt] at (4.65000,6.43000) {@8};
\node[anchor=north east,font=\fontsize{7.6}{8.6}\selectfont,rotate=45,inner sep=1pt] at (7.60000,5.11000) {@10};
\node[anchor=east,font=\fontsize{7.6}{8.6}\selectfont,inner sep=1pt] at (4.65000,5.61000) {@10};
\node[align=center,font=\fontsize{8.8}{9.8}\selectfont\bfseries,inner sep=0pt] at (6.37000,8.85000) {Cross-attention\\Blocks 3--4};
\fill[black!7] (8.81000,7.38667) rectangle ++(1.09333,1.09333);
\node[font=\fontsize{8.0}{9.0}\selectfont,inner sep=0pt] at (9.35667,7.93333) {\textendash};
\fill[GCRateMid!0.000!GCRateLow] (9.90333,7.38667) rectangle ++(1.09333,1.09333);
\node[font=\fontsize{8.0}{9.0}\selectfont,inner sep=0pt] at (10.45000,7.93333) {0.00};
\fill[GCRateMid!0.000!GCRateLow] (10.99667,7.38667) rectangle ++(1.09333,1.09333);
\node[font=\fontsize{8.0}{9.0}\selectfont,inner sep=0pt] at (11.54333,7.93333) {0.00};
\fill[GCRateMid!0.000!GCRateLow] (8.81000,6.29333) rectangle ++(1.09333,1.09333);
\node[font=\fontsize{8.0}{9.0}\selectfont,inner sep=0pt] at (9.35667,6.84000) {0.00};
\fill[black!7] (9.90333,6.29333) rectangle ++(1.09333,1.09333);
\node[font=\fontsize{8.0}{9.0}\selectfont,inner sep=0pt] at (10.45000,6.84000) {\textendash};
\fill[GCRateMid!0.000!GCRateLow] (10.99667,6.29333) rectangle ++(1.09333,1.09333);
\node[font=\fontsize{8.0}{9.0}\selectfont,inner sep=0pt] at (11.54333,6.84000) {0.00};
\fill[GCRateMid!0.000!GCRateLow] (8.81000,5.20000) rectangle ++(1.09333,1.09333);
\node[font=\fontsize{8.0}{9.0}\selectfont,inner sep=0pt] at (9.35667,5.74667) {0.00};
\fill[GCRateMid!0.000!GCRateLow] (9.90333,5.20000) rectangle ++(1.09333,1.09333);
\node[font=\fontsize{8.0}{9.0}\selectfont,inner sep=0pt] at (10.45000,5.74667) {0.00};
\fill[black!7] (10.99667,5.20000) rectangle ++(1.09333,1.09333);
\node[font=\fontsize{8.0}{9.0}\selectfont,inner sep=0pt] at (11.54333,5.74667) {\textendash};
\draw[black!60,line width=.25pt] (8.81000,5.20000) rectangle ++(3.28000,3.28000);
\node[anchor=north east,font=\fontsize{7.6}{8.6}\selectfont,rotate=45,inner sep=1pt] at (9.35667,5.11000) {@6};
\node[anchor=east,font=\fontsize{7.6}{8.6}\selectfont,inner sep=1pt] at (8.73000,7.93333) {@6};
\node[anchor=north east,font=\fontsize{7.6}{8.6}\selectfont,rotate=45,inner sep=1pt] at (10.45000,5.11000) {@8};
\node[anchor=east,font=\fontsize{7.6}{8.6}\selectfont,inner sep=1pt] at (8.73000,6.84000) {@8};
\node[anchor=north east,font=\fontsize{7.6}{8.6}\selectfont,rotate=45,inner sep=1pt] at (11.54333,5.11000) {@10};
\node[anchor=east,font=\fontsize{7.6}{8.6}\selectfont,inner sep=1pt] at (8.73000,5.74667) {@10};
\node[align=center,font=\fontsize{8.8}{9.8}\selectfont\bfseries,inner sep=0pt] at (10.45000,8.85000) {Cross-attention\\Blocks 5--6};
\fill[black!7] (12.89000,6.84000) rectangle ++(1.64000,1.64000);
\node[font=\fontsize{8.0}{9.0}\selectfont,inner sep=0pt] at (13.71000,7.66000) {\textendash};
\fill[GCRateMid!0.000!GCRateLow] (14.53000,6.84000) rectangle ++(1.64000,1.64000);
\node[font=\fontsize{8.0}{9.0}\selectfont,inner sep=0pt] at (15.35000,7.66000) {0.00};
\fill[GCRateMid!0.000!GCRateLow] (12.89000,5.20000) rectangle ++(1.64000,1.64000);
\node[font=\fontsize{8.0}{9.0}\selectfont,inner sep=0pt] at (13.71000,6.02000) {0.00};
\fill[black!7] (14.53000,5.20000) rectangle ++(1.64000,1.64000);
\node[font=\fontsize{8.0}{9.0}\selectfont,inner sep=0pt] at (15.35000,6.02000) {\textendash};
\draw[black!60,line width=.25pt] (12.89000,5.20000) rectangle ++(3.28000,3.28000);
\node[anchor=north east,font=\fontsize{7.6}{8.6}\selectfont,rotate=45,inner sep=1pt] at (13.71000,5.11000) {@8};
\node[anchor=east,font=\fontsize{7.6}{8.6}\selectfont,inner sep=1pt] at (12.81000,7.66000) {@8};
\node[anchor=north east,font=\fontsize{7.6}{8.6}\selectfont,rotate=45,inner sep=1pt] at (15.35000,5.11000) {@10};
\node[anchor=east,font=\fontsize{7.6}{8.6}\selectfont,inner sep=1pt] at (12.81000,6.02000) {@10};
\node[align=center,font=\fontsize{8.8}{9.8}\selectfont\bfseries,inner sep=0pt] at (14.53000,8.85000) {Cross-attention\\Blocks 7--8};
\fill[black!7] (6.77000,3.37400) rectangle ++(0.65600,0.65600);
\node[font=\fontsize{8.0}{9.0}\selectfont,inner sep=0pt] at (7.09800,3.70200) {\textendash};
\fill[GCRateMid!0.000!GCRateLow] (7.42600,3.37400) rectangle ++(0.65600,0.65600);
\node[font=\fontsize{8.0}{9.0}\selectfont,inner sep=0pt] at (7.75400,3.70200) {0.00};
\fill[GCRateMid!0.000!GCRateLow] (8.08200,3.37400) rectangle ++(0.65600,0.65600);
\node[font=\fontsize{8.0}{9.0}\selectfont,inner sep=0pt] at (8.41000,3.70200) {0.00};
\fill[GCRateMid!4.000!GCRateLow] (8.73800,3.37400) rectangle ++(0.65600,0.65600);
\node[font=\fontsize{8.0}{9.0}\selectfont,inner sep=0pt] at (9.06600,3.70200) {0.02};
\fill[GCRateMid!18.000!GCRateLow] (9.39400,3.37400) rectangle ++(0.65600,0.65600);
\node[font=\fontsize{8.0}{9.0}\selectfont,inner sep=0pt] at (9.72200,3.70200) {0.09};
\fill[GCRateMid!0.000!GCRateLow] (6.77000,2.71800) rectangle ++(0.65600,0.65600);
\node[font=\fontsize{8.0}{9.0}\selectfont,inner sep=0pt] at (7.09800,3.04600) {0.00};
\fill[black!7] (7.42600,2.71800) rectangle ++(0.65600,0.65600);
\node[font=\fontsize{8.0}{9.0}\selectfont,inner sep=0pt] at (7.75400,3.04600) {\textendash};
\fill[GCRateMid!0.000!GCRateLow] (8.08200,2.71800) rectangle ++(0.65600,0.65600);
\node[font=\fontsize{8.0}{9.0}\selectfont,inner sep=0pt] at (8.41000,3.04600) {0.00};
\fill[GCRateMid!0.000!GCRateLow] (8.73800,2.71800) rectangle ++(0.65600,0.65600);
\node[font=\fontsize{8.0}{9.0}\selectfont,inner sep=0pt] at (9.06600,3.04600) {0.00};
\fill[GCRateMid!0.000!GCRateLow] (9.39400,2.71800) rectangle ++(0.65600,0.65600);
\node[font=\fontsize{8.0}{9.0}\selectfont,inner sep=0pt] at (9.72200,3.04600) {0.00};
\fill[GCRateMid!0.000!GCRateLow] (6.77000,2.06200) rectangle ++(0.65600,0.65600);
\node[font=\fontsize{8.0}{9.0}\selectfont,inner sep=0pt] at (7.09800,2.39000) {0.00};
\fill[GCRateMid!0.000!GCRateLow] (7.42600,2.06200) rectangle ++(0.65600,0.65600);
\node[font=\fontsize{8.0}{9.0}\selectfont,inner sep=0pt] at (7.75400,2.39000) {0.00};
\fill[black!7] (8.08200,2.06200) rectangle ++(0.65600,0.65600);
\node[font=\fontsize{8.0}{9.0}\selectfont,inner sep=0pt] at (8.41000,2.39000) {\textendash};
\fill[GCRateMid!0.000!GCRateLow] (8.73800,2.06200) rectangle ++(0.65600,0.65600);
\node[font=\fontsize{8.0}{9.0}\selectfont,inner sep=0pt] at (9.06600,2.39000) {0.00};
\fill[GCRateMid!0.000!GCRateLow] (9.39400,2.06200) rectangle ++(0.65600,0.65600);
\node[font=\fontsize{8.0}{9.0}\selectfont,inner sep=0pt] at (9.72200,2.39000) {0.00};
\fill[GCRateMid!4.000!GCRateLow] (6.77000,1.40600) rectangle ++(0.65600,0.65600);
\node[font=\fontsize{8.0}{9.0}\selectfont,inner sep=0pt] at (7.09800,1.73400) {0.02};
\fill[GCRateMid!0.000!GCRateLow] (7.42600,1.40600) rectangle ++(0.65600,0.65600);
\node[font=\fontsize{8.0}{9.0}\selectfont,inner sep=0pt] at (7.75400,1.73400) {0.00};
\fill[GCRateMid!0.000!GCRateLow] (8.08200,1.40600) rectangle ++(0.65600,0.65600);
\node[font=\fontsize{8.0}{9.0}\selectfont,inner sep=0pt] at (8.41000,1.73400) {0.00};
\fill[black!7] (8.73800,1.40600) rectangle ++(0.65600,0.65600);
\node[font=\fontsize{8.0}{9.0}\selectfont,inner sep=0pt] at (9.06600,1.73400) {\textendash};
\fill[GCRateMid!0.000!GCRateLow] (9.39400,1.40600) rectangle ++(0.65600,0.65600);
\node[font=\fontsize{8.0}{9.0}\selectfont,inner sep=0pt] at (9.72200,1.73400) {0.00};
\fill[GCRateMid!18.000!GCRateLow] (6.77000,0.75000) rectangle ++(0.65600,0.65600);
\node[font=\fontsize{8.0}{9.0}\selectfont,inner sep=0pt] at (7.09800,1.07800) {0.09};
\fill[GCRateMid!0.000!GCRateLow] (7.42600,0.75000) rectangle ++(0.65600,0.65600);
\node[font=\fontsize{8.0}{9.0}\selectfont,inner sep=0pt] at (7.75400,1.07800) {0.00};
\fill[GCRateMid!0.000!GCRateLow] (8.08200,0.75000) rectangle ++(0.65600,0.65600);
\node[font=\fontsize{8.0}{9.0}\selectfont,inner sep=0pt] at (8.41000,1.07800) {0.00};
\fill[GCRateMid!0.000!GCRateLow] (8.73800,0.75000) rectangle ++(0.65600,0.65600);
\node[font=\fontsize{8.0}{9.0}\selectfont,inner sep=0pt] at (9.06600,1.07800) {0.00};
\fill[black!7] (9.39400,0.75000) rectangle ++(0.65600,0.65600);
\node[font=\fontsize{8.0}{9.0}\selectfont,inner sep=0pt] at (9.72200,1.07800) {\textendash};
\draw[black!60,line width=.25pt] (6.77000,0.75000) rectangle ++(3.28000,3.28000);
\node[anchor=north east,font=\fontsize{7.6}{8.6}\selectfont,rotate=45,inner sep=1pt] at (7.09800,0.66000) {@2};
\node[anchor=east,font=\fontsize{7.6}{8.6}\selectfont,inner sep=1pt] at (6.69000,3.70200) {@2};
\node[anchor=north east,font=\fontsize{7.6}{8.6}\selectfont,rotate=45,inner sep=1pt] at (7.75400,0.66000) {@4};
\node[anchor=east,font=\fontsize{7.6}{8.6}\selectfont,inner sep=1pt] at (6.69000,3.04600) {@4};
\node[anchor=north east,font=\fontsize{7.6}{8.6}\selectfont,rotate=45,inner sep=1pt] at (8.41000,0.66000) {@6};
\node[anchor=east,font=\fontsize{7.6}{8.6}\selectfont,inner sep=1pt] at (6.69000,2.39000) {@6};
\node[anchor=north east,font=\fontsize{7.6}{8.6}\selectfont,rotate=45,inner sep=1pt] at (9.06600,0.66000) {@8};
\node[anchor=east,font=\fontsize{7.6}{8.6}\selectfont,inner sep=1pt] at (6.69000,1.73400) {@8};
\node[anchor=north east,font=\fontsize{7.6}{8.6}\selectfont,rotate=45,inner sep=1pt] at (9.72200,0.66000) {@10};
\node[anchor=east,font=\fontsize{7.6}{8.6}\selectfont,inner sep=1pt] at (6.69000,1.07800) {@10};
\node[align=center,font=\fontsize{8.8}{9.8}\selectfont\bfseries,inner sep=0pt] at (8.41000,4.40000) {Shared exit FFN};
\begin{scope}[shift={(-0.60,-1.752)}]
\shade[bottom color=GCRateLow,top color=GCRateHigh,middle color=GCRateMid] (13.18000,2.50200) rectangle ++(0.19000,3.00600);
\draw[black!60,line width=.25pt] (13.18000,2.50200) rectangle ++(0.19000,3.00600);
\draw[black!65,line width=.25pt] (13.37000,2.50200) -- ++(.045,0);
\node[anchor=west,font=\fontsize{6.5}{7.5}\selectfont,inner sep=1pt] at (13.47000,2.50200) {0.0};
\draw[black!65,line width=.25pt] (13.37000,4.00500) -- ++(.045,0);
\node[anchor=west,font=\fontsize{6.5}{7.5}\selectfont,inner sep=1pt] at (13.47000,4.00500) {0.5};
\draw[black!65,line width=.25pt] (13.37000,5.50800) -- ++(.045,0);
\node[anchor=west,font=\fontsize{6.5}{7.5}\selectfont,inner sep=1pt] at (13.47000,5.50800) {1.0};
\node[rotate=90,font=\fontsize{6.8}{7.8}\selectfont,inner sep=0pt] at (14.00,4.00500) {Conflict rate};

\end{scope}
\end{tikzpicture}%
\endgroup%

%% file: artwork/gradient_conflict/further/declarevit_gradient_norm_epoch010.tex
\begingroup%
\definecolor{GNBlue}{RGB}{0,114,178}%
\definecolor{GNGreen}{RGB}{0,158,115}%
\definecolor{GNPink}{RGB}{204,121,167}%
\definecolor{GNOchre}{RGB}{181,137,0}%
\definecolor{GNGray}{RGB}{102,102,102}%
\begin{tikzpicture}[x=1cm,y=1cm]
\path[use as bounding box] (-0.10,0.00) rectangle (16.35,9.25);
\node[font=\fontsize{9.4}{10.4}\selectfont\bfseries,inner sep=0pt] at (8.125,9.04) {(B) Loss-weighted gradient norms};
\node[align=center,font=\fontsize{8.4}{9.4}\selectfont\bfseries,inner sep=0pt] at (2.290,8.400) {Cross-attention\\Blocks 1--2};
\node[anchor=east,font=\fontsize{7.6}{8.6}\selectfont,inner sep=.4pt] at (0.592,4.900) {0.00};
\draw[black!12,line width=.20pt] (0.650,5.417) -- (3.930,5.417);
\node[anchor=east,font=\fontsize{7.6}{8.6}\selectfont,inner sep=.4pt] at (0.592,5.417) {0.20};
\draw[black!12,line width=.20pt] (0.650,5.933) -- (3.930,5.933);
\node[anchor=east,font=\fontsize{7.6}{8.6}\selectfont,inner sep=.4pt] at (0.592,5.933) {0.40};
\draw[black!12,line width=.20pt] (0.650,6.450) -- (3.930,6.450);
\node[anchor=east,font=\fontsize{7.6}{8.6}\selectfont,inner sep=.4pt] at (0.592,6.450) {0.60};
\draw[black!12,line width=.20pt] (0.650,6.967) -- (3.930,6.967);
\node[anchor=east,font=\fontsize{7.6}{8.6}\selectfont,inner sep=.4pt] at (0.592,6.967) {0.80};
\draw[black!12,line width=.20pt] (0.650,7.483) -- (3.930,7.483);
\node[anchor=east,font=\fontsize{7.6}{8.6}\selectfont,inner sep=.4pt] at (0.592,7.483) {1.00};
\draw[black!12,line width=.20pt] (0.650,8.000) -- (3.930,8.000);
\node[anchor=east,font=\fontsize{7.6}{8.6}\selectfont,inner sep=.4pt] at (0.592,8.000) {1.20};
\draw[black!75,line width=.30pt] (0.650,4.900) -- (3.930,4.900);
\draw[black!75,line width=.30pt] (0.650,4.900) -- (0.650,8.000);
\fill[GNBlue] (0.7681,4.9000) rectangle ++(0.4198,2.7383);
\node[anchor=south,font=\fontsize{7.4}{8.4}\selectfont,inner sep=.25pt] at (0.9780,7.6983) {1.060};
\node[anchor=north east,rotate=45,font=\fontsize{7.6}{8.6}\selectfont,inner sep=.25pt] at (0.9780,4.8000) {@2};
\fill[GNGreen] (1.4241,4.9000) rectangle ++(0.4198,1.5087);
\node[anchor=south,font=\fontsize{7.4}{8.4}\selectfont,inner sep=.25pt] at (1.6340,6.4687) {0.584};
\node[anchor=north east,rotate=45,font=\fontsize{7.6}{8.6}\selectfont,inner sep=.25pt] at (1.6340,4.8000) {@4};
\fill[GNPink] (2.0801,4.9000) rectangle ++(0.4198,0.8422);
\node[anchor=south,font=\fontsize{7.4}{8.4}\selectfont,inner sep=.25pt] at (2.2900,5.8022) {0.326};
\node[anchor=north east,rotate=45,font=\fontsize{7.6}{8.6}\selectfont,inner sep=.25pt] at (2.2900,4.8000) {@6};
\fill[GNOchre] (2.7361,4.9000) rectangle ++(0.4198,0.6381);
\node[anchor=south,font=\fontsize{7.4}{8.4}\selectfont,inner sep=.25pt] at (2.9460,5.5981) {0.247};
\node[anchor=north east,rotate=45,font=\fontsize{7.6}{8.6}\selectfont,inner sep=.25pt] at (2.9460,4.8000) {@8};
\fill[GNGray] (3.3921,4.9000) rectangle ++(0.4198,0.4314);
\node[anchor=south,font=\fontsize{7.4}{8.4}\selectfont,inner sep=.25pt] at (3.6020,5.3914) {0.167};
\node[anchor=north east,rotate=45,font=\fontsize{7.6}{8.6}\selectfont,inner sep=.25pt] at (3.6020,4.8000) {@10};
\node[align=center,font=\fontsize{8.4}{9.4}\selectfont\bfseries,inner sep=0pt] at (6.370,8.400) {Cross-attention\\Blocks 3--4};
\draw[black!12,line width=.20pt] (4.730,5.417) -- (8.010,5.417);
\draw[black!12,line width=.20pt] (4.730,5.933) -- (8.010,5.933);
\draw[black!12,line width=.20pt] (4.730,6.450) -- (8.010,6.450);
\draw[black!12,line width=.20pt] (4.730,6.967) -- (8.010,6.967);
\draw[black!12,line width=.20pt] (4.730,7.483) -- (8.010,7.483);
\draw[black!12,line width=.20pt] (4.730,8.000) -- (8.010,8.000);
\draw[black!75,line width=.30pt] (4.730,4.900) -- (8.010,4.900);
\draw[black!75,line width=.30pt] (4.730,4.900) -- (4.730,8.000);
\fill[GNGreen] (4.8776,4.9000) rectangle ++(0.5248,1.4441);
\node[anchor=south,font=\fontsize{8.0}{9.0}\selectfont,inner sep=.25pt] at (5.1400,6.4041) {0.559};
\node[anchor=north east,rotate=45,font=\fontsize{7.6}{8.6}\selectfont,inner sep=.25pt] at (5.1400,4.8000) {@4};
\fill[GNPink] (5.6976,4.9000) rectangle ++(0.5248,0.6768);
\node[anchor=south,font=\fontsize{8.0}{9.0}\selectfont,inner sep=.25pt] at (5.9600,5.6368) {0.262};
\node[anchor=north east,rotate=45,font=\fontsize{7.6}{8.6}\selectfont,inner sep=.25pt] at (5.9600,4.8000) {@6};
\fill[GNOchre] (6.5176,4.9000) rectangle ++(0.5248,0.4521);
\node[anchor=south,font=\fontsize{8.0}{9.0}\selectfont,inner sep=.25pt] at (6.7800,5.4121) {0.175};
\node[anchor=north east,rotate=45,font=\fontsize{7.6}{8.6}\selectfont,inner sep=.25pt] at (6.7800,4.8000) {@8};
\fill[GNGray] (7.3376,4.9000) rectangle ++(0.5248,0.3074);
\node[anchor=south,font=\fontsize{8.0}{9.0}\selectfont,inner sep=.25pt] at (7.6000,5.2674) {0.119};
\node[anchor=north east,rotate=45,font=\fontsize{7.6}{8.6}\selectfont,inner sep=.25pt] at (7.6000,4.8000) {@10};
\node[align=center,font=\fontsize{8.4}{9.4}\selectfont\bfseries,inner sep=0pt] at (10.450,8.400) {Cross-attention\\Blocks 5--6};
\draw[black!12,line width=.20pt] (8.810,5.417) -- (12.090,5.417);
\draw[black!12,line width=.20pt] (8.810,5.933) -- (12.090,5.933);
\draw[black!12,line width=.20pt] (8.810,6.450) -- (12.090,6.450);
\draw[black!12,line width=.20pt] (8.810,6.967) -- (12.090,6.967);
\draw[black!12,line width=.20pt] (8.810,7.483) -- (12.090,7.483);
\draw[black!12,line width=.20pt] (8.810,8.000) -- (12.090,8.000);
\draw[black!75,line width=.30pt] (8.810,4.900) -- (12.090,4.900);
\draw[black!75,line width=.30pt] (8.810,4.900) -- (8.810,8.000);
\fill[GNPink] (9.0068,4.9000) rectangle ++(0.6997,0.9894);
\node[anchor=south,font=\fontsize{8.0}{9.0}\selectfont,inner sep=.25pt] at (9.3567,5.9494) {0.383};
\node[anchor=north east,rotate=45,font=\fontsize{7.6}{8.6}\selectfont,inner sep=.25pt] at (9.3567,4.8000) {@6};
\fill[GNOchre] (10.1001,4.9000) rectangle ++(0.6997,0.5373);
\node[anchor=south,font=\fontsize{8.0}{9.0}\selectfont,inner sep=.25pt] at (10.4500,5.4973) {0.208};
\node[anchor=north east,rotate=45,font=\fontsize{7.6}{8.6}\selectfont,inner sep=.25pt] at (10.4500,4.8000) {@8};
\fill[GNGray] (11.1935,4.9000) rectangle ++(0.6997,0.3307);
\node[anchor=south,font=\fontsize{8.0}{9.0}\selectfont,inner sep=.25pt] at (11.5433,5.2907) {0.128};
\node[anchor=north east,rotate=45,font=\fontsize{7.6}{8.6}\selectfont,inner sep=.25pt] at (11.5433,4.8000) {@10};
\node[align=center,font=\fontsize{8.4}{9.4}\selectfont\bfseries,inner sep=0pt] at (14.530,8.400) {Cross-attention\\Blocks 7--8};
\draw[black!12,line width=.20pt] (12.890,5.417) -- (16.170,5.417);
\draw[black!12,line width=.20pt] (12.890,5.933) -- (16.170,5.933);
\draw[black!12,line width=.20pt] (12.890,6.450) -- (16.170,6.450);
\draw[black!12,line width=.20pt] (12.890,6.967) -- (16.170,6.967);
\draw[black!12,line width=.20pt] (12.890,7.483) -- (16.170,7.483);
\draw[black!12,line width=.20pt] (12.890,8.000) -- (16.170,8.000);
\draw[black!75,line width=.30pt] (12.890,4.900) -- (16.170,4.900);
\draw[black!75,line width=.30pt] (12.890,4.900) -- (12.890,8.000);
\fill[GNOchre] (13.1852,4.9000) rectangle ++(1.0496,0.7492);
\node[anchor=south,font=\fontsize{8.0}{9.0}\selectfont,inner sep=.25pt] at (13.7100,5.7092) {0.290};
\node[anchor=north east,rotate=45,font=\fontsize{7.6}{8.6}\selectfont,inner sep=.25pt] at (13.7100,4.8000) {@8};
\fill[GNGray] (14.8252,4.9000) rectangle ++(1.0496,0.2971);
\node[anchor=south,font=\fontsize{8.0}{9.0}\selectfont,inner sep=.25pt] at (15.3500,5.2571) {0.115};
\node[anchor=north east,rotate=45,font=\fontsize{7.6}{8.6}\selectfont,inner sep=.25pt] at (15.3500,4.8000) {@10};
\node[align=center,font=\fontsize{8.4}{9.4}\selectfont\bfseries,inner sep=0pt] at (8.410,4.140) {Shared exit FFN};
\node[anchor=east,font=\fontsize{7.6}{8.6}\selectfont,inner sep=.4pt] at (6.712,0.850) {0.00};
\draw[black!12,line width=.20pt] (6.770,1.367) -- (10.050,1.367);
\node[anchor=east,font=\fontsize{7.6}{8.6}\selectfont,inner sep=.4pt] at (6.712,1.367) {0.20};
\draw[black!12,line width=.20pt] (6.770,1.883) -- (10.050,1.883);
\node[anchor=east,font=\fontsize{7.6}{8.6}\selectfont,inner sep=.4pt] at (6.712,1.883) {0.40};
\draw[black!12,line width=.20pt] (6.770,2.400) -- (10.050,2.400);
\node[anchor=east,font=\fontsize{7.6}{8.6}\selectfont,inner sep=.4pt] at (6.712,2.400) {0.60};
\draw[black!12,line width=.20pt] (6.770,2.917) -- (10.050,2.917);
\node[anchor=east,font=\fontsize{7.6}{8.6}\selectfont,inner sep=.4pt] at (6.712,2.917) {0.80};
\draw[black!12,line width=.20pt] (6.770,3.433) -- (10.050,3.433);
\node[anchor=east,font=\fontsize{7.6}{8.6}\selectfont,inner sep=.4pt] at (6.712,3.433) {1.00};
\draw[black!12,line width=.20pt] (6.770,3.950) -- (10.050,3.950);
\node[anchor=east,font=\fontsize{7.6}{8.6}\selectfont,inner sep=.4pt] at (6.712,3.950) {1.20};
\draw[black!75,line width=.30pt] (6.770,0.850) -- (10.050,0.850);
\draw[black!75,line width=.30pt] (6.770,0.850) -- (6.770,3.950);
\fill[GNBlue] (6.8881,0.8500) rectangle ++(0.4198,0.9843);
\node[anchor=south,font=\fontsize{7.4}{8.4}\selectfont,inner sep=.25pt] at (7.0980,1.8942) {0.381};
\node[anchor=north east,rotate=45,font=\fontsize{7.6}{8.6}\selectfont,inner sep=.25pt] at (7.0980,0.7500) {@2};
\fill[GNGreen] (7.5441,0.8500) rectangle ++(0.4198,0.7672);
\node[anchor=south,font=\fontsize{7.4}{8.4}\selectfont,inner sep=.25pt] at (7.7540,1.6772) {0.297};
\node[anchor=north east,rotate=45,font=\fontsize{7.6}{8.6}\selectfont,inner sep=.25pt] at (7.7540,0.7500) {@4};
\fill[GNPink] (8.2001,0.8500) rectangle ++(0.4198,0.5683);
\node[anchor=south,font=\fontsize{7.4}{8.4}\selectfont,inner sep=.25pt] at (8.4100,1.4783) {0.220};
\node[anchor=north east,rotate=45,font=\fontsize{7.6}{8.6}\selectfont,inner sep=.25pt] at (8.4100,0.7500) {@6};
\fill[GNOchre] (8.8561,0.8500) rectangle ++(0.4198,0.4340);
\node[anchor=south,font=\fontsize{7.4}{8.4}\selectfont,inner sep=.25pt] at (9.0660,1.3440) {0.168};
\node[anchor=north east,rotate=45,font=\fontsize{7.6}{8.6}\selectfont,inner sep=.25pt] at (9.0660,0.7500) {@8};
\fill[GNGray] (9.5121,0.8500) rectangle ++(0.4198,0.2971);
\node[anchor=south,font=\fontsize{7.4}{8.4}\selectfont,inner sep=.25pt] at (9.7220,1.2071) {0.115};
\node[anchor=north east,rotate=45,font=\fontsize{7.6}{8.6}\selectfont,inner sep=.25pt] at (9.7220,0.7500) {@10};
\end{tikzpicture}%
\endgroup%

%% file: artwork/gradient_conflict/further/declarevit_conflict_epoch100.tex
\begingroup%
\definecolor{GCRateLow}{RGB}{255,245,240}%
\definecolor{GCRateMid}{RGB}{251,106,74}%
\definecolor{GCRateHigh}{RGB}{165,0,38}%
\begin{tikzpicture}[x=1cm,y=1cm]
\path[use as bounding box] (-0.10,-0.05) rectangle (16.35,9.65);
\node[font=\fontsize{9.4}{10.4}\selectfont\bfseries,inner sep=0pt] at (8.125,9.43) {(A) Pairwise gradient conflict rates};
\fill[black!7] (0.65000,7.82400) rectangle ++(0.65600,0.65600);
\node[font=\fontsize{8.0}{9.0}\selectfont,inner sep=0pt] at (0.97800,8.15200) {\textendash};
\fill[GCRateMid!0.000!GCRateLow] (1.30600,7.82400) rectangle ++(0.65600,0.65600);
\node[font=\fontsize{8.0}{9.0}\selectfont,inner sep=0pt] at (1.63400,8.15200) {0.00};
\fill[GCRateMid!0.000!GCRateLow] (1.96200,7.82400) rectangle ++(0.65600,0.65600);
\node[font=\fontsize{8.0}{9.0}\selectfont,inner sep=0pt] at (2.29000,8.15200) {0.00};
\fill[GCRateMid!0.000!GCRateLow] (2.61800,7.82400) rectangle ++(0.65600,0.65600);
\node[font=\fontsize{8.0}{9.0}\selectfont,inner sep=0pt] at (2.94600,8.15200) {0.00};
\fill[GCRateMid!4.000!GCRateLow] (3.27400,7.82400) rectangle ++(0.65600,0.65600);
\node[font=\fontsize{8.0}{9.0}\selectfont,inner sep=0pt] at (3.60200,8.15200) {0.02};
\fill[GCRateMid!0.000!GCRateLow] (0.65000,7.16800) rectangle ++(0.65600,0.65600);
\node[font=\fontsize{8.0}{9.0}\selectfont,inner sep=0pt] at (0.97800,7.49600) {0.00};
\fill[black!7] (1.30600,7.16800) rectangle ++(0.65600,0.65600);
\node[font=\fontsize{8.0}{9.0}\selectfont,inner sep=0pt] at (1.63400,7.49600) {\textendash};
\fill[GCRateMid!0.000!GCRateLow] (1.96200,7.16800) rectangle ++(0.65600,0.65600);
\node[font=\fontsize{8.0}{9.0}\selectfont,inner sep=0pt] at (2.29000,7.49600) {0.00};
\fill[GCRateMid!0.000!GCRateLow] (2.61800,7.16800) rectangle ++(0.65600,0.65600);
\node[font=\fontsize{8.0}{9.0}\selectfont,inner sep=0pt] at (2.94600,7.49600) {0.00};
\fill[GCRateMid!0.000!GCRateLow] (3.27400,7.16800) rectangle ++(0.65600,0.65600);
\node[font=\fontsize{8.0}{9.0}\selectfont,inner sep=0pt] at (3.60200,7.49600) {0.00};
\fill[GCRateMid!0.000!GCRateLow] (0.65000,6.51200) rectangle ++(0.65600,0.65600);
\node[font=\fontsize{8.0}{9.0}\selectfont,inner sep=0pt] at (0.97800,6.84000) {0.00};
\fill[GCRateMid!0.000!GCRateLow] (1.30600,6.51200) rectangle ++(0.65600,0.65600);
\node[font=\fontsize{8.0}{9.0}\selectfont,inner sep=0pt] at (1.63400,6.84000) {0.00};
\fill[black!7] (1.96200,6.51200) rectangle ++(0.65600,0.65600);
\node[font=\fontsize{8.0}{9.0}\selectfont,inner sep=0pt] at (2.29000,6.84000) {\textendash};
\fill[GCRateMid!0.000!GCRateLow] (2.61800,6.51200) rectangle ++(0.65600,0.65600);
\node[font=\fontsize{8.0}{9.0}\selectfont,inner sep=0pt] at (2.94600,6.84000) {0.00};
\fill[GCRateMid!0.000!GCRateLow] (3.27400,6.51200) rectangle ++(0.65600,0.65600);
\node[font=\fontsize{8.0}{9.0}\selectfont,inner sep=0pt] at (3.60200,6.84000) {0.00};
\fill[GCRateMid!0.000!GCRateLow] (0.65000,5.85600) rectangle ++(0.65600,0.65600);
\node[font=\fontsize{8.0}{9.0}\selectfont,inner sep=0pt] at (0.97800,6.18400) {0.00};
\fill[GCRateMid!0.000!GCRateLow] (1.30600,5.85600) rectangle ++(0.65600,0.65600);
\node[font=\fontsize{8.0}{9.0}\selectfont,inner sep=0pt] at (1.63400,6.18400) {0.00};
\fill[GCRateMid!0.000!GCRateLow] (1.96200,5.85600) rectangle ++(0.65600,0.65600);
\node[font=\fontsize{8.0}{9.0}\selectfont,inner sep=0pt] at (2.29000,6.18400) {0.00};
\fill[black!7] (2.61800,5.85600) rectangle ++(0.65600,0.65600);
\node[font=\fontsize{8.0}{9.0}\selectfont,inner sep=0pt] at (2.94600,6.18400) {\textendash};
\fill[GCRateMid!0.000!GCRateLow] (3.27400,5.85600) rectangle ++(0.65600,0.65600);
\node[font=\fontsize{8.0}{9.0}\selectfont,inner sep=0pt] at (3.60200,6.18400) {0.00};
\fill[GCRateMid!4.000!GCRateLow] (0.65000,5.20000) rectangle ++(0.65600,0.65600);
\node[font=\fontsize{8.0}{9.0}\selectfont,inner sep=0pt] at (0.97800,5.52800) {0.02};
\fill[GCRateMid!0.000!GCRateLow] (1.30600,5.20000) rectangle ++(0.65600,0.65600);
\node[font=\fontsize{8.0}{9.0}\selectfont,inner sep=0pt] at (1.63400,5.52800) {0.00};
\fill[GCRateMid!0.000!GCRateLow] (1.96200,5.20000) rectangle ++(0.65600,0.65600);
\node[font=\fontsize{8.0}{9.0}\selectfont,inner sep=0pt] at (2.29000,5.52800) {0.00};
\fill[GCRateMid!0.000!GCRateLow] (2.61800,5.20000) rectangle ++(0.65600,0.65600);
\node[font=\fontsize{8.0}{9.0}\selectfont,inner sep=0pt] at (2.94600,5.52800) {0.00};
\fill[black!7] (3.27400,5.20000) rectangle ++(0.65600,0.65600);
\node[font=\fontsize{8.0}{9.0}\selectfont,inner sep=0pt] at (3.60200,5.52800) {\textendash};
\draw[black!60,line width=.25pt] (0.65000,5.20000) rectangle ++(3.28000,3.28000);
\node[anchor=north east,font=\fontsize{7.6}{8.6}\selectfont,rotate=45,inner sep=1pt] at (0.97800,5.11000) {@2};
\node[anchor=east,font=\fontsize{7.6}{8.6}\selectfont,inner sep=1pt] at (0.57000,8.15200) {@2};
\node[anchor=north east,font=\fontsize{7.6}{8.6}\selectfont,rotate=45,inner sep=1pt] at (1.63400,5.11000) {@4};
\node[anchor=east,font=\fontsize{7.6}{8.6}\selectfont,inner sep=1pt] at (0.57000,7.49600) {@4};
\node[anchor=north east,font=\fontsize{7.6}{8.6}\selectfont,rotate=45,inner sep=1pt] at (2.29000,5.11000) {@6};
\node[anchor=east,font=\fontsize{7.6}{8.6}\selectfont,inner sep=1pt] at (0.57000,6.84000) {@6};
\node[anchor=north east,font=\fontsize{7.6}{8.6}\selectfont,rotate=45,inner sep=1pt] at (2.94600,5.11000) {@8};
\node[anchor=east,font=\fontsize{7.6}{8.6}\selectfont,inner sep=1pt] at (0.57000,6.18400) {@8};
\node[anchor=north east,font=\fontsize{7.6}{8.6}\selectfont,rotate=45,inner sep=1pt] at (3.60200,5.11000) {@10};
\node[anchor=east,font=\fontsize{7.6}{8.6}\selectfont,inner sep=1pt] at (0.57000,5.52800) {@10};
\node[align=center,font=\fontsize{8.8}{9.8}\selectfont\bfseries,inner sep=0pt] at (2.29000,8.85000) {Cross-attention\\Blocks 1--2};
\fill[black!7] (4.73000,7.66000) rectangle ++(0.82000,0.82000);
\node[font=\fontsize{8.0}{9.0}\selectfont,inner sep=0pt] at (5.14000,8.07000) {\textendash};
\fill[GCRateMid!0.000!GCRateLow] (5.55000,7.66000) rectangle ++(0.82000,0.82000);
\node[font=\fontsize{8.0}{9.0}\selectfont,inner sep=0pt] at (5.96000,8.07000) {0.00};
\fill[GCRateMid!0.000!GCRateLow] (6.37000,7.66000) rectangle ++(0.82000,0.82000);
\node[font=\fontsize{8.0}{9.0}\selectfont,inner sep=0pt] at (6.78000,8.07000) {0.00};
\fill[GCRateMid!0.000!GCRateLow] (7.19000,7.66000) rectangle ++(0.82000,0.82000);
\node[font=\fontsize{8.0}{9.0}\selectfont,inner sep=0pt] at (7.60000,8.07000) {0.00};
\fill[GCRateMid!0.000!GCRateLow] (4.73000,6.84000) rectangle ++(0.82000,0.82000);
\node[font=\fontsize{8.0}{9.0}\selectfont,inner sep=0pt] at (5.14000,7.25000) {0.00};
\fill[black!7] (5.55000,6.84000) rectangle ++(0.82000,0.82000);
\node[font=\fontsize{8.0}{9.0}\selectfont,inner sep=0pt] at (5.96000,7.25000) {\textendash};
\fill[GCRateMid!0.000!GCRateLow] (6.37000,6.84000) rectangle ++(0.82000,0.82000);
\node[font=\fontsize{8.0}{9.0}\selectfont,inner sep=0pt] at (6.78000,7.25000) {0.00};
\fill[GCRateMid!0.000!GCRateLow] (7.19000,6.84000) rectangle ++(0.82000,0.82000);
\node[font=\fontsize{8.0}{9.0}\selectfont,inner sep=0pt] at (7.60000,7.25000) {0.00};
\fill[GCRateMid!0.000!GCRateLow] (4.73000,6.02000) rectangle ++(0.82000,0.82000);
\node[font=\fontsize{8.0}{9.0}\selectfont,inner sep=0pt] at (5.14000,6.43000) {0.00};
\fill[GCRateMid!0.000!GCRateLow] (5.55000,6.02000) rectangle ++(0.82000,0.82000);
\node[font=\fontsize{8.0}{9.0}\selectfont,inner sep=0pt] at (5.96000,6.43000) {0.00};
\fill[black!7] (6.37000,6.02000) rectangle ++(0.82000,0.82000);
\node[font=\fontsize{8.0}{9.0}\selectfont,inner sep=0pt] at (6.78000,6.43000) {\textendash};
\fill[GCRateMid!0.000!GCRateLow] (7.19000,6.02000) rectangle ++(0.82000,0.82000);
\node[font=\fontsize{8.0}{9.0}\selectfont,inner sep=0pt] at (7.60000,6.43000) {0.00};
\fill[GCRateMid!0.000!GCRateLow] (4.73000,5.20000) rectangle ++(0.82000,0.82000);
\node[font=\fontsize{8.0}{9.0}\selectfont,inner sep=0pt] at (5.14000,5.61000) {0.00};
\fill[GCRateMid!0.000!GCRateLow] (5.55000,5.20000) rectangle ++(0.82000,0.82000);
\node[font=\fontsize{8.0}{9.0}\selectfont,inner sep=0pt] at (5.96000,5.61000) {0.00};
\fill[GCRateMid!0.000!GCRateLow] (6.37000,5.20000) rectangle ++(0.82000,0.82000);
\node[font=\fontsize{8.0}{9.0}\selectfont,inner sep=0pt] at (6.78000,5.61000) {0.00};
\fill[black!7] (7.19000,5.20000) rectangle ++(0.82000,0.82000);
\node[font=\fontsize{8.0}{9.0}\selectfont,inner sep=0pt] at (7.60000,5.61000) {\textendash};
\draw[black!60,line width=.25pt] (4.73000,5.20000) rectangle ++(3.28000,3.28000);
\node[anchor=north east,font=\fontsize{7.6}{8.6}\selectfont,rotate=45,inner sep=1pt] at (5.14000,5.11000) {@4};
\node[anchor=east,font=\fontsize{7.6}{8.6}\selectfont,inner sep=1pt] at (4.65000,8.07000) {@4};
\node[anchor=north east,font=\fontsize{7.6}{8.6}\selectfont,rotate=45,inner sep=1pt] at (5.96000,5.11000) {@6};
\node[anchor=east,font=\fontsize{7.6}{8.6}\selectfont,inner sep=1pt] at (4.65000,7.25000) {@6};
\node[anchor=north east,font=\fontsize{7.6}{8.6}\selectfont,rotate=45,inner sep=1pt] at (6.78000,5.11000) {@8};
\node[anchor=east,font=\fontsize{7.6}{8.6}\selectfont,inner sep=1pt] at (4.65000,6.43000) {@8};
\node[anchor=north east,font=\fontsize{7.6}{8.6}\selectfont,rotate=45,inner sep=1pt] at (7.60000,5.11000) {@10};
\node[anchor=east,font=\fontsize{7.6}{8.6}\selectfont,inner sep=1pt] at (4.65000,5.61000) {@10};
\node[align=center,font=\fontsize{8.8}{9.8}\selectfont\bfseries,inner sep=0pt] at (6.37000,8.85000) {Cross-attention\\Blocks 3--4};
\fill[black!7] (8.81000,7.38667) rectangle ++(1.09333,1.09333);
\node[font=\fontsize{8.0}{9.0}\selectfont,inner sep=0pt] at (9.35667,7.93333) {\textendash};
\fill[GCRateMid!0.000!GCRateLow] (9.90333,7.38667) rectangle ++(1.09333,1.09333);
\node[font=\fontsize{8.0}{9.0}\selectfont,inner sep=0pt] at (10.45000,7.93333) {0.00};
\fill[GCRateMid!0.000!GCRateLow] (10.99667,7.38667) rectangle ++(1.09333,1.09333);
\node[font=\fontsize{8.0}{9.0}\selectfont,inner sep=0pt] at (11.54333,7.93333) {0.00};
\fill[GCRateMid!0.000!GCRateLow] (8.81000,6.29333) rectangle ++(1.09333,1.09333);
\node[font=\fontsize{8.0}{9.0}\selectfont,inner sep=0pt] at (9.35667,6.84000) {0.00};
\fill[black!7] (9.90333,6.29333) rectangle ++(1.09333,1.09333);
\node[font=\fontsize{8.0}{9.0}\selectfont,inner sep=0pt] at (10.45000,6.84000) {\textendash};
\fill[GCRateMid!0.000!GCRateLow] (10.99667,6.29333) rectangle ++(1.09333,1.09333);
\node[font=\fontsize{8.0}{9.0}\selectfont,inner sep=0pt] at (11.54333,6.84000) {0.00};
\fill[GCRateMid!0.000!GCRateLow] (8.81000,5.20000) rectangle ++(1.09333,1.09333);
\node[font=\fontsize{8.0}{9.0}\selectfont,inner sep=0pt] at (9.35667,5.74667) {0.00};
\fill[GCRateMid!0.000!GCRateLow] (9.90333,5.20000) rectangle ++(1.09333,1.09333);
\node[font=\fontsize{8.0}{9.0}\selectfont,inner sep=0pt] at (10.45000,5.74667) {0.00};
\fill[black!7] (10.99667,5.20000) rectangle ++(1.09333,1.09333);
\node[font=\fontsize{8.0}{9.0}\selectfont,inner sep=0pt] at (11.54333,5.74667) {\textendash};
\draw[black!60,line width=.25pt] (8.81000,5.20000) rectangle ++(3.28000,3.28000);
\node[anchor=north east,font=\fontsize{7.6}{8.6}\selectfont,rotate=45,inner sep=1pt] at (9.35667,5.11000) {@6};
\node[anchor=east,font=\fontsize{7.6}{8.6}\selectfont,inner sep=1pt] at (8.73000,7.93333) {@6};
\node[anchor=north east,font=\fontsize{7.6}{8.6}\selectfont,rotate=45,inner sep=1pt] at (10.45000,5.11000) {@8};
\node[anchor=east,font=\fontsize{7.6}{8.6}\selectfont,inner sep=1pt] at (8.73000,6.84000) {@8};
\node[anchor=north east,font=\fontsize{7.6}{8.6}\selectfont,rotate=45,inner sep=1pt] at (11.54333,5.11000) {@10};
\node[anchor=east,font=\fontsize{7.6}{8.6}\selectfont,inner sep=1pt] at (8.73000,5.74667) {@10};
\node[align=center,font=\fontsize{8.8}{9.8}\selectfont\bfseries,inner sep=0pt] at (10.45000,8.85000) {Cross-attention\\Blocks 5--6};
\fill[black!7] (12.89000,6.84000) rectangle ++(1.64000,1.64000);
\node[font=\fontsize{8.0}{9.0}\selectfont,inner sep=0pt] at (13.71000,7.66000) {\textendash};
\fill[GCRateMid!0.000!GCRateLow] (14.53000,6.84000) rectangle ++(1.64000,1.64000);
\node[font=\fontsize{8.0}{9.0}\selectfont,inner sep=0pt] at (15.35000,7.66000) {0.00};
\fill[GCRateMid!0.000!GCRateLow] (12.89000,5.20000) rectangle ++(1.64000,1.64000);
\node[font=\fontsize{8.0}{9.0}\selectfont,inner sep=0pt] at (13.71000,6.02000) {0.00};
\fill[black!7] (14.53000,5.20000) rectangle ++(1.64000,1.64000);
\node[font=\fontsize{8.0}{9.0}\selectfont,inner sep=0pt] at (15.35000,6.02000) {\textendash};
\draw[black!60,line width=.25pt] (12.89000,5.20000) rectangle ++(3.28000,3.28000);
\node[anchor=north east,font=\fontsize{7.6}{8.6}\selectfont,rotate=45,inner sep=1pt] at (13.71000,5.11000) {@8};
\node[anchor=east,font=\fontsize{7.6}{8.6}\selectfont,inner sep=1pt] at (12.81000,7.66000) {@8};
\node[anchor=north east,font=\fontsize{7.6}{8.6}\selectfont,rotate=45,inner sep=1pt] at (15.35000,5.11000) {@10};
\node[anchor=east,font=\fontsize{7.6}{8.6}\selectfont,inner sep=1pt] at (12.81000,6.02000) {@10};
\node[align=center,font=\fontsize{8.8}{9.8}\selectfont\bfseries,inner sep=0pt] at (14.53000,8.85000) {Cross-attention\\Blocks 7--8};
\fill[black!7] (6.77000,3.37400) rectangle ++(0.65600,0.65600);
\node[font=\fontsize{8.0}{9.0}\selectfont,inner sep=0pt] at (7.09800,3.70200) {\textendash};
\fill[GCRateMid!0.000!GCRateLow] (7.42600,3.37400) rectangle ++(0.65600,0.65600);
\node[font=\fontsize{8.0}{9.0}\selectfont,inner sep=0pt] at (7.75400,3.70200) {0.00};
\fill[GCRateMid!0.000!GCRateLow] (8.08200,3.37400) rectangle ++(0.65600,0.65600);
\node[font=\fontsize{8.0}{9.0}\selectfont,inner sep=0pt] at (8.41000,3.70200) {0.00};
\fill[GCRateMid!0.000!GCRateLow] (8.73800,3.37400) rectangle ++(0.65600,0.65600);
\node[font=\fontsize{8.0}{9.0}\selectfont,inner sep=0pt] at (9.06600,3.70200) {0.00};
\fill[GCRateMid!0.000!GCRateLow] (9.39400,3.37400) rectangle ++(0.65600,0.65600);
\node[font=\fontsize{8.0}{9.0}\selectfont,inner sep=0pt] at (9.72200,3.70200) {0.00};
\fill[GCRateMid!0.000!GCRateLow] (6.77000,2.71800) rectangle ++(0.65600,0.65600);
\node[font=\fontsize{8.0}{9.0}\selectfont,inner sep=0pt] at (7.09800,3.04600) {0.00};
\fill[black!7] (7.42600,2.71800) rectangle ++(0.65600,0.65600);
\node[font=\fontsize{8.0}{9.0}\selectfont,inner sep=0pt] at (7.75400,3.04600) {\textendash};
\fill[GCRateMid!0.000!GCRateLow] (8.08200,2.71800) rectangle ++(0.65600,0.65600);
\node[font=\fontsize{8.0}{9.0}\selectfont,inner sep=0pt] at (8.41000,3.04600) {0.00};
\fill[GCRateMid!0.000!GCRateLow] (8.73800,2.71800) rectangle ++(0.65600,0.65600);
\node[font=\fontsize{8.0}{9.0}\selectfont,inner sep=0pt] at (9.06600,3.04600) {0.00};
\fill[GCRateMid!0.000!GCRateLow] (9.39400,2.71800) rectangle ++(0.65600,0.65600);
\node[font=\fontsize{8.0}{9.0}\selectfont,inner sep=0pt] at (9.72200,3.04600) {0.00};
\fill[GCRateMid!0.000!GCRateLow] (6.77000,2.06200) rectangle ++(0.65600,0.65600);
\node[font=\fontsize{8.0}{9.0}\selectfont,inner sep=0pt] at (7.09800,2.39000) {0.00};
\fill[GCRateMid!0.000!GCRateLow] (7.42600,2.06200) rectangle ++(0.65600,0.65600);
\node[font=\fontsize{8.0}{9.0}\selectfont,inner sep=0pt] at (7.75400,2.39000) {0.00};
\fill[black!7] (8.08200,2.06200) rectangle ++(0.65600,0.65600);
\node[font=\fontsize{8.0}{9.0}\selectfont,inner sep=0pt] at (8.41000,2.39000) {\textendash};
\fill[GCRateMid!0.000!GCRateLow] (8.73800,2.06200) rectangle ++(0.65600,0.65600);
\node[font=\fontsize{8.0}{9.0}\selectfont,inner sep=0pt] at (9.06600,2.39000) {0.00};
\fill[GCRateMid!0.000!GCRateLow] (9.39400,2.06200) rectangle ++(0.65600,0.65600);
\node[font=\fontsize{8.0}{9.0}\selectfont,inner sep=0pt] at (9.72200,2.39000) {0.00};
\fill[GCRateMid!0.000!GCRateLow] (6.77000,1.40600) rectangle ++(0.65600,0.65600);
\node[font=\fontsize{8.0}{9.0}\selectfont,inner sep=0pt] at (7.09800,1.73400) {0.00};
\fill[GCRateMid!0.000!GCRateLow] (7.42600,1.40600) rectangle ++(0.65600,0.65600);
\node[font=\fontsize{8.0}{9.0}\selectfont,inner sep=0pt] at (7.75400,1.73400) {0.00};
\fill[GCRateMid!0.000!GCRateLow] (8.08200,1.40600) rectangle ++(0.65600,0.65600);
\node[font=\fontsize{8.0}{9.0}\selectfont,inner sep=0pt] at (8.41000,1.73400) {0.00};
\fill[black!7] (8.73800,1.40600) rectangle ++(0.65600,0.65600);
\node[font=\fontsize{8.0}{9.0}\selectfont,inner sep=0pt] at (9.06600,1.73400) {\textendash};
\fill[GCRateMid!0.000!GCRateLow] (9.39400,1.40600) rectangle ++(0.65600,0.65600);
\node[font=\fontsize{8.0}{9.0}\selectfont,inner sep=0pt] at (9.72200,1.73400) {0.00};
\fill[GCRateMid!0.000!GCRateLow] (6.77000,0.75000) rectangle ++(0.65600,0.65600);
\node[font=\fontsize{8.0}{9.0}\selectfont,inner sep=0pt] at (7.09800,1.07800) {0.00};
\fill[GCRateMid!0.000!GCRateLow] (7.42600,0.75000) rectangle ++(0.65600,0.65600);
\node[font=\fontsize{8.0}{9.0}\selectfont,inner sep=0pt] at (7.75400,1.07800) {0.00};
\fill[GCRateMid!0.000!GCRateLow] (8.08200,0.75000) rectangle ++(0.65600,0.65600);
\node[font=\fontsize{8.0}{9.0}\selectfont,inner sep=0pt] at (8.41000,1.07800) {0.00};
\fill[GCRateMid!0.000!GCRateLow] (8.73800,0.75000) rectangle ++(0.65600,0.65600);
\node[font=\fontsize{8.0}{9.0}\selectfont,inner sep=0pt] at (9.06600,1.07800) {0.00};
\fill[black!7] (9.39400,0.75000) rectangle ++(0.65600,0.65600);
\node[font=\fontsize{8.0}{9.0}\selectfont,inner sep=0pt] at (9.72200,1.07800) {\textendash};
\draw[black!60,line width=.25pt] (6.77000,0.75000) rectangle ++(3.28000,3.28000);
\node[anchor=north east,font=\fontsize{7.6}{8.6}\selectfont,rotate=45,inner sep=1pt] at (7.09800,0.66000) {@2};
\node[anchor=east,font=\fontsize{7.6}{8.6}\selectfont,inner sep=1pt] at (6.69000,3.70200) {@2};
\node[anchor=north east,font=\fontsize{7.6}{8.6}\selectfont,rotate=45,inner sep=1pt] at (7.75400,0.66000) {@4};
\node[anchor=east,font=\fontsize{7.6}{8.6}\selectfont,inner sep=1pt] at (6.69000,3.04600) {@4};
\node[anchor=north east,font=\fontsize{7.6}{8.6}\selectfont,rotate=45,inner sep=1pt] at (8.41000,0.66000) {@6};
\node[anchor=east,font=\fontsize{7.6}{8.6}\selectfont,inner sep=1pt] at (6.69000,2.39000) {@6};
\node[anchor=north east,font=\fontsize{7.6}{8.6}\selectfont,rotate=45,inner sep=1pt] at (9.06600,0.66000) {@8};
\node[anchor=east,font=\fontsize{7.6}{8.6}\selectfont,inner sep=1pt] at (6.69000,1.73400) {@8};
\node[anchor=north east,font=\fontsize{7.6}{8.6}\selectfont,rotate=45,inner sep=1pt] at (9.72200,0.66000) {@10};
\node[anchor=east,font=\fontsize{7.6}{8.6}\selectfont,inner sep=1pt] at (6.69000,1.07800) {@10};
\node[align=center,font=\fontsize{8.8}{9.8}\selectfont\bfseries,inner sep=0pt] at (8.41000,4.40000) {Shared exit FFN};
\begin{scope}[shift={(-0.60,-1.752)}]
\shade[bottom color=GCRateLow,top color=GCRateHigh,middle color=GCRateMid] (13.18000,2.50200) rectangle ++(0.19000,3.00600);
\draw[black!60,line width=.25pt] (13.18000,2.50200) rectangle ++(0.19000,3.00600);
\draw[black!65,line width=.25pt] (13.37000,2.50200) -- ++(.045,0);
\node[anchor=west,font=\fontsize{6.5}{7.5}\selectfont,inner sep=1pt] at (13.47000,2.50200) {0.0};
\draw[black!65,line width=.25pt] (13.37000,4.00500) -- ++(.045,0);
\node[anchor=west,font=\fontsize{6.5}{7.5}\selectfont,inner sep=1pt] at (13.47000,4.00500) {0.5};
\draw[black!65,line width=.25pt] (13.37000,5.50800) -- ++(.045,0);
\node[anchor=west,font=\fontsize{6.5}{7.5}\selectfont,inner sep=1pt] at (13.47000,5.50800) {1.0};
\node[rotate=90,font=\fontsize{6.8}{7.8}\selectfont,inner sep=0pt] at (14.00,4.00500) {Conflict rate};

\end{scope}
\end{tikzpicture}%
\endgroup%

%% file: artwork/gradient_conflict/further/declarevit_gradient_norm_epoch100.tex
\begingroup%
\definecolor{GNBlue}{RGB}{0,114,178}%
\definecolor{GNGreen}{RGB}{0,158,115}%
\definecolor{GNPink}{RGB}{204,121,167}%
\definecolor{GNOchre}{RGB}{181,137,0}%
\definecolor{GNGray}{RGB}{102,102,102}%
\begin{tikzpicture}[x=1cm,y=1cm]
\path[use as bounding box] (-0.10,0.00) rectangle (16.35,9.25);
\node[font=\fontsize{9.4}{10.4}\selectfont\bfseries,inner sep=0pt] at (8.125,9.04) {(B) Loss-weighted gradient norms};
\node[align=center,font=\fontsize{8.4}{9.4}\selectfont\bfseries,inner sep=0pt] at (2.290,8.400) {Cross-attention\\Blocks 1--2};
\node[anchor=east,font=\fontsize{7.6}{8.6}\selectfont,inner sep=.4pt] at (0.592,4.900) {0.00};
\draw[black!12,line width=.20pt] (0.650,5.288) -- (3.930,5.288);
\node[anchor=east,font=\fontsize{7.6}{8.6}\selectfont,inner sep=.4pt] at (0.592,5.288) {0.20};
\draw[black!12,line width=.20pt] (0.650,5.675) -- (3.930,5.675);
\node[anchor=east,font=\fontsize{7.6}{8.6}\selectfont,inner sep=.4pt] at (0.592,5.675) {0.40};
\draw[black!12,line width=.20pt] (0.650,6.062) -- (3.930,6.062);
\node[anchor=east,font=\fontsize{7.6}{8.6}\selectfont,inner sep=.4pt] at (0.592,6.062) {0.60};
\draw[black!12,line width=.20pt] (0.650,6.450) -- (3.930,6.450);
\node[anchor=east,font=\fontsize{7.6}{8.6}\selectfont,inner sep=.4pt] at (0.592,6.450) {0.80};
\draw[black!12,line width=.20pt] (0.650,6.838) -- (3.930,6.838);
\node[anchor=east,font=\fontsize{7.6}{8.6}\selectfont,inner sep=.4pt] at (0.592,6.838) {1.00};
\draw[black!12,line width=.20pt] (0.650,7.225) -- (3.930,7.225);
\node[anchor=east,font=\fontsize{7.6}{8.6}\selectfont,inner sep=.4pt] at (0.592,7.225) {1.20};
\draw[black!12,line width=.20pt] (0.650,7.613) -- (3.930,7.613);
\node[anchor=east,font=\fontsize{7.6}{8.6}\selectfont,inner sep=.4pt] at (0.592,7.613) {1.40};
\draw[black!12,line width=.20pt] (0.650,8.000) -- (3.930,8.000);
\node[anchor=east,font=\fontsize{7.6}{8.6}\selectfont,inner sep=.4pt] at (0.592,8.000) {1.60};
\draw[black!75,line width=.30pt] (0.650,4.900) -- (3.930,4.900);
\draw[black!75,line width=.30pt] (0.650,4.900) -- (0.650,8.000);
\fill[GNBlue] (0.7681,4.9000) rectangle ++(0.4198,2.5963);
\node[anchor=south,font=\fontsize{7.4}{8.4}\selectfont,inner sep=.25pt] at (0.9780,7.5563) {1.340};
\node[anchor=north east,rotate=45,font=\fontsize{7.6}{8.6}\selectfont,inner sep=.25pt] at (0.9780,4.8000) {@2};
\fill[GNGreen] (1.4241,4.9000) rectangle ++(0.4198,1.1315);
\node[anchor=south,font=\fontsize{7.4}{8.4}\selectfont,inner sep=.25pt] at (1.6340,6.0915) {0.584};
\node[anchor=north east,rotate=45,font=\fontsize{7.6}{8.6}\selectfont,inner sep=.25pt] at (1.6340,4.8000) {@4};
\fill[GNPink] (2.0801,4.9000) rectangle ++(0.4198,0.5367);
\node[anchor=south,font=\fontsize{7.4}{8.4}\selectfont,inner sep=.25pt] at (2.2900,5.4967) {0.277};
\node[anchor=north east,rotate=45,font=\fontsize{7.6}{8.6}\selectfont,inner sep=.25pt] at (2.2900,4.8000) {@6};
\fill[GNOchre] (2.7361,4.9000) rectangle ++(0.4198,0.3081);
\node[anchor=south,font=\fontsize{7.4}{8.4}\selectfont,inner sep=.25pt] at (2.9460,5.2681) {0.159};
\node[anchor=north east,rotate=45,font=\fontsize{7.6}{8.6}\selectfont,inner sep=.25pt] at (2.9460,4.8000) {@8};
\fill[GNGray] (3.3921,4.9000) rectangle ++(0.4198,0.1957);
\node[anchor=south,font=\fontsize{7.4}{8.4}\selectfont,inner sep=.25pt] at (3.6020,5.1557) {0.101};
\node[anchor=north east,rotate=45,font=\fontsize{7.6}{8.6}\selectfont,inner sep=.25pt] at (3.6020,4.8000) {@10};
\node[align=center,font=\fontsize{8.4}{9.4}\selectfont\bfseries,inner sep=0pt] at (6.370,8.400) {Cross-attention\\Blocks 3--4};
\draw[black!12,line width=.20pt] (4.730,5.288) -- (8.010,5.288);
\draw[black!12,line width=.20pt] (4.730,5.675) -- (8.010,5.675);
\draw[black!12,line width=.20pt] (4.730,6.062) -- (8.010,6.062);
\draw[black!12,line width=.20pt] (4.730,6.450) -- (8.010,6.450);
\draw[black!12,line width=.20pt] (4.730,6.838) -- (8.010,6.838);
\draw[black!12,line width=.20pt] (4.730,7.225) -- (8.010,7.225);
\draw[black!12,line width=.20pt] (4.730,7.613) -- (8.010,7.613);
\draw[black!12,line width=.20pt] (4.730,8.000) -- (8.010,8.000);
\draw[black!75,line width=.30pt] (4.730,4.900) -- (8.010,4.900);
\draw[black!75,line width=.30pt] (4.730,4.900) -- (4.730,8.000);
\fill[GNGreen] (4.8776,4.9000) rectangle ++(0.5248,1.3330);
\node[anchor=south,font=\fontsize{8.0}{9.0}\selectfont,inner sep=.25pt] at (5.1400,6.2930) {0.688};
\node[anchor=north east,rotate=45,font=\fontsize{7.6}{8.6}\selectfont,inner sep=.25pt] at (5.1400,4.8000) {@4};
\fill[GNPink] (5.6976,4.9000) rectangle ++(0.5248,0.6452);
\node[anchor=south,font=\fontsize{8.0}{9.0}\selectfont,inner sep=.25pt] at (5.9600,5.6052) {0.333};
\node[anchor=north east,rotate=45,font=\fontsize{7.6}{8.6}\selectfont,inner sep=.25pt] at (5.9600,4.8000) {@6};
\fill[GNOchre] (6.5176,4.9000) rectangle ++(0.5248,0.3778);
\node[anchor=south,font=\fontsize{8.0}{9.0}\selectfont,inner sep=.25pt] at (6.7800,5.3378) {0.195};
\node[anchor=north east,rotate=45,font=\fontsize{7.6}{8.6}\selectfont,inner sep=.25pt] at (6.7800,4.8000) {@8};
\fill[GNGray] (7.3376,4.9000) rectangle ++(0.5248,0.2422);
\node[anchor=south,font=\fontsize{8.0}{9.0}\selectfont,inner sep=.25pt] at (7.6000,5.2022) {0.125};
\node[anchor=north east,rotate=45,font=\fontsize{7.6}{8.6}\selectfont,inner sep=.25pt] at (7.6000,4.8000) {@10};
\node[align=center,font=\fontsize{8.4}{9.4}\selectfont\bfseries,inner sep=0pt] at (10.450,8.400) {Cross-attention\\Blocks 5--6};
\draw[black!12,line width=.20pt] (8.810,5.288) -- (12.090,5.288);
\draw[black!12,line width=.20pt] (8.810,5.675) -- (12.090,5.675);
\draw[black!12,line width=.20pt] (8.810,6.062) -- (12.090,6.062);
\draw[black!12,line width=.20pt] (8.810,6.450) -- (12.090,6.450);
\draw[black!12,line width=.20pt] (8.810,6.838) -- (12.090,6.838);
\draw[black!12,line width=.20pt] (8.810,7.225) -- (12.090,7.225);
\draw[black!12,line width=.20pt] (8.810,7.613) -- (12.090,7.613);
\draw[black!12,line width=.20pt] (8.810,8.000) -- (12.090,8.000);
\draw[black!75,line width=.30pt] (8.810,4.900) -- (12.090,4.900);
\draw[black!75,line width=.30pt] (8.810,4.900) -- (8.810,8.000);
\fill[GNPink] (9.0068,4.9000) rectangle ++(0.6997,0.7014);
\node[anchor=south,font=\fontsize{8.0}{9.0}\selectfont,inner sep=.25pt] at (9.3567,5.6614) {0.362};
\node[anchor=north east,rotate=45,font=\fontsize{7.6}{8.6}\selectfont,inner sep=.25pt] at (9.3567,4.8000) {@6};
\fill[GNOchre] (10.1001,4.9000) rectangle ++(0.6997,0.3739);
\node[anchor=south,font=\fontsize{8.0}{9.0}\selectfont,inner sep=.25pt] at (10.4500,5.3339) {0.193};
\node[anchor=north east,rotate=45,font=\fontsize{7.6}{8.6}\selectfont,inner sep=.25pt] at (10.4500,4.8000) {@8};
\fill[GNGray] (11.1935,4.9000) rectangle ++(0.6997,0.2286);
\node[anchor=south,font=\fontsize{8.0}{9.0}\selectfont,inner sep=.25pt] at (11.5433,5.1886) {0.118};
\node[anchor=north east,rotate=45,font=\fontsize{7.6}{8.6}\selectfont,inner sep=.25pt] at (11.5433,4.8000) {@10};
\node[align=center,font=\fontsize{8.4}{9.4}\selectfont\bfseries,inner sep=0pt] at (14.530,8.400) {Cross-attention\\Blocks 7--8};
\draw[black!12,line width=.20pt] (12.890,5.288) -- (16.170,5.288);
\draw[black!12,line width=.20pt] (12.890,5.675) -- (16.170,5.675);
\draw[black!12,line width=.20pt] (12.890,6.062) -- (16.170,6.062);
\draw[black!12,line width=.20pt] (12.890,6.450) -- (16.170,6.450);
\draw[black!12,line width=.20pt] (12.890,6.838) -- (16.170,6.838);
\draw[black!12,line width=.20pt] (12.890,7.225) -- (16.170,7.225);
\draw[black!12,line width=.20pt] (12.890,7.613) -- (16.170,7.613);
\draw[black!12,line width=.20pt] (12.890,8.000) -- (16.170,8.000);
\draw[black!75,line width=.30pt] (12.890,4.900) -- (16.170,4.900);
\draw[black!75,line width=.30pt] (12.890,4.900) -- (12.890,8.000);
\fill[GNOchre] (13.1852,4.9000) rectangle ++(1.0496,0.4437);
\node[anchor=south,font=\fontsize{8.0}{9.0}\selectfont,inner sep=.25pt] at (13.7100,5.4037) {0.229};
\node[anchor=north east,rotate=45,font=\fontsize{7.6}{8.6}\selectfont,inner sep=.25pt] at (13.7100,4.8000) {@8};
\fill[GNGray] (14.8252,4.9000) rectangle ++(1.0496,0.2461);
\node[anchor=south,font=\fontsize{8.0}{9.0}\selectfont,inner sep=.25pt] at (15.3500,5.2061) {0.127};
\node[anchor=north east,rotate=45,font=\fontsize{7.6}{8.6}\selectfont,inner sep=.25pt] at (15.3500,4.8000) {@10};
\node[align=center,font=\fontsize{8.4}{9.4}\selectfont\bfseries,inner sep=0pt] at (8.410,4.140) {Shared exit FFN};
\node[anchor=east,font=\fontsize{7.6}{8.6}\selectfont,inner sep=.4pt] at (6.712,0.850) {0.00};
\draw[black!12,line width=.20pt] (6.770,1.238) -- (10.050,1.238);
\node[anchor=east,font=\fontsize{7.6}{8.6}\selectfont,inner sep=.4pt] at (6.712,1.238) {0.20};
\draw[black!12,line width=.20pt] (6.770,1.625) -- (10.050,1.625);
\node[anchor=east,font=\fontsize{7.6}{8.6}\selectfont,inner sep=.4pt] at (6.712,1.625) {0.40};
\draw[black!12,line width=.20pt] (6.770,2.013) -- (10.050,2.013);
\node[anchor=east,font=\fontsize{7.6}{8.6}\selectfont,inner sep=.4pt] at (6.712,2.013) {0.60};
\draw[black!12,line width=.20pt] (6.770,2.400) -- (10.050,2.400);
\node[anchor=east,font=\fontsize{7.6}{8.6}\selectfont,inner sep=.4pt] at (6.712,2.400) {0.80};
\draw[black!12,line width=.20pt] (6.770,2.788) -- (10.050,2.788);
\node[anchor=east,font=\fontsize{7.6}{8.6}\selectfont,inner sep=.4pt] at (6.712,2.788) {1.00};
\draw[black!12,line width=.20pt] (6.770,3.175) -- (10.050,3.175);
\node[anchor=east,font=\fontsize{7.6}{8.6}\selectfont,inner sep=.4pt] at (6.712,3.175) {1.20};
\draw[black!12,line width=.20pt] (6.770,3.563) -- (10.050,3.563);
\node[anchor=east,font=\fontsize{7.6}{8.6}\selectfont,inner sep=.4pt] at (6.712,3.563) {1.40};
\draw[black!12,line width=.20pt] (6.770,3.950) -- (10.050,3.950);
\node[anchor=east,font=\fontsize{7.6}{8.6}\selectfont,inner sep=.4pt] at (6.712,3.950) {1.60};
\draw[black!75,line width=.30pt] (6.770,0.850) -- (10.050,0.850);
\draw[black!75,line width=.30pt] (6.770,0.850) -- (6.770,3.950);
\fill[GNBlue] (6.8881,0.8500) rectangle ++(0.4198,0.8292);
\node[anchor=south,font=\fontsize{7.4}{8.4}\selectfont,inner sep=.25pt] at (7.0980,1.7392) {0.428};
\node[anchor=north east,rotate=45,font=\fontsize{7.6}{8.6}\selectfont,inner sep=.25pt] at (7.0980,0.7500) {@2};
\fill[GNGreen] (7.5441,0.8500) rectangle ++(0.4198,0.6200);
\node[anchor=south,font=\fontsize{7.4}{8.4}\selectfont,inner sep=.25pt] at (7.7540,1.5300) {0.320};
\node[anchor=north east,rotate=45,font=\fontsize{7.6}{8.6}\selectfont,inner sep=.25pt] at (7.7540,0.7500) {@4};
\fill[GNPink] (8.2001,0.8500) rectangle ++(0.4198,0.4359);
\node[anchor=south,font=\fontsize{7.4}{8.4}\selectfont,inner sep=.25pt] at (8.4100,1.3459) {0.225};
\node[anchor=north east,rotate=45,font=\fontsize{7.6}{8.6}\selectfont,inner sep=.25pt] at (8.4100,0.7500) {@6};
\fill[GNOchre] (8.8561,0.8500) rectangle ++(0.4198,0.3197);
\node[anchor=south,font=\fontsize{7.4}{8.4}\selectfont,inner sep=.25pt] at (9.0660,1.2297) {0.165};
\node[anchor=north east,rotate=45,font=\fontsize{7.6}{8.6}\selectfont,inner sep=.25pt] at (9.0660,0.7500) {@8};
\fill[GNGray] (9.5121,0.8500) rectangle ++(0.4198,0.2344);
\node[anchor=south,font=\fontsize{7.4}{8.4}\selectfont,inner sep=.25pt] at (9.7220,1.1444) {0.121};
\node[anchor=north east,rotate=45,font=\fontsize{7.6}{8.6}\selectfont,inner sep=.25pt] at (9.7220,0.7500) {@10};
\end{tikzpicture}%
\endgroup%

%% file: sections/supplement/Exit_Queries_and_CLS_Token.tex
\origsection{Attention Visualization of [\textsc{cls}] and Exit Queries}
\label{app:attention_visualization}

Figures~\ref{fig:attention-cifar100-overview},
\ref{fig:attention-food101-overview}, and
\ref{fig:attention-imagenet1k-overview}
visualize the patch attention of the backbone
[\textsc{cls}] token and the corresponding exit query at different exit
depths. For each attention module, we aggregate the multi-head attention
weights by averaging them across all heads and retain only the weights
assigned to patch tokens. For each sample, the first two rows show the attention maps of
[\textsc{cls}] and the exit query, respectively, while the third row
shows their pointwise difference (\emph{Query minus CLS}). Red indicates
regions receiving greater relative attention from the exit query,
whereas blue indicates regions receiving greater relative attention
from [\textsc{cls}].

The examples show that the exit queries and backbone [\textsc{cls}] token
can exhibit different spatial attention patterns. In the shown cases, exit-query attention is more concentrated on
object-related regions, while the [\textsc{cls}] attention evolves across
backbone depth. These qualitative observations are consistent with the
intended separation between the dedicated early-classification state and
the backbone representation in DeclareViT.

\input{artwork/attention/figures/cifar100/overview.tex}
\input{artwork/attention/figures/food101/overview.tex}
\input{artwork/attention/figures/imagenet1k/overview.tex}

%% file: artwork/attention/figures/cifar100/overview.tex
\begin{figure*}[p]
\centering
\AttentionDatasetOverview{cifar100}{CIFAR-100}{0.07216409593820572}{3}{%
\input{artwork/attention/panels/cifar100/sample_00000958.tex}%
  \vspace{-2em}
\input{artwork/attention/panels/cifar100/sample_00002782.tex}%
  \vspace{-2em}
\input{artwork/attention/panels/cifar100/sample_00003292.tex}%
}%
\vspace{-2em}
\caption{Attention visualization for CIFAR-100 test samples. Within each sample block, the input is on the left; the three rows on the right show attention for Backbone CLS, Exit Query, and Query minus CLS. }
\label{fig:attention-cifar100-overview}
\end{figure*}

\begin{figure*}[p]
\ContinuedFloat
\centering
\AttentionDatasetOverview{cifar100}{CIFAR-100}{0.07216409593820572}{3}{%
  \input{artwork/attention/panels/cifar100/sample_00003507.tex}%
  \vspace{-2em}
\input{artwork/attention/panels/cifar100/sample_00003744.tex}%
\vspace{-2em}\input{artwork/attention/panels/cifar100/sample_00007178.tex}%
}%
\vspace{-2em}
\caption[]{Attention visualization for CIFAR-100 (continued). }
\end{figure*}

%% file: artwork/attention/panels/cifar100/sample_00000958.tex
\AttentionDatasetRow{0}{00000958}{otter}{0.0857214480638504}%

%% file: artwork/attention/panels/cifar100/sample_00002782.tex
\AttentionDatasetRow{1}{00002782}{poppy}{0.09480179846286774}%

%% file: artwork/attention/panels/cifar100/sample_00003292.tex
\AttentionDatasetRow{2}{00003292}{pear}{0.06106363609433174}%

%% file: artwork/attention/panels/cifar100/sample_00003507.tex
\AttentionDatasetRow{3}{00003507}{bowl}{0.08588717877864838}%

%% file: artwork/attention/panels/cifar100/sample_00003744.tex
\AttentionDatasetRow{4}{00003744}{cup}{0.06457071751356125}%

%% file: artwork/attention/panels/cifar100/sample_00007178.tex
\AttentionDatasetRow{5}{00007178}{willow tree}{0.0633401945233345}%

%% file: artwork/attention/figures/food101/overview.tex
\begin{figure*}[p]
\centering
\AttentionDatasetOverview{food101}{Food-101}{0.08914212882518768}{3}{%
\input{artwork/attention/panels/food101/sample_00001326.tex}%
  \vspace{-2em}
\input{artwork/attention/panels/food101/sample_00004553.tex}%
  \vspace{-2em}
\input{artwork/attention/panels/food101/sample_00006654.tex}%
}%
  \vspace{-2em}
\caption{Attention visualization for Food-101 test samples. Within each sample block, the input is on the left; the three rows on the right show attention for Backbone CLS, Exit Query, and Query minus CLS. }
\label{fig:attention-food101-overview}
\end{figure*}

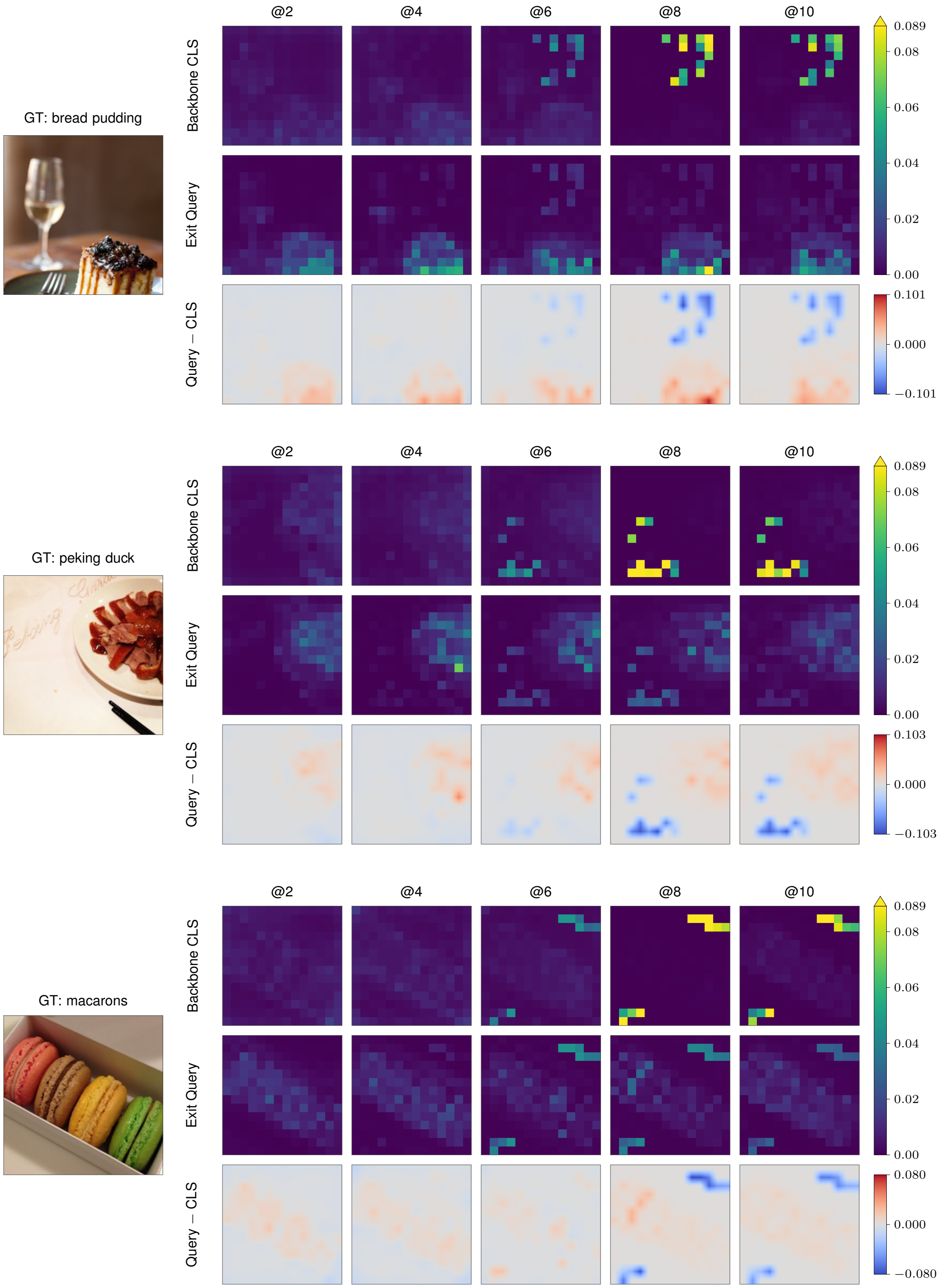
\begin{figure*}[p]
\ContinuedFloat
\centering
\AttentionDatasetOverview{food101}{Food-101}{0.08914212882518768}{3}{%
\input{artwork/attention/panels/food101/sample_00007147.tex}%
  \vspace{-2em}
\input{artwork/attention/panels/food101/sample_00013627.tex}%
\vspace{-2em}\input{artwork/attention/panels/food101/sample_00021388.tex}%
}%
  \vspace{-2em}
\caption[]{Attention visualization for Food-101 (continued). }
\end{figure*}

%% file: artwork/attention/panels/food101/sample_00001326.tex
\AttentionDatasetRow{0}{00001326}{spring rolls}{0.13342013955116272}%

%% file: artwork/attention/panels/food101/sample_00004553.tex
\AttentionDatasetRow{1}{00004553}{crab cakes}{0.11695902794599533}%

%% file: artwork/attention/panels/food101/sample_00006654.tex
\AttentionDatasetRow{2}{00006654}{carrot cake}{0.10099256038665771}%

%% file: artwork/attention/panels/food101/sample_00007147.tex
\AttentionDatasetRow{3}{00007147}{bread pudding}{0.1007891297340393}%

%% file: artwork/attention/panels/food101/sample_00013627.tex
\AttentionDatasetRow{4}{00013627}{peking duck}{0.10262875258922577}%

%% file: artwork/attention/panels/food101/sample_00021388.tex
\AttentionDatasetRow{5}{00021388}{macarons}{0.07963263988494873}%

%% file: artwork/attention/figures/imagenet1k/overview.tex
\begin{figure*}[p]
\centering
\AttentionDatasetOverview{imagenet1k}{ImageNet-1K}{0.07888570427894592}{3}{%
\input{artwork/attention/panels/imagenet1k/sample_00009107.tex}%
\vspace{-2em}\input{artwork/attention/panels/imagenet1k/sample_00017166.tex}%
\vspace{-2em}\input{artwork/attention/panels/imagenet1k/sample_00017381.tex}%
}%
\vspace{-2em}
\caption{Attention visualization for ImageNet-1K validation samples. Within each sample block, the input is on the left; the three rows on the right show attention for Backbone CLS, Exit Query, and Query minus CLS.}
\label{fig:attention-imagenet1k-overview}
\end{figure*}
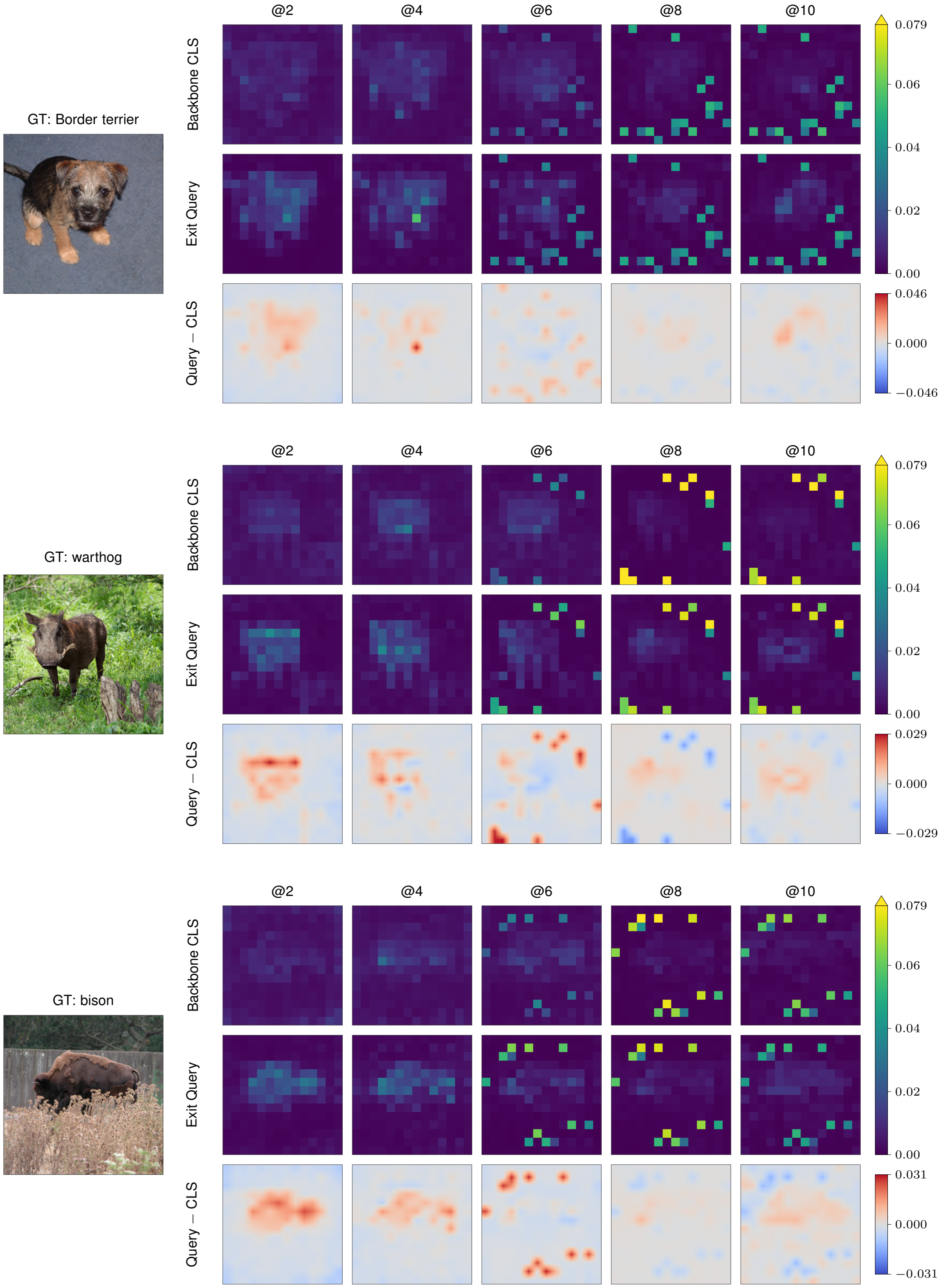

\begin{figure*}[p]
\ContinuedFloat
\centering
\AttentionDatasetOverview{imagenet1k}{ImageNet-1K}{0.07888570427894592}{3}{%
\input{artwork/attention/panels/imagenet1k/sample_00026875.tex}%
\vspace{-2em}
\input{artwork/attention/panels/imagenet1k/sample_00046434.tex}%
\vspace{-2em}
\input{artwork/attention/panels/imagenet1k/sample_00041690.tex}%
}%
\vspace{-2em}
\caption[]{Attention visualization for ImageNet-1K (continued).}
\end{figure*}
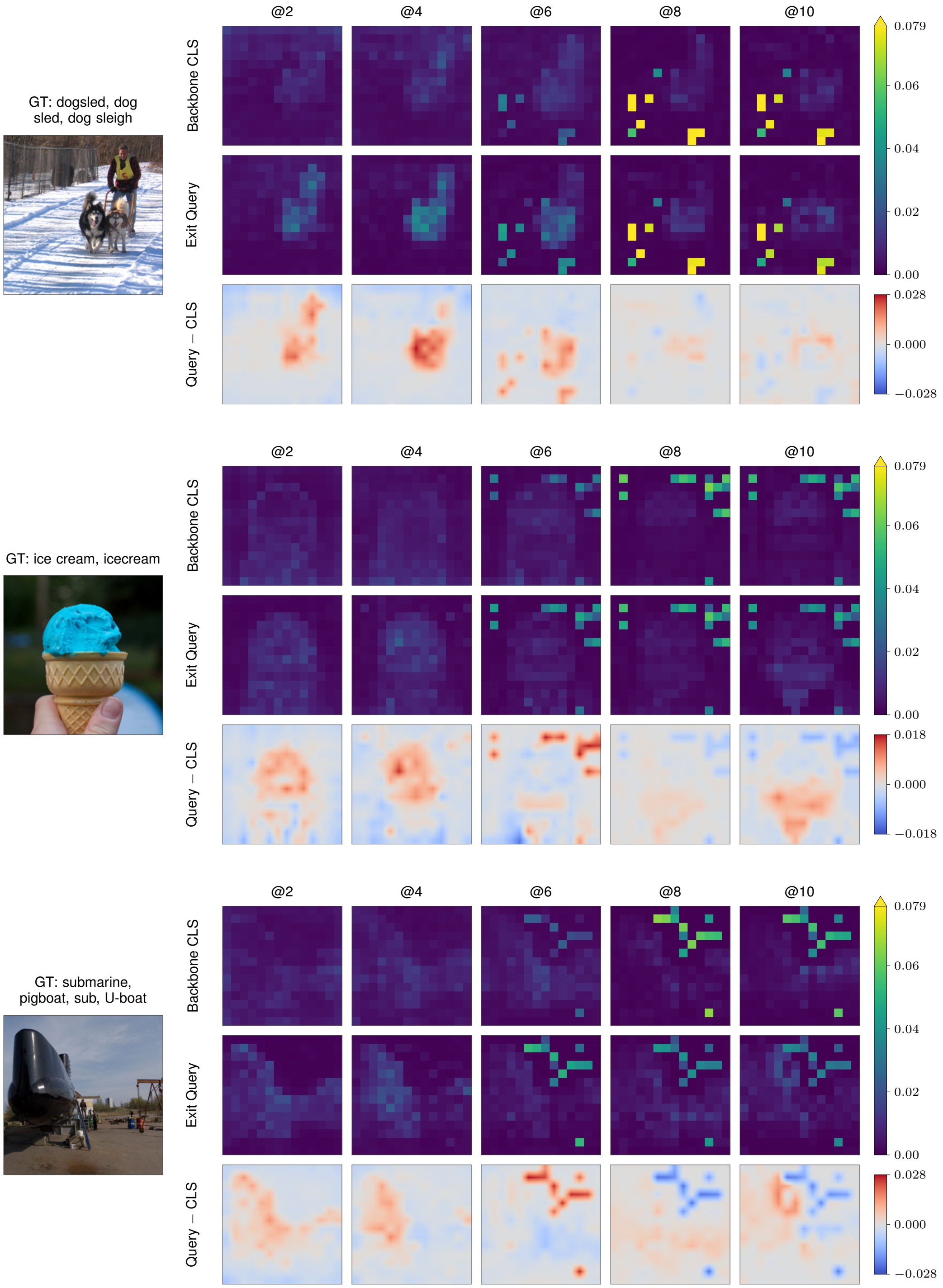

%% file: artwork/attention/panels/imagenet1k/sample_00009107.tex
\AttentionDatasetRow{0}{00009107}{Border terrier}{0.04638783633708954}%

%% file: artwork/attention/panels/imagenet1k/sample_00017166.tex
\AttentionDatasetRow{1}{00017166}{warthog}{0.02870018780231476}%

%% file: artwork/attention/panels/imagenet1k/sample_00017381.tex
\AttentionDatasetRow{2}{00017381}{bison}{0.030849911272525787}%

%% file: artwork/attention/panels/imagenet1k/sample_00026875.tex
\AttentionDatasetRow{3}{00026875}{dogsled, dog sled, dog sleigh}{0.02849351242184639}%

%% file: artwork/attention/panels/imagenet1k/sample_00046434.tex
\AttentionDatasetRow{4}{00046434}{ice cream, icecream}{0.017656926065683365}%

%% file: artwork/attention/panels/imagenet1k/sample_00041690.tex
\AttentionDatasetRow{5}{00041690}{submarine, pigboat, sub, U-boat}{0.02841564640402794}%